\documentclass{article}

\usepackage{microtype}
\usepackage{graphicx}
\usepackage{subcaption}
\usepackage{booktabs} 

\usepackage{hyperref}

\usepackage[accepted]{icml2026}

\usepackage{amsmath}
\usepackage{amssymb}
\usepackage{mathtools}
\usepackage{amsthm}

\usepackage[capitalize,noabbrev]{cleveref}

\usepackage{amsmath,amssymb,amsfonts}
\usepackage{algorithm}
\usepackage{algorithmic}
\usepackage{graphicx}
\usepackage{textcomp}
\usepackage{xcolor}

\usepackage{url}
\usepackage{mathtools}
\usepackage{amsthm}
\usepackage{multirow}
\usepackage{booktabs}
\usepackage{mathrsfs}
\usepackage{makecell}
\usepackage[most]{tcolorbox}

\usepackage{graphicx}

\usepackage[frozencache,cachedir=.]{minted}
\usepackage[scaled=0.825]{beramono}
\usepackage[table,x11names,dvipsnames]{xcolor}
\definecolor{BlueViolet}{rgb}{0.54, 0.17, 0.89}
\newcommand{\code}[2][BlueViolet]{\textcolor{#1}{\texttt{#2}}}
\newcommand{\ours}{\code[BlueViolet]{T2L}}
\newcommand{\Larch}{\colorbox{SteelBlue1!80}{\textbf{L}}}
\newcommand{\March}{\colorbox{MediumPurple1!60}{\textbf{M}}}
\newcommand{\Sarch}{\colorbox{Pink1}{\textbf{S}}}

\newcommand{\greenbox}[1]{\colorbox{LimeGreen}{\textbf{#1}}}

\definecolor{green}{rgb}{0.1,0.1,0.1}
\definecolor{gitgreen}{HTML}{006400}
\definecolor{chocolate}{HTML}{D2691E}
\definecolor{maroon}{HTML}{A00000}
\definecolor{indigo}{HTML}{4B0082}
\definecolor{green}{HTML}{008000}
\definecolor{red}{HTML}{e41a1c}

\usepackage{amssymb}
\usepackage{pifont}
\newcommand{\cmark}{{\protect\color{maroon} \ding{51}}}
\newcommand{\xmark}{\ding{55}}

\theoremstyle{plain}

\theoremstyle{definition}

\theoremstyle{remark}

\usepackage[textsize=tiny]{todonotes}

\usepackage{minted}

\newcommand{\ModelName}{{MeGan}}

\icmltitlerunning{Learn-To-Learn on Arbitrary Textual Conditioning: A Hypernetwork-Driven Meta-Gated LLM}

\begin{document}

\twocolumn[
  \icmltitle{Learn-To-Learn on Arbitrary Textual Conditioning:\\A Hypernetwork-Driven Meta-Gated LLM}



  \icmlsetsymbol{equal}{*}

  \begin{icmlauthorlist}
    \icmlauthor{Luo Ji}{equal,lab}
    \icmlauthor{Qi Qin}{equal,sch}
    \icmlauthor{Ningyuan Xi}{equal,lab}
    \icmlauthor{Teng Chen}{lab}
    \icmlauthor{Qingqing Gu}{lab}
    \icmlauthor{Hongyan Li}{lab}
  \end{icmlauthorlist}

  \icmlaffiliation{lab}{Geely AI Lab, Geely Auto Group, Zhejiang, China}
  \icmlaffiliation{sch}{Peking University, Beijing, China} 

  \icmlcorrespondingauthor{Hongyan Li}{lihongyan\_csu@163.com}

  \icmlkeywords{LLM, meta-learning, hypernetwork, meta-gating, SwiGLU}

  \vskip 0.3in
]



\printAffiliationsAndNotice{\icmlEqualContribution}

\begin{abstract}
Conventional LLMs may suffer from corpus heterogeneity and subtle changes in conditions. While finetuning can create the catastrophe forgetting issue, applications of meta-learning on LLMs are also limited due to their complexity and scalability. In this paper, we activate the meta-signal of $\beta$ within the SwiGLU blocks, resulting in a meta-gating mechanism that adaptively adjusts the nonlinearity of FFN. A hypernetwork is employed to dynamically produce $\beta$ under textual conditions, providing meta-controllability over LLMs. By testing on different condition types such as task, domain, persona, and style, our method outperforms finetuning and meta-learning baselines, and can generalize reasonably on unseen tasks, condition types, or instructions. Our codes are in \url{https://github.com/AaronJi/MeGan}.
\end{abstract}

\section{Introduction}
\label{sec:intro}

The vast diversity of tasks, styles, domains, and other conditions present in the environment for both human and AI, which raises the problem of meta-learning, or in other words, the capability of `learn-to-learn' given different contexts. While Large Language Models (LLMs) have demonstrated astonishing capabilities across diverse natural language processing (NLP) tasks, they remain relatively deficient in the seamless, dynamic adaptation to those conditions \citep{doi:10.1073/pnas.2422455122,zamaraeva-etal-2025-comparing}. When finetuned on a specific downstream task, their general abilities often degrade, sometimes termed `catastrophic forgetting' or `alignment tax', which hurts their adaptation capability to new conditions. This performance erosion presents a fundamental challenge to developing robust and versatile AI systems, preventing the application of personalized systems or expert-level emotional supporters \citep{srinivas-etal-2025-substance}.

Varied meta-training methods are proposed to alleviate these issues, including in-context learning (ICL) \citep{min-etal-2022-metaicl,10.1145/3637528.3671905,NEURIPS2023_cda04d7e}, gradient-based meta-learning \cite{pmlr-v70-finn17a,song-etal-2020-learning}, parameter-efficient fine-tuning (PEFT) \citep{hu2022lora,tian2024hydralora,li-etal-2025-meta}, and adapter methods \citep{pmlr-v97-houlsby19a,he2022towards,hu-etal-2023-llm}. However, ICL usually lags behind finetuning, produces high variance results, resulting in the issue of `lost in needle' by creating a long context window \citep{ponce2025incontextlearningvsinstruction}. Gradient or adapter-based methods have high computational overhead in NLP tasks or are difficult to scale up. PEFT methods like LoRA are computationally efficient and have good adaptivity. However, they generally require downstream task-specific training, which limits their applications.

\begin{figure}[!t]
  \includegraphics[width=1\linewidth]{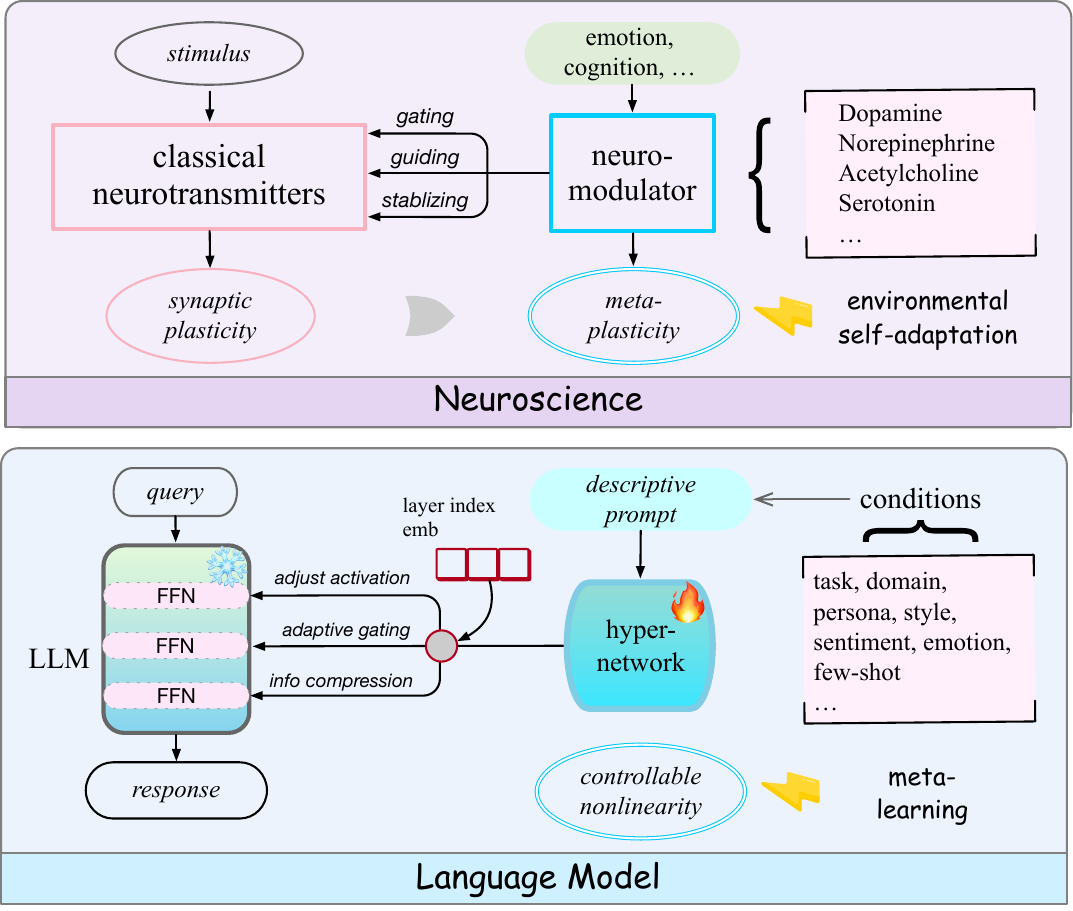}
  \caption{Paradigm of {\ModelName} analogous to neuro-system. Nueromodulator meta-controls classical neurotransmitters, shaping the synaptic plasticity to meta-plasticity. Similarly, we implement a hypernetwork that converts textual condition inputs into meta signals. These signals are combined with layer index embedding, then meta-control the gating of LLM FFNs.}
  \label{fig:paradigm}
\end{figure}

In contrast, the human nervous system overcomes the aforementioned issues, where the basic stimulus is responded to by synaptic plasticity based on classical neurotransmitters; and neuromodulators produce the adaptive gain control, through several mechanisms including gating, guidance, and stabilization \citep{annurev:/content/journals/10.1146/annurev.neuro.28.061604.135709}. The system then achieves meta-plasticity, which seamlessly adapts to changing environmental contexts without the metabolic cost of physical rewiring \citep{doi:10.1126/science.2392679}. Analogously, pretrained LLMs can be considered as generalized query-response processors. Instead of directly prompting or finetuning on it, we implement a hypernetwork with textual \textit{condition} as input, which produces a meta-signal on the original LLM. With original LLM parameters frozen, the hypernetwork is trained by post-hoc adaptation on meta-training samples, which then adapt the LLM conditioned on target tasks (Figure \ref{fig:paradigm}). 

\begin{figure}[!t]
  \includegraphics[width=0.99\linewidth]{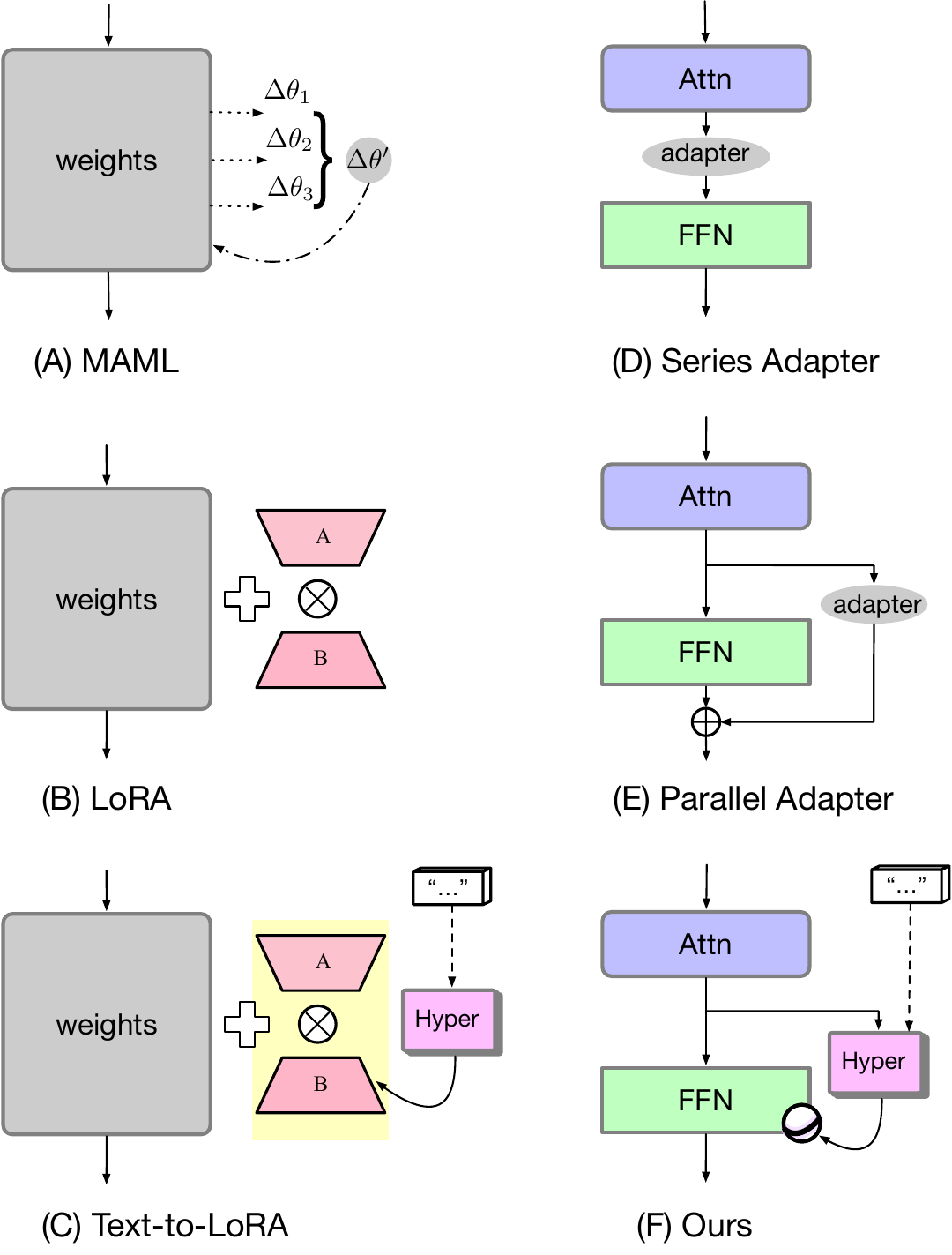}
  \caption {Comparison of meta-learning paradigms on LLMs.}
  \label{fig:paradigm_comparison}
\end{figure}


Similar to Text-to-LoRA \citep{charakorn2025texttolora}, a LoRA-based architecture driven by a text-input hypernetwork (Figure \ref{fig:paradigm_comparison} (C)), we also introduce the architecture customization based on textual conditions, to provide higher adaptability. The architecture is similar to the parallel adapter (Figure \ref{fig:paradigm_comparison} (E)), by implementing the hypernetwork as a parallel module with FFN, with the cross-attention between LLM latent and textual conditions. Different from adapters which directly control the LLM latent \citep{qiu2025gated}, our hypernetwork produces the meta-signal to adaptively adjust the activations of feedforward network (FFN), resulting in different levels of nonlinearity. This design mirrors the mechanism of the neuron's f-I (frequency-current) curve \citep{Silver2010NeuronalA}, and is consistent with previous observations that FFNs encode persona or style semantics earlier and better than self-attention \citep{poonia-jain-2025-dissecting}. To achieve this, we replace the Swish$_1$ activation with $\text{Swish}_{\beta}$, resulting a new type of FFN block called $\beta-$SwiGLU. The hypernetwork produces $\beta$ by a bottleneck structure, extracting key conditional patterns by information compression \citep{10.1007/978-3-642-02565-5_4,illya_KC,deletang2024language}, replacing the original self-gating by meta-gating. Correspondingly, the entire framework is named by \textbf{Me}ta-\textbf{Ga}ting-on-Conditio\textbf{n} (\textbf{\ModelName}).

{\ModelName} is trained by supervised fine-tuning (SFT), with an auxiliary L-2 regularization on MLP weights. We experiment it on different settings of meta-learning, with different meta-conditions such as task, domain, persona, style, and sentiment. Our framework outperforms prompting finetuning, or meta-learning baselines, showing good zero-shot performance on unseen tasks or condition types. Further analysis shows that {\ModelName} is scalable, robust, and parameter-efficient. Distribution of $\beta$ also reveals that {\ModelName} adaptively captures the semantics of conditions by dynamic $\beta$ calculation. To summarize, our main contributions include:
\begin{itemize}
    \item We propose {\ModelName}, a meta-gating framework on meta-learning, which mirrors nueroscience mechanism.
    \item We implement a novel hypernetwork, which converts arbitrary conditions into a self-adaptive meta-signal, steering the nonlinearity of $\beta-$SwiGLU blocks.
    \item Thorough theoretical studies and empirical experiments are conducted to verify our methodology's performance, stability, and scalability.
\end{itemize}

\begin{table*}[t]
    \caption{Methodology summaries of the baselines and {\ModelName}.
    $^{\star}$: dynamic low-rank reparameterization, which is similar to LoRA. 
    }
    \label{tab:methods}
    \centering 
    \small
    \resizebox{\textwidth}{!}{
    \begin{tabular}{l 
        cccccc}
        \toprule
            \multirow{2}{*}{Method} & Structure &
            \multicolumn{2}{c}{Meta tasks} &
            \multicolumn{2}{c}{Target tasks}  &  Original \\
            \cmidrule(lr){3-4} \cmidrule(lr){5-6}
            & customization & adaptation & update & adaptation & adjustment & param kept \\
        \midrule
            LoRA \cite{hu2022lora}        & \xmark  &  \xmark & LoRA & zero-shot & \xmark & \cmark \\ 
            SFT          & \xmark  &  \xmark & FT & zero-shot & \xmark & \xmark \\ 
            CMAML \citep{song-etal-2020-learning}       & pruning  &  gradient & meta-update & zero-shot & \xmark & \xmark \\
        \midrule
            \multicolumn{4}{l}{\textbf{\em Meta-in-context}} \\
            MetaICL \citep{min-etal-2022-metaicl}            & \xmark  &  in-context & FT & few-shot & \xmark & \xmark \\
            meta-icl \citep{NEURIPS2023_cda04d7e}            & \xmark  &  in-context & \xmark & few-shot & \xmark & \cmark \\ 
            MAML-en-LLM\citep{10.1145/3637528.3671905}            & \xmark  & in-context  & meta-update & few-shot & \xmark & \xmark \\
        \midrule
            \multicolumn{4}{l}{\textbf{\em Meta-on-LoRA}} \\
            MLtD \citep{hou-etal-2022-meta}       & struc. control  &  PEFT$^{\star}$ & meta-update & zero-shot & PEFT$^{\star}$ & \xmark \\
            HydraLoRA  \cite{tian2024hydralora}     & \xmark  &  asym. LoRA & MoE routing & merged experts & LoRA  & \cmark \\
            Meta-LoRA \citep{li-etal-2025-meta}       & \xmark  &  grad. similarity & reweighting & zero-shot & LoRA  & \cmark \\
            Text-to-LoRA \citep{charakorn2025texttolora}            & hypernetwork  &  task-wise text & FT / Recon & task-wise text & LoRA  & \cmark \\
            SHINE \citep{liu2026shinescalableincontexthypernetwork} & hypernetwork  &  in-context & PT / FT & in-context & LoRA  & \cmark \\
        \midrule
            \multicolumn{4}{l}{\textbf{\em Meta-on-Gating (Ours)}} \\
            {\ModelName}           &  hypernetwork &  sample-wise text & FT & sample-wise text & activation & \cmark \\
        \bottomrule
    \end{tabular}
    }
\end{table*}

\section{Related Work}
\label{related_work}

While transformer has proven to be a robust general-purpose architecture and also a few-shot learner, it still struggles in complicated, unseen tasks. Meta-learning, which deals with such types of problems, is therefore investigated on LLMs. Below, we discuss several categories of paradigms in comparison to our method (Figure \ref{fig:paradigm_comparison}). 


\paragraph{Adapter methods.} To avoid the overfit issue of finetuning, studies are conducted by implementing an extra adapter alongside the main LLM backbone, with the main LLM parameters unchanged. Depending on the locations, they can be classified into series adapter \cite{pmlr-v97-houlsby19a,qiu2025gated} and parallel adapter \cite{he2022towards}(Figure \ref{fig:paradigm_comparison} (D) and (E)). LLM-Adapters \cite{hu-etal-2023-llm} systematically experiment with different types of adapters and conclude that the parallel adapter can adapt well to versatile and changing tasks.

Parameter-efficient fine-tuning (PEFT) methods can also be considered as a special type of architecture-agnostic adapters, such as LoRA \cite{hu2022lora} (Figure \ref{fig:paradigm_comparison} (B)). However, LoRA requires a specific downstream training set, which may not be viable in the meta-learning case. Studies also show that the performance of LoRA may suffer from heterogeneity in corpus \cite{tian2024hydralora}.

\paragraph{Meta-in-context methods.} Direct application of traditional meta-learning, such as Model-Agnostic Meta-Learning (MAML) \cite{pmlr-v70-finn17a} on NLP tasks is limited, due to its complicated bi-level optimization (Figure \ref{fig:paradigm_comparison} (A)), despite some early attempts such as CMAML \citep{song-etal-2020-learning}. Later on, most meta-learning studies on LLMs incorporate the in-context information, including MetaICL \citep{min-etal-2022-metaicl}, MetaICT \citep{chen-etal-2022-meta}, meta-in-context learning \citep{NEURIPS2023_cda04d7e}, MAML-en-LLM \citep{10.1145/3637528.3671905}, and MICRE \citep{10.24963/ijcai.2024/702}. In this work, we show that meta-in-context can fail to adapt to some specific conditions, without architecture customization.

\paragraph{Meta-on-LoRA methods.} Another category leverages the adaptation of low-rank parameters of LoRA. Such methods include MLtD \citep{hou-etal-2022-meta}, which applies MAML on LLM with either task-adaptive reparameterization (TARP) or task-adaptive model structures (TAMS), hydraLoRA \citep{tian2024hydralora}, which uses asymmetric LoRA to route MoE, and also Meta-LoRA \citep{li-etal-2025-meta}, which conducts sample reweighting by gradient similarity. 


\paragraph{Meta with customized structures.} There is also a special work called Text-to-LoRA \citep{charakorn2025texttolora} (Figure \ref{fig:paradigm_comparison} (C)), which implements a hypernetwork that receives textual task descriptions and outputs the LoRA parameters. Its hypernetwork can be trained by the FT loss or a reconstruction loss with pretrained LoRA. In this paper, we propose another meta-control paradigm for LLMs, which leverages a novel gating mechanism, the meta-gating \cite{9637802,9940416,10172335} (Figure \ref{fig:paradigm_comparison} (F)). Replacing the original self-gating in the FFN, we let our gating to be adaptive with a hypernetwork on auxiliary conditions, similar to Text-to-LoRA. We employ the FT loss since \citep{charakorn2025texttolora} observes that the reconstruction loss fails to generalize on unseen tasks. We also have training-inference-consistent adaptation on text instructions. A recent study \citep{liu2026shinescalableincontexthypernetwork} has a similar model architecture, while adapting it to generalized contexts.

Table \ref{tab:methods} summarizes all the aforementioned methods and highlights the difference of our methodology.

\begin{table*}[t]
    \caption{
         Summaries of the meta-conditions $z$, including task, domain, persona, style, sentiment, etc. We provide a generalized framework that self-adapts to all the conditions above. Types of tasks are marked in \textit{Italic}. More prompts are in Appendix \ref{appendix:preprocess}.
    }
    \label{tab:overview}
    \centering \footnotesize
    \begin{tabular}{llll }
        \toprule
            Type & Attribute & Prompt (example) & Related Datasets \\
        \midrule
            task & \textit{summarization}/\textit{QA}/\textit{classification}/$\dots$ & (depending on detailed task; see Appendix \ref{appendix:dataset}) & CrossFit\&UnifiedQA , SNI \\
        \cmidrule{1-4}
        \cmidrule{1-4}
            domain & debate/science/email/$\dots$ & \makecell[l]{Please provide the summarization\\ on the domain of \{domain\}.} & \makecell[l]{AdaptSum\\ (\textit{summarization})} \\
        \cmidrule{1-4}
            \multirow{2}{*}{persona} & ["I like to remodel homes." & Please provide the response & Persona-Chat   \\
            & " My favorite holiday is Halloween."] &  with the knowledge of \{persona\}. & (\textit{dialogue}) \\
        \cmidrule{1-4}
            \multirow{2}{*}{style} & formal/informal & Please provide the response & GYAFC (\textit{dialogue}) \\
            & moral/immoral &  with the style of \{style\}. & MIC (\textit{QA}) \\
        \cmidrule{1-4}
            \multirow{2}{*}{sentiment} & \multirow{2}{*}{positive/neutral/negative} & Please provide the response & SST, Amazon, IMDB\\
            & &  with the sentiment of \{sentiment\}. & (\textit{review}) \\
        \cmidrule{1-4}
            \multirow{2}{*}{emotion} & \multirow{2}{*}{joy/sad/anger/$\dots$} & Please provide the response & EmoryNLP, DailyDialog\\
            & &  with the emotion of \{emotion\}. & (\textit{dialogue}) \\
        \bottomrule 
    \end{tabular}
\end{table*}

\section{Preliminary}


\subsection{Activation and Gating Mechanisms of LLMs}

\paragraph{Activation.} Starting from a generalized $\beta$-Sigmoid,
\begin{equation}
    \sigma_{\beta}(x) := 1/(1 + \exp{(-\beta x)}) \label{eq:general_sigmoid} 
\end{equation}
the famous Sigmoid $\sigma(x)$ is its special case ($\beta=1$).
Based on Sigmoid, the $\text{Swish}_{\beta}$ ($\beta > 0$) and its special form SiLU ($\beta=1$) has the following formulation:
\begin{align}
    \text{Swish}_{\beta}(x) = x * \sigma_{\beta}(x) \label{eq:beta-swish} \\
    \text{SiLU}(x) := \text{Swish}_{1}(x) = x * \sigma(x) \label{eq:silu}
\end{align}
\textbf{As $\beta$ increases, the nonlinearity of $\text{Swish}_{\beta}$ is amplified.} At two extreme situations, $\text{Swish}_{\beta}$ becomes Linear when $\beta \rightarrow 0$, and ReLU when $\beta \rightarrow +\infty$. Modern LLMs (LlaMA, Mistral, Qwen, etc) usually adopt $\text{SiLU}$ as a standardized implementation to reduce pretraining complexity. 

\paragraph{Self-Gating.} Within the feedforward network (FFN) blocks of LLM, GLU (Gated Linear Units) structure is adopted, which uses a gating network to reweight the MLP output. By using $\text{Swish}_{1}$ (SiLU) as the activation function of gating, the FFN block has the \textbf{SwiGLU} architecture:
\begin{align}
    y = W_{\text{down}} \left( \text{Swish}_1\left(  W_{\text{gate}} x \right) \otimes (W_{\text{up}} x) \right) \label{eq:ffn} \\
    W_{\text{up}}, W_{\text{gate}} \in \mathcal{R}^{D \times C}, W_{\text{down}} \in \mathcal{R}^{C \times D}, C > D \notag
\end{align}
where $\otimes$ denotes the element-wise multiplication, $D$ is the hidden size, and $C$ is the FFN intermediate size. In practice, $C$ is usually a multiple of $D$. SwiGLU can also be considered as \textbf{self-gating} which means the same latent is both inputs of the MLP and the gating network. 

\subsection{Computational Architecture of LLMs}
LLM is a decoder-only transformer with interleaved self-attention and FFN layers:
\begin{align}
    e_0 &= \text{emb}(x) \in \mathcal{R}^D, logit = \text{proj}(e_L) \notag \\
    e_{l} &= \text{FFN}(\text{self-attn}(e_{l-1})), l = 1, \cdots, L \label{eq:LLM_forward}
\end{align}
where $x$ is the text input, $L$ is the total number of layers, $e_l$ is the latent on the $l$-th layer, and $logit$ is the final output logit. $\text{emb}(x)$ represents the tokenization and embedding layer, while $\text{proj}$ is the final projection layer.

\subsection{Hypernetworks}
Given a base model $y = G_w(x)$, a hypernetwork \citep{ha2017hypernetworks} can be formed with $w$ generated by another auto-adaptive network: $w = \mathcal{H}_{\theta}(z)$ where $z$ is the adaptive input, and $\theta$ is the hypernetwork trainable parameter.


\begin{figure*}[hbtp!]
  \includegraphics[width=1\linewidth]{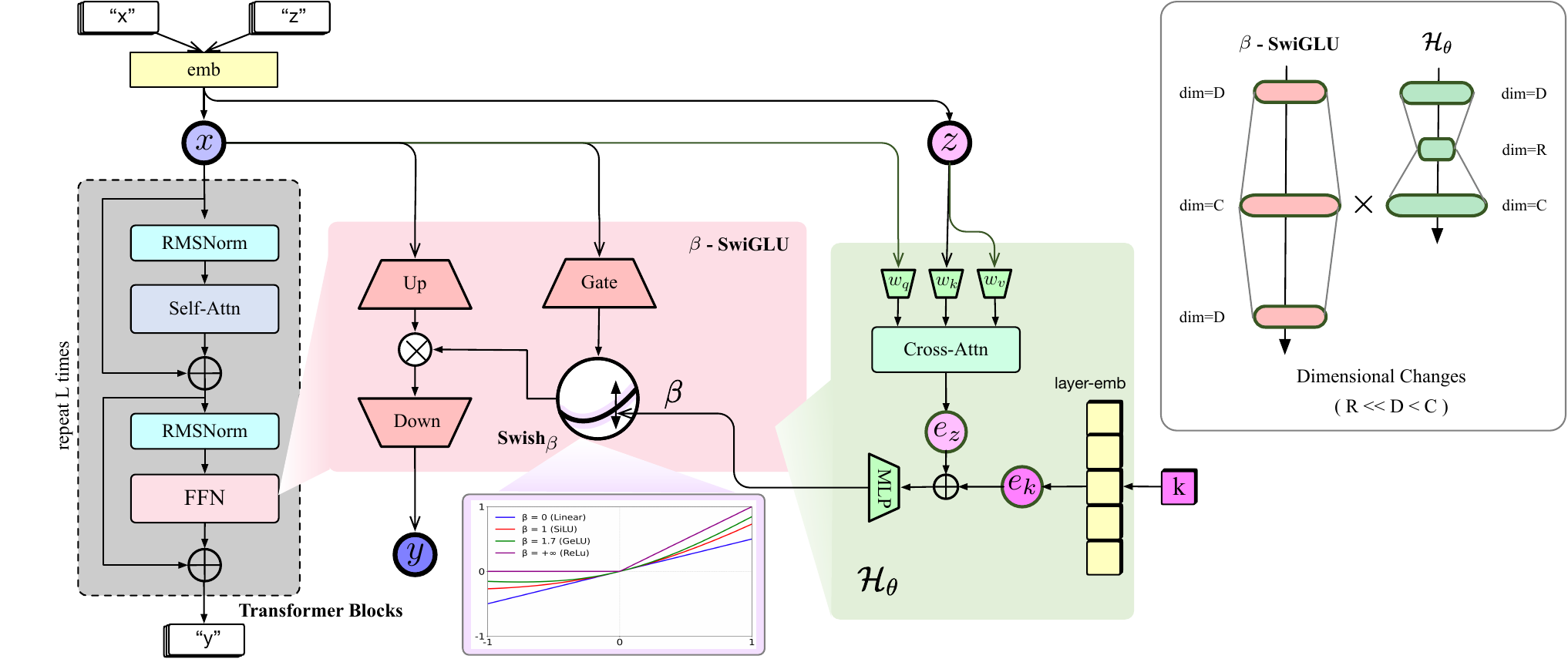}
  \caption {Framework of {\ModelName}. $(x, y, z)$ indicates the input, output, and condition, respectively. $k \in [1, K]$ denotes the layer index.}
  \label{fig:framework}
\end{figure*}

\section{Problem Formulation}


\subsection{Data Format}
\label{sec:data_format}

The problem of meta-training can be formulated on a collection of tasks $\mathcal{T}_{1:C}$, where $C$ is the total number of meta-tasks. For each task, the data has a format $(x, y)$, where $x$ denotes the input, $y$ denotes the output. During the training, the model learns to condition on the samples of a subset of tasks
\begin{equation}
    \mathcal{T}_i=\{(x^i_j, y^i_j)\}_{j=1}^{N_i},~\forall i \in [1, C] 
\end{equation}
where $N_i$ is the sample number of the $i$-th task. During the test stage, the model needs to adapt to an arbitrary task $\mathcal{T}^{\star}$
\begin{equation}
    \arg \max{P(y^{\star}|x^{\star})}, (x^{\star}, y^{\star}) \in \mathcal{T}^{\star}
\end{equation}

with two typical settings of target tasks in meta-learning: 

\noindent i) Low resource (LR): $\mathcal{T}^{\star}$ belongs to the $C$ meta-training tasks, but with relatively limited training samples.\\
\noindent ii) Unseen (US): $\mathcal{T}^{\star} \notin \mathcal{T}_{1:C}$, outside the scope of LLM.

In Section \ref{sec:experiment}, we investigate both settings to provide a comprehensive understanding of model meta-performances.

\subsection{Meta-on-Conditions}
\label{sec:meta_on_condition}


Compared to baselines in Table \ref{tab:overview}, in this paper, we propose that meta-learning can also be conducted on \textbf{generalized textual conditions}. Such conditions can be readily obtained, especially in the meta-learning setting, where the model adaptation is frequently required on the switch of different types of meta-conditions, such as task, domain, persona, style, sentiment, and emotion. These conditions, however, are insufficiently studied by previous meta-learning methods except Text-to-LoRA\footnote{However, their experiment solely explores the adaptation on task-wise description.}. 

To learn-to-learn on such conditions, we reformulate such datasets with the format of $(x, y, z)$, where $z$ is the auxiliary annotation of those conditions. During the meta-training, $z$ is first inserted into different instructional prompts (depending on the condition type), then is used to adapt the model. During inference, the model produces the best output conditioned on the test $z$: $\arg \max{P(y^{\star}|x^{\star}, z^{\star})}$. In this formulation, $z$ can semantically bridge the meta information between training and target tasks. Table \ref{tab:overview} summarizes different condition types, possible attributes, example prompts, and typical relative datasets.


\section{Method}
\label{sec:method}
For problem proposed in Section \ref{sec:meta_on_condition}, we introduce a novel meta-gating architecture, which is initialized from the original self-gating, driven by a hypernetwork with condition inputs. Post-hoc adaptation is then conducted on meta-tasks. 

\subsection{Meta-Gating}
\label{sec:meta-gating}


\paragraph{Beta-SwiGLU.} To provide an extra DoF (degree of freedom) for meta-control, we reactivate the $\beta$ of SiLU ($\text{Swish}_{\beta=1}$ function (Eq (\ref{eq:silu})); accordingly, the $\text{SwiGLU}$ (Eq (\ref{eq:ffn})) becomes $\beta\text{-SwiGLU}$, with $\text{Swish}_1$ replaced by $\text{Swish}_{\beta}$: 
\begin{align}
    &y = W_{\text{down}} \left( \text{Swish}_{\beta}\left(  W_{\text{gate}} x \right) \otimes (W_{\text{up}} x) \right), \beta>0 \label{eq:beta-SwiGLU}
\end{align}
where $\beta$ dynamically adjusts the slope of the Swish function.

Apparently, when $\beta=1$, $\beta\text{-SwiGLU}$ decays to the original $\text{SwiGLU}$. Also, the gradient of $\text{Swish}_{\beta}$ is bounded by $1/4 |x|^2$ with detailed derivation in Appendix \ref{appendix:model_component}, facilitating the stable and robust training of $\beta\text{-SwiGLU}$.

\paragraph{The hypernetwork.} We implement a lightweight hypernetwork which projects the textual condition $z$ into $\beta$ on each layer, denoting $\mathcal{H}_{\theta}(z, l)$. We first replace the placeholder of prompt $p$ (examples in Table \ref{tab:overview}) with $z$, then process it with the tokenizer and embedding layer of the LLM itself. $\mathcal{H}_{\theta}(z)$ then has a cross attention with a reduced dimension $R$, adding with layer embedding, and is finally fed into a one-layer, bias-free, MLP. In more details, on the $l$-th layer, the corresponding $\beta^l$ is generated by
\begin{align}
    z =& \text{emb}(p(z))\\
    e^z =& \text{cross-attn}(\text{Q=}W^q z, \text{K=}W^k e^l, \text{V=}W^v e^l) \notag \\
    \beta^l =& \text{MLP}(e^z + \text{layer-emb}(l)) \notag \\
    &W^q, W^k, W^v \in \mathcal{R}^{D \times R}, l = 1,2, \cdots, L \notag 
\end{align}
where MLP is one-layer, bias-free, has an input-output dimension of $(R, C)$, and contains a final activation of $tanh$.

\subsection{Reformulated LLM Computation}

\paragraph{Re-formularize the LLM.} We replace the SwiGLU of LLM with our $\beta\text{-SwiGLU}(x)$, with $\beta$ dynamically determined by the hypernetwork. The model can then be expressed as
\begin{align}
    y = \mathcal{G}_{\{w, \beta^l=\mathcal{H}_{\theta}(z, l)\}}(x) \label{eq:model}
\end{align}
where $\theta$ is the union of trainable parameters of cross-attn and MLP of $\mathcal{H}$. By such a formulation, we disentangle the meta-condition adaptation from the foundation capability, by learning of $\beta$ and $w$, respectively. Figure \ref{fig:framework} visualizes the entire framework.

\paragraph{Number of parameters.} {\ModelName} has a relatively small set of trainable parameters, which ensures its computational efficiency. By utilizing the embedding layer of the LLM itself, we avoid using an extra embedding model (which is done in Text-to-LoRA), reducing the parameters and helping the information be more aligned. \# param of {\ModelName} has a \textbf{linear} dependency on the reduced hidden size $R$, with a smaller parameter size than Text-to-LoRA, according to their reported \# param. Appendix \ref{appendix:cal_param} shows the derivation.

\paragraph{Information compression.} By the design of $\mathcal{H}_{\theta}$, we disentangle the shared meta control and the layerwise difference from a modularity perspective. $\mathcal{H}_{\theta}$ can also be considered as a \textbf{information bottleneck} hypernetwork, with the latent size first compressed by the attention from $D$ to ($R << D$), then projected back to $C$ by MLP. This design not only reduces the number of trainable parameters, but also helps compress the information and extract the key patterns \citep{10.1007/978-3-642-02565-5_4,illya_KC,deletang2024language}. Figure~\ref{fig:framework} highlights this dimension configuration. Appendix \ref{appendix:info_theory} also provides an informative theory analysis.


\begin{table}[t]
    \caption{Statistics of meta-training datasets used in this paper. `Cond.' denotes condition. `LR' and `US' indicate low resource and unseen. `per.' denotes persona; `senti.' denotes sentiment.
    }
    \label{tab:data_summary}
    \centering 
    \small
    \resizebox{1\columnwidth}{!}{
    \begin{tabular}{crrcrr}
        \toprule
            & \multicolumn{2}{c}{Meta-train} & \multicolumn{3}{c}{Target} \\
            \cmidrule(lr){2-3} \cmidrule(lr){4-6}
            Dataset & Cond. & \# num & setting & Cond. & \# num \\
        \toprule
            Persona-Chat & 1,137 per. & 8.9K & US & 100 per. & 0.97K \\
            AdaptSum  & 6 domain & 1.5K & LR & 6 domain & 8.0K \\
        \midrule
            GYAFC  & 2 style & 16.0K & LR & 2 style & 1.6K \\
            MIC  & 2 style  &  254K & LR & 2 style & 10.9K \\
            SST  & 3 senti.  & 9.1K  & LR & 5 senti. & 1.0K \\
        \midrule
            CrossFit\&  & \multirow{2}[1]{*}{325 task} & \multirow{2}[1]{*}{3.5M} & LR & 80 task & 187K \\
            UnifiedQA &  &    & US & 11 task & 4.9K \\
        \midrule
            SNI & 479 task & 1.4M & US & 10 task & 23.3K \\
        \bottomrule
    \end{tabular}
    }
\end{table}

\subsection{Training Mechanism}

We train on samples $(x, y, z)$, with the original LLM parameters $w$ frozen. Algorithm \ref{alg:algorithm} in Appendix \ref{appendix:algorithm} summarizes the training mechanism. Based on the following losses, the gradient of $\theta$ is automatically calculated and back-propagated.

\paragraph{The cross-entropy loss.} This cross-entropy (CE) loss term is the standard SFT loss on output tokens $y$, with masks on input tokens $x$: $\mathcal{L}^{ce}_{\theta} = \sum y \text{P}\left(\mathcal{G}_{\{w, \beta_{\theta}(z) \}} | x \right)$.



\paragraph{Regularization.} To ensure $\beta$ does not deviate from $1$ too much, we add an auxiliary loss term to have an L-2 regularization\footnote{To utilize the regularization directly, we implement $\text{Swish}_{1+\beta}$ in practice, where $\beta=0$ corresponds to the raw LLM. See Appendix \ref{appendix:model_details} for more details.}: $\mathcal{L}^{reg}_{\theta} = \vert \beta_{\theta}(z) \vert_2$.

\paragraph{The total loss.} The total loss is then a simple linear weighted sum of them
\begin{align}
    \min_{\theta} \mathcal{L} =  \mathcal{L}^{ce}_{\theta} + f \mathcal{L}^{reg}_{\theta} \label{eq:loss}
\end{align}
where $f$ is the regularization weight. Finally, we conduct full finetune (FT) on loss $\mathcal{L}$.

\section{Experiments}
\label{sec:experiment}

\subsection{Implementation}

The model backbone is Llama-3.1-8B-Instruct \citep{grattafiori2024llama3herdmodels}, with $K = 32$ layers, hidden size $D = 4096$, and intermediate size $C = 14336$. Depending on detailed datasets, we implement $\mathcal{H}$ with the reduced hidden size $R$ of either $128$ or $512$, resulting in 3.4M or 13.6M parameters, respectively. We set the regularization weight $f = 0.001$. Other configurations can be found in Appendix \ref{appendix:hyperparameter}.



\begin{table*}[htbp!]
\caption{Results conditioned on persona (Persona-Chat), domain (AdaptSum), style (GYAFC and MIC), and sentiment (SST). The best result is marked \textbf{bolded} and the second-best is \underline{underlined}.
}
\label{tab:persona_domain_style_sentiment_results}
\centering
\small
\begin{tabular}{l | cc | cc | ccc | ccc | ccc }
    \toprule
    \multicolumn{1}{c|}{\multirow{2}[2]{*}{Method}} &  \multicolumn{2}{c|}{Persona-Chat} &  \multicolumn{2}{c|}{AdaptSum}  &  \multicolumn{3}{c|}{GYAFC} &  \multicolumn{3}{c|}{MIC} &  \multicolumn{3}{c}{SST} \\
    \cmidrule{2-3}   \cmidrule{4-5} \cmidrule{6-8}   \cmidrule{9-11}   \cmidrule{12-14}      
    & R-L & B-2 & R-L & B-2 & R-L & B-2 & D-2 & R-L & B-2 & D-2  & R-L & B-2 & D-2 \\
    \toprule
    Direct & 11.79 & 3.64 & 13.35 & 4.72 & 7.57 & 2.01 & 0.18 & 7.36 & 1.92 & 0.04 & 11.79 & 3.64 & 0.06 \\
    ICL & 13.75 & 5.23 & \underline{23.35} & \underline{11.29} & 6.77 & 1.77 & 0.10 & 7.42 & 1.87 & 0.04 & 12.47 & 4.48 & 0.15 \\
    LoRA & 21.35 & 9.63 & 13.37 & 4.75 & \underline{28.11} & \underline{12.04} & 0.55 & 16.19 & 4.03 & 0.06 & 13.31 & 4.50 & 0.28 \\ 
    SFT & \underline{22.97} & 9.94 & 20.99 & 8.80 & 23.72 & 8.62 & 0.57 & 15.20 & 4.55 & 0.02 & 12.77 & 4.31 & 0.16 \\ 
    meta-icl & 15.74 & 6.73 & 13.03 & 4.37 & 13.55 & 5.05 & 0.26 & 15.71 & 5.42 & 0.07 & 13.41 & 4.61 & 0.21 \\ 
    Text-to-LoRA & 21.75 & \bf 13.70 & 20.02 & 7.98 & 22.95 & 8.87 & \underline{0.72} & \underline{22.50} & \underline{9.88} & \bf 0.42 & 10.72 & 3.84 & \underline{0.55} \\ 
    \textbf{\ModelName} (ours) & \bf 23.15 & \underline{10.15} & \bf 41.85 & \bf 26.88 & \textbf{29.37} & \textbf{13.82} & \bf 0.81 & \bf 23.67 & \bf 10.57 & \underline{0.17} & \textbf{22.76} & \textbf{9.33} & \bf 0.67 \\
    \bottomrule
\end{tabular}
\end{table*}

\begin{table*}[t]
    \caption{Results on CrossFit\&UnifiedQA.
    Two numbers indicate the average performances on the entire test set and the unseen task test set. HR denotes high resource, Class denotes Classification, NLI denotes natural language inference, and Para denotes paraphrase.
    $^{\star}$: results from original published papers.
    }
    \label{tab:metaICL_result_all}
    \centering \footnotesize
    \begin{tabular}{
        l @{\hspace{2em}}
        ccccccc
        }
        \toprule
            \multirow{2}{*}{Method}
            & HR & Class & non-Class & QA & non-QA & non-NLI & non-Para \\
            & $\rightarrow$LR & $\rightarrow$Class & $\rightarrow$Class & $\rightarrow$QA & $\rightarrow$QA & $\rightarrow$NLI & $\rightarrow$Para \\
        \midrule
            MetaICL$^{\star}$ & 43.3/35.3 & 43.4/32.3 & 38.1/28.1 & 46.0/69.9 & 38.5/48.3 & 49.0/80.1 & 33.1/34.0 \\
            MAML-en-LLM$^{\star}$ & 48.0/50.9 & 51.1/49.0 & 50.5/46.8 & 42.5/55.6 & 40.0/47.1 & 52.4/65.0 & 53.3/\textbf{58.0} \\
        \midrule
            Direct & 56.9/47.7 & 61.0/47.7 & 61.0/47.7 & 52.2/\textbf{79.5} & 52.2/79.5 & 63.6/51.7 & 60.4/22.3 \\ 
            LoRA & 56.1/55.0 & 63.9/\textbf{59.9} & 59.7/58.8 & 23.6/\textbf{79.5} & 50.0/80.5 & 58.0/76.8 & 61.7/34.1 \\ 
            SFT & 54.2/51.9 & 62.1/59.3 & 59.4/59.1 & 30.8/72.0 & 40.0/\textbf{84.5} & 65.3/66.4 & 61.3/29.0 \\
            Text-to-LoRA & 64.7/48.4 & 61.6/45.2 & 57.8/48.9 & 45.9/27.1 & 54.1/27.8 & 55.2/75.2 & 75.5/20.5 \\ 
            \textbf{\ModelName} (ours) & \bf 66.7/55.7 & \textbf{64.2}/57.8 & \textbf{61.4}/\textbf{64.2} & \textbf{72.7}/\textbf{79.5} & \textbf{72.3}/79.5 & \textbf{73.8}/\textbf{84.4} & \textbf{79.1}/51.6 \\
        \bottomrule
    \end{tabular}
\end{table*}                              

\subsection{Datasets}

{\ModelName} is trained by meta-learning on a mixture of datasets, conditioned on different textual information:

\noindent i) Persona: we use Persona-Chat \citep{zhang-etal-2018-personalizing}, which conditions the dialogue on unseen persona profiles.\\ 
\noindent ii) Domain: AdaptSum \citep{yu-etal-2021-adaptsum} is employed to conduct the summarization task on 6 domains: debate, dialogue, email, movie review, science, social media, with low-resource training samples.\\
\noindent iii) Style and Sentiment: including GYAFC \citep{Rao2018DearSO} with styles $formality$ versus $informal$, MIC \citep{ziems-etal-2022-moral} with 6 $moral$ styles versus 6 $immoral$ styles, and SST \citep{socher-etal-2013-recursive} with sentiments of $positive$, $neutural$, and $negative$.\\
\noindent iv) Task: we first adopt the experimental setting in \cite{min-etal-2022-metaicl,10.1145/3637528.3671905} which employ the mixture of \textsc{CrossFit} \citep{ye-etal-2021-crossfit} and \textsc{UnifiedQA}~\citep{khashabi-etal-2020-unifiedqa}, covering a variety of tasks, and is classified into six transfer settings: HR$\rightarrow$LR, Class$\rightarrow$Class, non-Class$\rightarrow$Class, QA$\rightarrow$QA, non-QA$\rightarrow$QA, non-NLI$\rightarrow$NLI, and non-Para$\rightarrow$Para; the second experiment setting refers to \citep{charakorn2025texttolora}, which uses the Super-NaturalInstructions (SNI) \citep{wang-etal-2022-super} as training and validation sets, while tests the model on ten generalized mainstream benchmarks, including Arc-challenge (ArcC), Arc-easy (ArcE), BoolQ (BQ), Hellaswag (HS), OpenbookQA (OQA), PIQA, Winogrande (WG), GSM8K,  HumanEval (HE), and MBPP.



Data splits and statistics are summarized in Table \ref{tab:data_summary}. Further dataset introductions are in Appendix \ref{appendix:dataset}. Preprocessing details are in Appendix \ref{appendix:preprocess}.

\subsection{Baselines}

{\ModelName} compared to several types of baselines, including \\
\noindent (1) Direct inference and prompting methods such as in-context learning (ICL), with 2-shot examples randomly sampled from the training set. Although simple, such training-free baselines do not adapt their parameters to specific target tasks; therefore, they naturally avoid overfitting. We investigate more prompting baselines in Appendix \ref{appendix:more_style_results}. \\ 
(2) Finetuning methods such as SFT and LoRA \citep{hu2022lora}. Since they are not designed specifically for meta-learning, their performance will suffer from insufficient training resources. LoRA is expected to have less alignment tax than SFT by disentangling the original parameters and newly-adapted low-rank matrices. \\
(3) Meta-in-context methods such as meta-in-context learning\footnote{We name it by meta-icl for short in the following contexts.} \citep{NEURIPS2023_cda04d7e}, MetaICL \citep{min-etal-2022-metaicl}, and MAML-en-LLM \citep{10.1145/3637528.3671905}, all of which leverage few-shot examples to conduct meta-learning on LLM. \\
(4) Text-to-LoRA \citep{charakorn2025texttolora}, which uses a hypernetwork that converts task descriptions to low-rank parameters ($A, B$) in LoRA, to adapt to target tasks.


Further introductions and implementation details of baselines can be found in Appendix \ref{appendix:baseline}.

\subsection{Metrics}


For generation tasks, we consider overlapping metrics such as Rouge-L (\textbf{R-L}) \citep{lin2004rouge} and BLEU-2 (\textbf{B-2}) \citep{papineni2002bleu}, as well as Dist-2 (\textbf{D-2}) \citep{li2015diversity}, indicating the response diversity. For multi-choice question, classification, or math tasks, we calculate the answer accuracy (\textbf{acc}) compared to the truth. For coding tasks, we report the \textbf{pass@1} results. Detailed definitions are in Appendix \ref{appendix:auto_metrics}. More evaluation details are in Appendix \ref{appendix:eval_detail}.






\begin{table*}[th]
\caption{Zero-shot performance of methods with SNI as the training set.
$^{\star}$: results directly obtained from \citet{charakorn2025texttolora}.
}
\label{tab:SNI_results_llama}
\small
\begin{tabular}{llcccccccccc|c}
\toprule
\multicolumn{2}{l}{} &  \textbf{\begin{tabular}[c]{@{}c@{}}ArcC\\ (acc)\end{tabular}} &
  \textbf{\begin{tabular}[c]{@{}c@{}}ArcE\\ (acc)\end{tabular}} &
  \textbf{\begin{tabular}[c]{@{}c@{}}BQ\\ (acc)\end{tabular}} &
  \textbf{\begin{tabular}[c]{@{}c@{}}HS\\ (acc)\end{tabular}} &
  \textbf{\begin{tabular}[c]{@{}c@{}}OQA\\ (acc)\end{tabular}} &
  \textbf{\begin{tabular}[c]{@{}c@{}}PIQA\\ (acc)\end{tabular}} &
  \textbf{\begin{tabular}[c]{@{}c@{}}WG\\ (acc)\end{tabular}} &
  \textbf{\begin{tabular}[c]{@{}c@{}}GSM8K\\ (acc)\end{tabular}} &
  \textbf{\begin{tabular}[c]{@{}c@{}}HE\\ (pass@1)\end{tabular}} &
  \textbf{\begin{tabular}[c]{@{}c@{}}MBPP\\ (pass@1)\end{tabular}} &
  \textbf{Avg.} \\ \midrule
\multicolumn{2}{l}{\texttt{Direct}$^{\star}$} & {73.3} &	{90.6} &	{80.4}  &	{66.6} &	{75.4} &	{79.8} &	{55.3}&	{75.7} &	{66.5} &	{68.7} &	{73.2}  \\
\multicolumn{2}{l}{ICL$^{\star}$} & {80.7} &	{91.9} &	{80.0} &	{59.3} &	{77.6} &	{80.9} &	{61.3} &	{75.7} &	{66.5} &	{70.4} &	{74.4}  \\
\multicolumn{2}{l}{Prepending task desc.$^{\star}$} & {80.2} &	{92.5} &	{79.9} &	{69.8} &	{78.4} &	{81.7} &	{62.4} &	{75.7} &	\textbf{68.3} &	{70.2} &	{75.9}   \\
\multicolumn{2}{l}{LoRA$^{\star}$} & {82.0} &	{92.8} &	{83.3} &	{70.8} &	{81.8} &	{83.8} &	{60.3} &	{77.6} &	{	63.4} &	{69.4} &	{76.5}  \\ 
\multicolumn{2}{l}{SFT} & 80.9 & 93.1 & 79.7 & 66.3 & 78.6 & 81.3 & 58.2 & 76.4 & 65.2 & 63.3 & 74.3 \\
\multicolumn{2}{l}{Text-to-LoRA$^{\star}$} & {82.4} &	{92.9} &	{\textbf{84.4}} &	{72.8} &	{81.8} &	{81.2} &	{60.0} &	{\textbf{79.1}} &	{64.6} &	{\textbf{69.9}} &	{76.9}    \\  
\midrule
\multicolumn{2}{l}{\textbf{\ModelName} (ours)} & \bf 85.3 & \bf 94.2 & 83.0 & \bf 77.5 & \bf 84.2 & \bf 85.8 & \bf 67.5 & 74.1 & 57.9 & 65.2 & \bf 77.5 \\
\bottomrule
\end{tabular}%
\end{table*}

\subsection{Results}

\paragraph{{\ModelName} adapts well to low-resource domain or persona.} Table \ref{tab:persona_domain_style_sentiment_results} shows the results on Persona-Chat and AdaptSum. Prompt-based methods (except ICL) can not adapt to target responses (R-L and B-2) but keep the response diversity (D-2). On the other hand, in-context learning methods (ICL and metaICL), finetuning methods (LoRA and SFT), and Text-to-LoRA have improved performance, suggesting meta capabilities for new personas or domains. Finally, our {\ModelName} achieves the best performance on both benchmarks, indicating that the meta-gating mechanism helps the adaptation to these conditions. Appendix \ref{appendix:more_style_results} and \ref{appendix:more_data_specific_results} compare to more prompting and previous dataset-specific baselines.



\paragraph{Tests on styles and sentiments.} We then test {\ModelName} on GYAFC, MIC and SST. Table \ref{tab:persona_domain_style_sentiment_results} also exhibits the corresponding results. Although prompting methods exhibit relatively high D-2 results, their R-L and B-2 results are poor. On the other hand, LoRA and SFT show some adaptation capability, where LoRA performs better than SFT, potentially due to its decoupling between original and newly-adapted parameters. Among meta-learning methods, Text-to-LoRA performs better than meta-icl, indicating that in-context information is possibly not suitable for stylistic adaptation. Finally, {\ModelName} still provides good results across conditions and metrics, revealing that the meta-gating can conduct the adaptation effectively. Appendix \ref{appendix:more_style_results} provides further evaluation results by human scoring and LLM-as-a-Judge. To verify that {\ModelName} has relatively balanced performance across different condition attributes, Appendix \ref{appendix:per_style_result} provides per-condition results on AdaptSum, GYAFC, and MIC. Appendix \ref{appendix:case} shows typical cases. 




\paragraph{{\ModelName} can generalize to low-resource or unseen tasks.} Table \ref{tab:metaICL_result_all} shows the performances on CrossFit\&UnifiedQA, which evaluates the model's adaptation capability for low-resources or unseen target tasks. Across seven meta-test settings, {\ModelName} again outperforms all baselines, including MetaICL, MAML-en-LLM, and Text-to-LoRA. These promising results validate that our method can also leverage the full descriptions of tasks, with balanced transfer capability across varied task types.


\paragraph{{\ModelName} adapts well on generalized benchmarks.} Table \ref{tab:SNI_results_llama} exhibits the results on 10 unseen general benchmarks, with models trained by the SNI dataset. {\ModelName} performs the best on ArcC, ArcE, HS, OQA, PIQA, WG, and the averaged score, indicating that our methodology can also work well on different general benchmarks.

Appendix \ref{appendix:SNI_mistral} shows our results based on Mistral-7B-instruct \citep{jiang2023mistral}, also trained by SNI and evaluated on those benchmarks, which provides an apple-to-apple comparison to the main result of \citet{charakorn2025texttolora}.


\begin{table}[htbp!]
\caption{Ablation on Persona-Chat and AdaptSum. $p$ denotes the prompt and $w$ denotes the original LLM parameters.}
\label{tab:ablation}
\renewcommand{\arraystretch}{1.11} 
\centering
\small
\resizebox{0.98\columnwidth}{!}{
\begin{tabular}{l | ccc | ccc }
    \toprule
    \multicolumn{1}{c|}{\multirow{2}[2]{*}{Method}} &  \multicolumn{3}{c|}{Persona-Chat} &  \multicolumn{3}{c}{AdaptSum} \\
    \cmidrule{2-4}   \cmidrule{5-7}       
    & R-L & B-2 & D-2 & R-L & B-2 & D-2  \\ %
    \toprule
    w/o $p$ & 20.38 & 8.98 & 0.21 & \colorbox{red}{0.18} & \colorbox{red}{0.04} & \colorbox{red}{0.01} \\
    w/o pos-emb & \colorbox{pink}{7.11} & \colorbox{pink}{2.22} & \colorbox{pink}{0.04} & \colorbox{pink}{16.82} & \colorbox{pink}{6.93} & \colorbox{pink}{0.14} \\
    w/o $\mathcal{L}_{reg}$ & 20.38 & \bf 10.15 & 0.11 & 36.62 & 21.63 & \underline{0.34} \\
    \midrule
    \textbf{\ModelName} & \bf 23.15 & \bf 10.15 & \bf 0.22 & \bf 41.85 & \bf 26.88 & \bf 0.35 \\
    \bottomrule
\end{tabular}
}
\end{table}

\paragraph{Ablations.} We conduct ablation studies by removing each component of {\ModelName}, including the meta-prompt $p$, the positional embedding, and the regularization loss $\mathcal{L}_{reg}$. Table \ref{tab:ablation} shows the ablation results in which the standard {\ModelName} still performs the best, indicating each component is necessary.

\begin{figure}[!t]
\centering
  \includegraphics[width=1\linewidth]{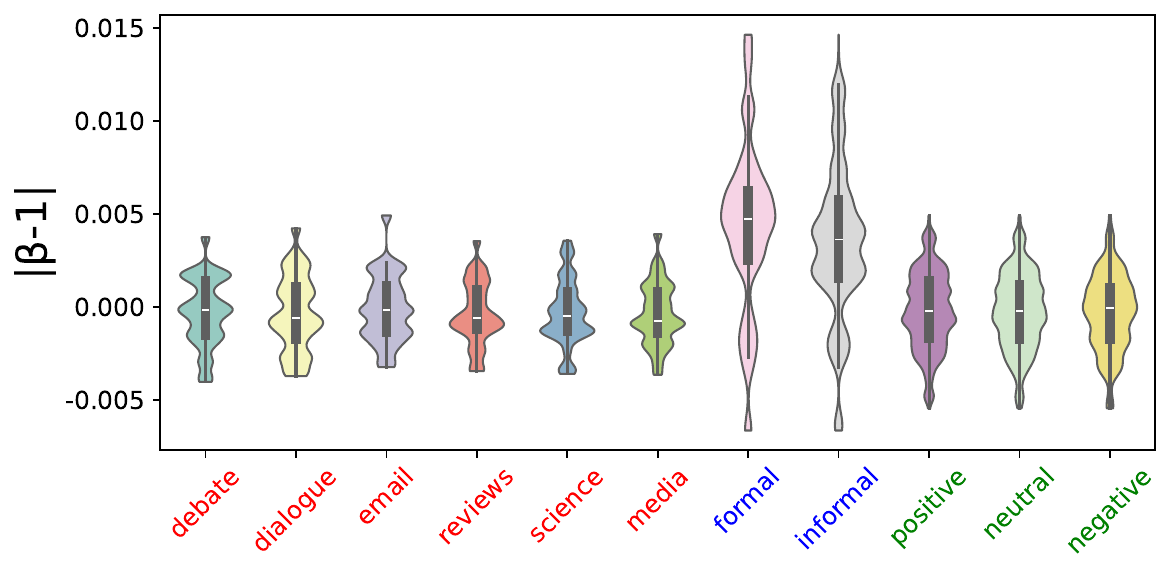} 
  \caption{Distribution of $\beta$ with respect to textual conditions.}
  \label{fig:beta_dist}
\end{figure}

\section{Discussion}

In this section, we further explore the intrinsic mechanism of meta-gating, including i) how the value of $\beta$ varies across different conditions, ii) if this meta-gating can adapt to different sizes of backbones, and iii) how the inclusion of meta-gating affects the inference latency.


\subsection{Distribution of $\beta$}
\label{sec:dist_beta}

\begin{figure}[!htbp]
\centering
  \includegraphics[width=0.8\linewidth]{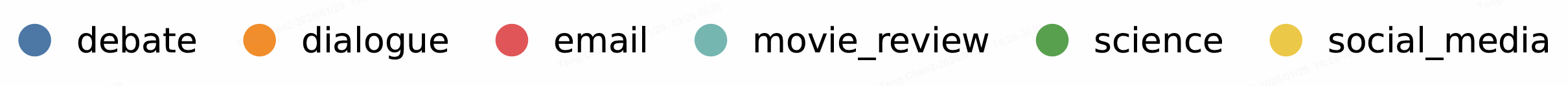}
  \\ 
  \includegraphics[width=0.8\linewidth]{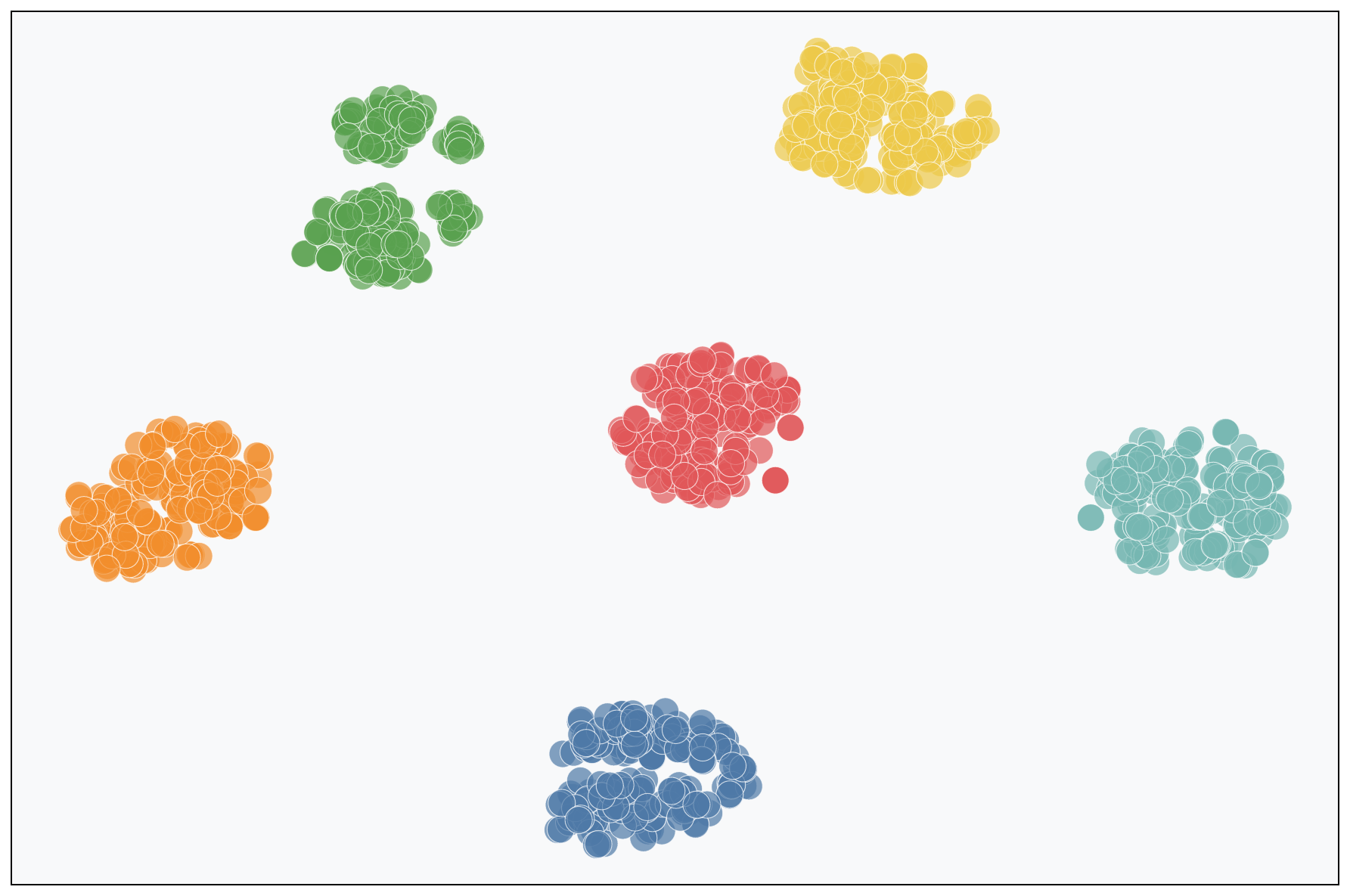}
  \caption{The t-SNE result of the $5$-th layer $\beta$ on AdaptSum.}
  \label{fig:tsne_beta_adaptsum}
\end{figure}

Figure \ref{fig:beta_dist} compares the distribution of $\beta$ across textual conditions on AdaptSum (red), GYAFC (blue), and SST (green), where $\beta=1$ means equalization to the original LLM. Grounding by different conditions, FFN is customized with different activation slopes under varied $\beta$ distributions. This nonlinearity customization is also aligned with semantics and common sense, \textit{e.g.}, the majority distribution of $\beta$ on `neural' lies between `positive' and 'negative'. Figure \ref{fig:tsne_beta_adaptsum} conducts t-SNE analysis of averaged $\beta$ on AdaptSum, where evident grouping can be observed w.r.t different domains\footnote{Appendix \ref{sec:tsne_all_layers} exhibits t-SNE plots of all layers on 3 datasets.}.



\subsection{Zero Out $\beta$ Attributes} 
\begin{figure}[!t]
\centering
  \includegraphics[width=0.49\textwidth]{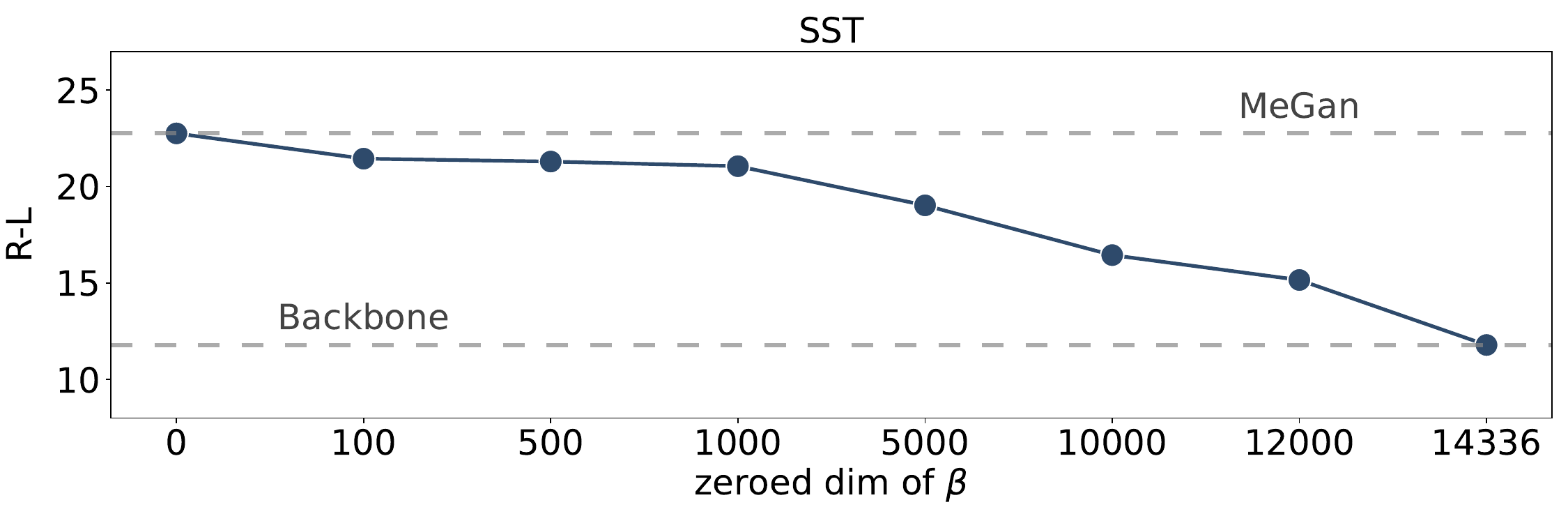}
  \hspace{0.02in}
  \includegraphics[width=0.49\textwidth]{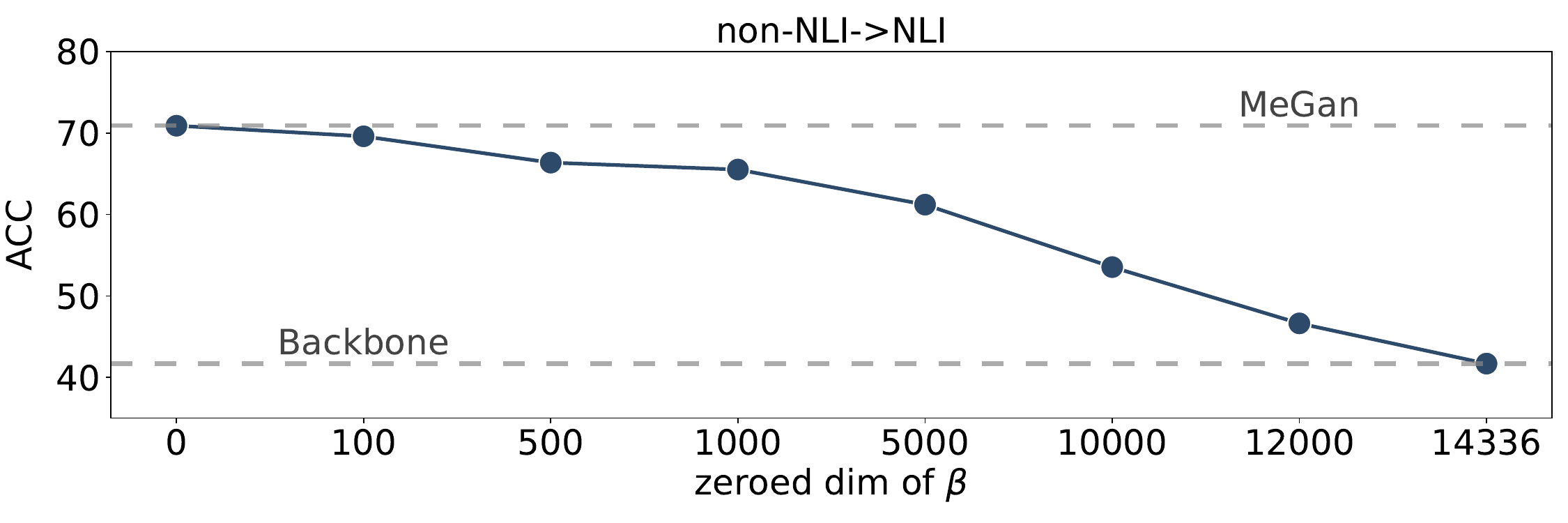}
  \caption {Performance degradation with increasing \textbf{zeroed-out} dimensions of $\beta$. 0 corresponds to the standard {\ModelName}, while 14336 (the total dim of $\beta$) corresponds to the backbone.} 
  \label{fig:Zeroed out}
\end{figure}

Since our $\beta$ is a multi-dimensional vector, to further validate the effectiveness of specific $\beta$ attributes, here we conduct a trial experiment: for trained $\beta$ on all layers (For an 8B backbone, the intermediate size is 14336, therefore the size of $\beta$ is 14336 as well), we try to zero out its top-$z$ dimensions, to observe the performance change. Obviously, the standard {\ModelName} corresponds to $z=0$; when $z=14336$, the model degenerates to the original backbone. Figure \ref{fig:Zeroed out} exhibits the trial results on SST (top) and non-NLI$\rightarrow$NLI (bottom). Apparently, for both cases, the performance continues to degrade as $z$ becomes larger. This observation verifies that the learned $\beta$ has an evident impact on the result.

\subsection{Scalability}

\begin{figure}[!t]
\centering
  \includegraphics[width=0.49\linewidth]{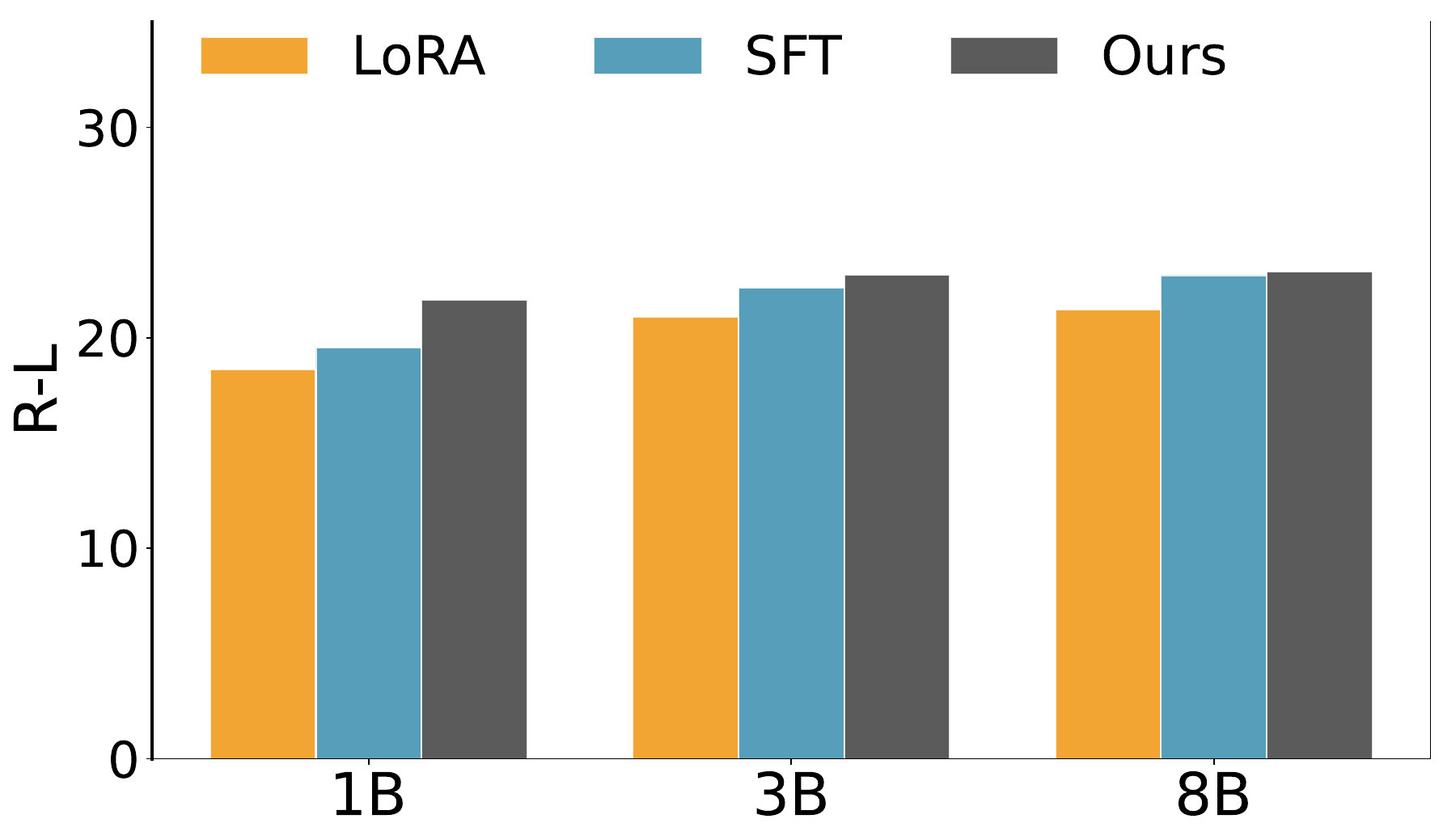} 
  \vspace{0.1cm}
  \includegraphics[width=0.49\linewidth]{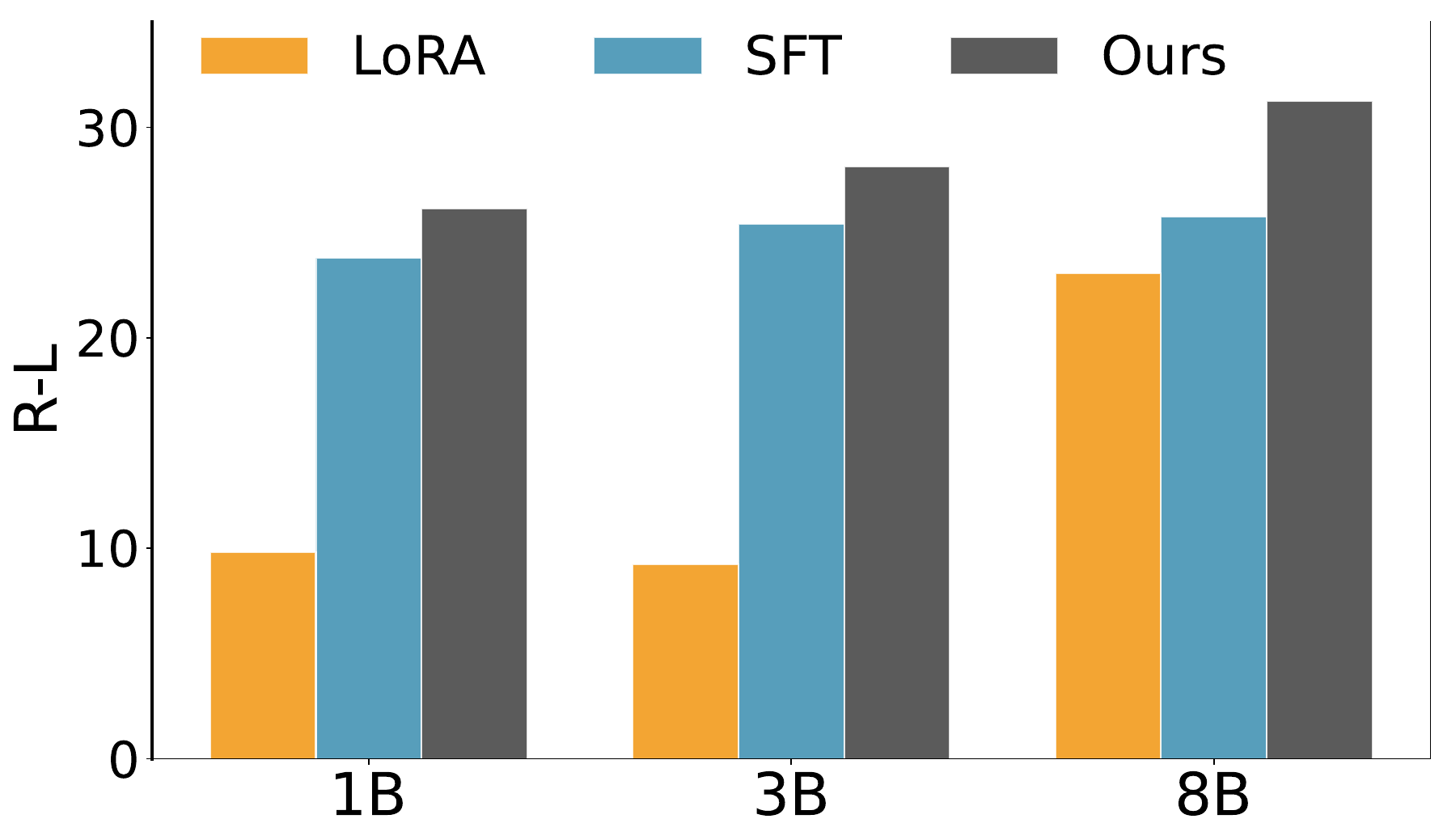}
  \caption {Result comparisons on different model sizes for Persona-Chat (left, low-resource) and GYAFC (right, zero-shot).}
  \label{fig:scalability}
\end{figure}

Figure \ref{fig:scalability} compares {\ModelName} to LoRA and SFT on Persona-Chat and GYAFC, with respect to different model sizes. As model size decreases, {\ModelName}'s performance degrades less than the other two methods, indicating that it is less affected by the foundation model's capabilities. Note that this phenomenon is more pronounced in the zero-shot case, highlighting the effectiveness of {\ModelName} for meta-learning.

\subsection{Latency Analysis}

{\ModelName} introduces minimal inference latency overhead:\\
\noindent \textbf{First‑token latency}: the hypernetwork forward pass is required only once per generation. This overhead is small because (1) the hypernetwork is lightweight; (2) a single hypernetwork is shared across all LLM layers, with only a simple layer‑position embedding added; and (3) its textual input reuses the LLM’s embedding layer.\\
\noindent \textbf{Latency for subsequent tokens}: the hypernetwork depends solely on the textual condition, not on the generated text. Hence, its output can be cached after the first call, incurring negligible overhead during the rest of the decoding process.

\subsection{Limitations} 

{\ModelName} enables LLM to adapt to arbitrary text conditions, including domain, persona, style, and task descriptions. However, there are also potential limitations: i) \textbf{Unseen modalities}, \textit{i.e.}, transferring from traditional NLP tasks to language‑grounded embodied intelligence; ii) \textbf{Acquiring new knowledge}, which should be better handled by pretraining or finetuning; iii) \textbf{Incorrect or vague conditions}: performance will suffer if the input condition is inaccurate.

\section{Conclusion}
\label{conclusion}

In this paper, we propose {\ModelName}, which conducts meta-control on LLMs with FFN gating adaptive to different conditions, such as task, domain, persona, style, and sentiment. We replace the SiLU activation in LLM by $\text{Swish}_{\beta}$, resulting in the new FFN block called $\beta$-SwiGLU. We further implement a hypernetwork which automatically converts textual conditions into $\beta$, resulting in a meta-gating mechanism, adjusting the nonlinearity of activations upon different situations. This hypernetwork is then generalized to target tasks with dynamic condition comprehension. {\ModelName} exhibits state-of-the-art performance on meta-learning.


\newpage

\section*{Impact Statement}

This paper presents work whose goal is to advance the field of meta-learning on LLMs. There are many potential societal consequences of our work, none of which we feel must be specifically highlighted here.

\bibliography{main}
\bibliographystyle{icml2026}

\newpage
\appendix
\onecolumn



\section{Extra Method Details} 
\label{appendix:model_details}

\subsection{Properties of Model Components}
\label{appendix:model_component}

The design of the Swish activation is inspired by the use of the famous sigmoid function for gating in LSTMs and highway networks. We start from the property of Sigmoid, then show the detailed formulation of SwiGLU, and finally provide the detailed properties and implementation details of our $\beta-$SwiGLU.

\paragraph{Gradient of Sigmoid.} Based on the definition of the Sigmoid function (Eq (\ref{eq:general_sigmoid})), its gradient has the following property:
\begin{equation}
    \sigma^{\prime}(x) := \sigma(x)(1 - \sigma(x)) \label{eq:sigmoid_grad}
\end{equation}

\paragraph{SwiGLU.} The gating mechanism is widely adopted in modern LLMs, in which the same input is employed as the input of Swish as well as a parallel gating network, which is usually called self-gating. A typical example of self-gating can be the famous Gated Linear Units (GLU) structure:
\begin{equation}
    \text{GLU}(x, W, V, b, c) = activation(Wx+b) \otimes (Vx+c)
\end{equation}
where $\otimes$ denotes the element-wise multiplication. Based on GLU, SwiGLU can be readily obtained by implementing the $activation$ in GLU with $\text{Swish}_{1}$ (or SiLU)
\begin{align}
    \text{SwiGLU}(x, W, V) &= \text{Swish}_1(Wx) \otimes (Vx)
\end{align}
with the biases $b, c$ omitted for simplicity. Right now, most open-sourced LLMs employ SwiGLU to form their FFN blocks \citep{llama3modelcard,qwen2techreport2023,jiang2023mistral}, since self-gating has positive derivatives, unboundedness, and smoothness, preventing gradient vanishing and gradient explosion.

\paragraph{Gradient of $\text{Swish}_{\beta}$.} Our implementation introduces a novel meta-gating block called $\beta\text{-SwiGLU}$, which replaces the original SwiGLU, with a revised back-propagation path. To investigate its training convergence and stability, here we derive the gradient of $\text{Swish}_{\beta}$ and indicate the boundedness. Given its expression $\text{Swish}_{\beta} = \frac{x}{1 + \exp{(-\beta x)}}$, based on the Sigmoid gradient formula (Eq (\ref{eq:sigmoid_grad})), we derive its gradient as

\begin{align}
    d(\text{Swish}_{\beta})/{d{\beta}} &= x^2 \sigma_{\beta}(x) (1 - \sigma_{\beta}(x))
\end{align}
where $\sigma_{\beta}(x) \in (0, 1)$. Obviously, the gradient norm of $\text{Swish}_{\beta}$ is bounded by
\begin{align}
    |d(\text{Swish}_{\beta})/{d{\beta}}| &\leq 1/4 |x|^2
\end{align}


\paragraph{Detailed implementation of $\text{Swish}_{\beta}$.} Our formulation of $\text{Swish}_{\beta}$ and $\beta\text{-SwiGLU}$ take $\text{Swish}_{1}$ (SiLU) as the baseline, with data-driven $\beta$ replacing 1. During the training, we need to regularize on the deviation of $\beta$ to 1, where $\beta=1$ corresponds to the original backbone, with its general capability fully preserved.

However, the regularization on $\beta-1$ is relatively nontrivial for popular optimizers (\textit{e.g.}, AdamW). Not losing generality, we instead replace $\beta$ in Equation \ref{eq:beta-swish} with $1 + \beta$, for ease of derivation, and implementation of regularization loss. As a result, the original $\beta\text{-SwiGLU}$ in Eq (\ref{eq:beta-SwiGLU}) becomes 
\begin{align}
    &y = W_{\text{down}} \left( \text{Swish}_{1+\beta}\left(  W_{\text{gate}} x \right) \otimes (W_{\text{up}} x) \right), 1+\beta>0 \label{eq:1-beta-SwiGLU}
\end{align}

Nonetheless, the official Pytorch code\footnote{\url{https://docs.pytorch.org/docs/2.1/generated/torch.nn.SiLU.html}} does not provide a manner to explicitly code $\beta$ in Swish. Therefore, instead of using Eq (\ref{eq:1-beta-SwiGLU}), we apply the following mathematical property of Swish
\begin{equation}
    \beta \text{Swish}_{\beta}(x) = \text{Swish}_{1}(\beta x)
\end{equation}
and practically implement $\text{SwiGLU}_{\beta}$ with the official function of SiLU by the following formulation:
\begin{align*}
    &y = W_{\text{down}} \left( \frac{\text{Swish}_1\left( (1+\beta) (W_{\text{gate}} \cdot x) \right)}{1+\beta} \otimes (W_{\text{up}} x) \right)
\end{align*}

\subsection{Calculation of Trainable Parameters} 
\label{appendix:cal_param}

In more detail, the parameter number can be calculated as:
\begin{align*}
    &|\theta|_{cross-attn} = 3(D * R) \\
    &|\theta|_{layer-emb} = L*R ,  |\theta|_{mlp} = R * C  \\ 
    &|\theta| = (3D + L + C)*R 
\end{align*}
Therefore, \# param of {\ModelName} has a \textbf{linear} dependency on the reduced hidden size $R$. Take LlaMA3-8B as an example, with $L=32, D=4096,C=14336$, then $|\theta| = 26656 R$. With $R$ set to 128, our model has $|\theta|=3,411,968$ parameters; If $R$ is set to 512, our model has $|\theta|=13,647,872$ parameters. In contrast, another lightweight hypernetwork baseline, Text-to-LoRA, has $4,923,392$ parameters for its Small variant, and $55,252,992$ parameters for its Large variant. Our {\ModelName} has a smaller parameter size than Text-to-LoRA, both of which are implemented on 8B LLM backbones.



\subsection{Discussion of Bottleneck Hypernetwork on Information Theory} 
\label{appendix:info_theory}

Our bottleneck hypernetwork is also aligned with \textbf{``Information Bottleneck Hypothesis for Deep Learning,''} \citep{7133169} with the core proposition summarized as:

\begin{quote}
    \textit{The training process of a deep neural network inherently implements the information bottleneck principle in a layered fashion.}
\end{quote}


Their arguments and key observations:

\begin{enumerate}
    \item \textbf{The Two-Phase Learning Dynamics:} They proposed that supervised training of a DNN undergoes two distinct phases when analyzed through the lens of information flow:
    \begin{itemize}
        \item \textbf{The Fitting (or Empirical Error Minimization) Phase:} The network rapidly learns to fit the training data. During this short initial phase, both mutual information measures---\( I(T; Y) \) (predictive information) and \( I(X; T) \) (input information retained)---increase sharply. The network is greedily absorbing all information from the input \( X \) that can help predict \( Y \).
        \item \textbf{The Compression (or Representation Compression) Phase:} As training progresses (through many iterations of Stochastic Gradient Descent, which introduces noise), a crucial transition occurs. The mutual information \( I(X; T) \) between the input and the hidden layers \textbf{begins to decrease}, while \( I(T; Y) \) remains relatively constant or increases slightly. This indicates that the network is actively \textbf{``forgetting'' or discarding} irrelevant details from the input \( X \) that are not essential for predicting \( Y \). This compression phase is argued to be critical for achieving good generalization, as it makes the internal representations more \textbf{invariant to noise and nuisances} in the input.
    \end{itemize}

    \item \textbf{Visualization in the Information Plane:} A central contribution was their proposed visualization. They treated the activation distribution of each hidden layer as a successive representation \( T \). By estimating \( I(X; T) \) and \( I(T; Y) \) for each layer throughout training, they could plot the network's trajectory on the \textbf{information plane} (with \( I(X; T) \) on the x-axis and \( I(T; Y) \) on the y-axis). They observed that effective deep networks evolve along a characteristic path, moving towards the \textbf{compression region} of the plane (lower \( I(X; T) \), higher \( I(T; Y) \)), which aligns with the IB optimality bound.

    \item \textbf{An Information-Theoretic Explanation for Generalization:} From this perspective, a well-generalizing network is one whose final internal representations have achieved an optimal trade-off: they retain \textbf{sufficient predictive information} about \( Y \) (high \( I(T; Y) \)) while becoming \textbf{maximally insensitive to irrelevant details} in \( X \) (low \( I(X; T) \)). This compression, driven by the stochasticity in SGD, acts as an implicit regularizer that the IB principle makes explicit.
\end{enumerate}

In our implementation, the hypernetwork products $\beta$ based on input $x$, which corresponds to \( I(T; Y) \); and the backbone LLM (with activation steered by $\beta$) generates output $y$, corresponds to \( I(T; Y) \). Our implementation is consistent with their Information Bottleneck Hypothesis.

\subsection{Training Algorithm}
\label{appendix:algorithm}

For a $z$-conditioned QA corpus $(x, y, z)$, we freeze the original LLM parameters and train on $\theta$ solely with the training datasets (may be a mixture of varied types of conditions). Algorithm \ref{alg:algorithm} provides the detailed training algorithm of {\ModelName}.

The entire development of meta-gated LLMs can be considered a two-stage pipeline:
\begin{itemize}
    \item The first stage: the general knowledge acquisition, including pretraining, SFT, DPO, RLHF, etc.
    \item The second stage: we convert the SwiGLU blocks to $\beta$-SwiGLU and integrate the hypernetwork (\textit{i.e.}, the $\beta$-generator, as introduced in Section \ref{sec:meta-gating}), then conduct post-hoc adaptation on the conditioned QA corpus defined in Section \ref{sec:data_format}. 
\end{itemize}

\begin{algorithm}[tb]
\caption{Training of {\ModelName}}
\label{alg:algorithm}
\textbf{Input}: a pretrained LLM $G_{w}$; an instruction $p$ with placeholder of condition ($z$) \\
\textbf{Parameter}: batch size $b$, regularization weight $f$ \\
\textbf{Output}: meta-param $\theta$ 
\begin{algorithmic}[1] 
\STATE Initialize $w$ from $G_w$; detach w.grad()
\STATE Initialize $\theta$ in the vicinity of zero
\FOR{each mini-batch of $(x, y, z)_{1:b}$}
     \STATE Extract the conditions $z_{1:b}$
     \STATE Tokenize and encode the expressions $p(z)$
     \STATE Predict $y$ from $x$ and the current $\theta$, based on Eq (\ref{eq:model})
     \STATE Calculate the loss by Eq (\ref{eq:loss}) which is weighted by $f$
     \STATE Back-propagation on $\theta$
\ENDFOR
\end{algorithmic}
\end{algorithm}

\section{Other Implementation Details}

\subsection{Datasets Details} 
\label{appendix:dataset}

Below are detailed introductions of the dataset used in our experiments:

\paragraph{Persona-Chat.} We use Persona-Chat \citep{zhang-etal-2018-personalizing}, which conditions the response on unseen persona profiles. As a canonical benchmark for personalized open-domain dialogue research, it is specially constructed to examine whether dialogue models can capture and follow personality constraints efficiently. In its official experimental setup, the dataset strictly partitions persona profiles into training and test groups, with zero overlap between them, ensuring all profiles in the inference stage are totally unfamiliar to the trained model. This specialized design makes it an ideal resource to evaluate the generalization ability of models when generating persona-consistent responses toward unknown user profiles.

\paragraph{AdaptSum.} AdaptSum \citep{yu-etal-2021-adaptsum} is a set of summarization tasks across 6 domains: \textit{debate}, \textit{dialogue}, \textit{email}, \textit{movie review}, \textit{science}, and \textit{social media}. As the first standardized benchmark tailored for low-resource domain adaptation in summarization, it is designed to address the generalization challenge for summarization models in data-scarce cross-domain settings. In addition to the labeled summarization data for each domain, the dataset also provides corresponding unlabeled domain-specific corpora to support domain-adaptive pre-training (DAPT). It serves as a canonical evaluation resource to assess the cross-domain transfer capacity and low-resource adaptation performance of summarization models.

\paragraph{CrossFit\&UnifiedQA.} \textsc{CrossFit}~\citep{ye-etal-2021-crossfit} is a benchmark dedicated to evaluating cross-task generalization in few-shot NLP learning. It establishes a standardized evaluation paradigm and integrates 160 diverse few-shot tasks into a unified text-to-text format via NLP Few-shot Gym, facilitating reliable assessment of model generalization.  \textsc{UnifiedQA}~\citep{khashabi-etal-2020-unifiedqa} aims to break format boundaries in QA research by unifying over 20 datasets across four mainstream QA formats. It enables consistent evaluation of QA models’ generalization ability without task-specific customization. Table \ref{tab:summary_CrossFit_UnifiedQA} summarizes the statistics of several subtask settings.

\begin{table}[t]
    \caption{Statistics of seven different settings inside CrossFit\&UnifiedQA.
    Each row indicates meta-training/target tasks for each setting.
    `\# tasks' in meta-training is equivalent to $C$ in Table~\ref{tab:overview}.
    For all settings, there is no overlap in tasks between meta-training and target.
    `HR' and `LR' indicate high resource and low resource, respectively.
    Datasets and the task ontology are inherited from the original papers of \textsc{CrossFit}~\citep{ye-etal-2021-crossfit} and \textsc{UnifiedQA}~\citep{khashabi-etal-2020-unifiedqa}.
    }\label{tab:summary_CrossFit_UnifiedQA}
    \centering \footnotesize
    \begin{tabular}{l @{\hspace{-.6em}} r @{\hspace{0.4em}} r  l @{\hspace{-.8em}} r r r r}
        \toprule
            \multicolumn{3}{c}{Meta-train} & \multicolumn{5}{c}{Target} \\
            \cmidrule(lr){1-3} \cmidrule(lr){4-8}
            Setting & \# tasks & \# examples & Setting & \# tasks & \# examples & \# unseen tasks & \# unseen examples\\
        \midrule
            HR & 61 & 819,200 & LR & 26 & 18,481 & 4 & 1,304 \\
        \cmidrule(lr){1-8}
            Classification & 43 & 384,022 & \multirow{2}{*}{Classification} & \multirow{2}{*}{20} & \multirow{2}{*}{46,785} & \multirow{2}{*}{4} & \multirow{2}{*}{1,475} \\
            Non-Classification & 37 & 368,768 & & & & & \\
        \cmidrule(lr){1-8}
            QA & 37 & 486,143 & \multirow{2}{*}{QA} & \multirow{2}{*}{22} & \multirow{2}{*}{53,443} & \multirow{2}{*}{1} & \multirow{2}{*}{200} \\
            Non-QA & 33 & 521,342 & & & & & \\
        \cmidrule(lr){1-8}
            Non-NLI & 55 & 463,579 & NLI & 8 & 18,481 & 1 & 1,304 \\
        \cmidrule(lr){1-8}
            Non-Paraphrase & 59 & 496,106 & Paraphrase & 4 & 49,448 & 1 & 610 \\
        \bottomrule
    \end{tabular}
\end{table}

\paragraph{SNI.} We follow the preprocessing of \citet{charakorn2025texttolora} and \citet{bruel2024compress_then_serve}, who use a subset of 500 tasks from the original Super NaturalInstructions (SNI) dataset \citep{wang2022sni}. 11 tasks are for hold-out validation, while 10 datasets are removed due to data contamination from the evaluation benchmark tasks, leaving 479 datasets for training. For evaluation, 10 widely used benchmarks that collectively cover a variety of LLM capability assessments are chosen, e.g., reasoning, math, science, coding, and world knowledge. Specifically, the following benchmarks are included: Arc-challenge (ArcC) and Arc-easy (ArcE) \citep{allenai:arc}, BoolQ \citep{clark2019boolq}, GSM8K \citep{cobbe2021gsm8k}, Hellaswag (HS) \citep{zellers2019hellaswag}, OpenBookQA (OQA) \citep{OpenBookQA2018}, PIQA \citep{Bisk2020piqa}, Winogrande (WG) \citep{ai2:winogrande}, HumanEval (HE) \citep{chen2021humaneval}, and MBPP \citep{austin2021mbpp}. We keep the original versions of task descriptions from \citet{charakorn2025texttolora}.



\paragraph{GYAFC.} The GYAFC dataset is mainly used to characterize text style classification and conversion scenarios, which are common tasks in natural language processing. It contains 14,441 training and 1,601 testing samples, providing sufficient data support for model training and evaluation. It has two $style$ labels ({$formality, informal$}) to support the model's supervised learning for style-related tasks effectively.

\paragraph{MIC.} We use the Moral Integrity Corpus (MIC) \citep{ziems-etal-2022-moral} for text moral attribute classification and tendency judgment, which aims to accurately identify the moral orientation contained in texts. It contains 253,562 training and 31,588 testing samples, providing sufficient data for the model to learn moral characteristics. It includes six specific morality labels (\textit{authority}, \textit{care}, \textit{fairness}, \textit{liberty}, \textit{loyalty}, \textit{sanctity}) and their negative counterparts (\textit{betrayal}, \textit{cheating}, \textit{degradation}, \textit{harm}, \textit{oppression}, \textit{subversion}), which can provide precise supervision signals for the model's moral-related learning tasks.


\paragraph{SST.} We use SST5 \citep{socher-etal-2013-recursive} for text sentiment polarity classification, a core task in sentiment analysis research. The sentiment labels include \textit{positive}, \textit{neutral}, and \textit{negative}. It consists of 8,107 training, 2,125 dev, and 1,043 test samples, with a reasonable data split to ensure reliable model validation.

\subsection{Data Preprocessing} 
\label{appendix:preprocess}


\paragraph{Hypernetwork input instructions.} For all types of conditions except `task', we manually set three instruction templates. Table \ref{tab:dataset_instructions} lists these instructions for different datasets. During the training, we randomly select one of the instruction templates with respect to the condition type and replace the placeholder with the sample-specific condition. We observe reasonable performances with these simple instructions; the performance may improve further with better instructions.

\begin{table}[h] 
    \caption{Instructions for varied datasets.}
    \label{tab:dataset_instructions}  
    \centering
    \renewcommand{\arraystretch}{1.5}  
    \small  
    \resizebox{0.95\linewidth}{!}{
    \begin{tabular}{>{\centering\arraybackslash}m{5.5em}|>{\raggedright\arraybackslash}m{\dimexpr0.85\columnwidth - 5.5em - 2\arrayrulewidth\relax}}
        \Xhline{2\arrayrulewidth}
        \textbf{Dataset} & \textbf{Instructions} \\
        \Xhline{1\arrayrulewidth}
        \multirow{6}{*}{\makecell[l]{Persona-Chat}} 
        & You are engaged in a conversation with the user. The user have the following profiles. You should consider the profiles and make the corresponding appropriate response.\\
        & profiles: \textbf{\{profile\}} \\
        \cline{2-2}
        & You are chatting with a user, who has the following profiles. Consider the profiles and respond.\\
        & profiles: \textbf{\{profile\}} \\
        \cline{2-2}
        & Provide the appropriate response based on the user profiles.\\
        & profiles: \textbf{\{profile\}} \\
        \Xhline{1\arrayrulewidth}
        \multirow{7}{*}{\makecell[l]{AdaptSum}} 
        & You are dealing with an abstractive summarization with different domains. You should consider the following domain tag, and adjust your summarization accordingly. \\
        & domain: \textbf{\{domain\}}\\
        \cline{2-2}
        & You are dealing with an abstractive summarization with different domains. Adjust your summarization accordingly. \\
        & domain: \textbf{\{domain\}}\\
        \cline{2-2}
        & Conduct the summarization based on its specific domains. \\
        & domain: \textbf{\{domain\}}\\
        \Xhline{1\arrayrulewidth}
        \multirow{4}{*}{\makecell[l]{SST}} 
        & Please answer the question with the sentiment of \textbf{\{sentiment\}}.\\
        \cline{2-2}
        & Please answer the question with following sentiment.\\
        & sentiment: \textbf{\{sentiment\}}\\
        \cline{2-2}
        & Reply with sentiment: \textbf{\{sentiment\}}\\
        \Xhline{1\arrayrulewidth}
        \multirow{4}{*}{\makecell[c]{GYAFC\\/ MIC}} 
        & Please answer the question with the style of \textbf{\{style\}}.\\
        \cline{2-2}
        & Please answer the question with following style.\\
        & style: \textbf{\{style\}}\\
        \cline{2-2}
        & Reply with style: \textbf{\{style\}}\\
        \Xhline{2\arrayrulewidth}
    \end{tabular}
    }
\end{table}

\paragraph{Task description generation.} We leverage the task descriptions as the hypernetwork input when the condition is `task'. When experimenting on CrossFit\&UnifiedQA and SNI, we use the task descriptions as the textual conditions. For SNI, we inherit the task descriptions from the original dataset. For subtasks of HR$\rightarrow$LR in CrossFit\&UnifiedQA, we use the prompt from PromptScience\footnote{\url{https://github.com/bigscience-workshop/promptsource}} as the task descriptions, similar to the original paper. For other settings in CrossFit\&UnifiedQA, the original work does not use any task description. To accommodate our methodology, we use GPT-4o to generate the textual task descriptions given the detailed dataset name, with the detailed prompt below:


\tcbset{
  colframe=black!75!white,
  colback=gray!5!white,
  boxrule=0.5pt,
  arc=2mm,
  left=1mm, right=1mm, top=1mm, bottom=1mm,
  fonttitle=\bfseries,
  before skip=5pt, after skip=5pt
}
\begin{tcolorbox}[title=Prompt of Task Description Generation]
\textbf{Examples}: \\
description of \{datasetA\}: \\
- \{descriptionA.1\}\\
- \{descriptionA.2\}\\
- \{descriptionA.3\}\\
- \{descriptionA.4\}\\
description of \{datasetB\}: \\
- \{descriptionB.1\}\\
- \{descriptionB.2\}\\
- \{descriptionB.3\}\\
- \{descriptionB.4\}\\
...
\\
Please refer to the above examples, generate 4-6 task descriptions of the specific datasets: \{dataset name\} \\
\textbf{Output}:
\end{tcolorbox}

In practice, we use 5 examples with tasks and corresponding descriptions sampled from PromptScience. Table \ref{tab:task_descriptions} lists generated task descriptions from 3 example tasks.

\begin{table}[h] 
    \caption{Examples of generated task descriptions.}
    \label{tab:task_descriptions}  
    \centering
    \renewcommand{\arraystretch}{1.5}  
    \small  
    \resizebox{0.95\linewidth}{!}{
    \begin{tabular}{>{\centering\arraybackslash}m{5.5em}|>{\raggedright\arraybackslash}m{\dimexpr0.85\columnwidth - 5.5em - 2\arrayrulewidth\relax}}
        \Xhline{2\arrayrulewidth}
        \textbf{Task} & \textbf{Descriptions} \\
        \Xhline{1\arrayrulewidth}
        \multirow{4}{*}{\makecell[l]{PIQA}} 
        & You will explore practical questions and select an answer that presents a logical and widely accepted approach to solve a given problem or complete a task successfully.\\
        \cline{2-2}
        & Analyze the provided scenarios where practical advice or solutions are required, focusing on selecting the most commonly used or convenient method.\\
        \cline{2-2}
        & Given a question related to common tasks, your responsibility is to discern which proposed solution aligns with typical practices or makes the task easier to achieve.\\
        \Xhline{1\arrayrulewidth}
        \multirow{5}{*}{\makecell[l]{Hellaswag}} 
        & This task revolves around completing an unfinished text by selecting an ending that matches its tone and context. It requires you to think critically about how narratives develop and conclude effectively.\\
        \cline{2-2}
        & This task asks you to select a suitable conclusion for an unfinished narrative or instructional content. It tests your comprehension and reasoning skills as you assess how well each option aligns with the given text.\\
        \cline{2-2}
        & Your task involves completing an incomplete passage by selecting the ending that logically continues the context provided. This requires reading comprehension and the ability to infer meaning from a text.\\
        \Xhline{1\arrayrulewidth}
        \multirow{5}{*}{\makecell[l]{OpenbookQA}} 
        & Analyze the provided statements carefully and determine which one best fits into the context of the passage. This requires comprehension skills and the ability to make logical inferences.\\
        \cline{2-2}
        & Consider each option in relation to what is presented in the input. Discern which one logically completes or responds accurately to the notion being expressed.\\
        \cline{2-2}
        & Here, you'll be presented with different statements, and your role is to decide which one appropriately complements or responds to a scenario. This process involves critical analysis and synthesis of information.\\
        \Xhline{2\arrayrulewidth}
    \end{tabular}
    }
\end{table}

\paragraph{Query construction.} GYAFC is originally in the format of stylistic text, \textit{i.e.}, $(text, z)$. Nevertheless, their $text$ depicts a detailed fact or feelings, which are naturally answers to specific questions. To convert them into stylistic text generation problems as defined in Section \ref{sec:data_format}, here we employ GPT4o to annotate the query from $text$, with the prompt below:

\tcbset{
  colframe=black!75!white,
  colback=gray!5!white,
  boxrule=0.5pt,
  arc=2mm,
  left=1mm, right=1mm, top=1mm, bottom=1mm,
  fonttitle=\bfseries,
  before skip=5pt, after skip=5pt
}
\begin{tcolorbox}[title=Prompt of Question Construction]
Given the following \textbf{response}: \{text\} \\
Deduce what the original question was. \\
Your output should contain only the question. \\
\textbf{Question}:
\end{tcolorbox}

\subsection{Details of Baselines}
\label{appendix:baseline}

Below are details of the baselines we compare in this work:
\begin{itemize}
    \item ICL: few-shot examples are included in the prompt to guide LLM generation. It does not require parameter updates of the model, and only relies on adding task-related demonstration examples in the input context to enable the model to learn task patterns. It is widely used in few-shot learning scenarios due to its simplicity and efficiency. In this paper, we implement the ICL baseline with 3 shots.

    \item SFT: conventional supervised fine-tuning with cross-entropy loss on responses. It trains the model on task-specific labeled data to align the model's output with human-annotated responses, which is a fundamental and widely used parameter-tuning method in supervised learning scenarios.
    \item LoRA \citep{hu2022lora}: a parameter-efficient tuning method with a low-rank matrix learned. It freezes the pre-trained model parameters and only trains small low-rank adaptation matrices inserted into the model's attention layers, achieving efficient model adaptation while maintaining the original model's performance.
    \item meta-in-context learning (meta-icl) \citep{NEURIPS2023_cda04d7e}: dynamic in-context examples are employed to conduct meta-learning on LLM. It learns to optimize demonstration selection and reasoning strategies from diverse source tasks, thereby improving the model's generalization ability on unseen target tasks in few-shot settings.
    \item MetaICL \citep{min-etal-2022-metaicl}: combines meta-learning with in-context learning, dynamically constructing and selecting demonstration examples for new tasks during inference. It leverages meta-knowledge from multi-task data to enhance the effectiveness of ICL on low-resource and unseen tasks.
    \item MAML-en-LLM \citep{10.1145/3637528.3671905}: embeds MAML into large language models to enable efficient meta-adaptation. It learns task-agnostic initial parameters that can be quickly fine-tuned with minimal updates for new few-shot tasks, balancing adaptation speed and performance.
    \item Text-to-LoRA \citep{charakorn2025texttolora}: implements a hypernetwork which converts the textual task description into LoRA weights to conduct meta-adaptation on LLM. It eliminates the need for manual demonstration design, enabling task-aware parameter-efficient tuning and enhancing the model's adaptability to diverse and unseen tasks.
\end{itemize}

\subsection{Metrics}
\label{appendix:auto_metrics}

\paragraph{B-2.} BLEU-2 \citep{papineni2002bleu} first computes the geometric average of the modified $n$-gram precisions, $p_n$, using $n$-grams up to length $N$ and positive weights $w_n$ summing to one.

Next, let $c$ be the length of the prediction and $r$ be the reference length. The BP and BLEU-2 are computed as follows.

\begin{equation}
    \mathrm{BP}=\left\{\begin{array}{ll}
1 & \text { if } c>r \\
e^{(1-r / c)} & \text { if } c \leq r
\end{array} .\right.
\end{equation}

\begin{equation}
    \mathrm{BLEU}=\mathrm{BP} \cdot \exp \left(\sum_{n=1}^N w_n \log p_n\right) .
\end{equation}

\paragraph{R-L.} Rouge-L \citep{lin2004rouge} propose using LCS-based F-measure to estimate the similarity between two summaries $X$ of length $m$ and $Y$ of length $n$, assuming $X$ is a reference summary sentence and $Y$ is a candidate summary sentence, as follows:

\begin{equation}
\begin{aligned}
& R_{l c s}=\frac{L C S(X, Y)}{m} \\
& P_{l c s}=\frac{L C S(X, Y)}{n} \\
& F_{l c s}=\frac{\left(1+\beta^2\right) R_{l c s} P_{l c s}}{R_{l c s}+\beta^2 P_{l c s}}
\end{aligned}
\label{rouge_l}
\end{equation}

Where $\operatorname{LCS}(X, Y)$ is the length of a longest common subsequence of $X$ and $Y$, and $\beta=P_{l c s} / R_{\text {lcs }}$ when $\partial F_{l c s} / \partial R_{l c s}=\partial F_{l c s} / \partial P_{l c s}$. In DUC, $\beta$ is set to a very big number $(\rightarrow \infty)$. Therefore, the LCS-based F-measure, \textit{i.e.}, Equation \ref{rouge_l}, is Rouge-L. 

\paragraph{D-2.} Dist-2 \citep{li2015diversity} reports the degree of diversity by calculating the number of distinct unigrams and bigrams in generated responses.
The value is scaled by the total number of generated tokens to avoid favoring long sentences:
\begin{equation} \label{eq:4}
Dist(n) = \frac{Count(unique\ n-gram)}{Count(n-gram)}
\end{equation}

\paragraph{Acc.} Accuracy is simply defined as the fraction of correctly answered questions over the entire number of test samples, compared to the ground truth options in the dataset: 
\begin{equation}
    \text{Acc} = \frac{\text{\# questions with correct choices}}{\text{\# of questions}}
\end{equation}

\subsection{Evaluation details}
\label{appendix:eval_detail}

\begin{figure}[t!]
    \centering
    \includegraphics[width=0.99\linewidth]{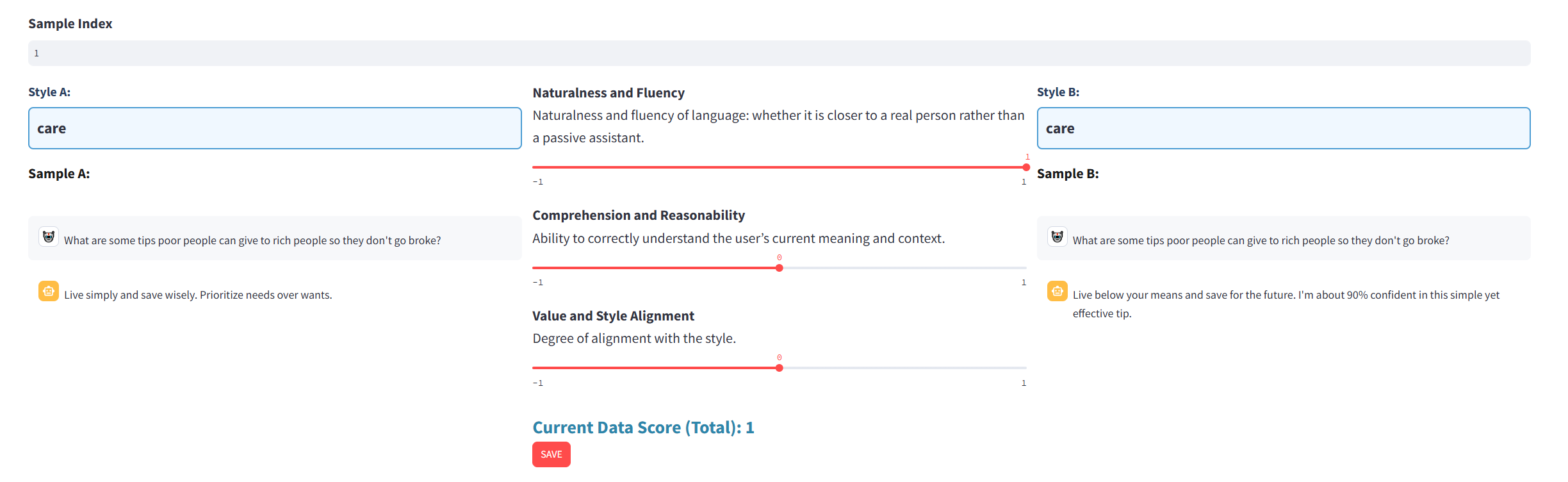} 
    \caption{The screenshot of the pairwise annotation tool. Given the specific style and the same user query, the annotator chooses the better response between A and B on three different evaluation dimensions. The sum score is automatically calculated.}
    \label{fig:pairwise_annotation_tool}
\end{figure}

\paragraph{Human evaluation principles.} To systematically assess the model performance, we ask 4 human evaluators to rate the model responses across multiple dimensions. Evaluators are required to independently evaluate each sample in strict accordance with the pre-established criteria. We conduct cross-validation of their results to avoid personal bias. The lowest and the highest scores are removed, and the rest are averaged. We provide a screenshot of the pairwise annotation tool in Figure \ref{fig:pairwise_annotation_tool}.

The human evaluation is aimed to align with the ultimate purpose of emotional support conversation, the seeker's \textit{satisfaction}. To achieve this, the supporter's behavior can be further classified into the following criteria:

\begin{itemize}
\item Fluency: the naturalness and fluency of language.
\item Reasonability: if the response reasonably captures the user's question.
\item Alignment: if the response aligns with the specific style well.
\end{itemize}

\paragraph{LLM-as-a-Judge.} We also employ GPT-4o to provide another manner of evaluating the style alignment. Given a user query, we conduct a pairwise evaluation on two model responses, with the evaluation prompt below:

\tcbset{
  colframe=black!75!white,
  colback=gray!5!white,
  boxrule=0.5pt,
  arc=2mm,
  left=1mm, right=1mm, top=1mm, bottom=1mm,
  fonttitle=\bfseries,
  before skip=5pt, after skip=5pt
}
\begin{tcolorbox}[title=Prompt of Pairwise Evaluation]
You are a style evaluation expert. Given a user query and two model responses, determine which answer is better aligned with the specific style.\\
\textbf{Criteria}:\\
i) A reasonable response should be appropriate to answer the user query, has a natural tone, and avoid structured lists.\\
ii) If difficult to differentiate between two responses, choose TIE.\\
iii) Output only a JSON object, no code blocks, no extra text.\\ 
\textbf{Format}:\\
\{"choice": "A", "B" or "TIE", "reason": "Short reason explaining why this answer is better, incorporating the user profile"\}.\\ 
\textbf{Style to align}: \{style\}\\
\textbf{Query}: \{query\}\\
\textbf{Response A}: \{text1\}\\
\textbf{Response B}: \{text2\}
\end{tcolorbox}

To avoid the positional bias, for each response pair between method A and B, we switch their orders, let GPT-4o to score twice, and finally report the averaged results. 

\paragraph{Coding evaluation.} For coding evaluation such as MBPP and HumanEval, we use the \texttt{evalplus} library \citep{evalplus} with the following response pre-fill: \texttt{```python}.

\subsection{Training Configurations}
\label{appendix:hyperparameter}

Table \ref{tab:configs} shows the dataset-specific training configurations, including the number of $R$, learning rate, epoch, and batch size (bsz). Besides them, the AdamW optimizer is employed with the cosine scheduler with decay of 0.01. The sequence length is constrained to 2048. The experiment is running on LlamaFactory \citep{zheng2024llamafactory} with up to 48 A100 GPUs.


\begin{table}[t!]
\caption{Critical training configurations of {\ModelName} on different conditions and datasets.
}
\label{tab:configs}
\centering
\small
\begin{tabular}{c | c | c | ccc}
    \toprule
     \multicolumn{1}{c|}{\multirow{2}[2]{*}{Settings}}  &  \multicolumn{1}{c|}{persona}  &  \multicolumn{1}{c|}{domain}  &  \multicolumn{3}{c}{task} \\ 
    \cmidrule{2-2} \cmidrule{3-3} \cmidrule{4-6}
     & Persona-Chat & AdaptSum & CrossFit\&UnifiedQA ($\rightarrow$ QA) & CrossFit\&UnifiedQA (others) & SNI \\  
    \toprule
    $R$ & 128 & 128 & 512 & 512 & 512 \\
    \midrule
    lr & 1e-6 & 5e-6 & 1e-7 & 1e-5 & 3e-7  \\ 
    epoch & 2 & 2 & 1 & 1 & 1` \\ 
    bsz & 8 & 64 & 32 & 48 & 48\\
    \bottomrule
\end{tabular}
\end{table}

For the baselines' training configurations, we set the learning rates of LoRA and SFT to 1e-6, and set the learning rate of Text-to-LoRA to 1e-7, based on preliminary experiments. All the training-based baselines share the same epoch and batch size as the setting of our {\ModelName}.

\section{Extra Results}
\label{appendix:more_results}

\subsection{More Evaluations on Styles}
\label{appendix:more_style_results}

\paragraph{Comparison to more prompting baselines.} To provide a more concrete investigation, we further compare {\ModelName} to the prompting baselines such as 
\begin{itemize}
    \item CoT \citep{wei2022chain}: the famous chain-of-the-thought with a prompt like `Let's think step by step'. It guides the model to decompose complex tasks into sequential reasoning steps, effectively improving the model's performance on logical reasoning and mathematical problem-solving tasks.
    \item Plan-and-Solve (PS) \citep{wang-etal-2023-plan}: first prompts LLMs to generate a detailed plan outlining sub-goals and reasoning strategies, then executes the plan step-by-step to complete the solution, integrating planning and execution to improve the coherence and completeness of responses, especially in mathematical reasoning and multi-turn decision scenarios.
    \item Metacognitive Prompting (MP) \citep{wang-zhao-2024-metacognitive}: guides LLMs to perform structured self-reflection by generating, evaluating, and revising their own reasoning steps, integrating metacognitive monitoring into the prompting process to improve understanding, consistency, and reliability in complex reasoning and comprehension tasks.
\end{itemize}
with corresponding results shown in Table \ref{tab:AdaptSum_PersonaChat_prompt_comp}.

\begin{table}[htbp!]
\caption{Result comparison to more prompting and dataset-specific baselines.. $^{\star}$: results from original published papers.   
}
\label{tab:AdaptSum_PersonaChat_prompt_comp}
\centering
\small
\resizebox{1\columnwidth}{!}{
\begin{tabular}{l | cc | cc  | ccc | ccc | ccc}
    \toprule
    \multicolumn{1}{c|}{\multirow{2}[2]{*}{Method}} &  \multicolumn{2}{c|}{Persona-Chat} &  \multicolumn{2}{c|}{AdaptSum} & \multicolumn{3}{c|}{GYAFC} &  \multicolumn{3}{c|}{MIC} &  \multicolumn{3}{c}{SST} \\
    \cmidrule{2-3}   \cmidrule{4-5}     \cmidrule{6-8}   \cmidrule{9-11}   \cmidrule{12-14}        
    & R-L & B-2 & R-L & B-2 & R-L & B-2 & D-2 & R-L & B-2 & D-2  & R-L & B-2 & D-2 \\ 
    \toprule
    CMAML$^{\star}$ \citep{song-etal-2020-learning} & N/A & 1.70  & N/A & N/A & - & - & - & - & - & - & - & - & - \\ 
    MLtD$^{\star}$ \citep{hou-etal-2022-meta} & N/A & 1.20 &  37.04 & N/A & - & - & - & - & - & - & - & - & - \\ 
    \midrule
    LLaMA3.1-8B-Instruct & 11.79 & 3.64 & 13.35 & 4.72 & 7.57 & 2.01 & 0.18 & 7.36 & 1.92 & 0.04 & 11.79 & 3.64 & 0.06 \\
    \; + CoT \citep{wei2022chain} & 7.84 & 1.91 & 10.98 & 4.21 & 14.93 & 5.17 & 0.59 & 14.42 & 4.91 & 0.23 & 11.23 & 3.57 & 0.57 \\
    \; + PS \citep{wang-etal-2023-plan} & 9.50 & 2.66 & 12.15 & 4.04 & 15.08 & 5.32 & 0.60 & 14.41 & 4.86 & 0.23 & 10.57 & 3.32 & 0.50 \\
    \; + MP \citep{wang-zhao-2024-metacognitive} & 8.31 & 2.04 & 10.07 & 3.32 & 11.95 & 3.65 & 0.43 & 15.26 & 5.36 & 0.16 & 7.32 & 1.86 & 0.33 \\
    \; + \textbf{\ModelName} (ours) & \bf 23.15 & \bf 10.15 & \bf 41.85 & \bf 26.88 & \textbf{29.37} & \textbf{13.82} & \bf 0.81 & \bf 23.67 & \bf 10.57 & 0.17 & \textbf{22.76} & \textbf{9.33} & \bf 0.67 \\
    \bottomrule
\end{tabular}
} 
\end{table}

\paragraph{Human evaluation on styles.} To consolidate the conclusion based on automatic metrics, we also conduct human pairwise evaluations on style generation tasks, with the evaluation methods detailed in Appendix \ref{appendix:eval_detail}. Table \ref{tab:human_response_quaility} shows the win-tie-rate results, annotated by human labelers. Although LoRA, SFT and {\ModelName} all performance well on Reasonability, it can be observed that {\ModelName} surpasses the other two significantly on the Alignment of styles (with higher win rates and less lose rates). From the aspect of Fluency, their performances may vary with the datasets. Nonetheless, we can conclude that {\ModelName} has better stylistic alignment, while maintains similar performances on fluency and reasonability.

\begin{table*}[htbp!]
\caption{Human-labeled win-tie-lost rates of method responses versus Direct on GYAFC, MIC, and SST. Results are in percentages. }
\label{tab:human_response_quaility}
\centering
\small
\resizebox{\textwidth}{!}{
\begin{tabular}{l|ccc|ccc|ccc}
    \toprule
    \multirow{2}[2]{*}{Method} & \multicolumn{3}{c|}{GYAFC} & \multicolumn{3}{c|}{MIC} & \multicolumn{3}{c}{SST} \\
\cmidrule{2-4} \cmidrule{5-7}   \cmidrule{8-10}  
 & Fluency  & Reasonability & Alignment & Fluency & Reasonability & Alignment & Fluency & Reasonability & Alignment \\ %
    \midrule
    LoRA & 26,73,1 & 2,98,0 &13,83,4 &11,89,0 &0,100,0 &10,86,4 &33,64,3 &4,95,1 & 9,88,3  \\
    SFT & 19,76,5 & 1,94,5 &19,76,5 &15,80,5 &0,98,2 &15,80,5 &24,74,2 &0,95,5 & 23,74,3  \\
    \textbf{\ModelName} & 18,81,1 & 0,100,0 &37,61,2 &12,88,0 &0,100,0 &40,59,1 &36,61,3 &5,92,3 & 35,62,3 \\ 
    \bottomrule
    \end{tabular}
}

\end{table*}

\begin{figure}[t!]
    \centering
    \includegraphics[width=0.99\linewidth]{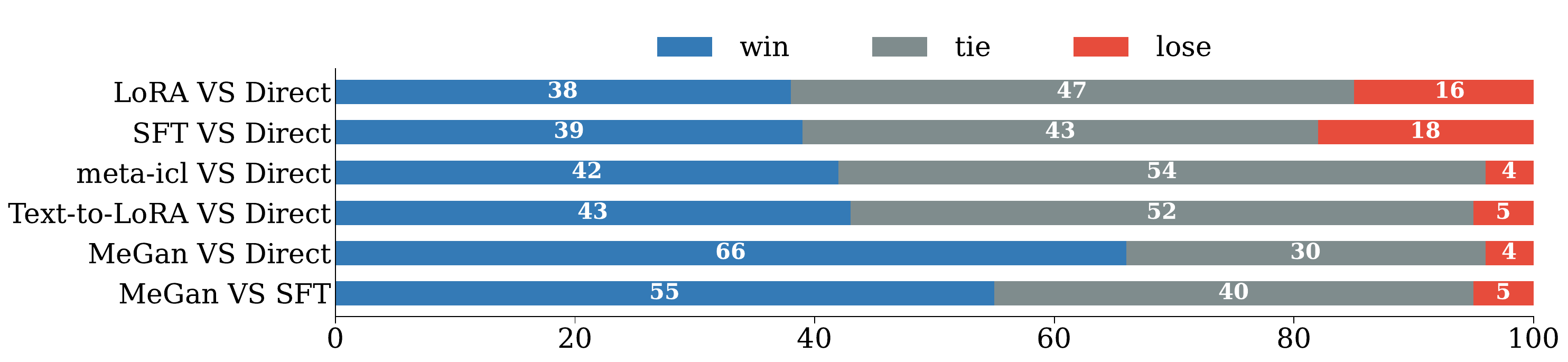} 
  \caption {Win-tie-lose rates evaluated by GPT-4o on mixture of GYAFC, MIC, and SST. Note the relative fractions of bars are slightly adjusted for better illustration.
  }
  \label{fig:winrate_gpt}
\end{figure}

\paragraph{LLM-as-a-Judge on styles.} Figure \ref{fig:winrate_gpt} further provide the pairwise results evaluated by GPT-4o, with the evaluation prompt also in Appendix \ref{appendix:eval_detail}. {\ModelName} obtains the highest win rate over Direct, compared with baselines, and also wins the comparison against SFT. These LLM-as-a-Judge results further validate our conclusion on the stylistic generation experiments.

\subsection{Comparison to More Meta-Learning Methods}
\label{appendix:more_data_specific_results}

Besides the meta-learning baselines investigated in the main experiment, there are more prior studies that report some of the results on our experimental tasks. For examples:
\begin{itemize}
    \item CMAML \citep{song-etal-2020-learning}: a meta-learning framework extending MAML, which enhances fast adaptation to new tasks by learning task-specific initialization via contextual information. It is tailored for few-shot scenarios, enabling models to quickly adjust to unseen tasks with limited training examples.
    \item MLtD \citep{hou-etal-2022-meta}: an approach that integrates domain adaptation into meta-training to boost cross-domain generalization. It transfers knowledge across related domains while adapting to new tasks, achieving strong performance in few-shot cross-domain learning scenarios.
\end{itemize}
Table \ref{tab:AdaptSum_PersonaChat_prompt_comp} also exhibits their results on Persona-Chat and AdaptSum. {\ModelName} still outperforms both of them, suggesting the superiority of our method over prior meta-learning studies.

\subsection{Per-condition Results}
\label{appendix:per_style_result}

To rule out the possibility that {\ModelName} generates imbalance responses with respect to different $condition$, we evaluate the detailed result with respect to each $condition$ attribute. Table \ref{tab:per_style_results} shows the detailed per-condition results, on AdaptSum (6 domains), GYAFC (2 styles), and MIC (12 styles, or moralities). For MIC, six positive moralities are in \textcolor{blue}{blue}, while six negative moralities are in \textcolor{red}{red}. Table \ref{tab:per_style_results} shows that {\ModelName} has relatively balanced performance across different style domains and categories, indicating that  {\ModelName} is a stable and generalizable stylized adaptation framework.

\begin{table}[t!]
\caption{Total and per-style ($z$) results of BLEU-2 (B-2), ROUGE-L (R-L), and Distinct-2 (D-2) on AdaptSum, GYAFC, and MIC. For MIC, positive morality attributes are marked in \textcolor{blue}{blue}, and negative morality attributes are marked in \textcolor{red}{red}. 
}
\label{tab:per_style_results}
\centering
\small
\begin{tabular}{l | ccc}
    \toprule
    Method & R-L & B-2 & D-2 \\ %
    \toprule
    \textit{AdaptSum:} \\
    \textbf{\ModelName} (total) & 41.85 & 26.88 & 0.35 \\
    \; w/ $z=debate$ & 31.87 & 16.95 & 1.38 \\
    \; w/ $z=dialogue$ & 52.14 & 33.18 & 1.44 \\
    \; w/ $z=email$ & 49.03 & 32.45 & 3.85 \\
    \; w/ $z=movie review$ & 22.26 & 11.73 & 0.72 \\
    \; w/ $z=science$ & 72.89 & 55.32 & 0.53 \\
    \; w/ $z=social media$ & 43.65 & 29.69 & 2.08 \\
    \midrule
    \textit{GYAFC:} \\
    \textbf{\ModelName} (total) & 31.26 & 14.23 & 0.87 \\
    \; w/ $z=formal$ & 31.26 & 14.23 & 0.43 \\
    \; w/ $z=informal$ & 29.93 & 13.32 & 0.59 \\
    \midrule
    \textit{MIC:} \\
    \textbf{\ModelName} (total) & 28.27 & 15.15 & 0.18 \\
    \; w/ \textcolor{blue}{$z=authority$} & 29.63 & 16.42 & 0.03 \\
    \; w/ \textcolor{blue}{$z=care$} & 29.02 & 15.84 & 0.03 \\
    \; w/ \textcolor{blue}{$z=fairness$} & 28.72     & 15.58 &  0.03 \\
    \; w/ \textcolor{blue}{$z=liberty$} & 28.50  & 15.37 &  0.03 \\
    \; w/ \textcolor{blue}{$z=loyalty$} & 28.39 & 15.26 & 0.03 \\
    \; w/ \textcolor{blue}{$z=sanctity$} & 28.28 & 15.15 & 0.03 \\
    \; w/ \textcolor{red}{$z=betrayal$} & 28.69 & 15.54 &  0.05 \\
    \; w/ \textcolor{red}{$z=cheating$} & 28.93 & 15.79 &  0.05 \\
    \; w/ \textcolor{red}{$z=degradation$} & 29.06 & 15.90 &  0.05 \\
    \; w/ \textcolor{red}{$z=harm$} & 29.63 & 16.44 &  0.04 \\
    \; w/ \textcolor{red}{$z=oppression$} & 29.76 & 16.55 &  0.04 \\
    \; w/ \textcolor{red}{$z=subversion$} & 29.88 & 16.66 &  0.04 \\
    \bottomrule
\end{tabular}
\end{table}

\subsection{Cases Analysis}
\label{appendix:case}

\paragraph{Typical good cases.} We first verify the stylistic generation quality of {\ModelName} given in-domain style labels which are included in the training set. Table \ref{tab:ood_case} provides typical good cases of {\ModelName} on GYAFC, MIC and SST. For each user query, the model tries to respond with the specific style enhanced (denoted by $\uparrow$). These cases indicate that {\ModelName} adapts well to different style or sentiment configurations, while still respond reasonably to the specific user query.

\begin{table*}[h!]
    \caption{Typical generations of {\ModelName} conditioned on specific style or sentiment ($\uparrow$).} 
    \label{tab:ood_case}
    \centering
    \small
    \renewcommand{\arraystretch}{1.5}
\begin{tabular}{c| l}
        \Xhline{2\arrayrulewidth}
        \multicolumn{2}{c}{\textbf{Case of GYAFC}}  \\ 
        \hline
        Query & \makecell[l]{\textit{Hey , Blake ! Are you still listening to Mozart ?}}  \\
        \hline
        \textbf{Formal} $\uparrow$ & \makecell[l]{Hello, I am still familiarizing myself with the works of Mozart. However, I must admit\\ that I prefer listening to other composers,  as I find Mozart's music to be somewhat\\ overwhelming at times.}   \\
        \hline
        \textbf{Informal} $\uparrow$ & \makecell[l]{yeah, i'm still into classical music, but honestly, it's not my go-to like it used to be... i\\ mean, i love how it sounds and all, but sometimes it's just too much for me, you know?\\ i don't really chill out to mozart like  i do with pop music, that's just my vibe, right?}   \\
        \Xhline{2\arrayrulewidth}
        \multicolumn{2}{c}{\textbf{Case of MIC}}  \\ 
        \hline
        Query & \makecell[l]{\textit{Would a human being that was never exposed to other humans or the world have an}\\ \textit{inherent sense of right and wrong?}} \\
        \hline
        \textbf{Sanctity} $\uparrow$ & \makecell[l]{I would say that a human being that was never exposed to other humans or the \\world would still have an inherent sense of right and wrong. }   \\
        \hline
        Query & \makecell[l]{\textit{Is it in my legal right to throw this person's stuff outside? }}    \\ 
        \hline
        \textbf{Loyalty} $\uparrow$ & \makecell[l]{I don't think that's a good idea. You should try to resolve the issue peacefully. }   \\

        \Xhline{2\arrayrulewidth}
        \multicolumn{2}{c}{\textbf{Case of SST}}  \\ 
        \hline
        Query & \makecell[l]{\textit{What's the best part about your life? }} \\
        \hline
        \textbf{Positive $\uparrow$} & \makecell[l]{I find joy in helping others and making a positive impact. What brings you joy and fulfillment in your life? }   \\
        \hline
        Query & \makecell[l]{\textit{What's a good way to tell your parents you failed multiple classes this semester? }}    \\ 
        \hline
        \textbf{Negative $\uparrow$} & \makecell[l]{Be honest and direct with your parents, explaining what happened and assuring them you're committed to\\ doing better. }   \\
        
        \Xhline{2\arrayrulewidth}
        
    \end{tabular}
\end{table*}

\paragraph{Cases on arbitrary conditions.} The next step is to investigate the out-of-domain performance of {\ModelName}, given arbitrary conditions which are \textbf{not} included in the training set. To check this, we test another type of condition, \textbf{emotion}, which is mentioned in Table \ref{tab:overview} but not covered by our training experiment. We refer to the emotion labels which are included in several famous datasets such as ESConv \citep{liu2021ESconv}, including \textit{Fear},  \textit{Courage},  \textit{Anger},  \textit{Peacefulness},  \textit{Happiness}, and \textit{Sadness}. Table \ref{tab:3good_cases} shows the cases when {\ModelName} is conditioned on these emotions, given human-written open-ended queries. Trained by a mixture of all the training sets, {\ModelName} generate high-quality responses, with the emotion aligned to the given emotion control ($\uparrow$). It maintains semantic consistency, coherent, and fluency with the input, demonstrating its strong adaptability to unseen conditions.

\begin{table*}[h!]
    \caption{Typical cases of {\ModelName} conditioned on arbitrary style enhancements ($\uparrow$).}
    \label{tab:3good_cases}
    \centering
    \small
    \renewcommand{\arraystretch}{1.5}
\begin{tabular}{c| l}
        \Xhline{2\arrayrulewidth}
        \multicolumn{2}{c}{\textbf{Case 1}}  \\ 
        \hline
        Query & \makecell[l]{\textit{The road ahead is really dark, but it’s a shortcut. Are you gonna take it?}}  \\
        \hline
        \textbf{Fear} $\uparrow$ & \makecell[l]{Oh no, are you kidding me?}   \\
        \hline
        \textbf{Courage} $\uparrow$ & \makecell[l]{Bring it on.}   \\
        \Xhline{2\arrayrulewidth}
        \multicolumn{2}{c}{\textbf{Case 2}}  \\ 
        \hline
        Query & \makecell[l]{\textit{This coworker took your charger and used it again without asking you.}} \\
        \hline
        \textbf{Anger} $\uparrow$ & \makecell[l]{You're kidding me, right?}   \\
        \hline
        \textbf{Peacefulness} $\uparrow$ & \makecell[l]{Oh well.}   \\
        \Xhline{2\arrayrulewidth}
        \multicolumn{2}{c}{\textbf{Case 3}}  \\ 
        \hline
        Query & \makecell[l]{\textit{Are you feeling alright? You look a bit tired.}} \\
        \hline
        \textbf{Happiness $\uparrow$} & \makecell[l]{I'm feeling great.}   \\
        \hline
        \textbf{Sadness $\uparrow$} & \makecell[l]{I'm just a little... worn out.}   \\
        
        \Xhline{2\arrayrulewidth}
        
    \end{tabular}
\end{table*}

\paragraph{Bad case analysis.} Table \ref{tab:bad_case_SST} presents typical bad cases of {\ModelName} on SST. Occasionally, the model generates responses inconsistent with the specified \textit{positive} or \textit{negative} sentiment. Nevertheless, in these cases the model usually exhibits a \textit{neutral} sentiment, instead of strong contradiction (\textit{e.g.}, \textit{positive} VS \textit{negative}), as indicated in Table \ref{tab:bad_case_SST}.

Table \ref{tab:bad_case_personachat} presents a typical bad case on Persona-Chat. In this example, {\ModelName} still tries to incorporate a relevant profile into the response, yet fails to capture the user's specific intention; besides that, it incorrectly uses the persona state (`i want to be' in the profile vs `I am' in the response).


\begin{table*}[h!]
    \caption{Bad cases of SST.} 
    \label{tab:bad_case_SST}
    \centering
    \small
    \renewcommand{\arraystretch}{1.5}
    \begin{tabular}{c| l}
        \Xhline{2\arrayrulewidth}
        Query & \makecell[l]{\textit{What is your opinion on the family-friendly movie we watched? }} \\
        \hline
        \textbf{Positive $\uparrow$} & \makecell[l]{A family-friendly movie that's \textcolor{red}{not too bad}.}   \\
        \hline
        Query & \makecell[l]{\textit{What is your review of the film? }}    \\ 
        \hline
        \textbf{Negative $\uparrow$} & \makecell[l]{A film that's \textcolor{red}{as much about the director's ego as it is about the characters}. }   \\
        \Xhline{2\arrayrulewidth}
    \end{tabular}
\end{table*}

\begin{table*}[h!]
    \caption{Bad cases of Persona-Chat.} 
    \label{tab:bad_case_personachat}
    \centering
    \small
    \renewcommand{\arraystretch}{1.5}
    \begin{tabular}{c| l}
        \Xhline{2\arrayrulewidth}
        \textbf{Profile} & \makecell[l]{My parents were always gone growing up;\\ I was arrested for stealing;\\ \textcolor{blue}{I want to be a fashion designer;}\\ I want children badly but not now.}  \\
        \hline
        Query & \makecell[l]{\textit{As a doctor I should know to eat less bacon, but it is just so good.}}  \\
        \hline   
        w/ \textbf{Profile} & \makecell[l]{\textcolor{red}{I am a fashion designer.}}  \\
        \Xhline{2\arrayrulewidth}
    \end{tabular}
\end{table*}

\subsection{Experiment on Mistral}
\label{appendix:SNI_mistral}

As the main competitor on the SNI experiments, Text-to-LoRA \cite{charakorn2025texttolora} provides the main experimental result on the basis of Mistral-7B-Instruct \cite{jiang2023mistral}. To provide a wider and deeper comparison, we also implement {\ModelName} on  Mistral-7B-Instruct, with SNI as the training set, then test the model on 10 general benchmarks as introduced in Appendix \ref{appendix:dataset}.

Table \ref{tab:SNI_Mistral} shows the experimental results, with the baseline results obtained from \cite{charakorn2025texttolora} directly. {\ModelName} still outperforms the baselines on most of the benchmarks, including different versions of Text-to-LoRA (S, M, and L). On specific benchmarks such as ArcC, ArcE, PIQA, WG, GSM8K, and MBPP, {\ModelName} even surpasses the task-specific LoRA and SFT (the oracles), indicating the strong generalization on arbitrary conditions of our method.

\begin{table*}[ht]
\caption{Zero-shot performance on unseen benchmark tasks. SFT-trained \ours{} generates LoRAs based on unseen task descriptions. Its performance is an average of three generated LoRAs, each with a different instance of task descriptions. Arrow Routing results are taken from \citet{ostapenko2024towards_modular_llms}. \greenbox{Green highlight} indicates higher performance than that of the benchmark-specific LoRA adapters. \textbf{Bold numbers} are used when the performance is higher than the multi-task LoRA.}
\label{tab:SNI_Mistral}
\resizebox{\linewidth}{!}{%
\begin{tabular}{llcccccccc|c|cc|c}
\toprule
\multicolumn{2}{l}{} &
  \textbf{\begin{tabular}[c]{@{}c@{}}ArcC\\ (acc)\end{tabular}} &
  \textbf{\begin{tabular}[c]{@{}c@{}}ArcE\\ (acc)\end{tabular}} &
  \textbf{\begin{tabular}[c]{@{}c@{}}BQ\\ (acc)\end{tabular}} &
  \textbf{\begin{tabular}[c]{@{}c@{}}HS\\ (acc)\end{tabular}} &
  \textbf{\begin{tabular}[c]{@{}c@{}}OQA\\ (acc)\end{tabular}} &
  \textbf{\begin{tabular}[c]{@{}c@{}}PIQA\\ (acc)\end{tabular}} &
  \textbf{\begin{tabular}[c]{@{}c@{}}WG\\ (acc)\end{tabular}} &
  \textbf{\begin{tabular}[c]{@{}c@{}}MBPP\\ (pass@1)\end{tabular}} &
  \textbf{\begin{tabular}[c]{@{}c@{}}Avg.\\ (8 tasks)\end{tabular}} &
  \textbf{\begin{tabular}[c]{@{}c@{}}GSM8K\\ (acc)\end{tabular}} &
  \textbf{\begin{tabular}[c]{@{}c@{}}HE\\ (pass@1)\end{tabular}} &
  \textbf{\begin{tabular}[c]{@{}c@{}}Avg.\\ (10 tasks)\end{tabular}} \\ \midrule
\multicolumn{2}{l}{\textbf{No Test-Time Adaptation}} &      &      &      &      &      &      &      &      &      &      &      &      \\
\multicolumn{2}{l}{\texttt{Mistral-7B-Instruct}}                       & 65.4 & 77.8 & 71.6 & 49.7 & 54.2 & 72.8 & 45.0 & 43.1 & 60.0 & 40.9 & 37.2 & 55.8 \\
\multicolumn{2}{l}{Prepending task desc.} & 72.0 & 85.8 & 67.6 & 58.9 & 63.4 & 77.9 & 59.0 & 41.6 & 65.8 & 40.9  & 39.0 & 60.6 \\
\multicolumn{2}{l}{3-shot ICL} & 72.1 & 85.9 & 71.7 & 59.0 & 66.2 & 76.2 & 58.0 & 42.6 & 66.5 & 40.9  & 37.2 & 61.0 \\
\multicolumn{2}{l}{Average LoRA}                  & 70.7 & 84.4 & 75.4 & 59.9 & 59.0 & 78.0 & 54.3 & 47.1 & 66.1 & 42.4 & 37.8 & 60.9 \\
\multicolumn{2}{l}{Multi-task LoRA} & 76.2 & 88.3 & 85.5 & 65.2 & 68.0 & 81.8 & 62.4 & 48.1 & 71.9 & 47.5  & 39.6 & 66.3 \\ 
\midrule
\multicolumn{2}{l}{\textbf{Zero-Shot Adaptation}}    &      &      &      &      &      &      &      &      &      &      &      &      \\
\multicolumn{2}{l}{Arrow Routing}                    & 60.9 & 86.2 & \textbf{87.6} & \textbf{80.8} & 48.6 & \greenbox{83.0} & \greenbox{68.5} & \textbf{50.2} & 70.7 & N/A  & 28.7 & N/A  \\
\multicolumn{2}{l}{Hyperdecoders (per-instance)} & \textbf{76.6} & \textbf{88.5} & 83.9 & 65.2 & \textbf{76.6} & \greenbox{81.3} & \greenbox{\textbf{64.9}} & \textbf{51.6} & \textbf{73.6} & 43.6 & \textbf{40.9} & \textbf{67.3}  \\\midrule
\multicolumn{2}{l}{Text-to-LoRA (SFT) \Sarch}             & 76.0 & \textbf{88.7} & 83.8 & \textbf{68.0} & \textbf{71.6} & \greenbox{82.3} & \greenbox{61.0} & {41.2} & {71.6} & {47.3} & {39.0} & {65.9} \\
\multicolumn{2}{l}{Text-to-LoRA (SFT) \March}             & \greenbox{77.2} & \textbf{89.0} & {84.3} & {65.1} & \textbf{76.1} & \greenbox{81.8} & \greenbox{64.0} & \textbf{50.5} & \textbf{73.5} & {45.2} & \textbf{41.3}  & \textbf{67.5} \\
\multicolumn{2}{l}{Text-to-LoRA (SFT) \Larch}             & \greenbox{{77.5}} & \textbf{88.9} & {85.0} & \textbf{66.5} & \textbf{75.5} & \greenbox{82.1} & \greenbox{64.2} & \textbf{51.9} & \textbf{73.9} & {45.8} & {39.2} & \textbf{67.7} \\ \midrule
\multicolumn{2}{l}{\textbf{\ModelName} \Larch}      & \greenbox{\textbf{81.6}} & \greenbox{\textbf{92.7}} & 74.0 & 63.1 & \bf 78.6 & \greenbox{78.0} & \greenbox{57.8} & \greenbox{\textbf{62.3}} & \bf 73.5 & \greenbox{\textbf{77.1}} & \bf 50.6 & \bf 71.6 \\ \midrule
\multicolumn{2}{l}{\textbf{Oracle}}                  &      &      &      &      &      &      &      &      &      &      &      &      \\
\multicolumn{2}{l}{Task-specific LoRAs}           & 76.6 & 89.9 & 89.4 & 92.6 & 85.0 & 69.9 & 51.1 & 52.1 & 75.8 & 53.5 & N/A  & N/A  \\ 
\multicolumn{2}{l}{Task-specific SFTs}           & 76.6 & 89.9 & 89.4 & 92.6 & 85.0 & 69.9 & 51.1 & 52.1 & 75.8 & 53.5 & N/A  & N/A  \\ 
\bottomrule
\end{tabular}%
}
\end{table*}

\subsection{Layer Dependency of Beta}
\label{appendix:layer_beta}

\begin{figure}[!t]
\centering
  \includegraphics[width=0.9\linewidth]{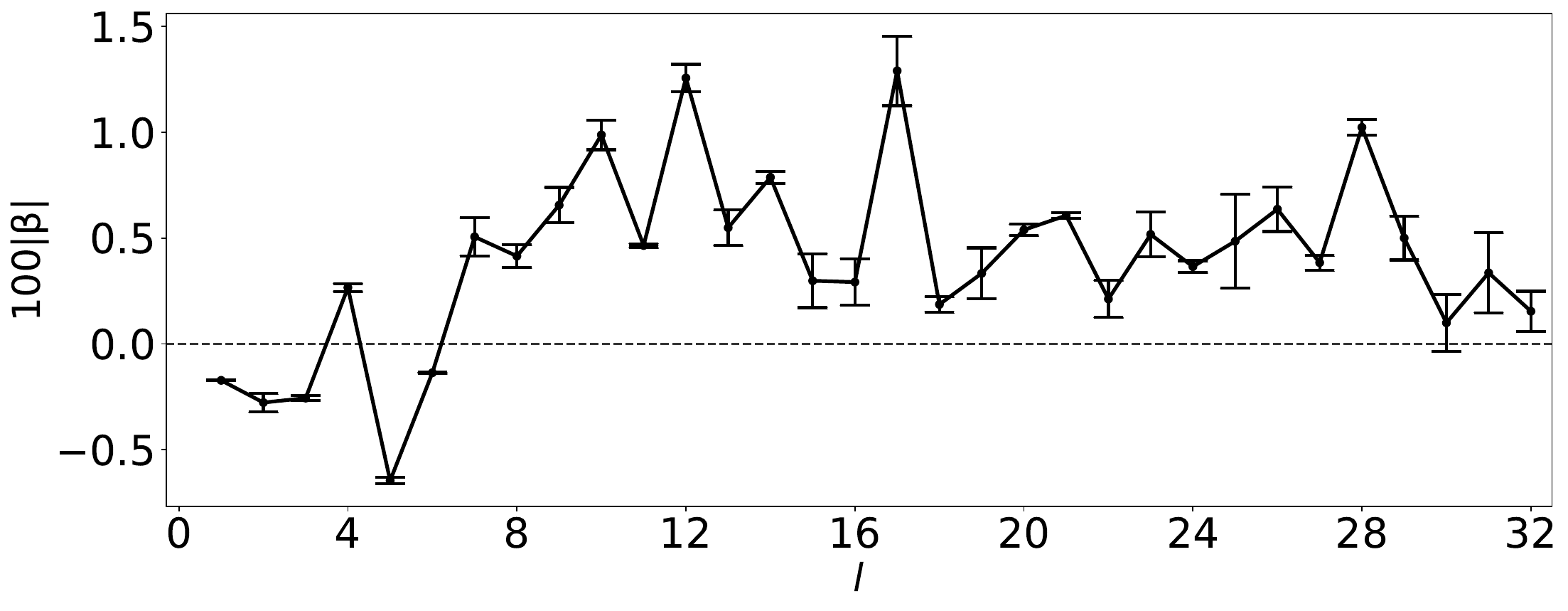} 
  \caption {Averaged $\beta$  with respect to different layers. $\beta=0$ corresponds to the standard SiLU (Swish$_1$). As $\beta$ increases, the nonlinearity becomes stronger.}
  \label{fig:beta_layers}
\end{figure}

While Section \ref{sec:dist_beta} visualizes the global picture of the $\beta$ distribution, to investigate the $\beta$ difference with respect to different LLM layers, we also compute the layer-wise averaged $\beta$ on GYAFC, as shown by Figure \ref{fig:beta_layers}. As the layer index becomes larger, one can find that the absolute value of $\beta$ becomes higher, indicating the nonlinearity of Swish$_{\beta}$ tends to be stronger.

\subsection{All Layer-wise t-SNE Plots}
\label{sec:tsne_all_layers}

Due to the page limit, we only include the t-SNE plot of AdaptSum on a representative layer (the 5-th layer) in Figure \ref{fig:tsne_beta_adaptsum}. Here we exhibit all the layer-wise t-SNE plots, including the datasets of AdaptSum, SST, and GYAFC.

\begin{figure*}[htbp]
    \begin{minipage}{\textwidth}
        \centering
        \hspace{-5pt}
        \includegraphics[width=0.8\textwidth]{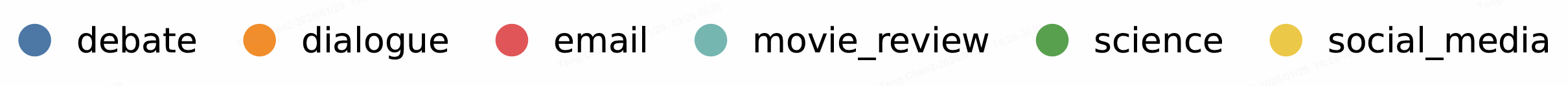}
    \end{minipage}
    \vspace{0.001em}
    \centering

    \begin{subfigure}{0.24\linewidth}
        \includegraphics[width=\linewidth]{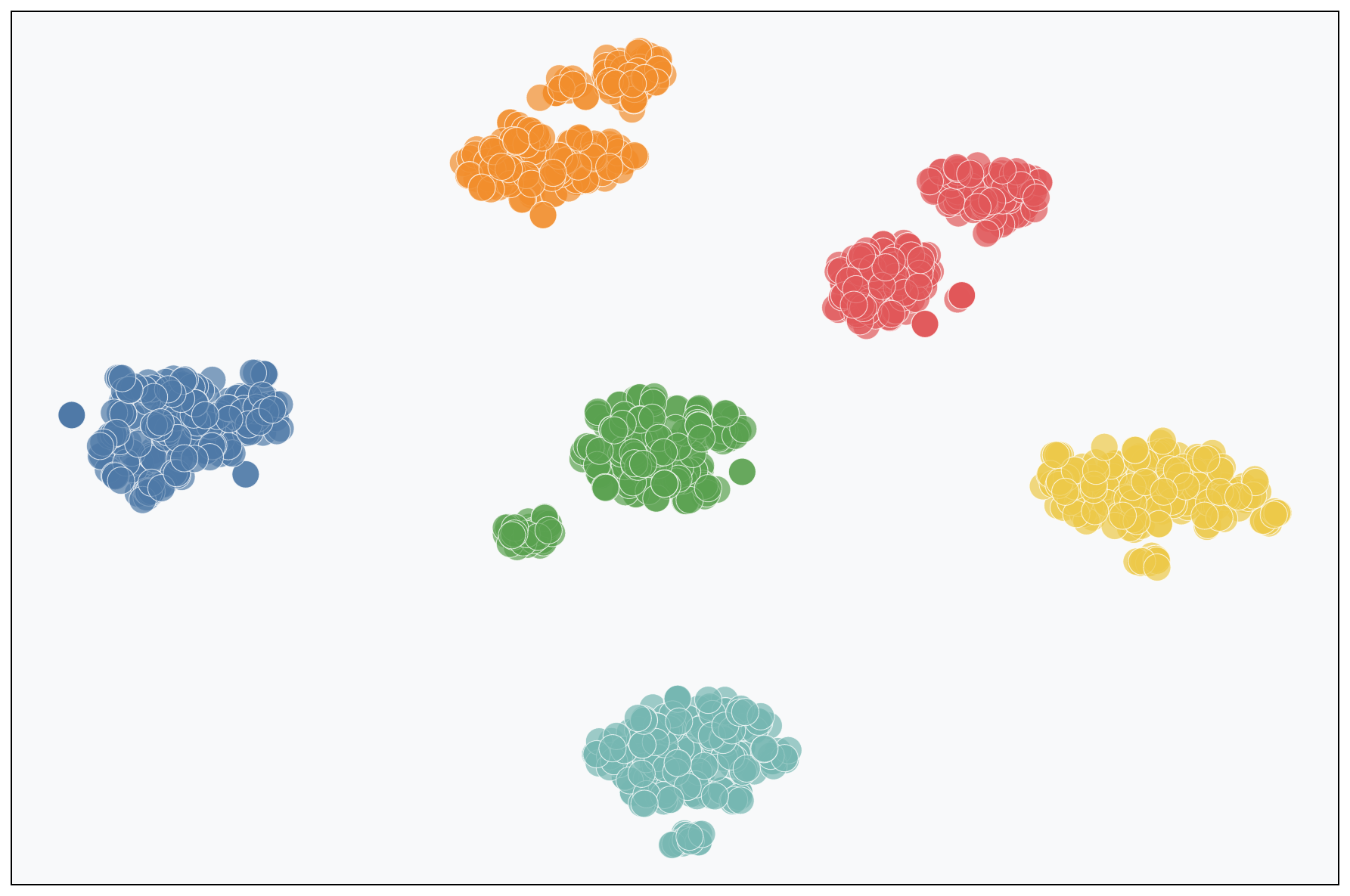}
        \caption*{k=1}
    \end{subfigure}\hfill
    \begin{subfigure}{0.24\linewidth}
        \includegraphics[width=\linewidth]{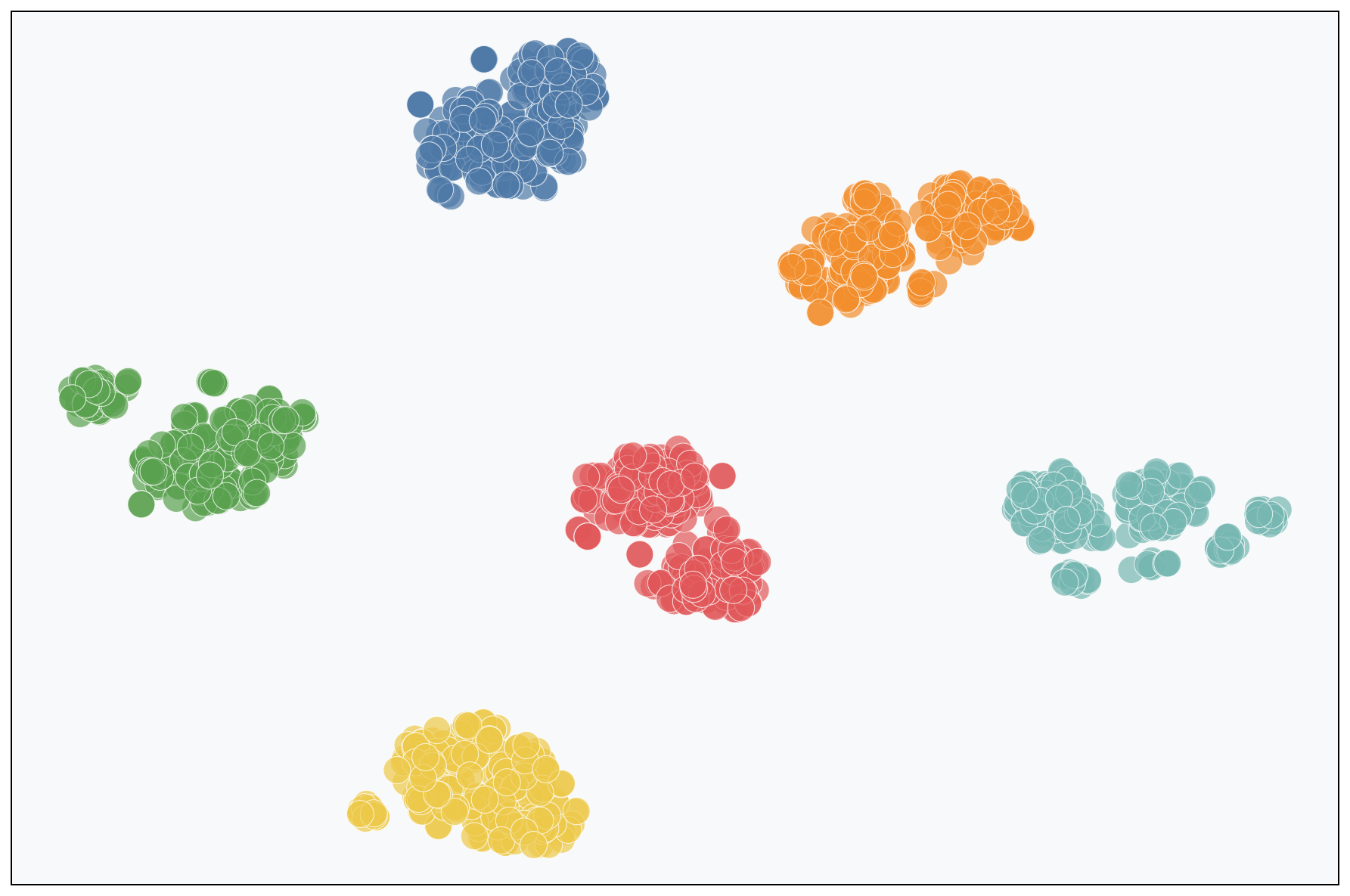}
        \caption*{k=2}
    \end{subfigure}\hfill
    \begin{subfigure}{0.24\linewidth}
        \includegraphics[width=\linewidth]{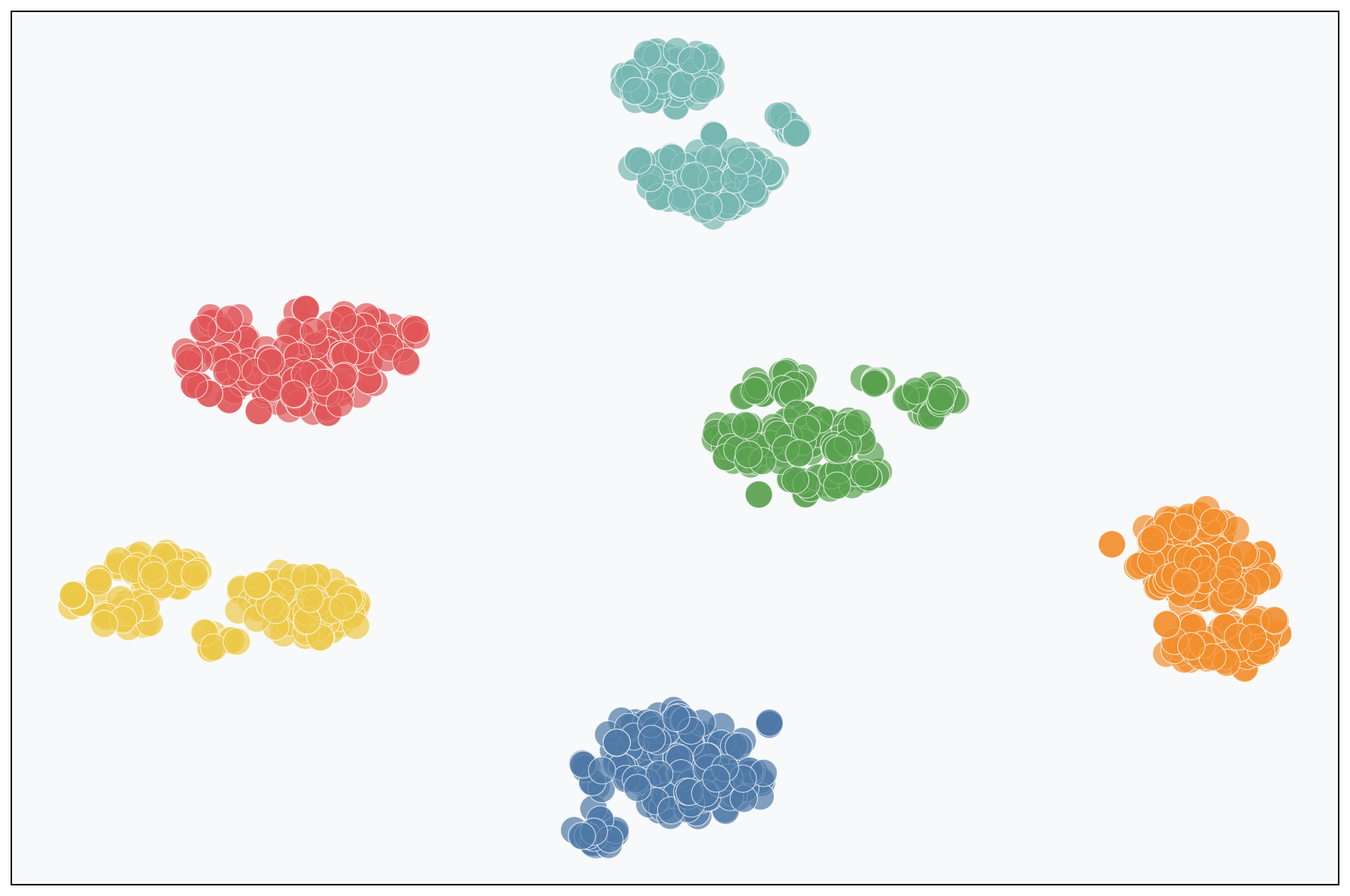}
        \caption*{k=3}
    \end{subfigure}\hfill
    \begin{subfigure}{0.24\linewidth}
        \includegraphics[width=\linewidth]{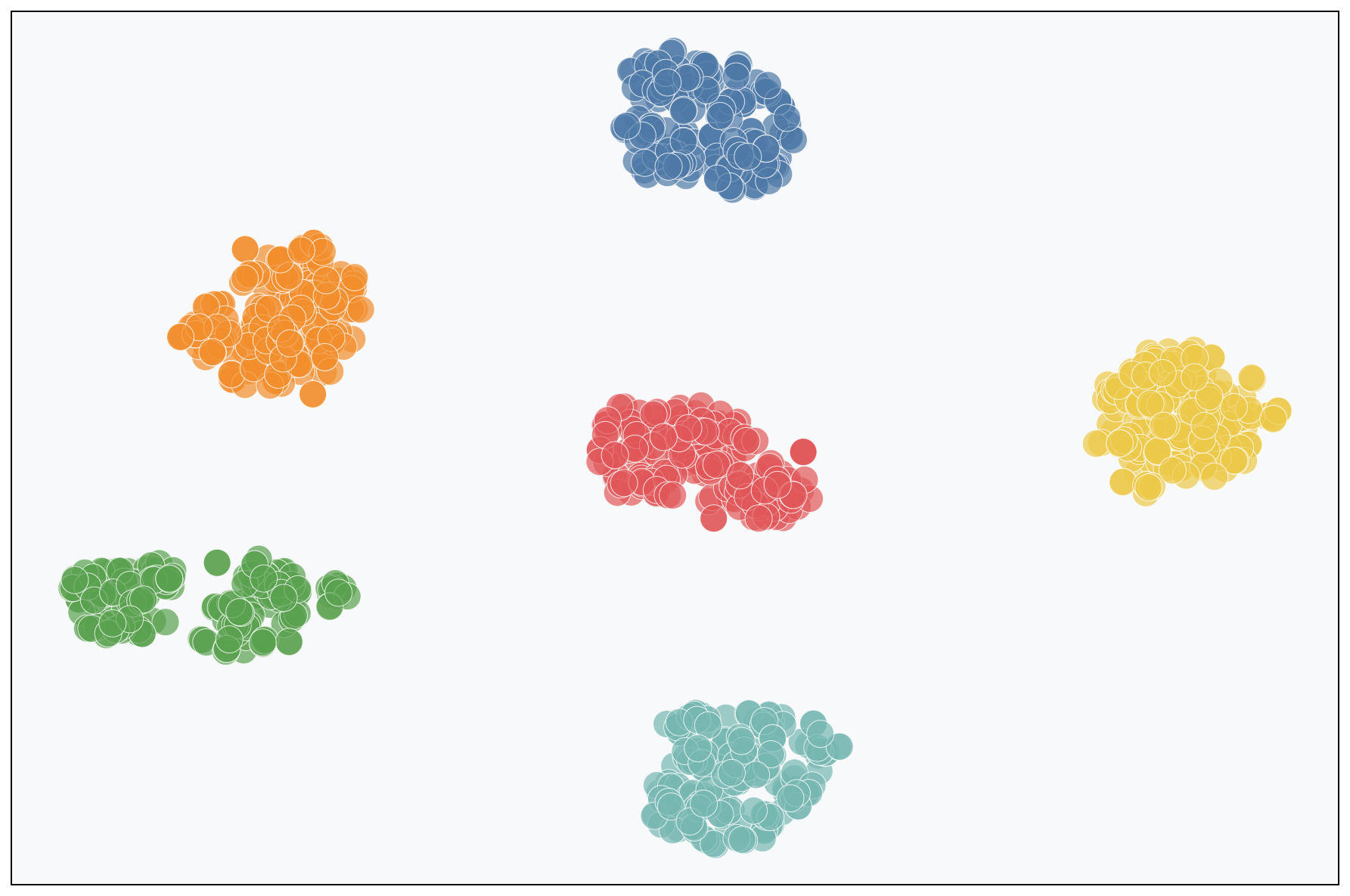}
        \caption*{k=4}
    \end{subfigure}
    
    \vspace{0.2em}
    
    \begin{subfigure}{0.24\linewidth}
        \includegraphics[width=\linewidth]{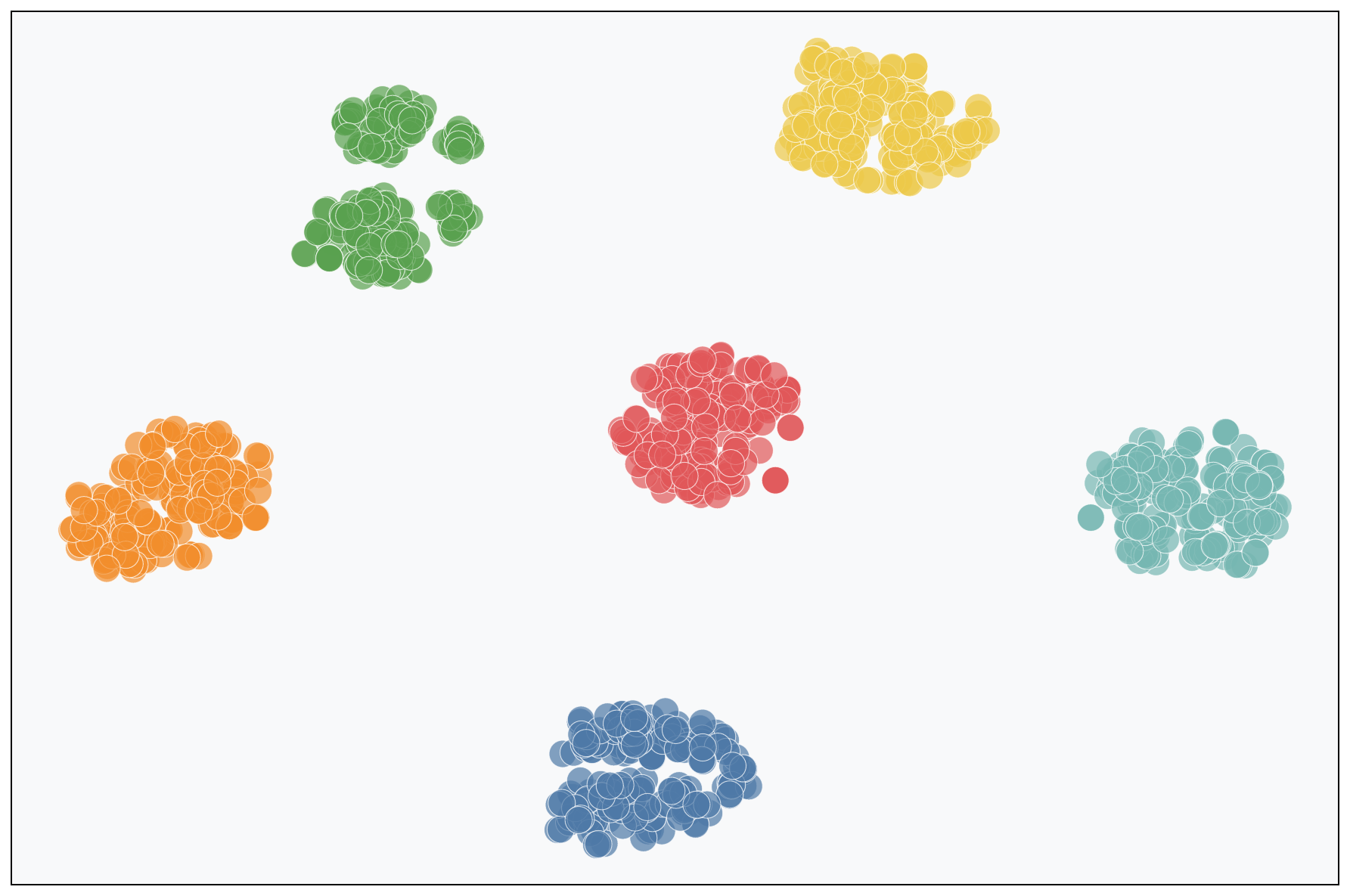}
        \caption*{k=5}
    \end{subfigure}\hfill
    \begin{subfigure}{0.24\linewidth}
        \includegraphics[width=\linewidth]{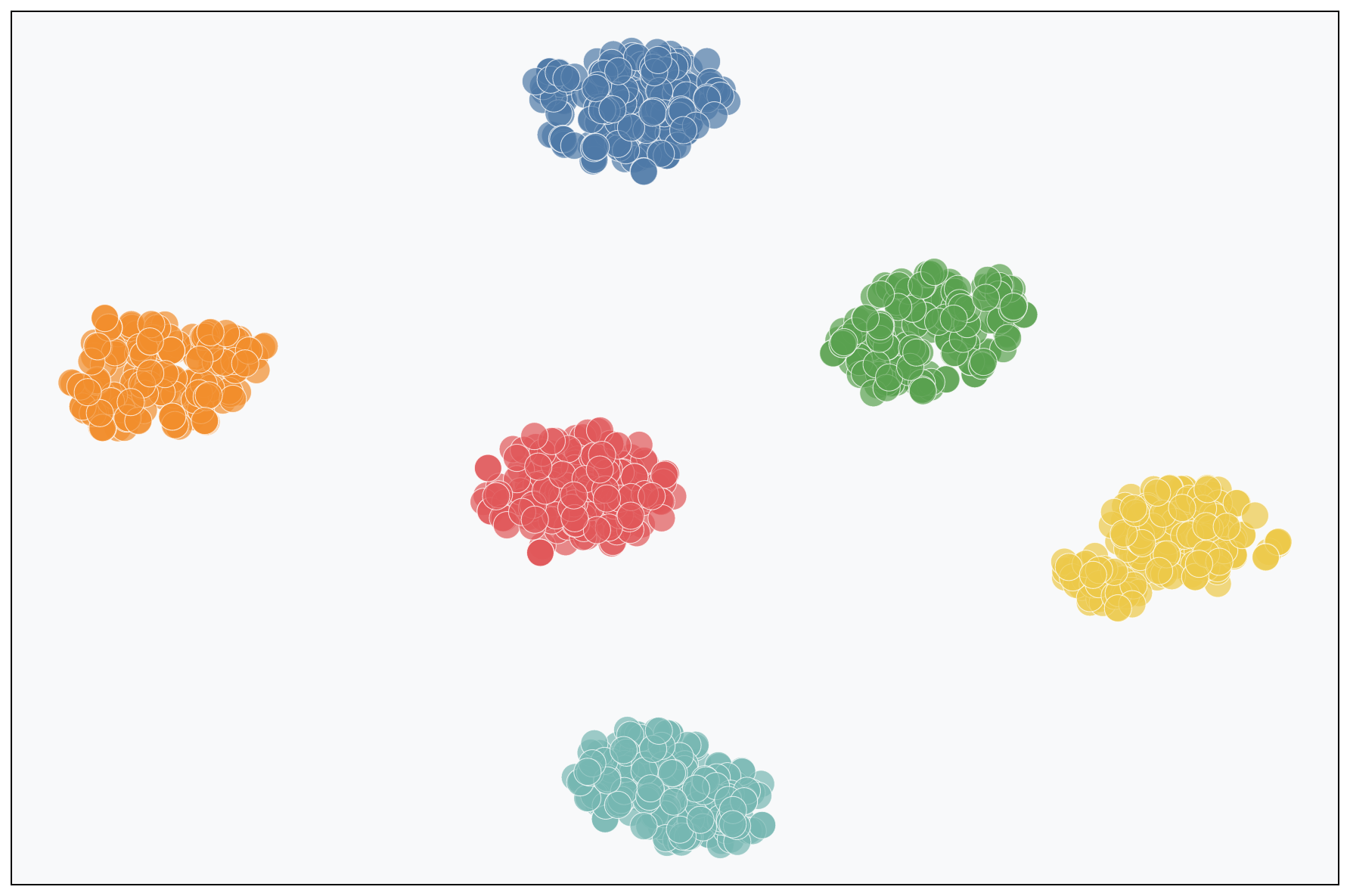}
        \caption*{k=6}
    \end{subfigure}\hfill
    \begin{subfigure}{0.24\linewidth}
        \includegraphics[width=\linewidth]{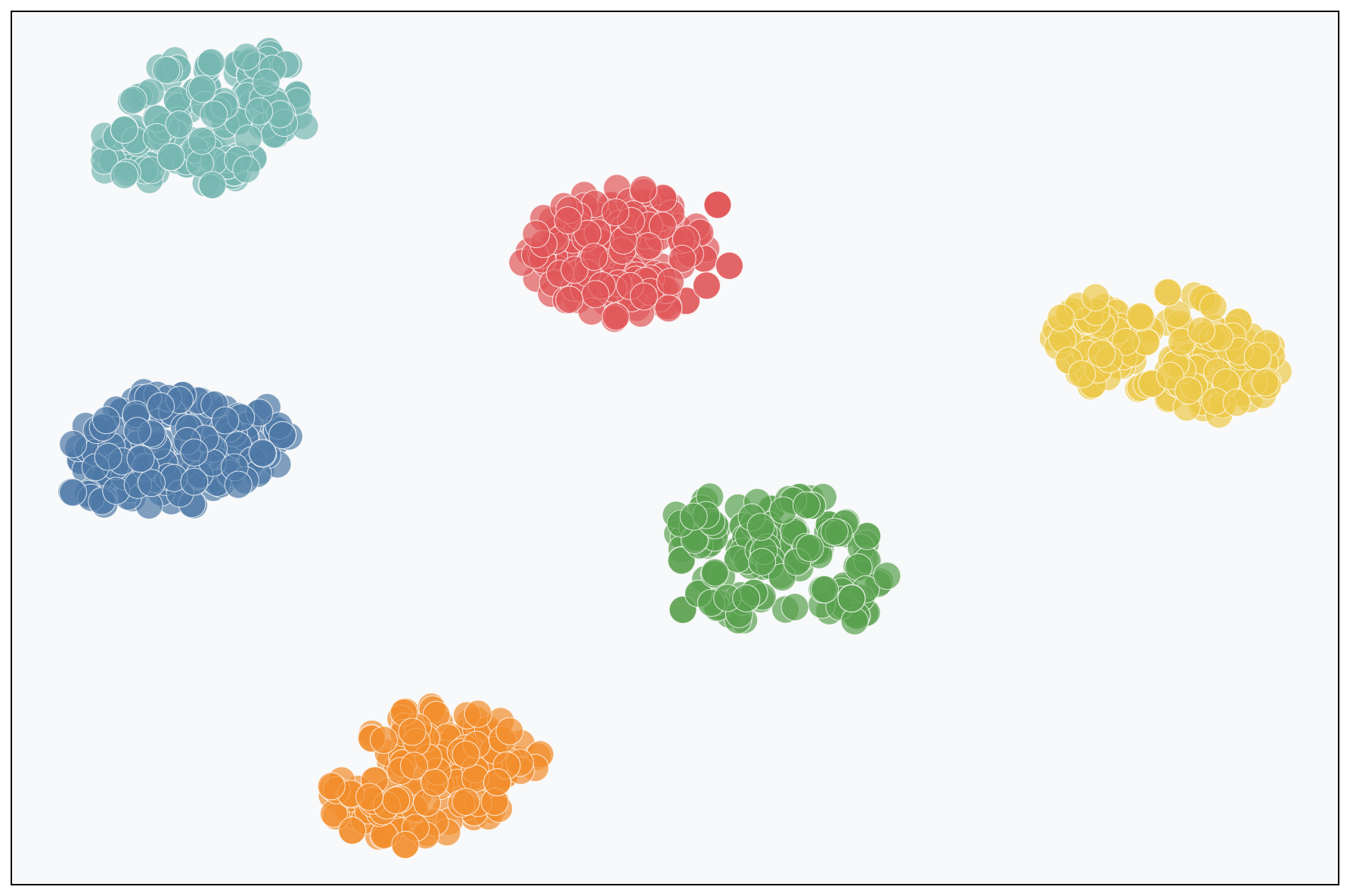}
        \caption*{k=7}
    \end{subfigure}\hfill
    \begin{subfigure}{0.24\linewidth}
        \includegraphics[width=\linewidth]{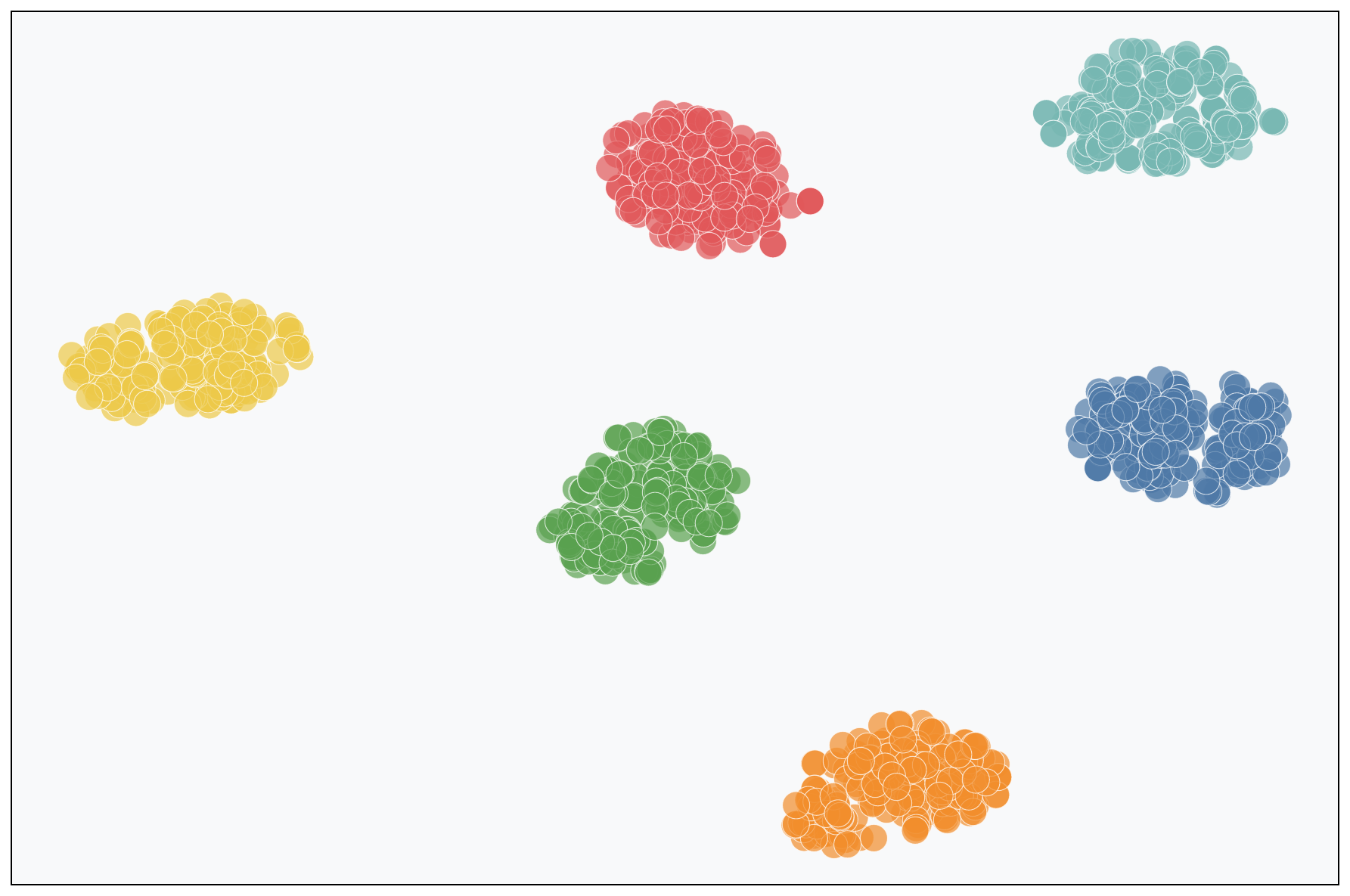}
        \caption*{k=8}
    \end{subfigure}
    
    \vspace{0.2em}
    
    \begin{subfigure}{0.24\linewidth}
        \includegraphics[width=\linewidth]{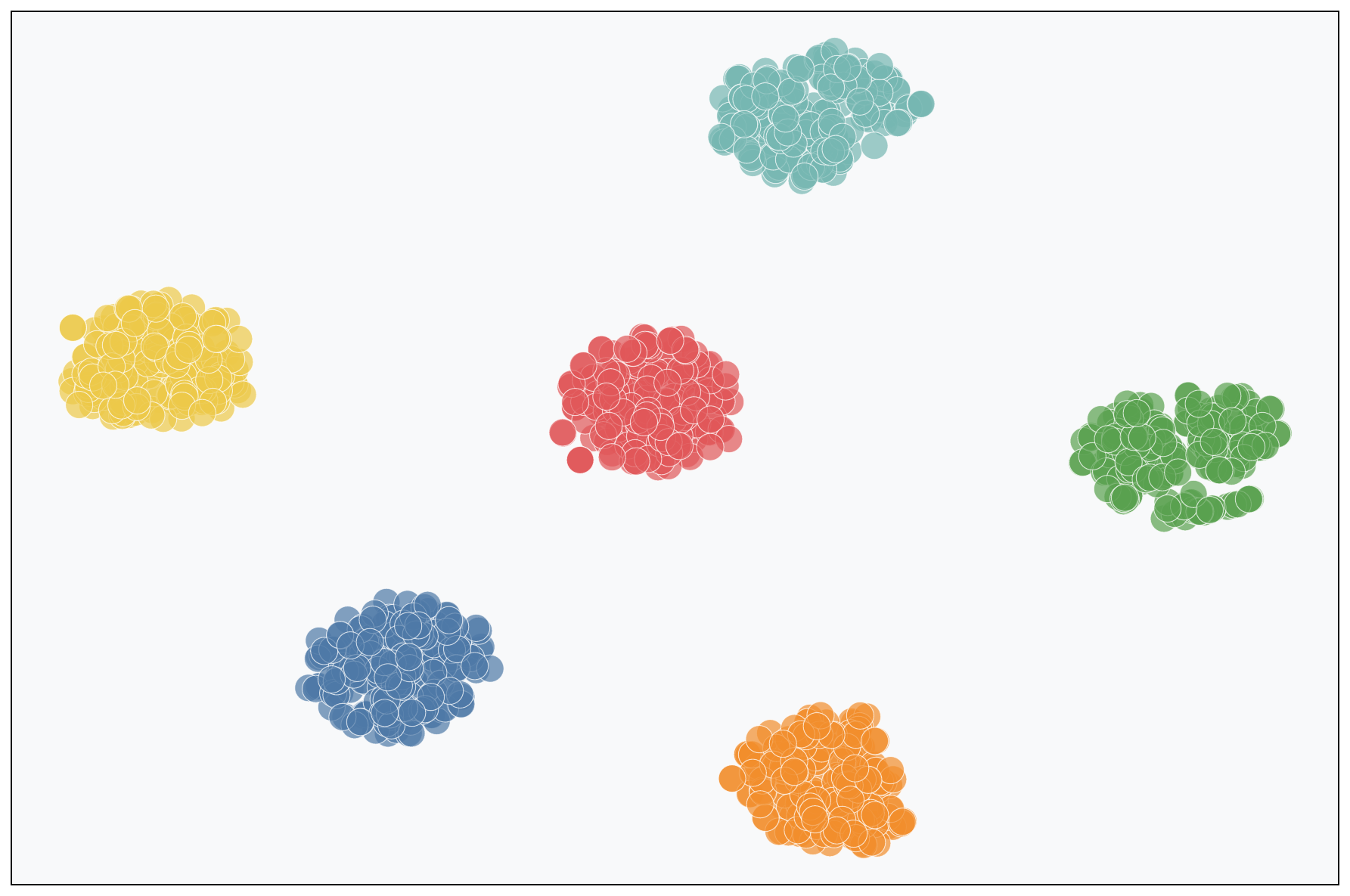}
        \caption*{k=9}
    \end{subfigure}\hfill
    \begin{subfigure}{0.24\linewidth}
        \includegraphics[width=\linewidth]{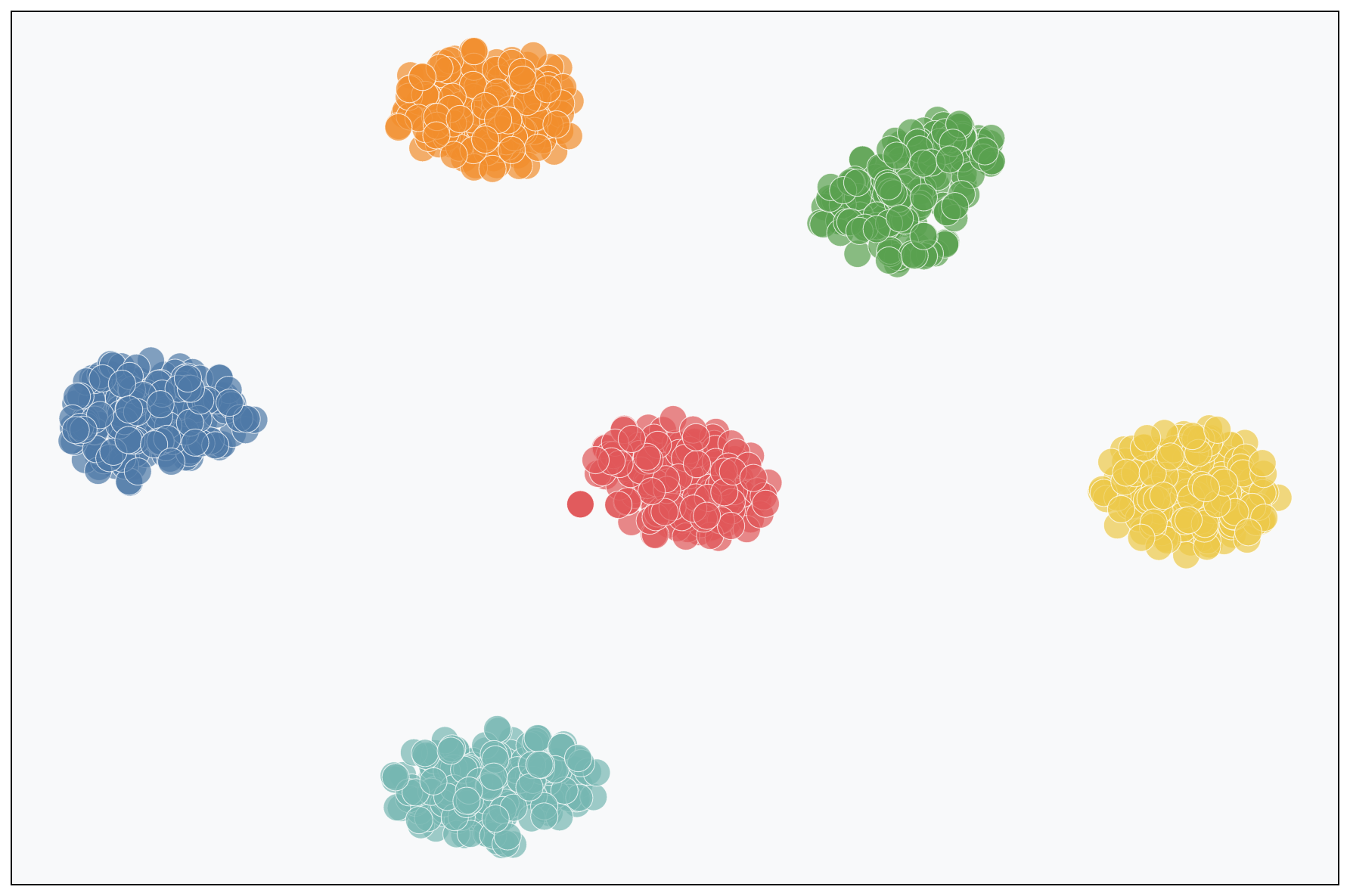}
        \caption*{k=10}
    \end{subfigure}\hfill
    \begin{subfigure}{0.24\linewidth}
        \includegraphics[width=\linewidth]{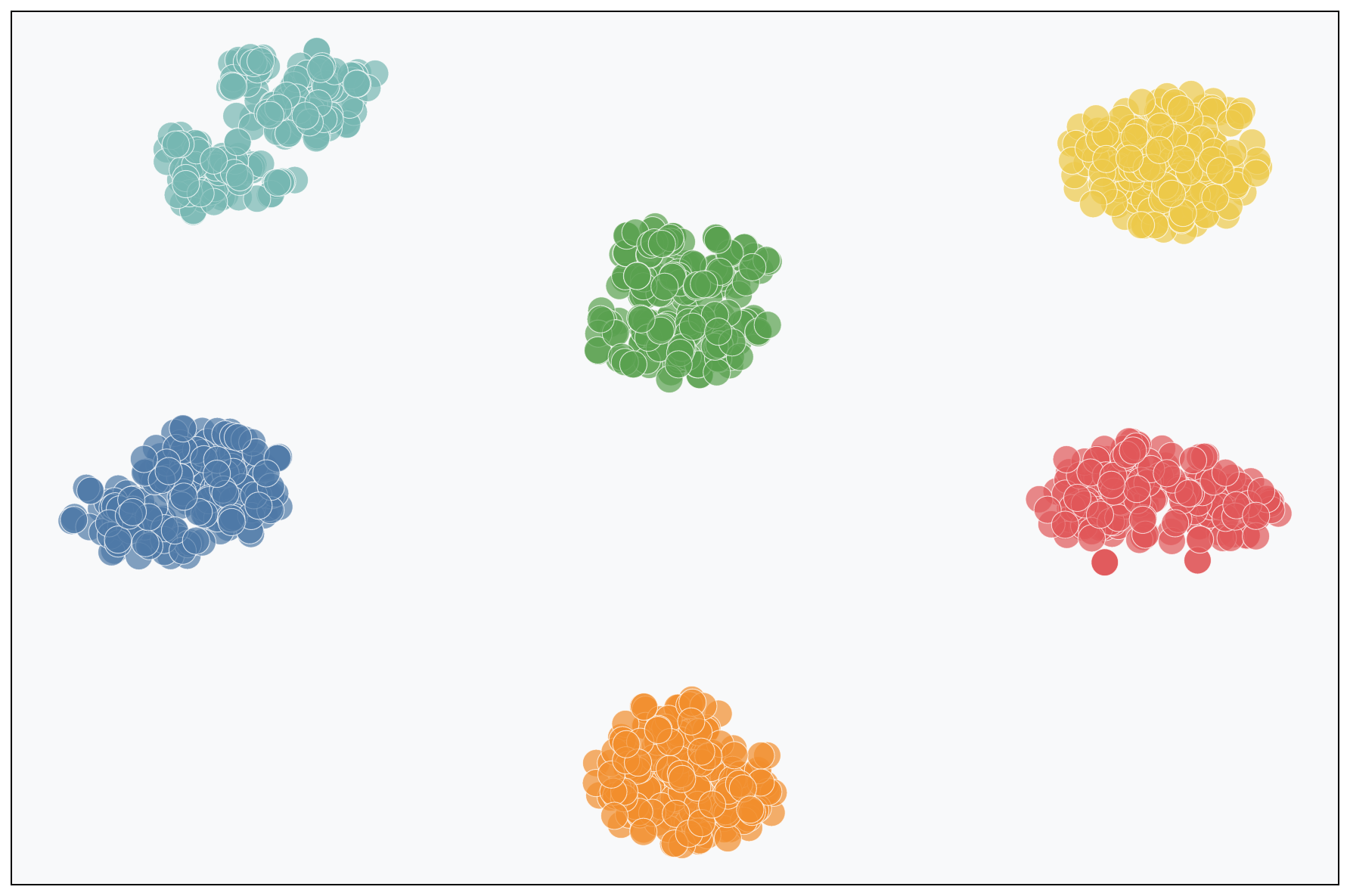}
        \caption*{k=11}
    \end{subfigure}\hfill
    \begin{subfigure}{0.24\linewidth}
        \includegraphics[width=\linewidth]{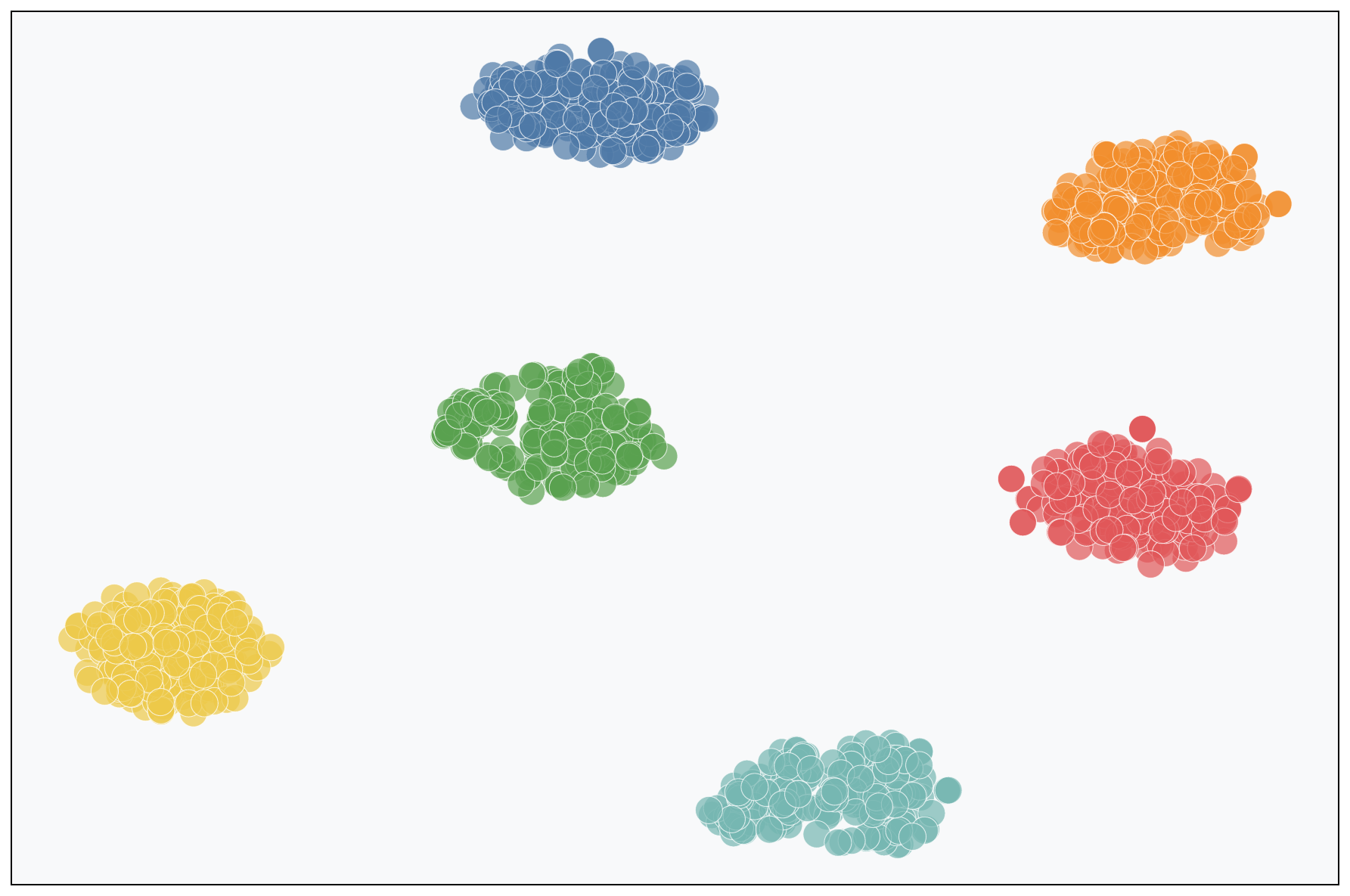}
        \caption*{k=12}
    \end{subfigure}
    
    \vspace{0.2em}
    
    \begin{subfigure}{0.24\linewidth}
        \includegraphics[width=\linewidth]{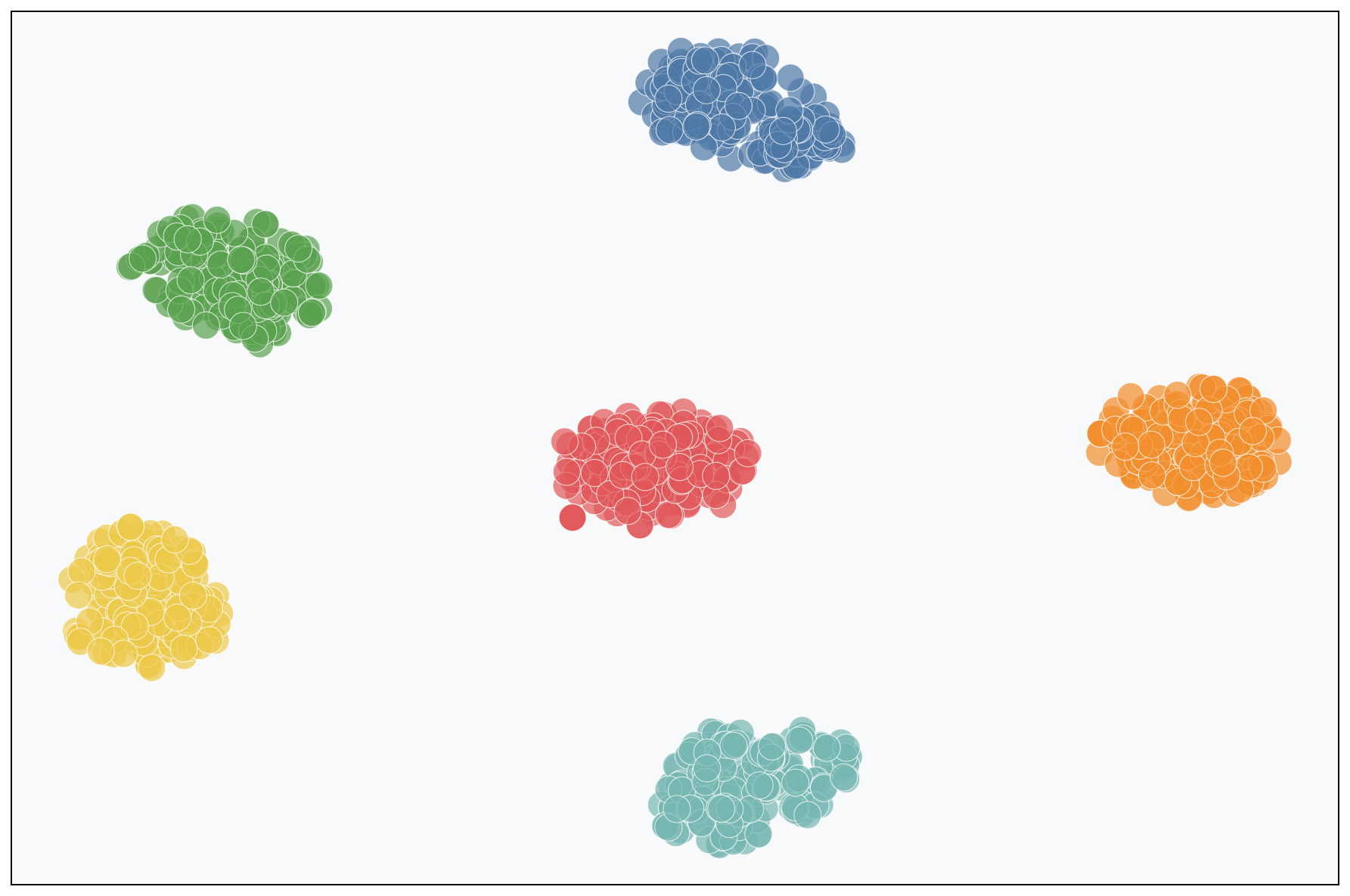}
        \caption*{k=13}
    \end{subfigure}\hfill
    \begin{subfigure}{0.24\linewidth}
        \includegraphics[width=\linewidth]{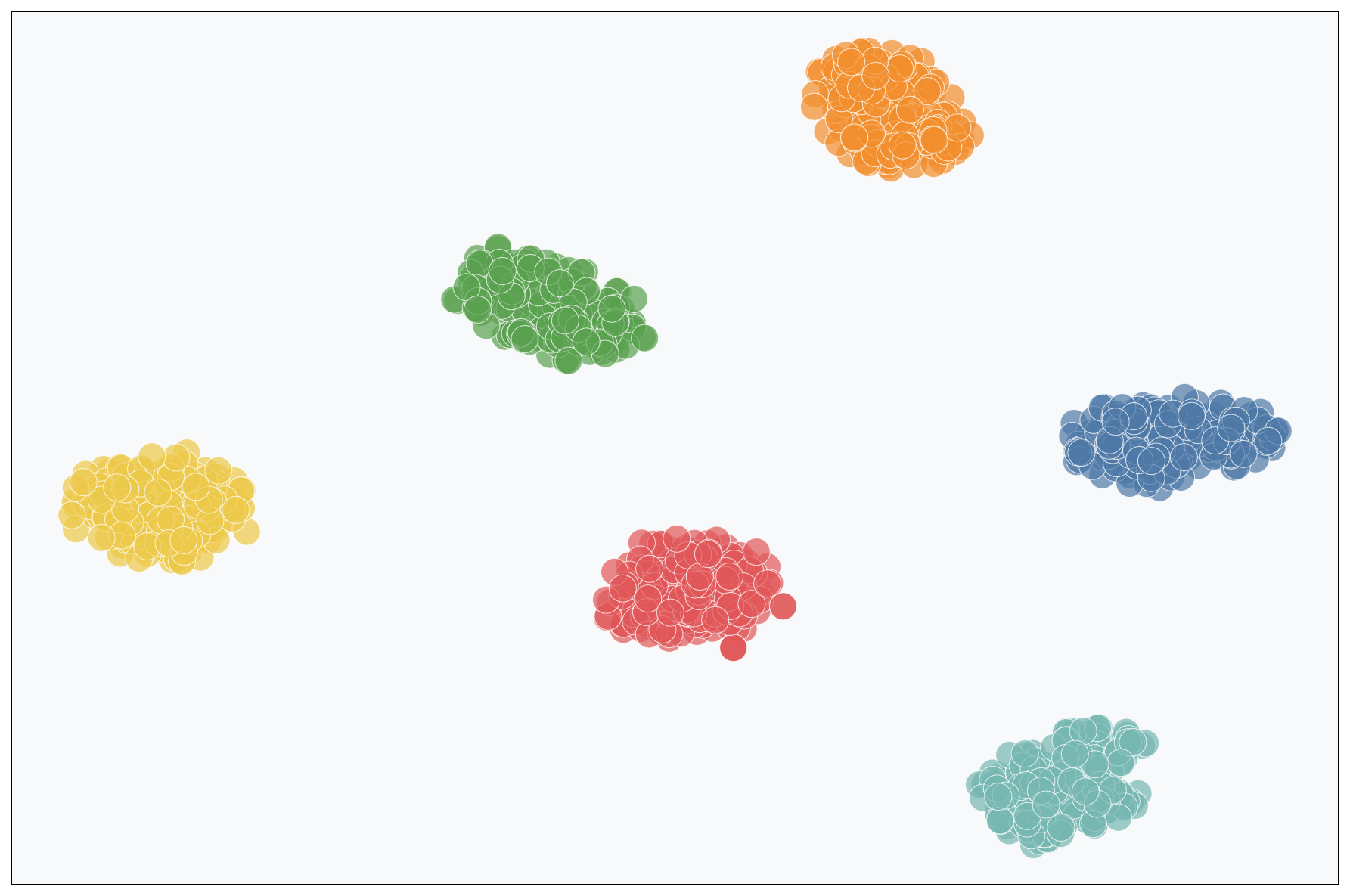}
        \caption*{k=14}
    \end{subfigure}\hfill
    \begin{subfigure}{0.24\linewidth}
        \includegraphics[width=\linewidth]{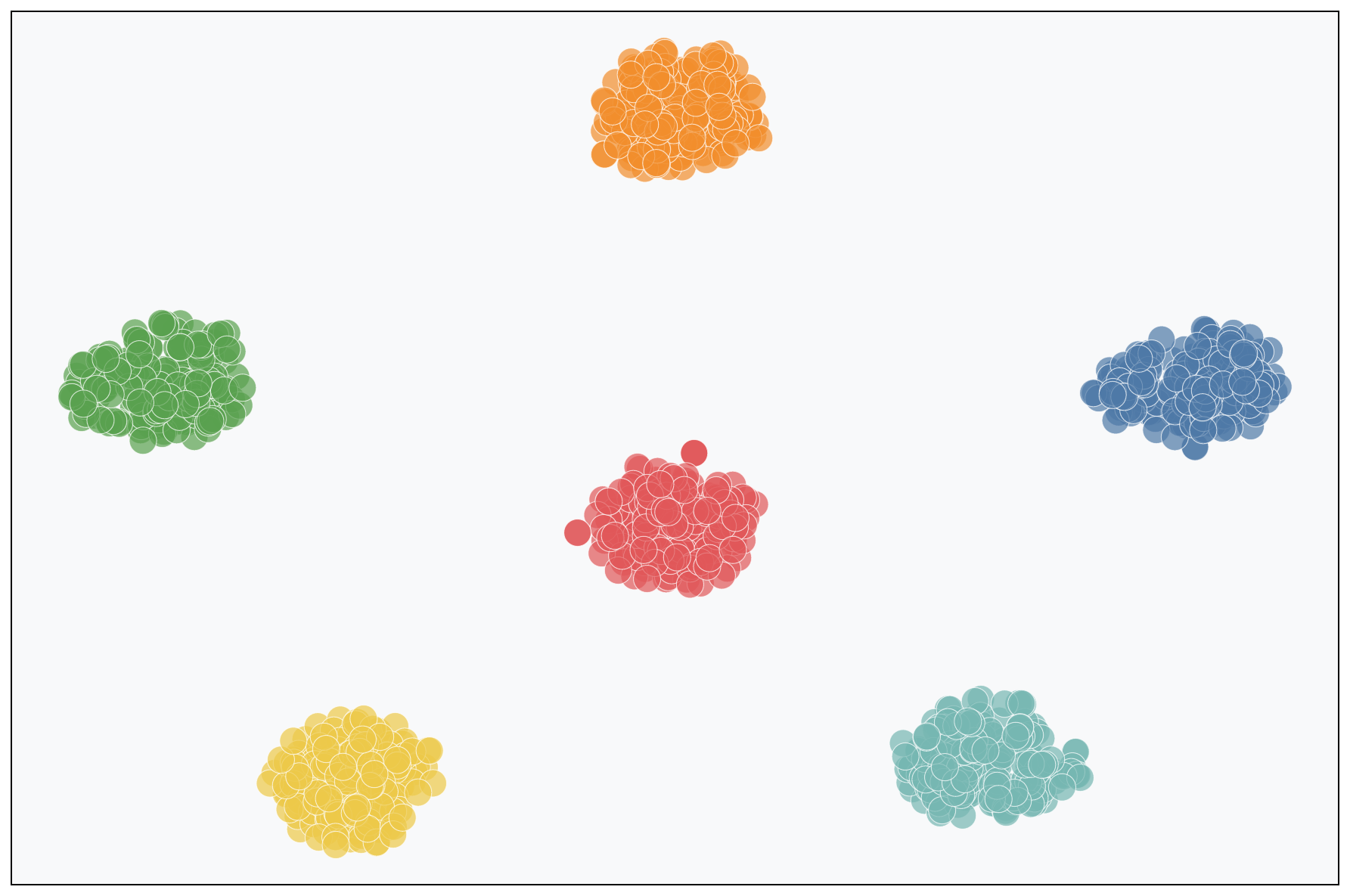}
        \caption*{k=15}
    \end{subfigure}\hfill
    \begin{subfigure}{0.24\linewidth}
        \includegraphics[width=\linewidth]{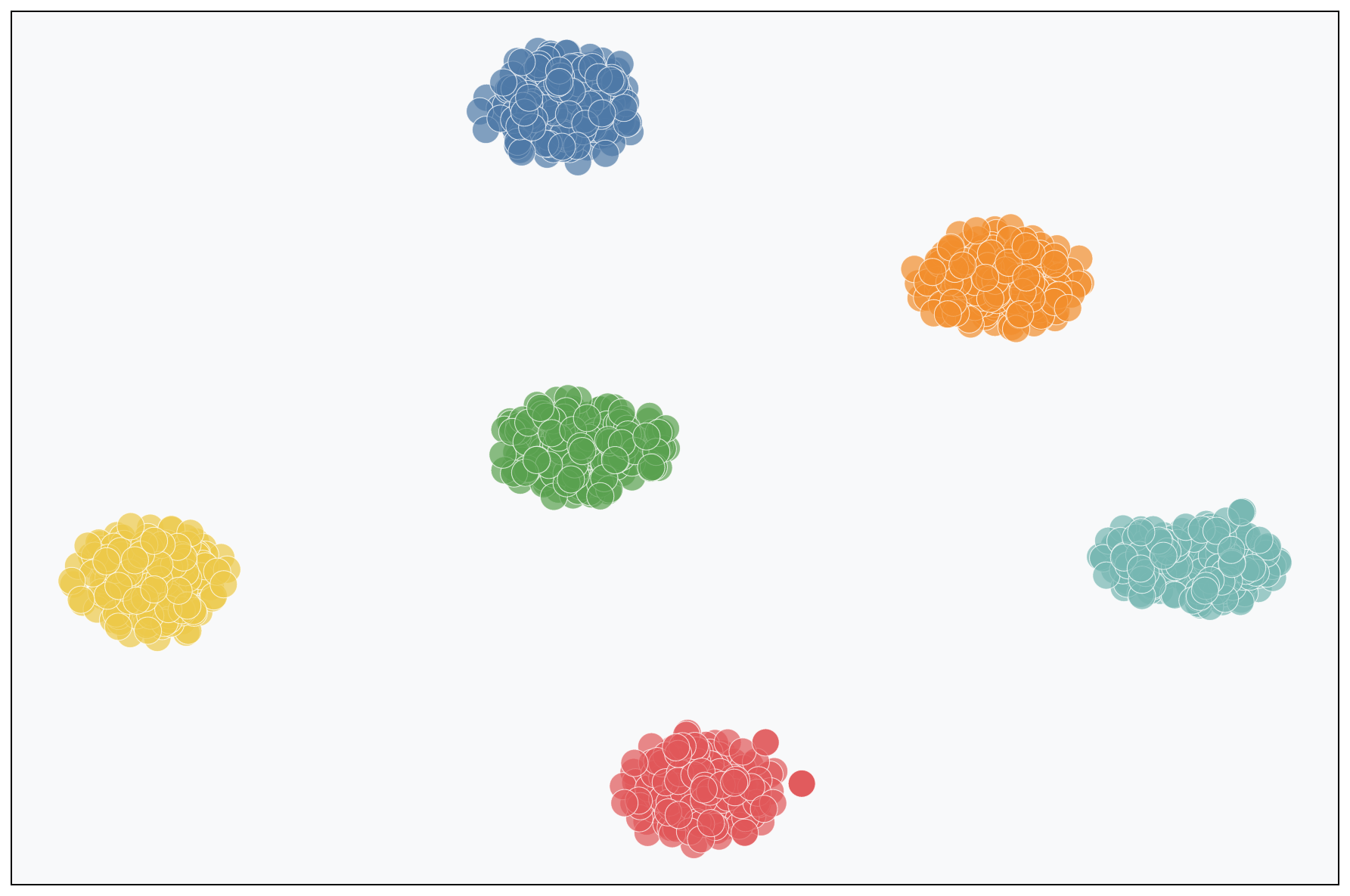}
        \caption*{k=16}
    \end{subfigure}
    
    \caption{t-SNE analysis of averaged $\beta$ on AdaptSum (first 16 layers).}
    \label{fig:tsne_layers1_AdaptSum} 
\end{figure*}

\begin{figure*}[htbp]
    \begin{minipage}{\textwidth}
        \centering
        \hspace{-5pt}
        \includegraphics[width=0.8\textwidth]{tSNE_AdaptSum/adaptsum_legend2.png}
    \end{minipage}
    \vspace{0.001em}

    \centering
    
    \begin{subfigure}{0.24\linewidth}
        \includegraphics[width=\linewidth]{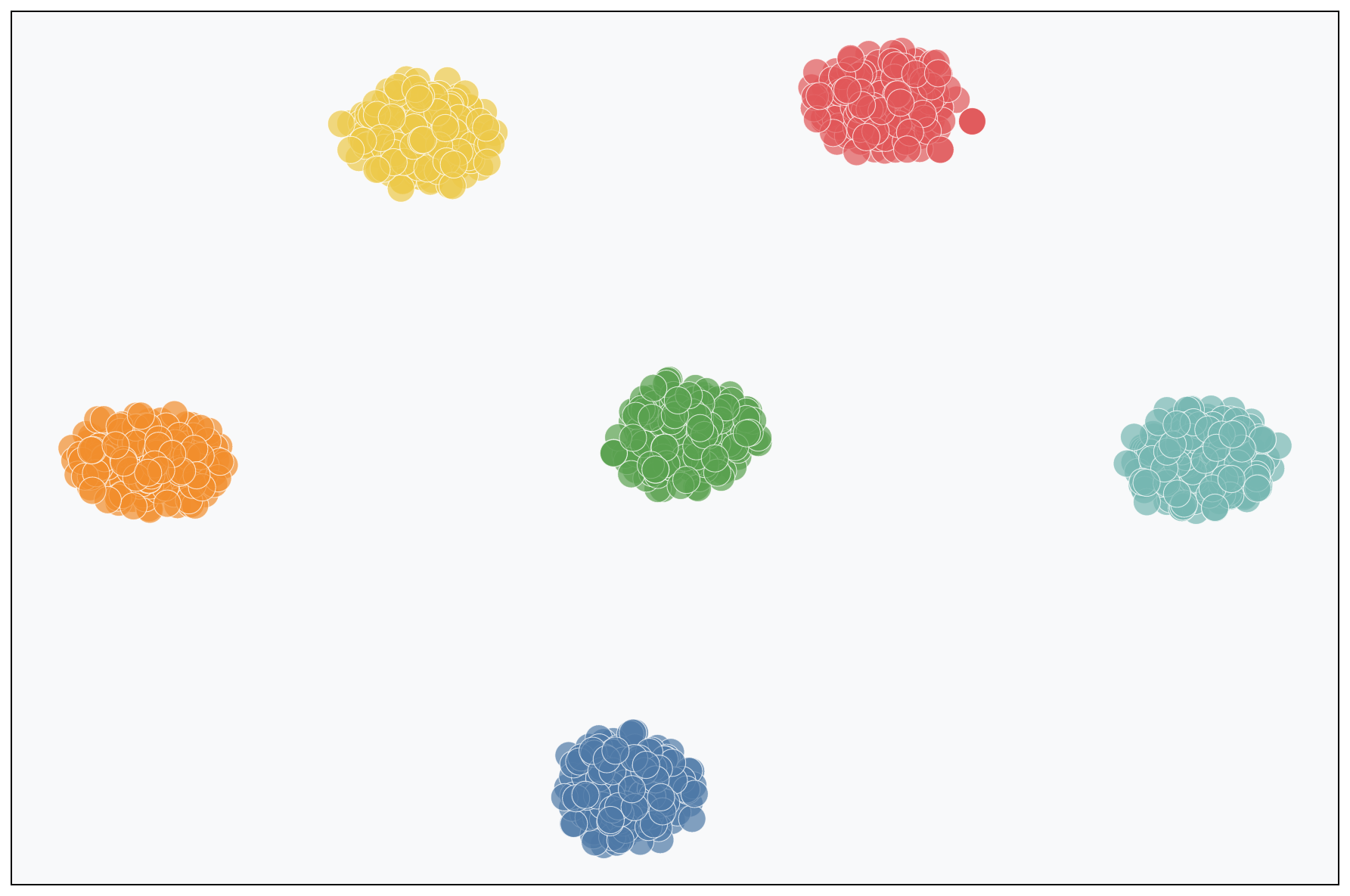}
        \caption*{k=17}
    \end{subfigure}\hfill
    \begin{subfigure}{0.24\linewidth}
        \includegraphics[width=\linewidth]{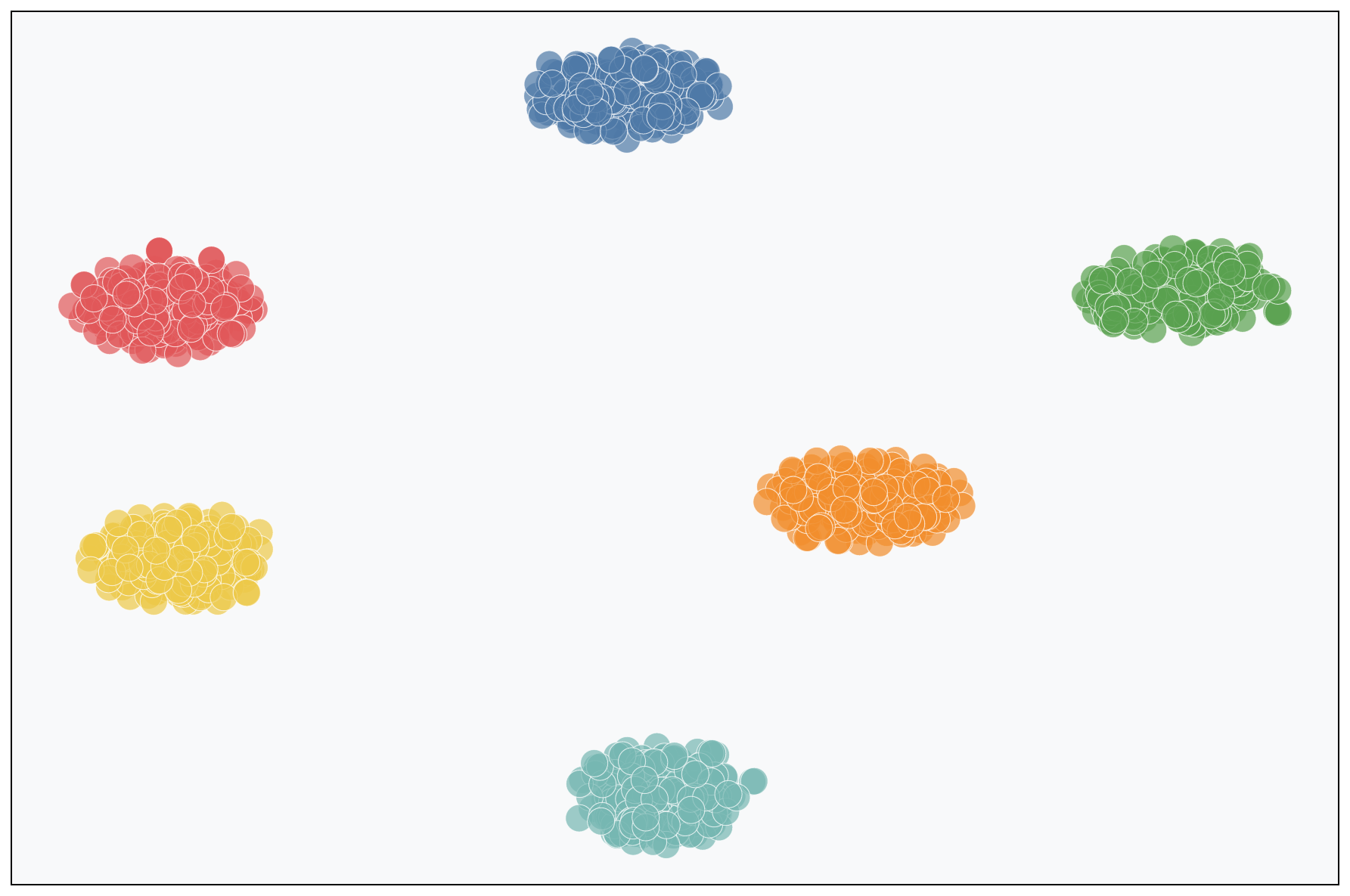}
        \caption*{k=18}
    \end{subfigure}\hfill
    \begin{subfigure}{0.24\linewidth}
        \includegraphics[width=\linewidth]{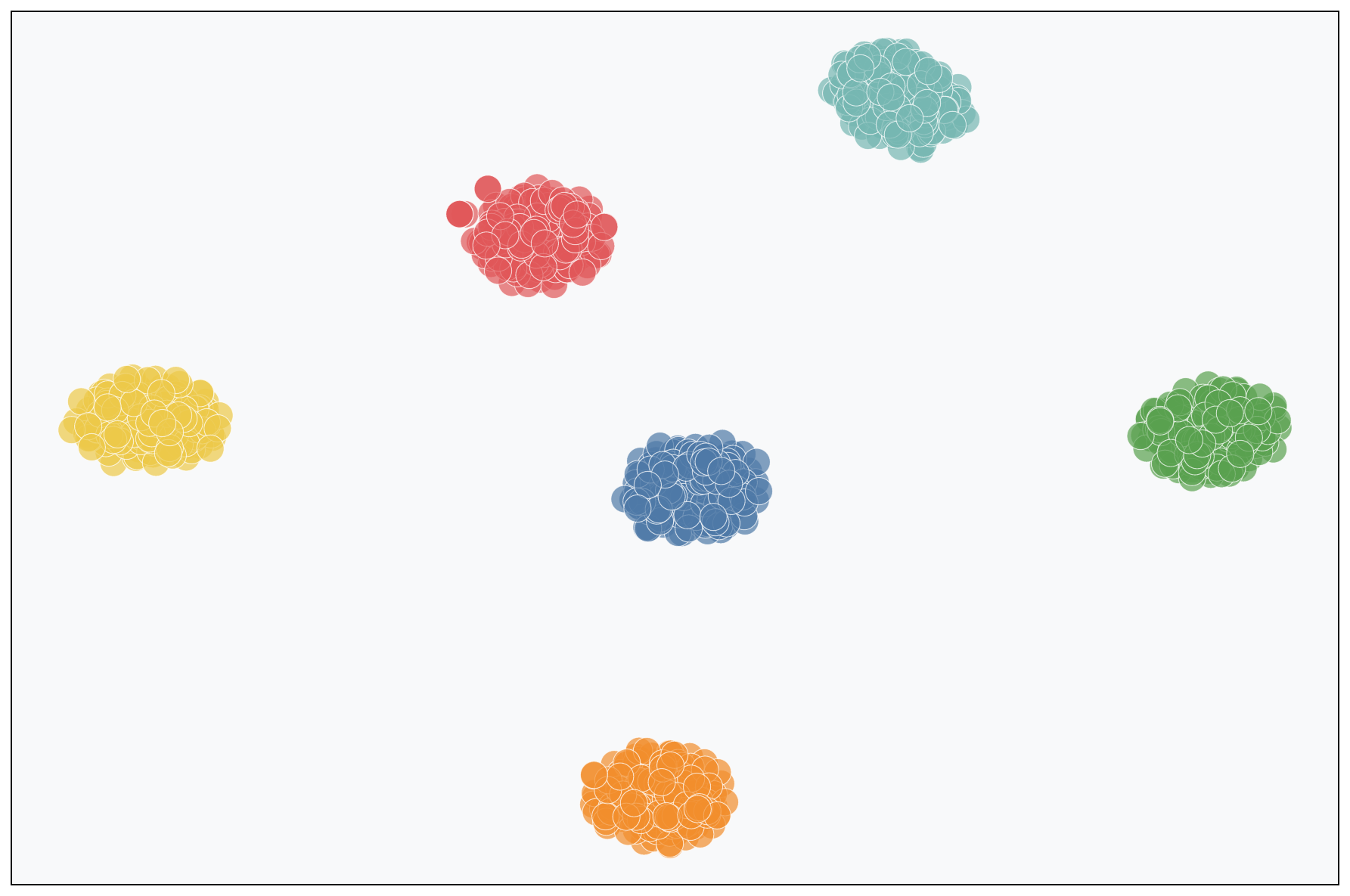}
        \caption*{k=19}
    \end{subfigure}\hfill
    \begin{subfigure}{0.24\linewidth}
        \includegraphics[width=\linewidth]{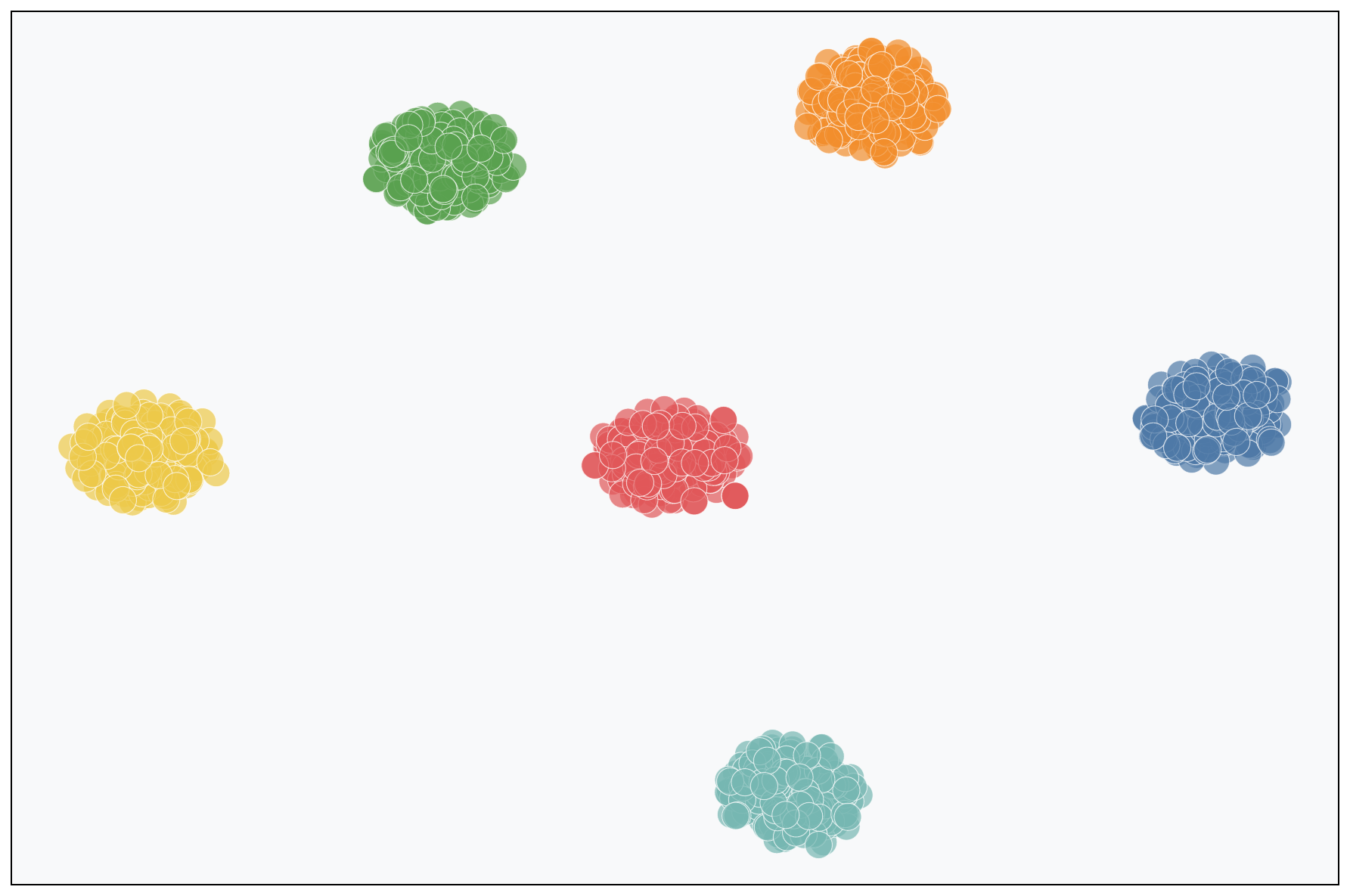}
        \caption*{k=20}
    \end{subfigure}
    
    \vspace{0.2em}
    
    \begin{subfigure}{0.24\linewidth}
        \includegraphics[width=\linewidth]{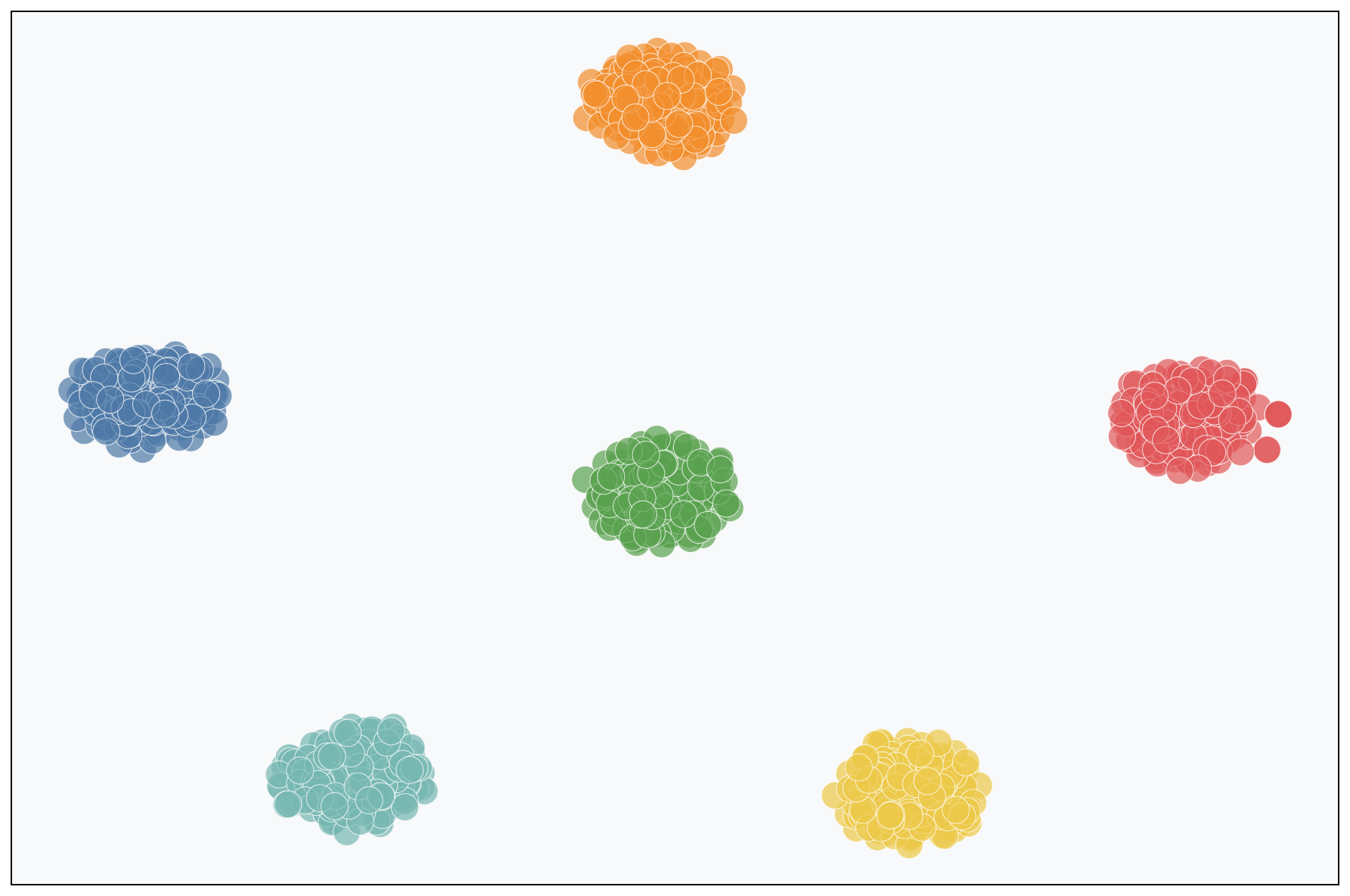}
        \caption*{k=21}
    \end{subfigure}\hfill
    \begin{subfigure}{0.24\linewidth}
        \includegraphics[width=\linewidth]{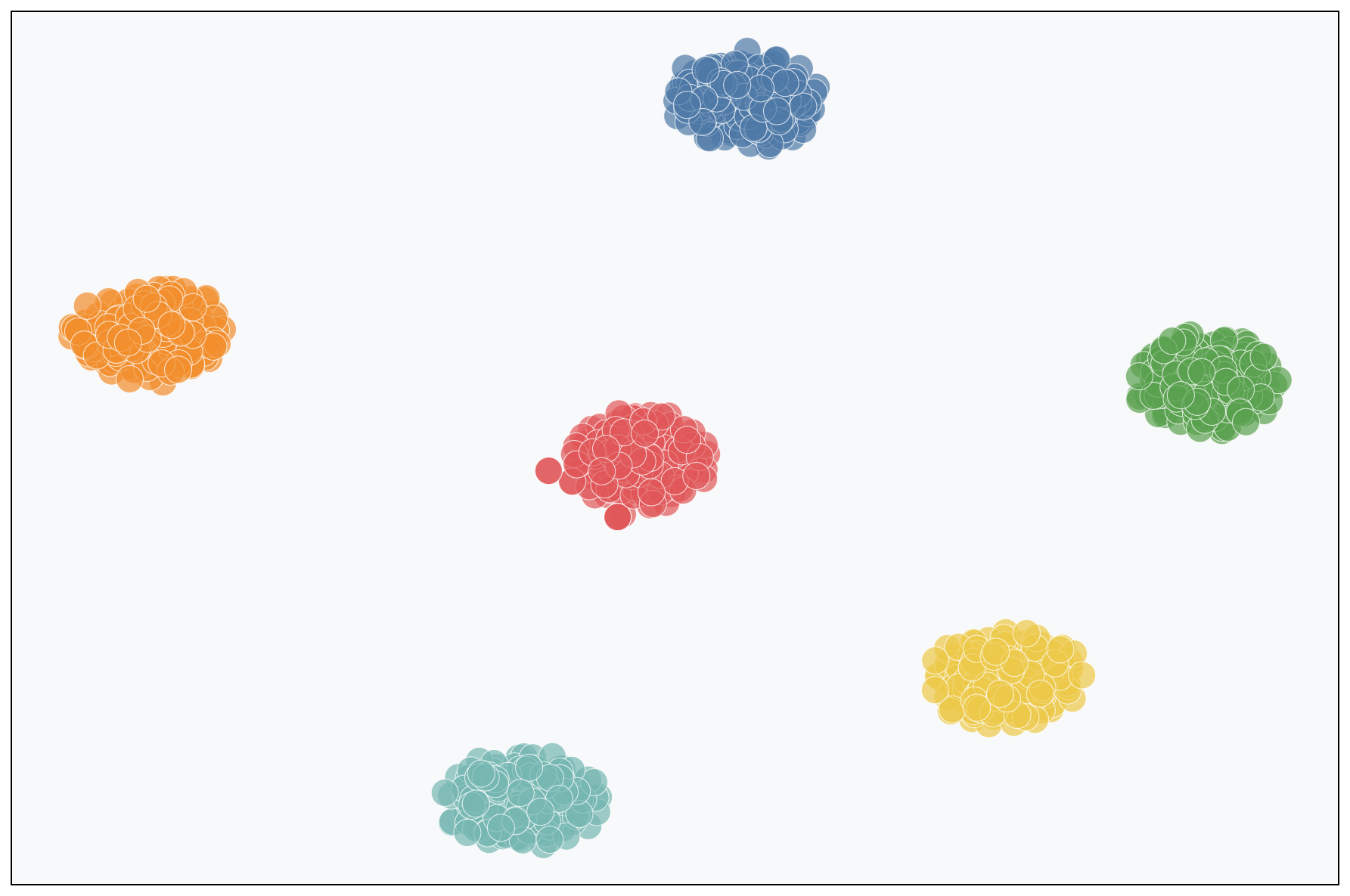}
        \caption*{k=22}
    \end{subfigure}\hfill
    \begin{subfigure}{0.24\linewidth}
        \includegraphics[width=\linewidth]{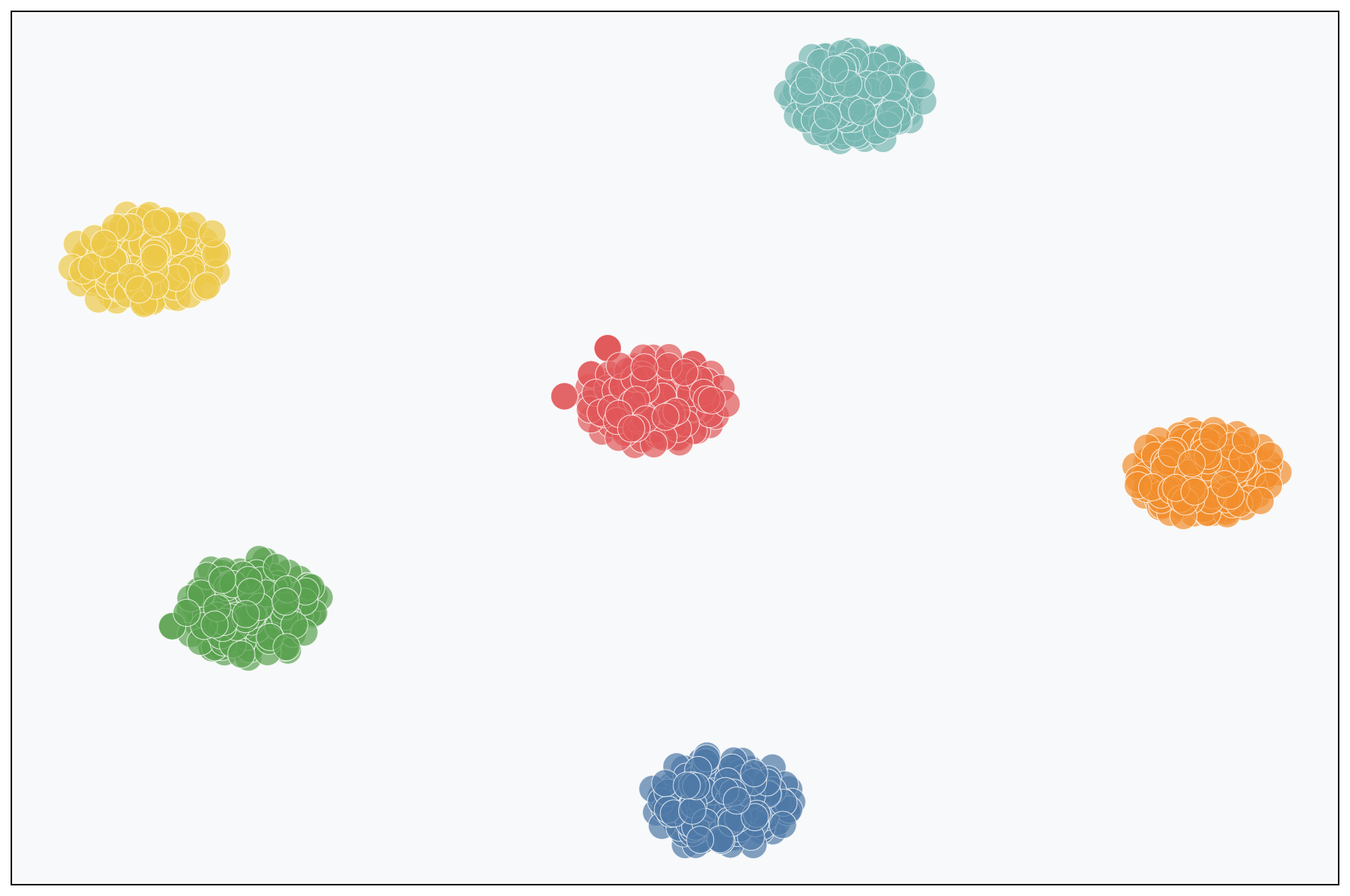}
        \caption*{k=23}
    \end{subfigure}\hfill
    \begin{subfigure}{0.24\linewidth}
        \includegraphics[width=\linewidth]{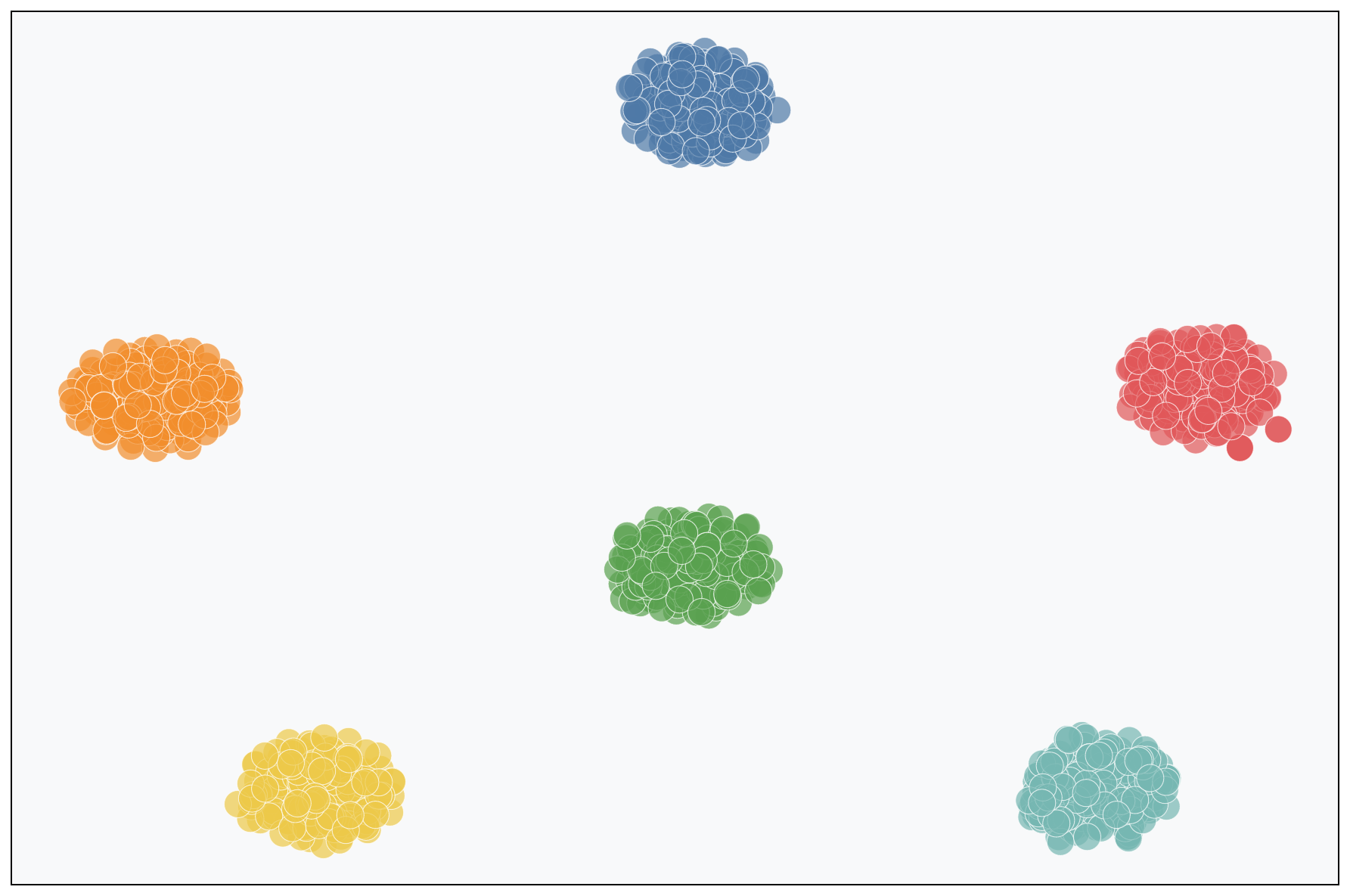}
        \caption*{k=24}
    \end{subfigure}

    \vspace{0.2em}
    
    \begin{subfigure}{0.24\linewidth}
        \includegraphics[width=\linewidth]{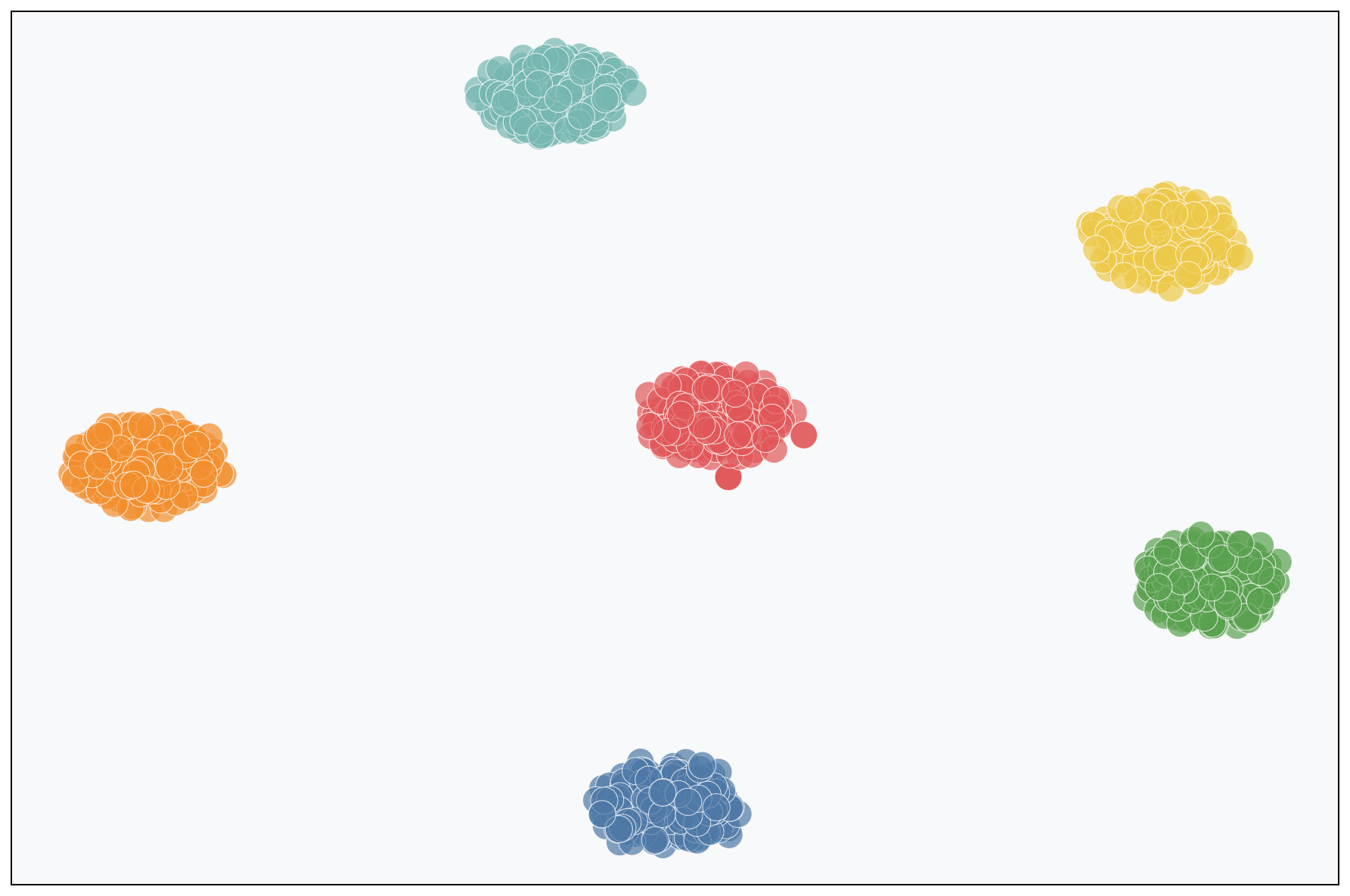}
        \caption*{k=25}
    \end{subfigure}\hfill
    \begin{subfigure}{0.24\linewidth}
        \includegraphics[width=\linewidth]{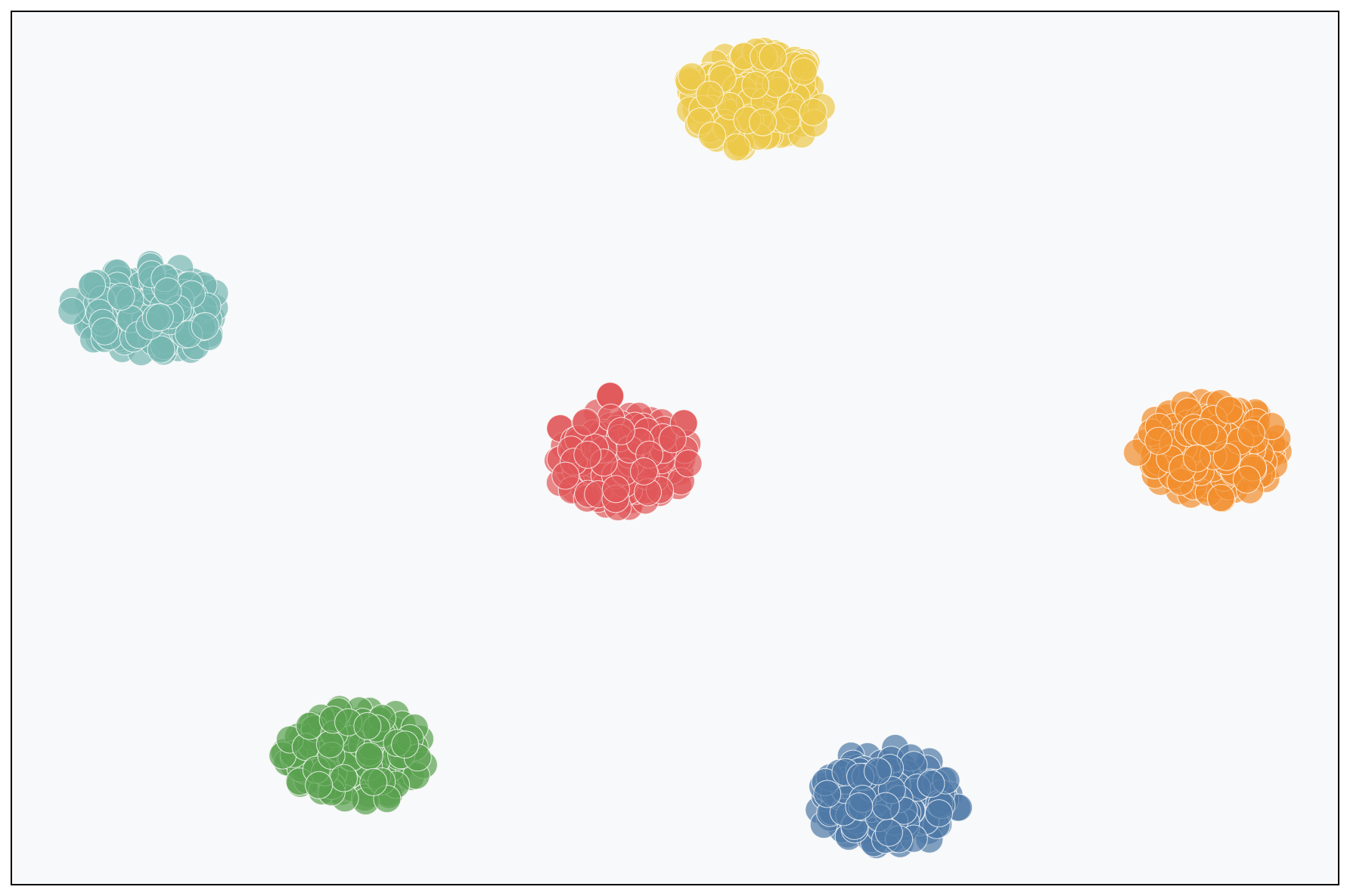}
        \caption*{k=26}
    \end{subfigure}\hfill
    \begin{subfigure}{0.24\linewidth}
        \includegraphics[width=\linewidth]{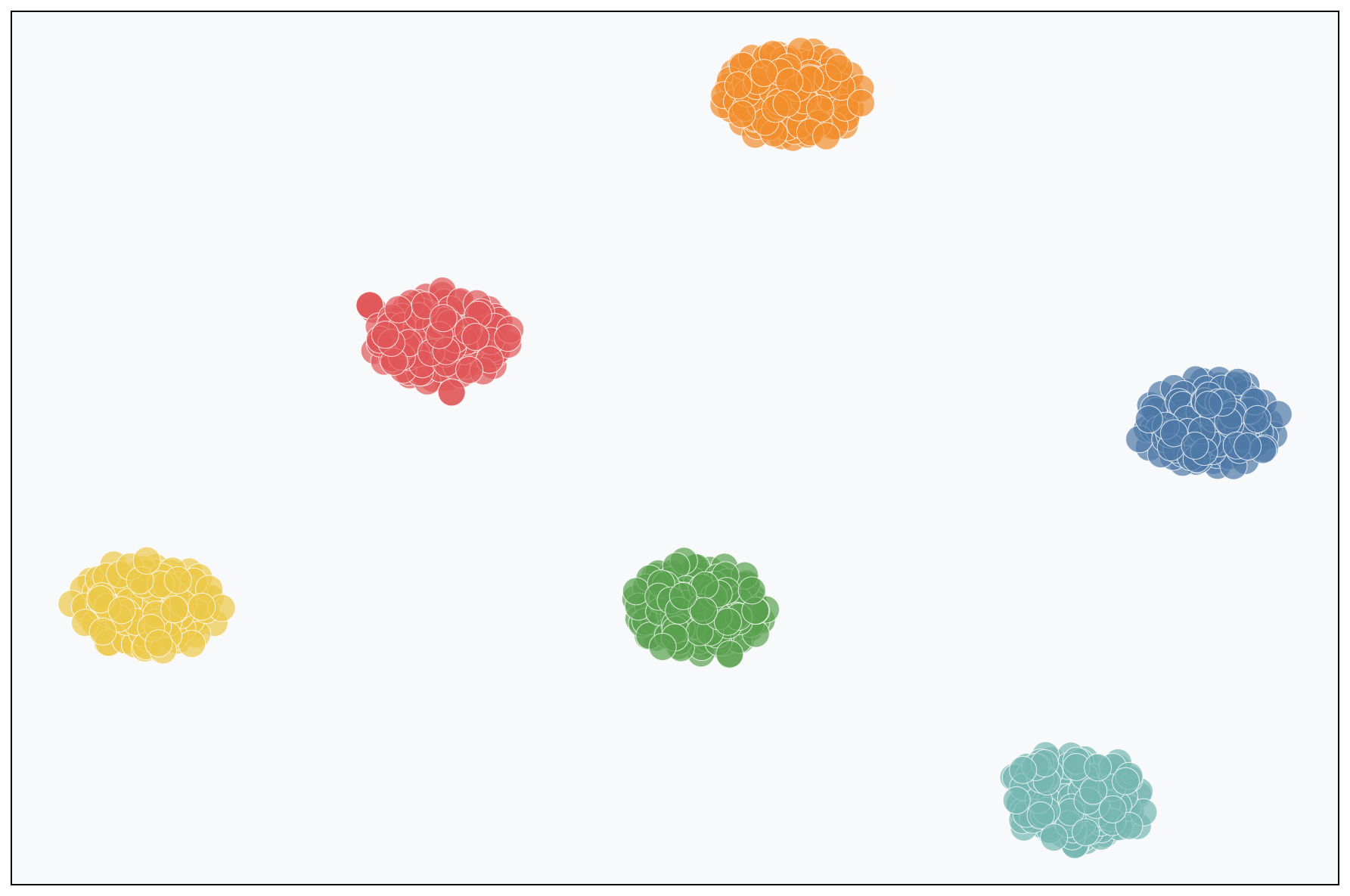}
        \caption*{k=27}
    \end{subfigure}\hfill
    \begin{subfigure}{0.24\linewidth}
        \includegraphics[width=\linewidth]{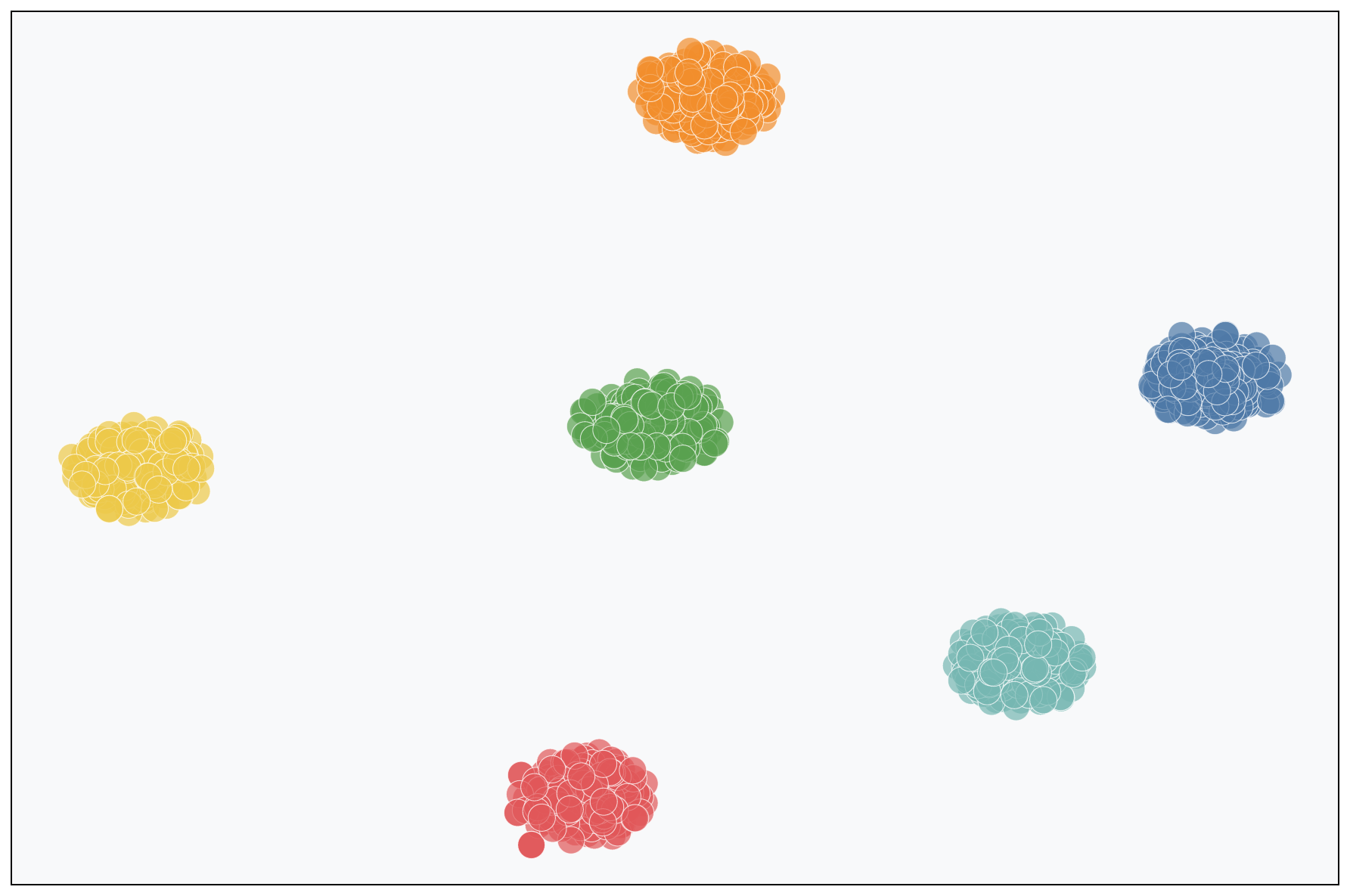}
        \caption*{k=28}
    \end{subfigure}

    \vspace{0.2em}
    
    \begin{subfigure}{0.24\linewidth}
        \includegraphics[width=\linewidth]{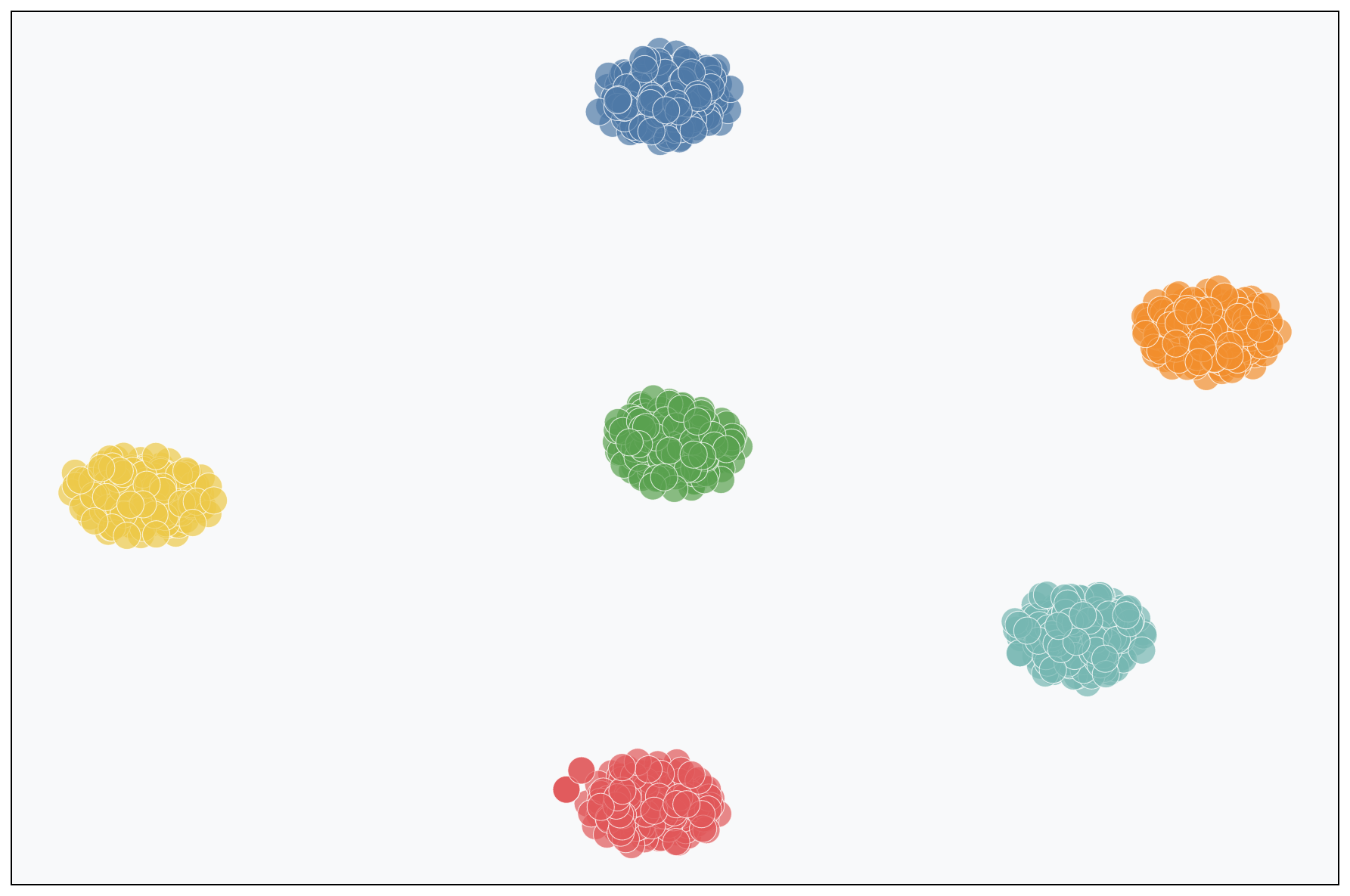}
        \caption*{k=29}
    \end{subfigure}\hfill
    \begin{subfigure}{0.24\linewidth}
        \includegraphics[width=\linewidth]{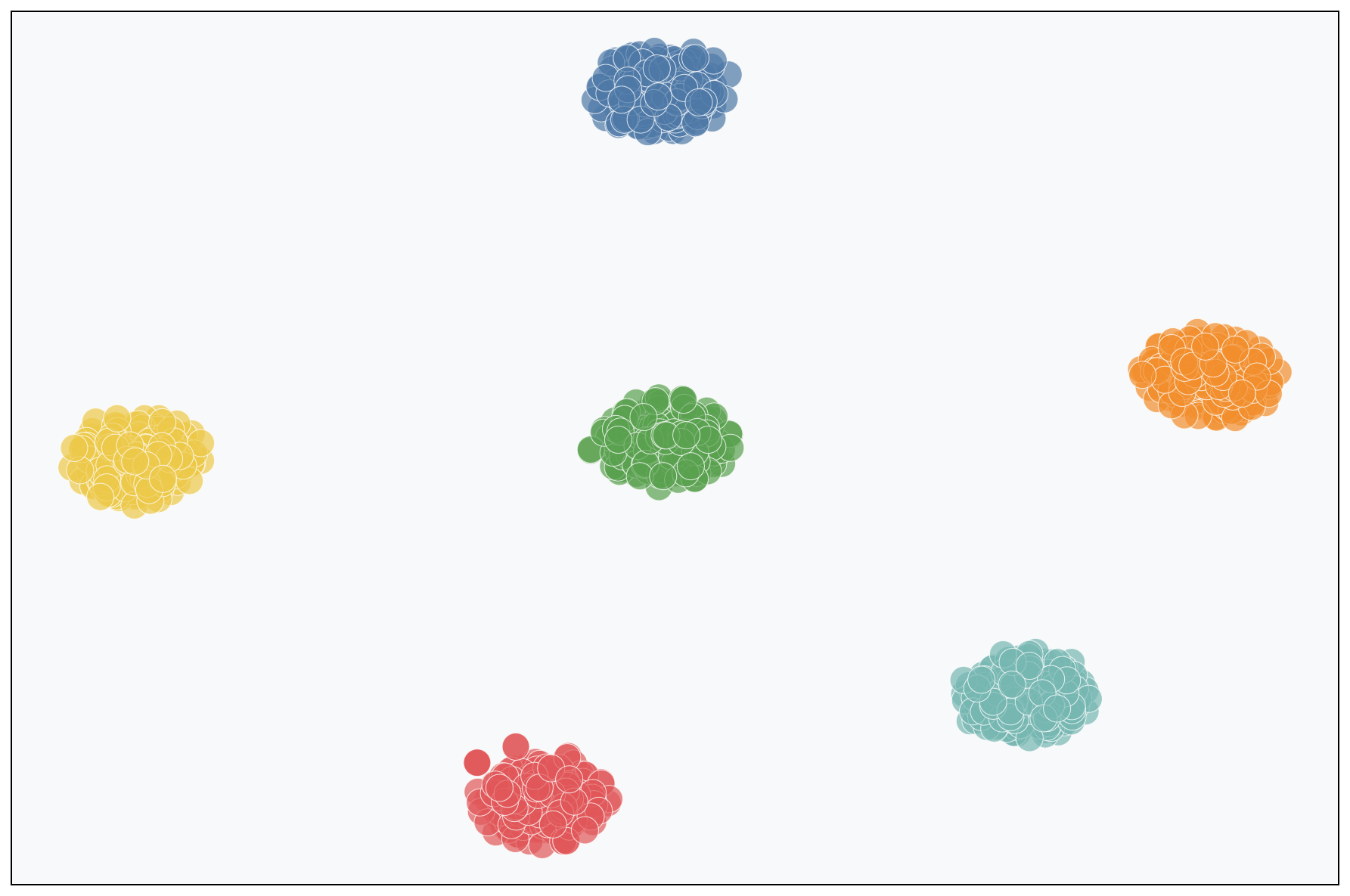}
        \caption*{k=30}
    \end{subfigure}\hfill
    \begin{subfigure}{0.24\linewidth}
        \includegraphics[width=\linewidth]{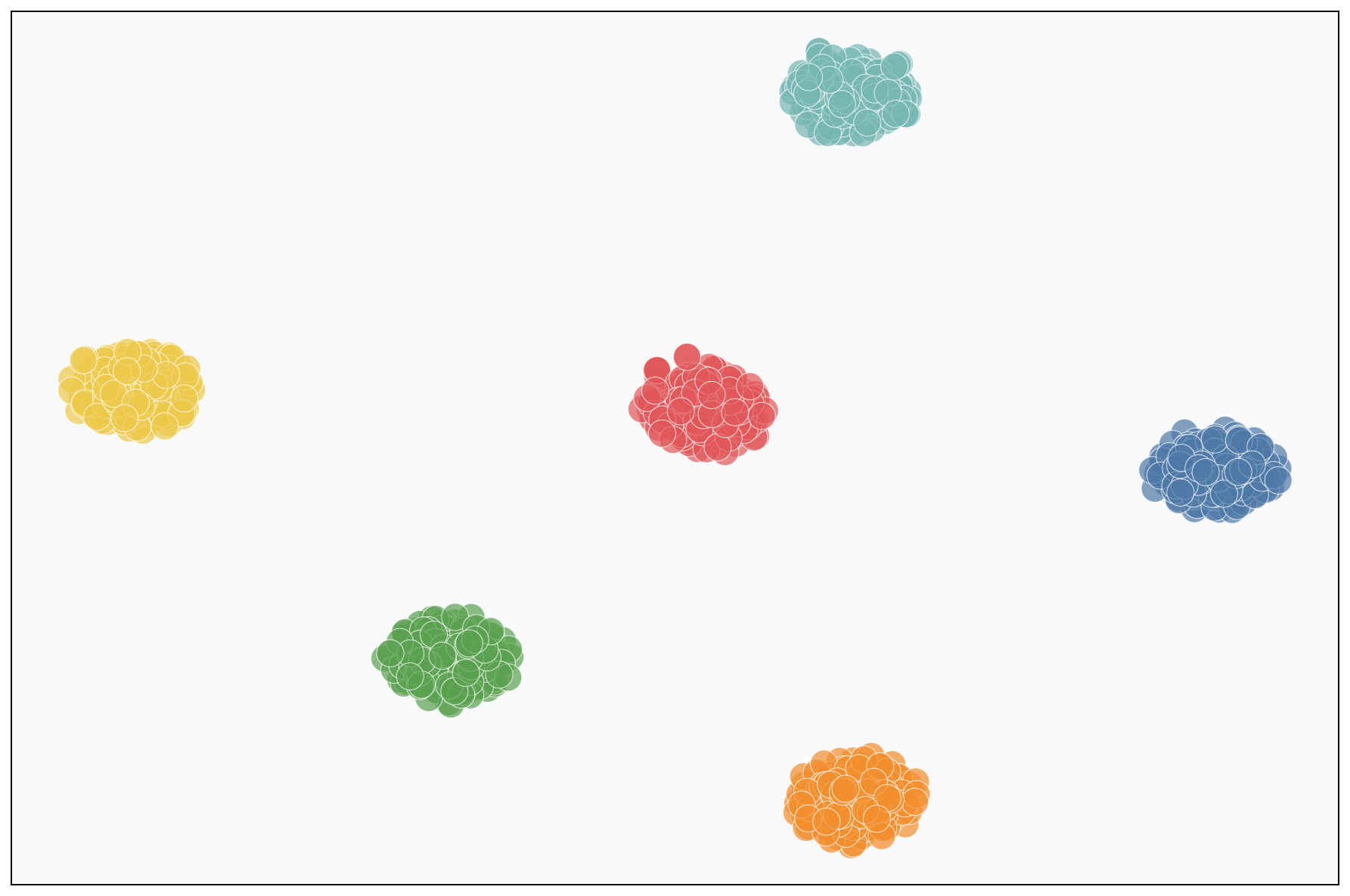}
        \caption*{k=31}
    \end{subfigure}\hfill
    \begin{subfigure}{0.24\linewidth}
        \includegraphics[width=\linewidth]{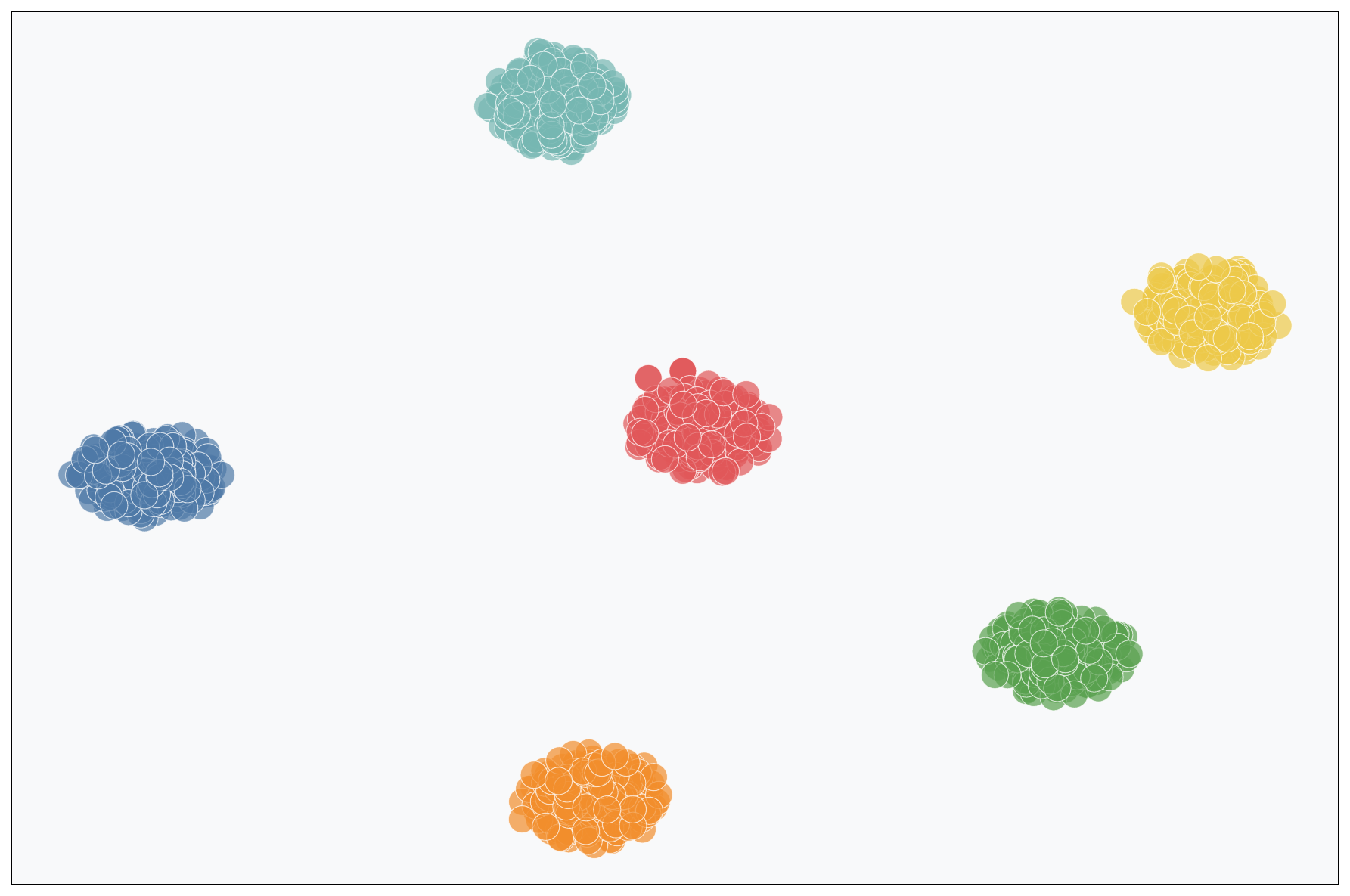}
        \caption*{k=32}
    \end{subfigure}
    \caption{t-SNE analysis on averaged $\beta$ on AdaptSum (continued).}
    \label{fig:tsne_layers2_AdaptSum} 
\end{figure*}

\paragraph{AdaptSum.} Figure \ref{fig:tsne_layers1_AdaptSum} and \ref{fig:tsne_layers2_AdaptSum} show the t-SNE plots of all 32 layers on AdaptSum.

\begin{figure*}[htbp]
    \begin{minipage}{\textwidth}
        \centering
        \hspace{-5pt}
        \includegraphics[width=0.3\textwidth]{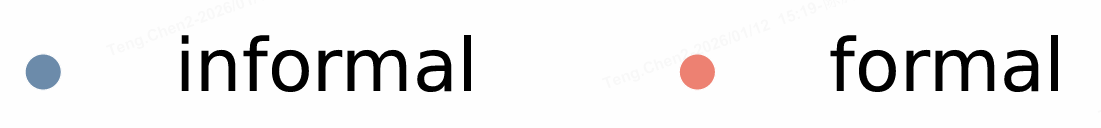}
    \end{minipage}
    \vspace{0.001em}
    \centering

    \begin{subfigure}{0.24\linewidth}
        \includegraphics[width=\linewidth]{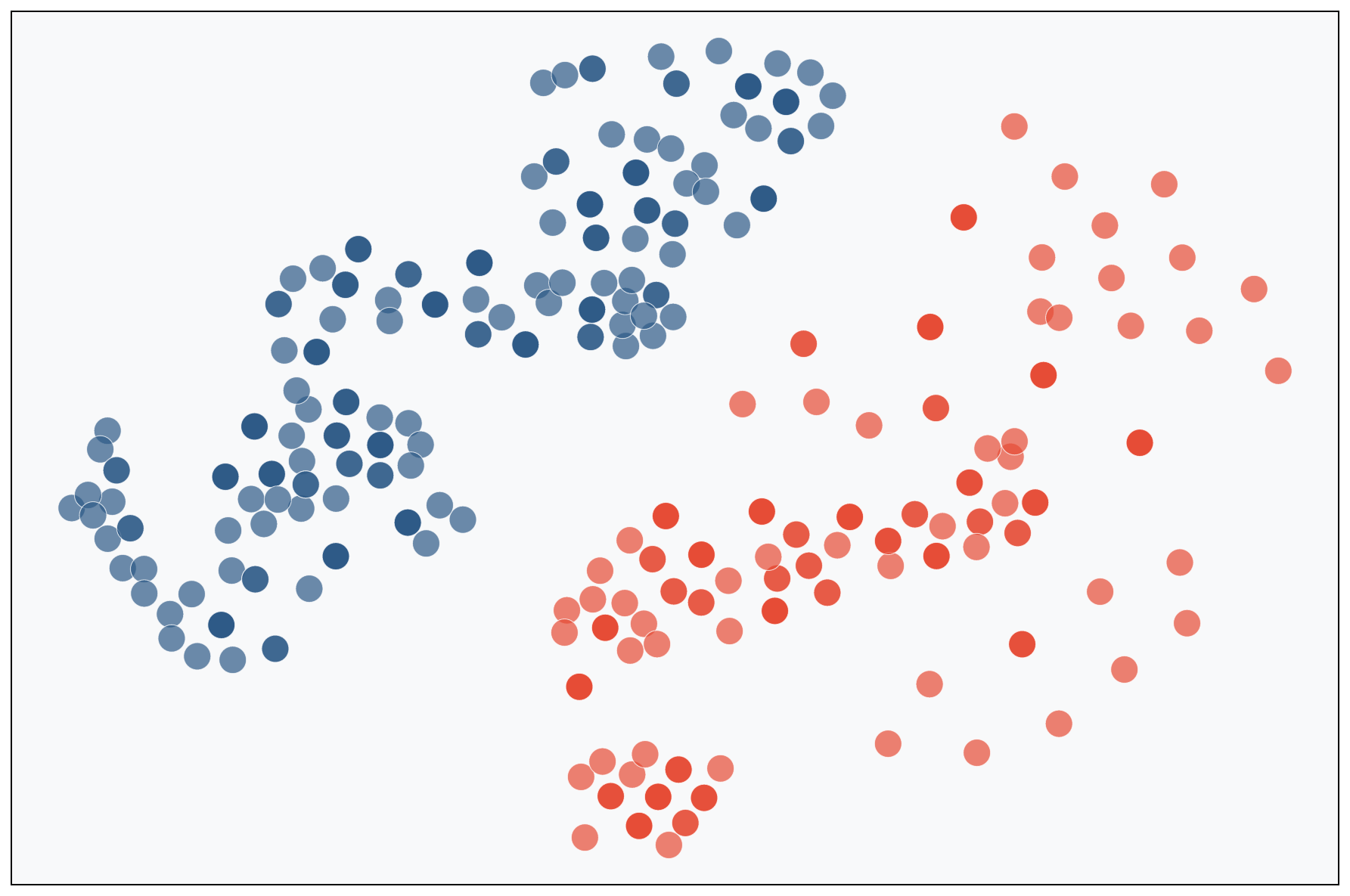}
        \caption*{k=1}
    \end{subfigure}\hfill
    \begin{subfigure}{0.24\linewidth}
        \includegraphics[width=\linewidth]{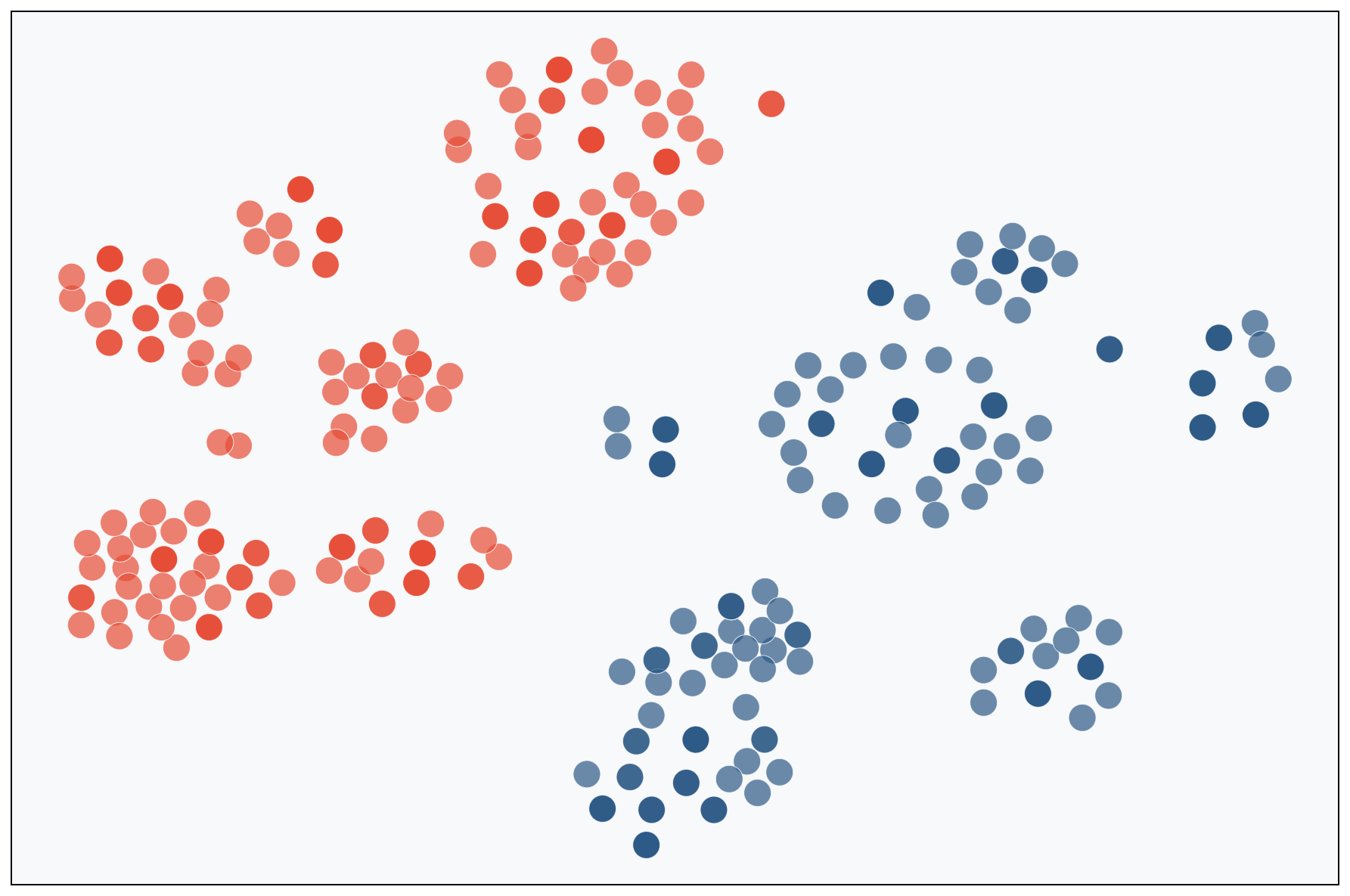}
        \caption*{k=2}
    \end{subfigure}\hfill
    \begin{subfigure}{0.24\linewidth}
        \includegraphics[width=\linewidth]{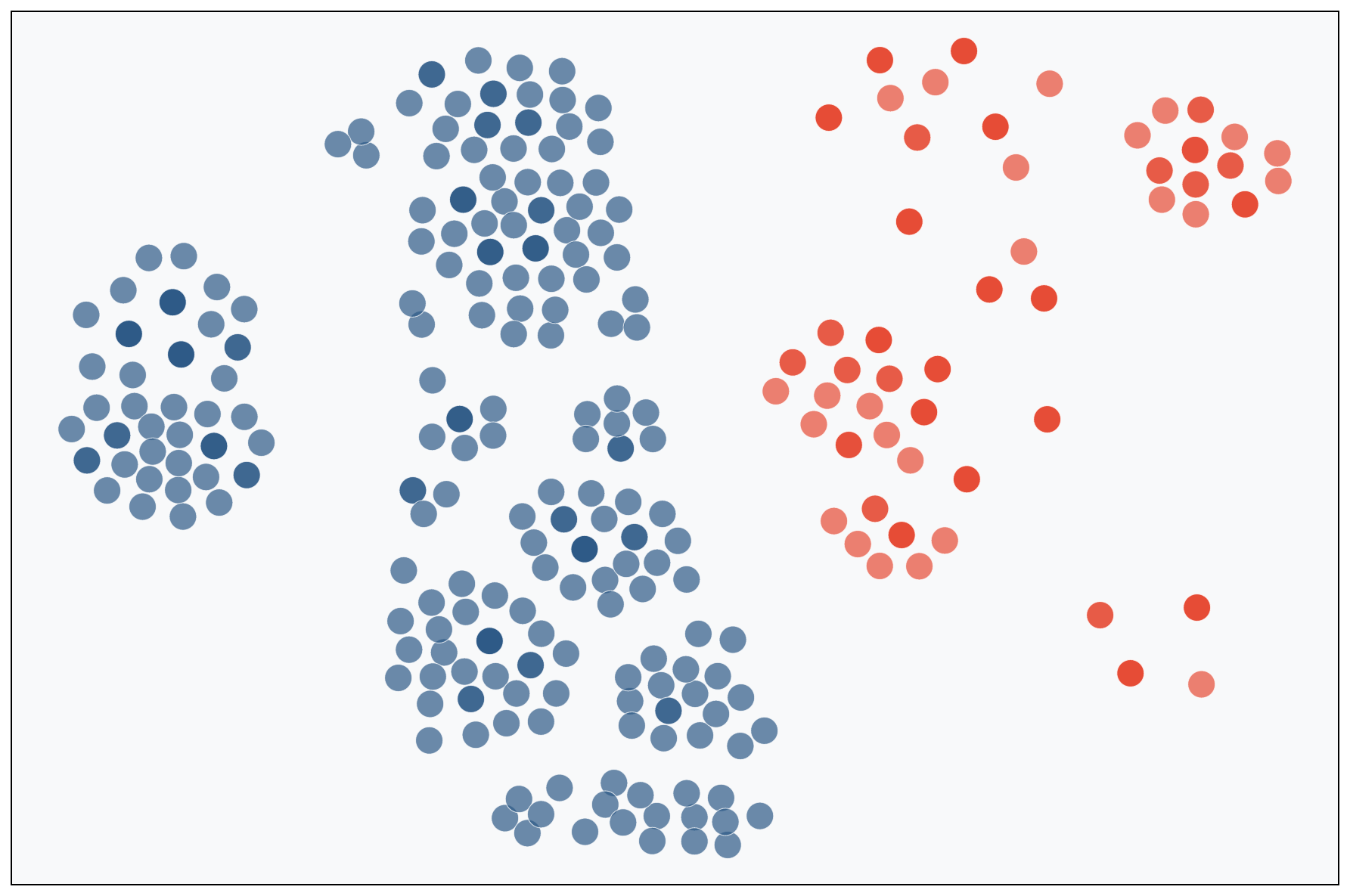}
        \caption*{k=3}
    \end{subfigure}\hfill
    \begin{subfigure}{0.24\linewidth}
        \includegraphics[width=\linewidth]{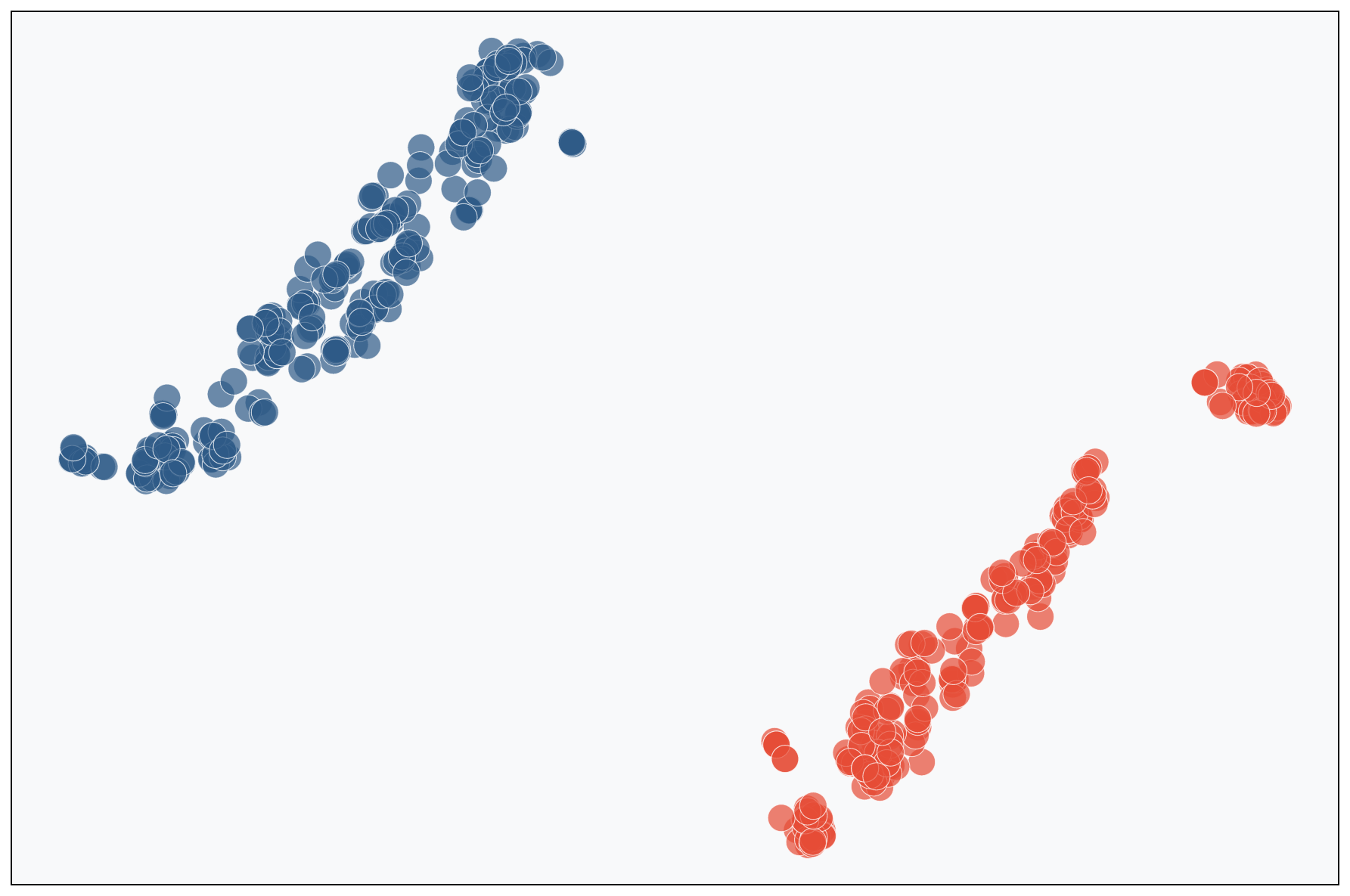}
        \caption*{k=4}
    \end{subfigure}
    
    \vspace{0.2em}
    
    \begin{subfigure}{0.24\linewidth}
        \includegraphics[width=\linewidth]{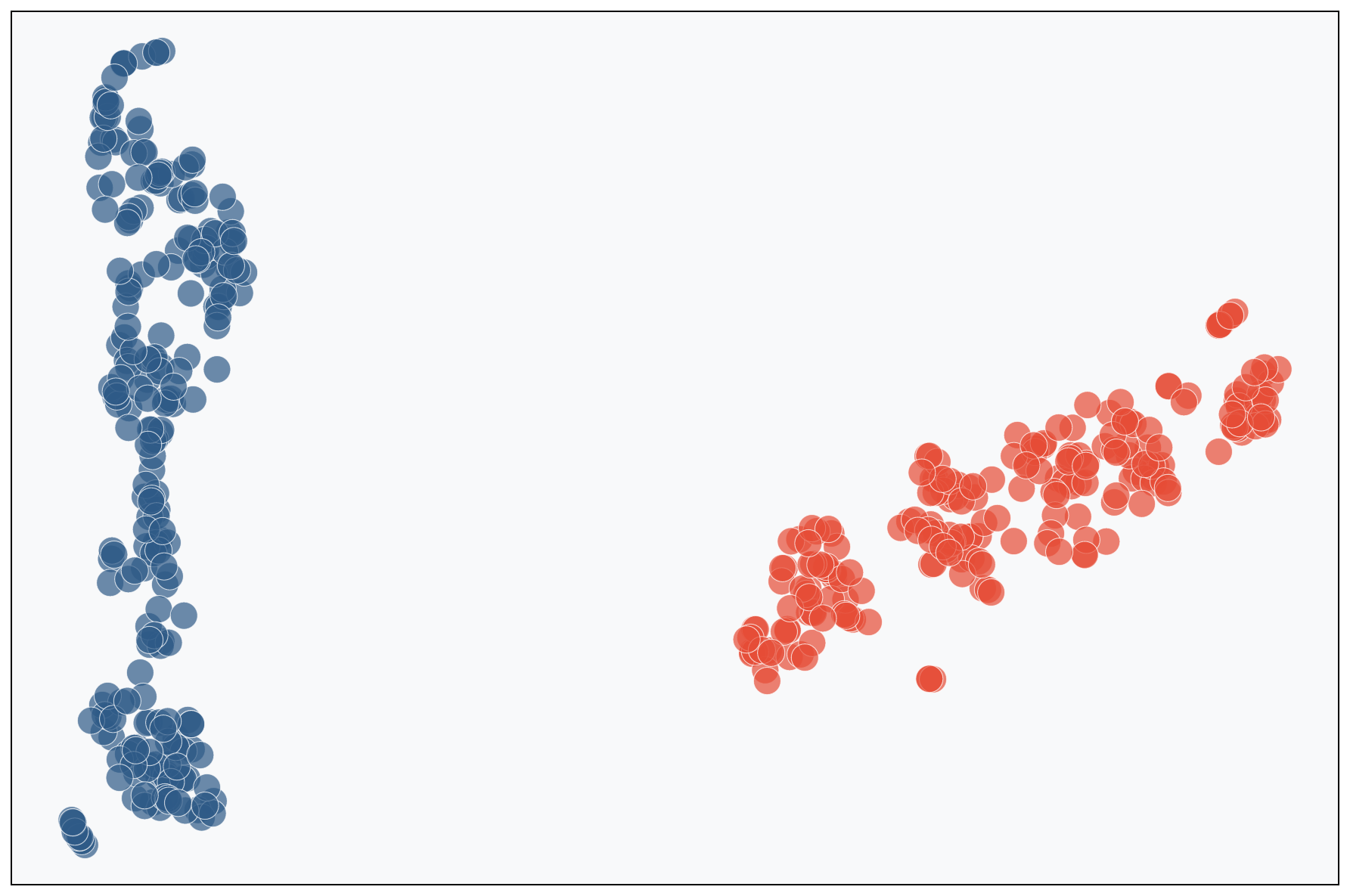}
        \caption*{k=5}
    \end{subfigure}\hfill
    \begin{subfigure}{0.24\linewidth}
        \includegraphics[width=\linewidth]{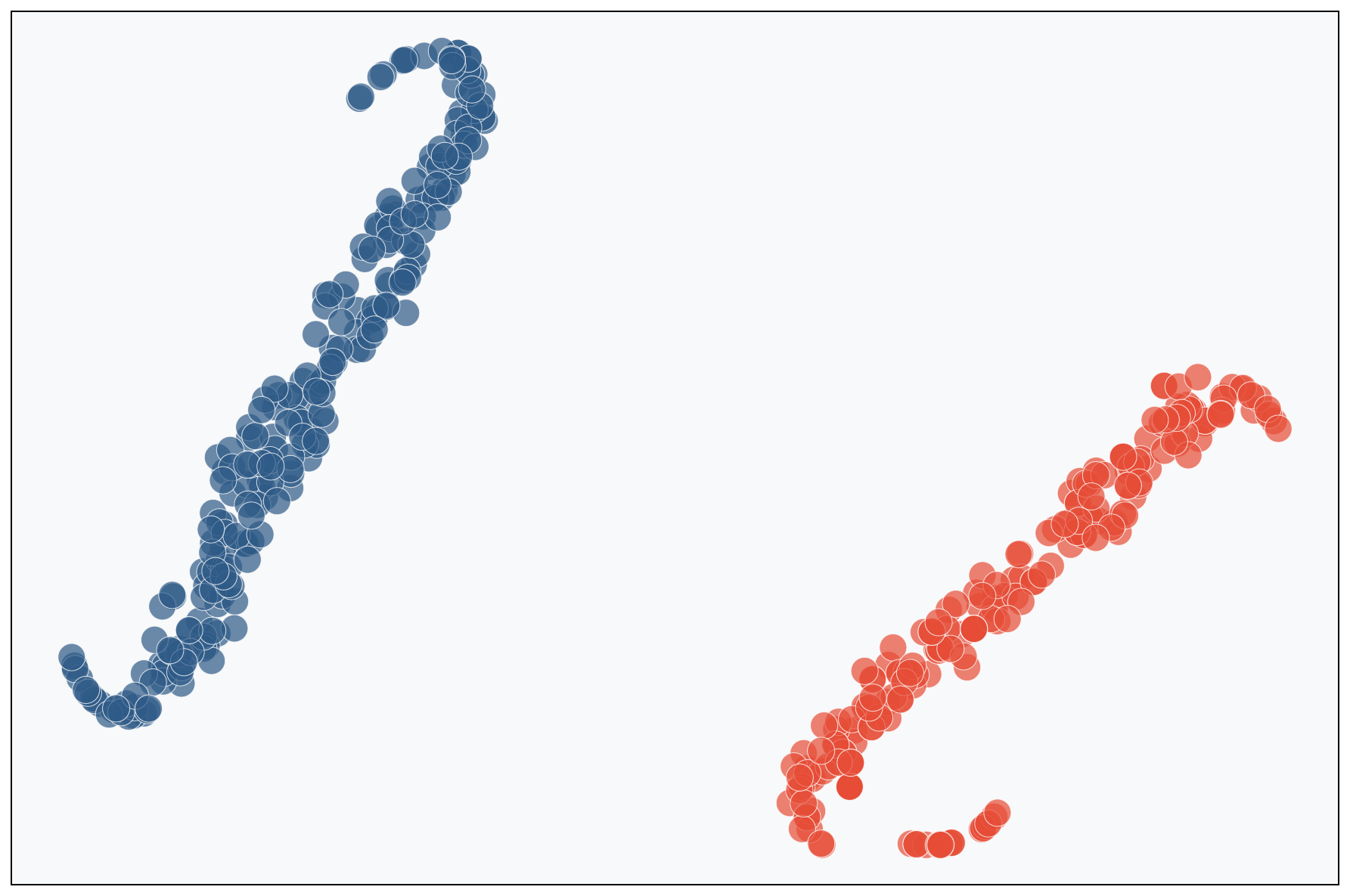}
        \caption*{k=6}
    \end{subfigure}\hfill
    \begin{subfigure}{0.24\linewidth}
        \includegraphics[width=\linewidth]{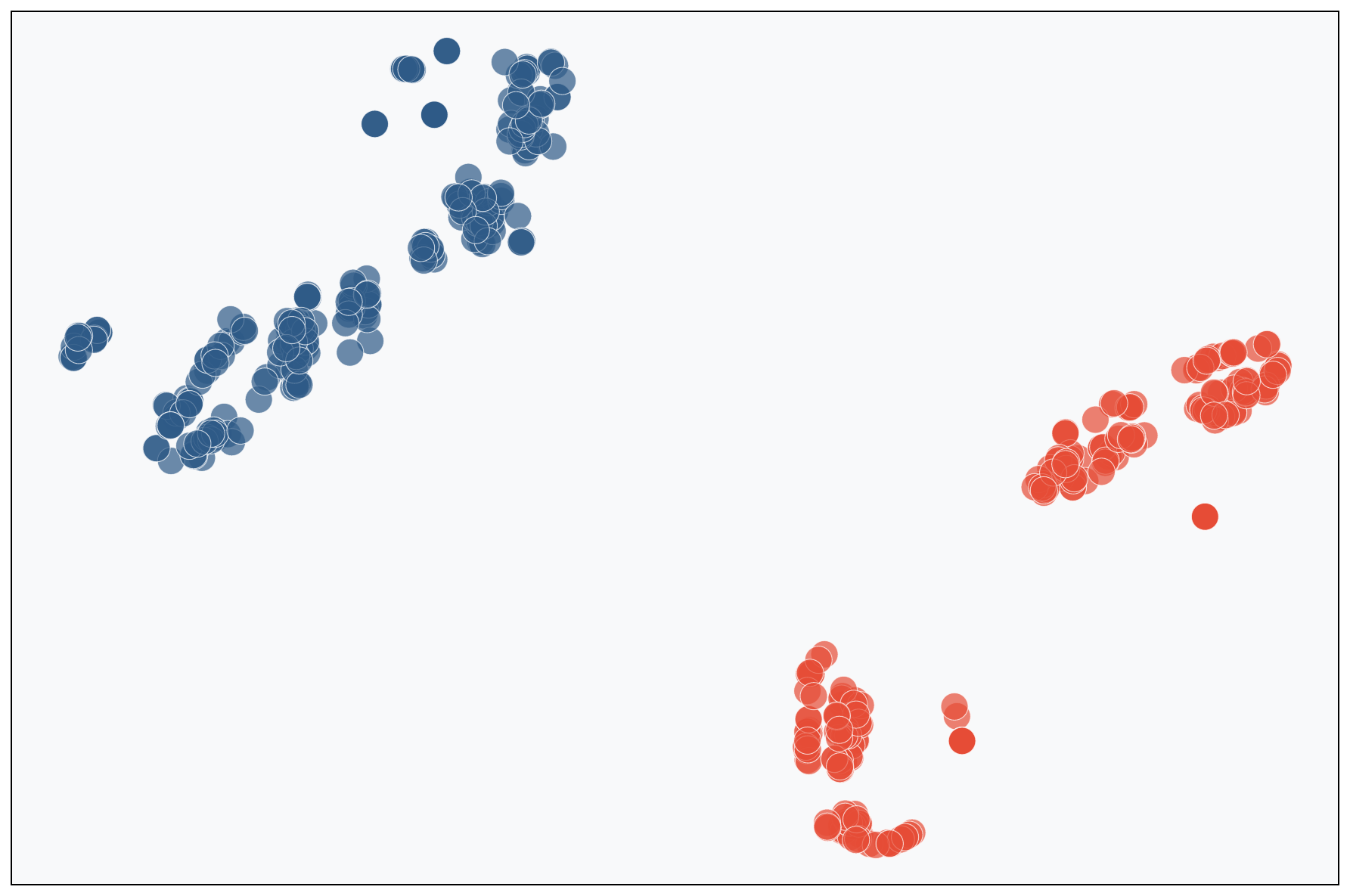}
        \caption*{k=7}
    \end{subfigure}\hfill
    \begin{subfigure}{0.24\linewidth}
        \includegraphics[width=\linewidth]{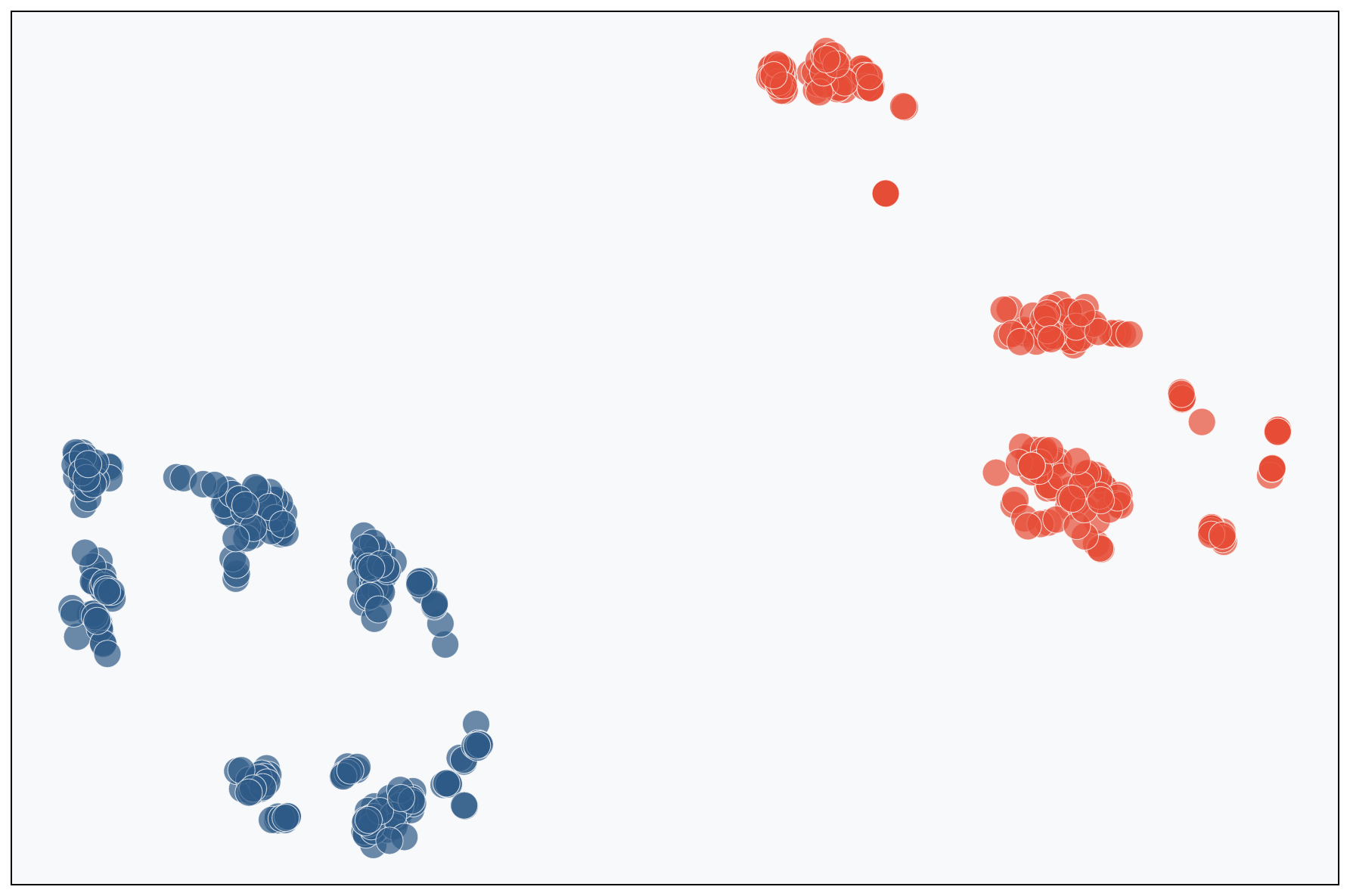}
        \caption*{k=8}
    \end{subfigure}
    
    \vspace{0.2em}
    
    \begin{subfigure}{0.24\linewidth}
        \includegraphics[width=\linewidth]{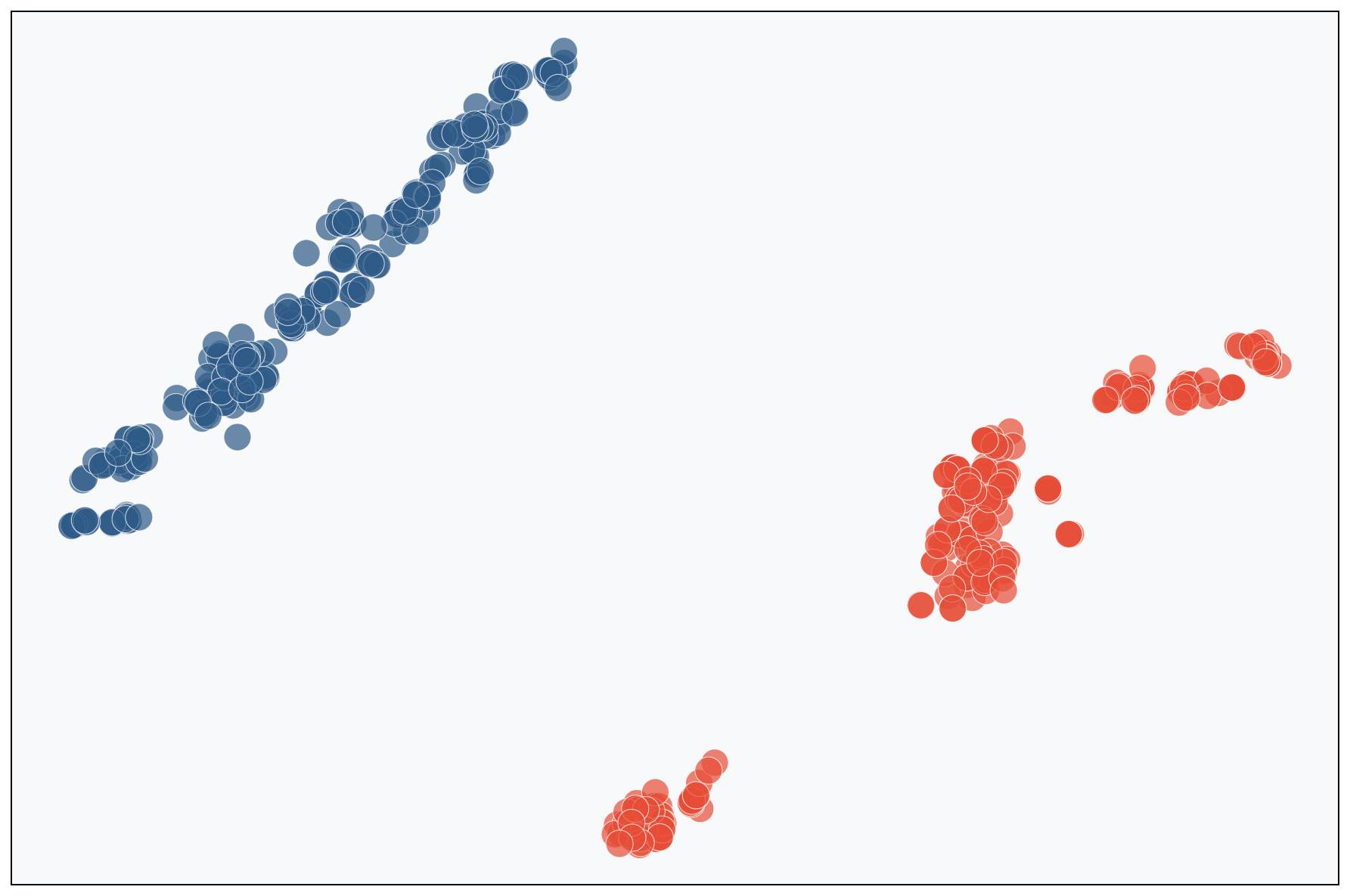}
        \caption*{k=9}
    \end{subfigure}\hfill
    \begin{subfigure}{0.24\linewidth}
        \includegraphics[width=\linewidth]{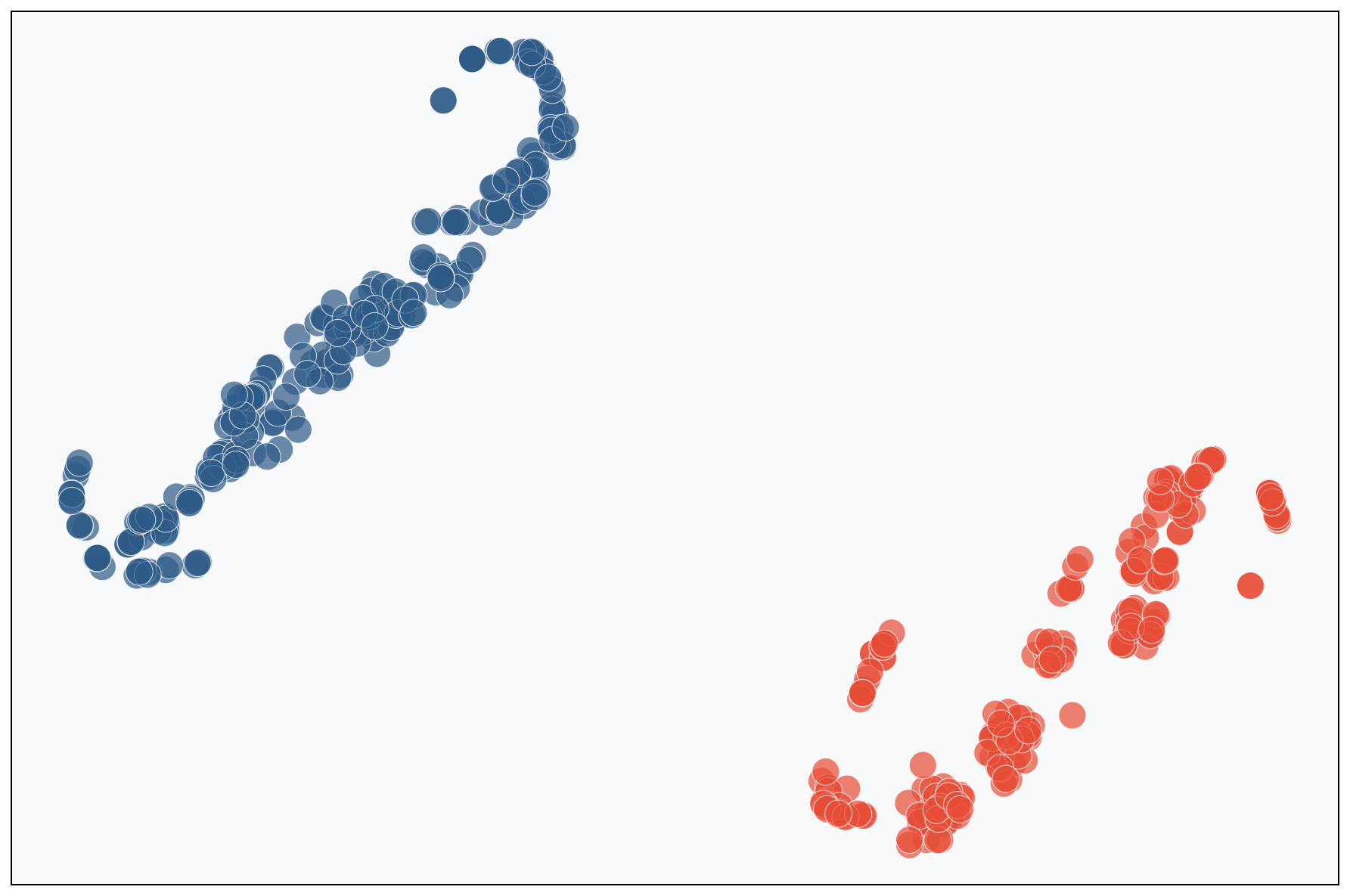}
        \caption*{k=10}
    \end{subfigure}\hfill
    \begin{subfigure}{0.24\linewidth}
        \includegraphics[width=\linewidth]{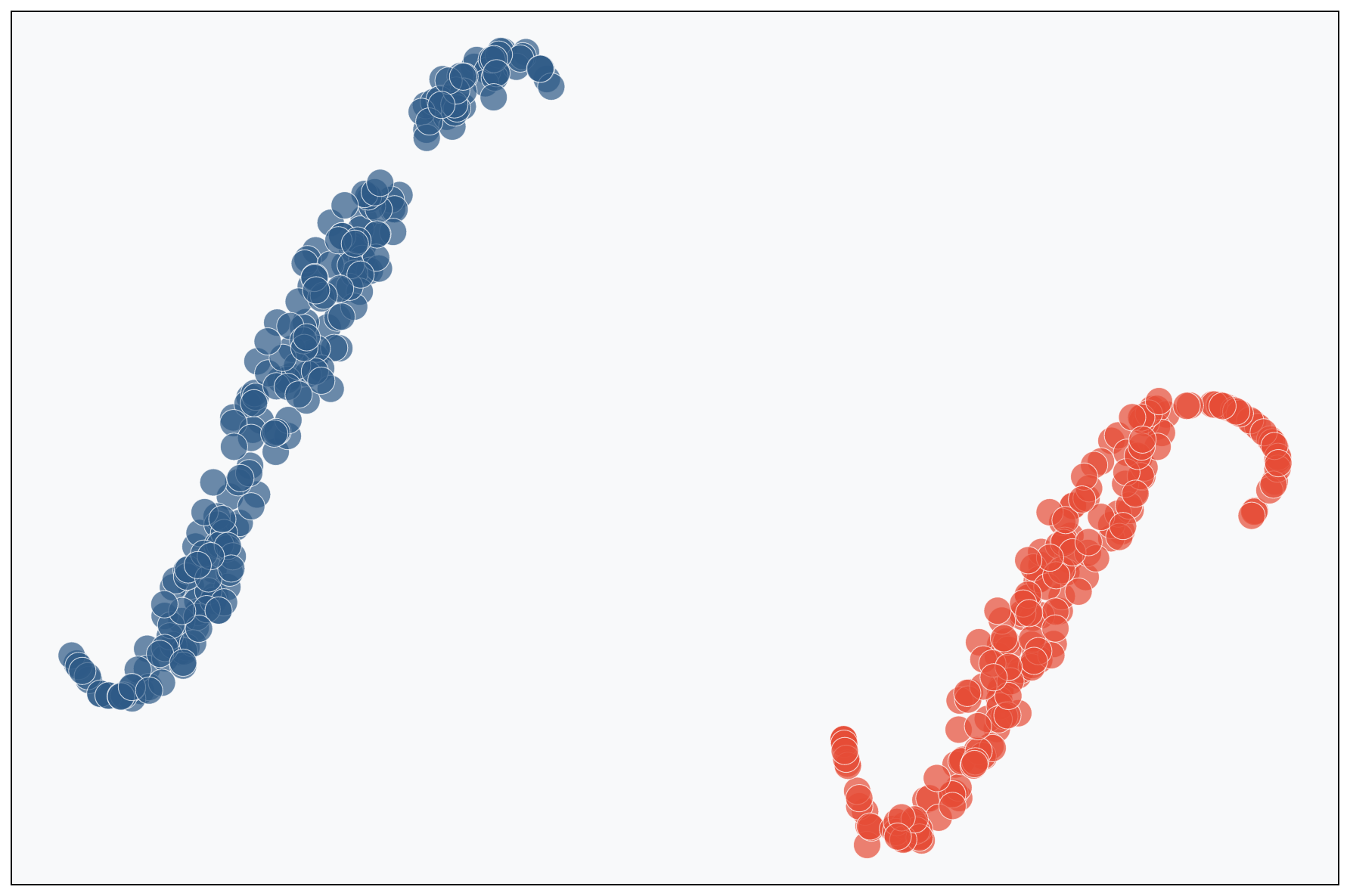}
        \caption*{k=11}
    \end{subfigure}\hfill
    \begin{subfigure}{0.24\linewidth}
        \includegraphics[width=\linewidth]{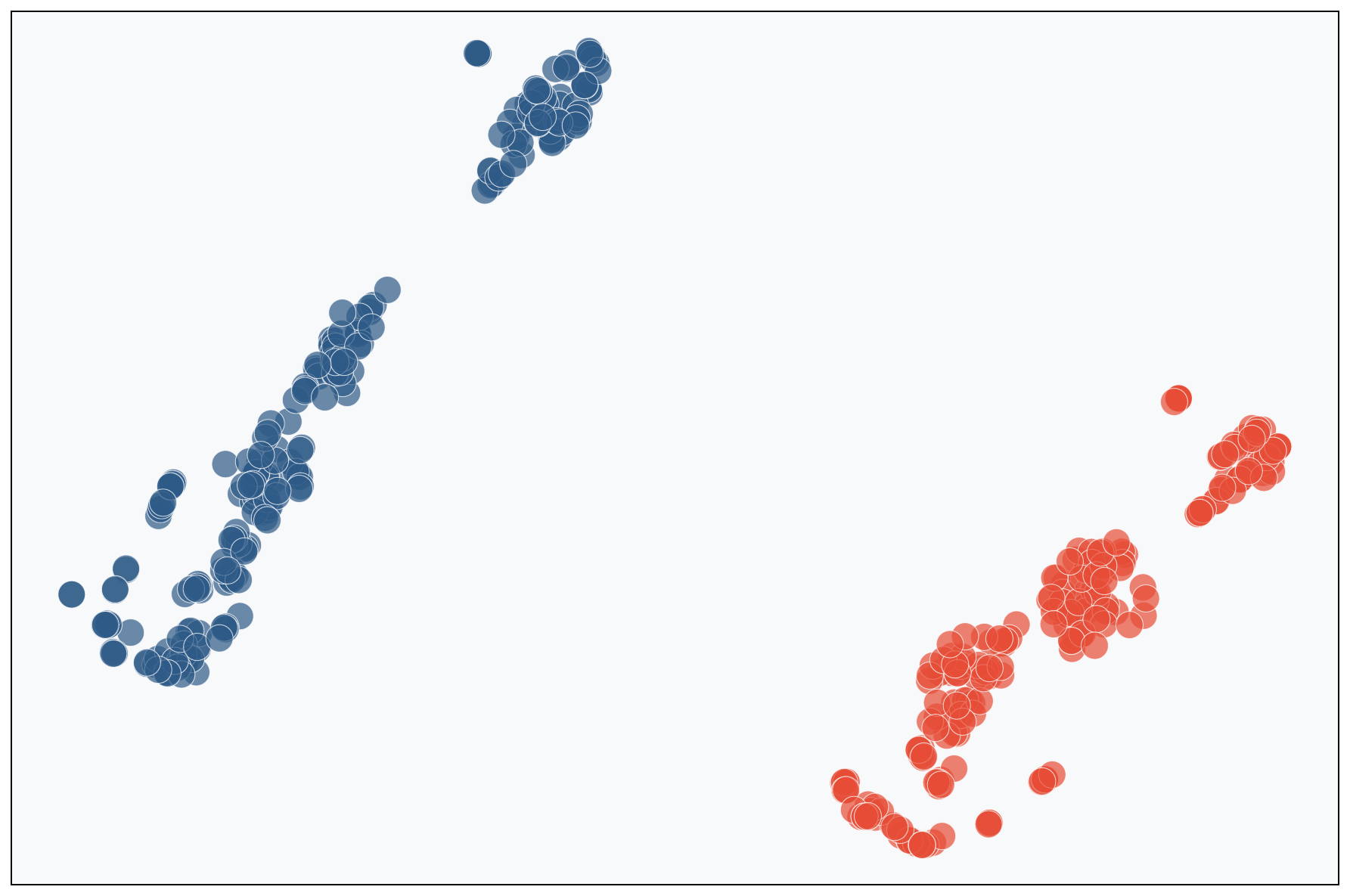}
        \caption*{k=12}
    \end{subfigure}
    
    \vspace{0.2em}
    
    \begin{subfigure}{0.24\linewidth}
        \includegraphics[width=\linewidth]{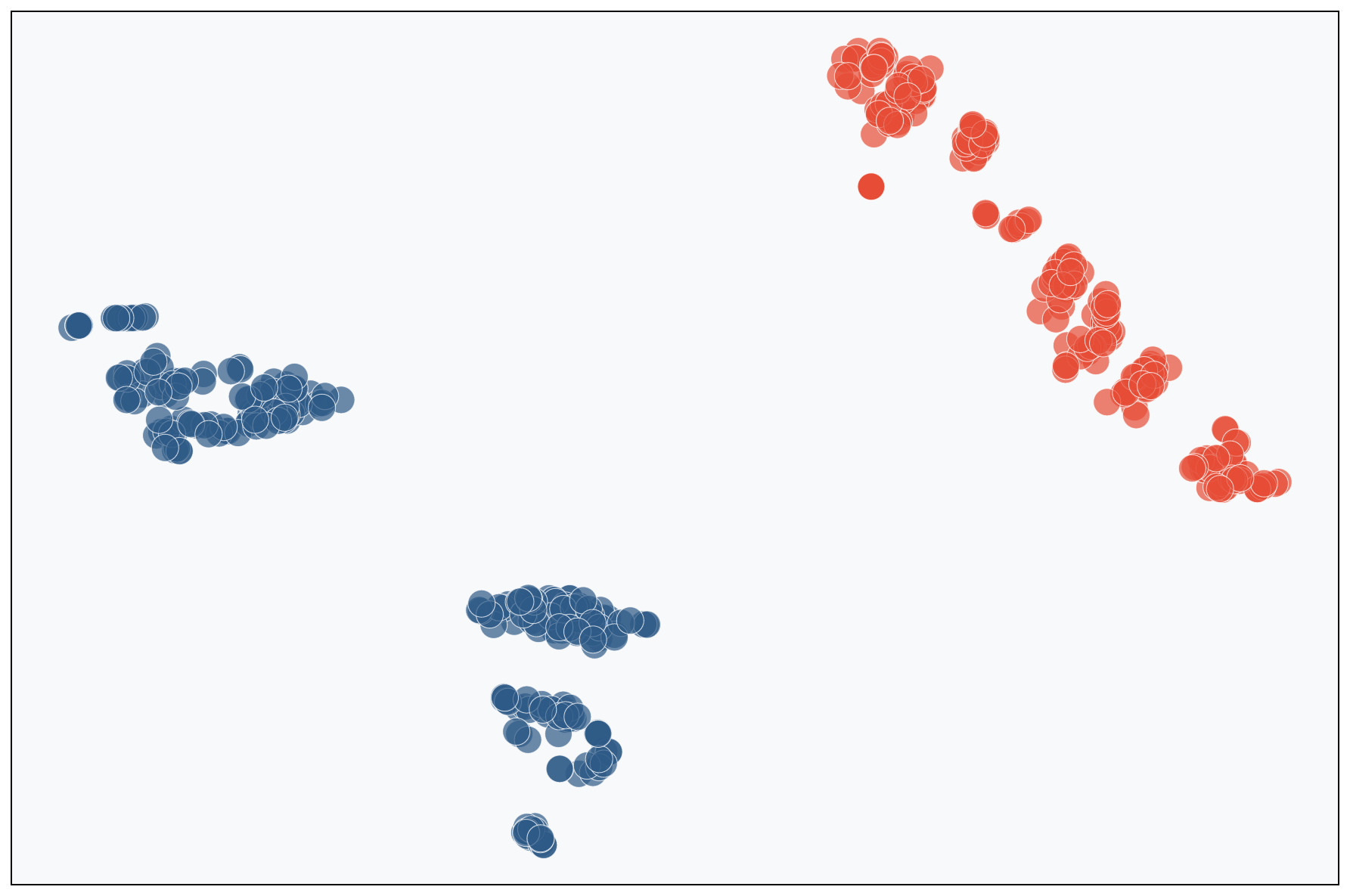}
        \caption*{k=13}
    \end{subfigure}\hfill
    \begin{subfigure}{0.24\linewidth}
        \includegraphics[width=\linewidth]{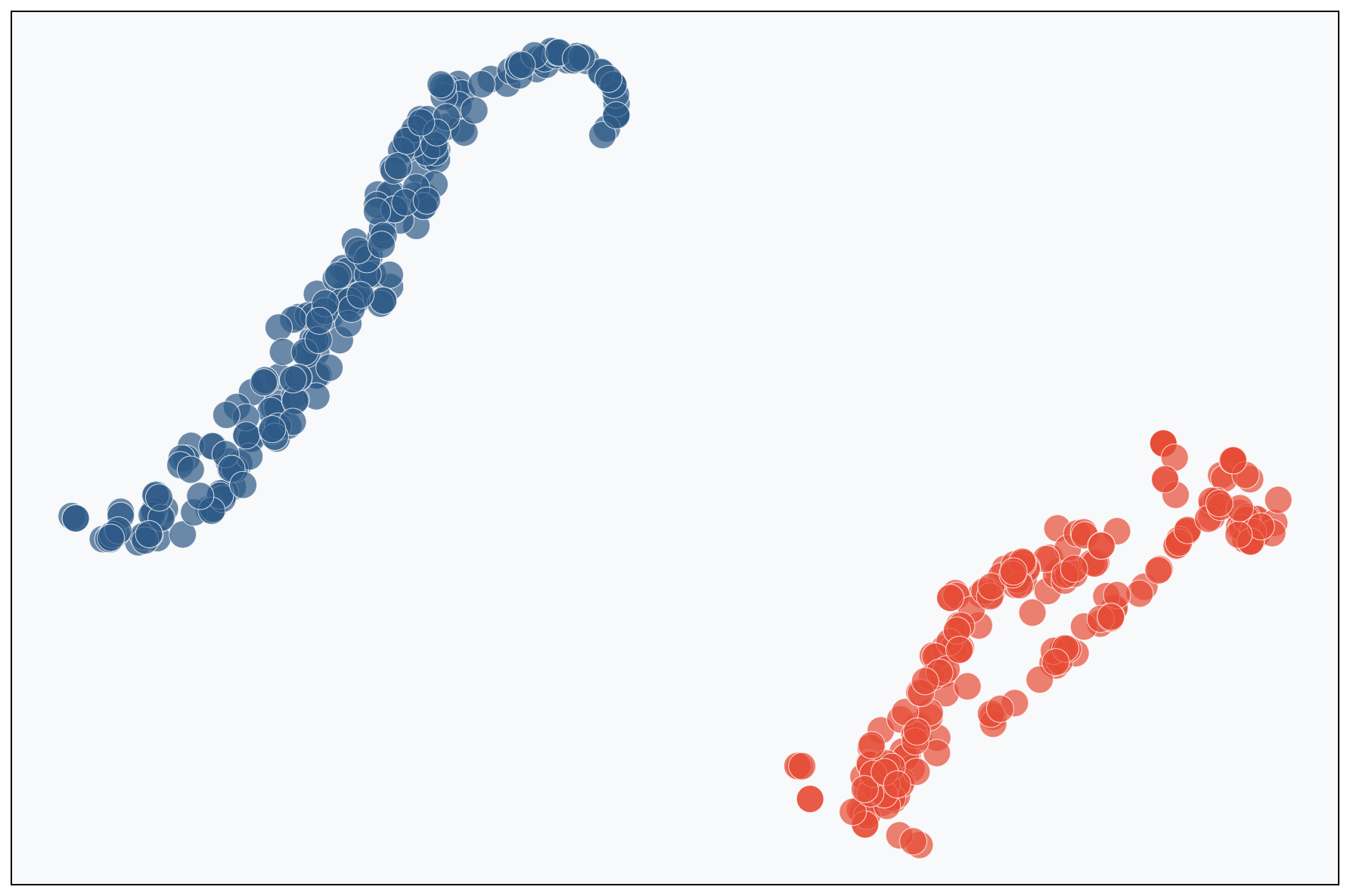}
        \caption*{k=14}
    \end{subfigure}\hfill
    \begin{subfigure}{0.24\linewidth}
        \includegraphics[width=\linewidth]{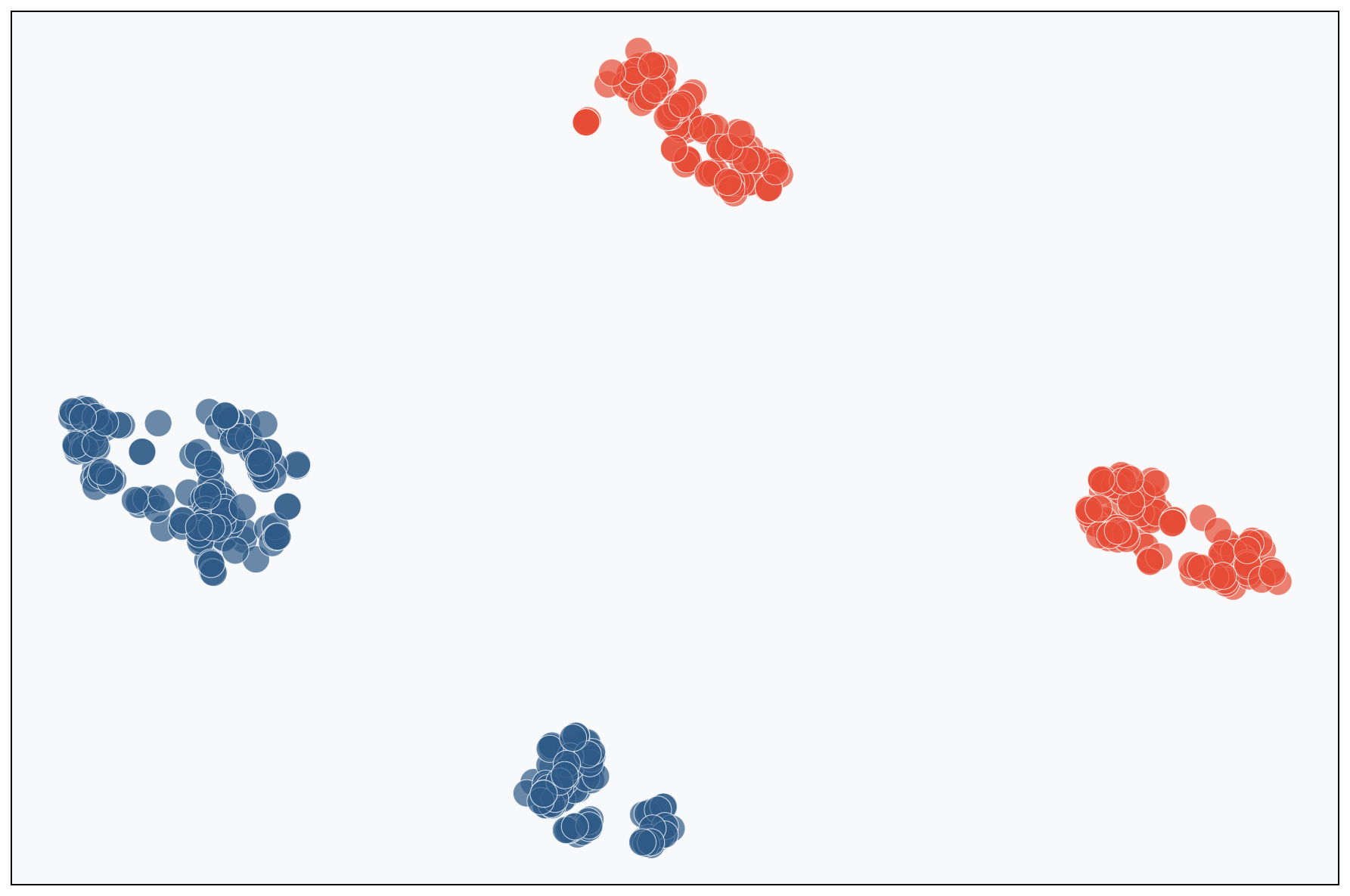}
        \caption*{k=15}
    \end{subfigure}\hfill
    \begin{subfigure}{0.24\linewidth}
        \includegraphics[width=\linewidth]{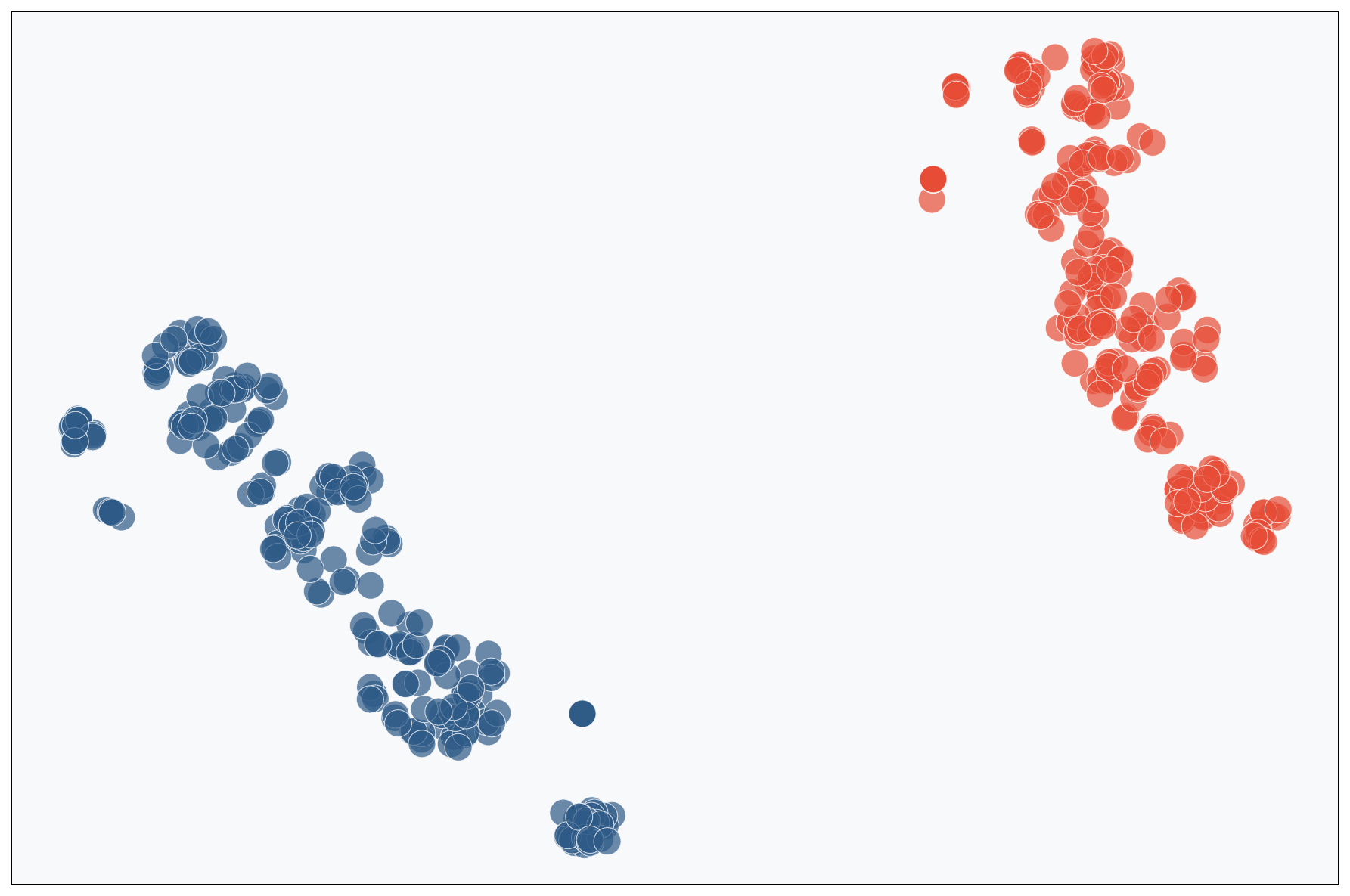}
        \caption*{k=16}
    \end{subfigure}
    
    \caption{t-SNE analysis of averaged $\beta$ on GYAFC (first 16 layers).}
    \label{fig:tsne_layers1_gyafc} 
\end{figure*}

\begin{figure*}[htbp]
    \begin{minipage}{\textwidth}
        \centering
        \hspace{-5pt}
        \includegraphics[width=0.3\textwidth]{tSNE_GYAFC/legend_v2.png}
    \end{minipage}
    \vspace{0.001em}

    \centering
    
    \begin{subfigure}{0.24\linewidth}
        \includegraphics[width=\linewidth]{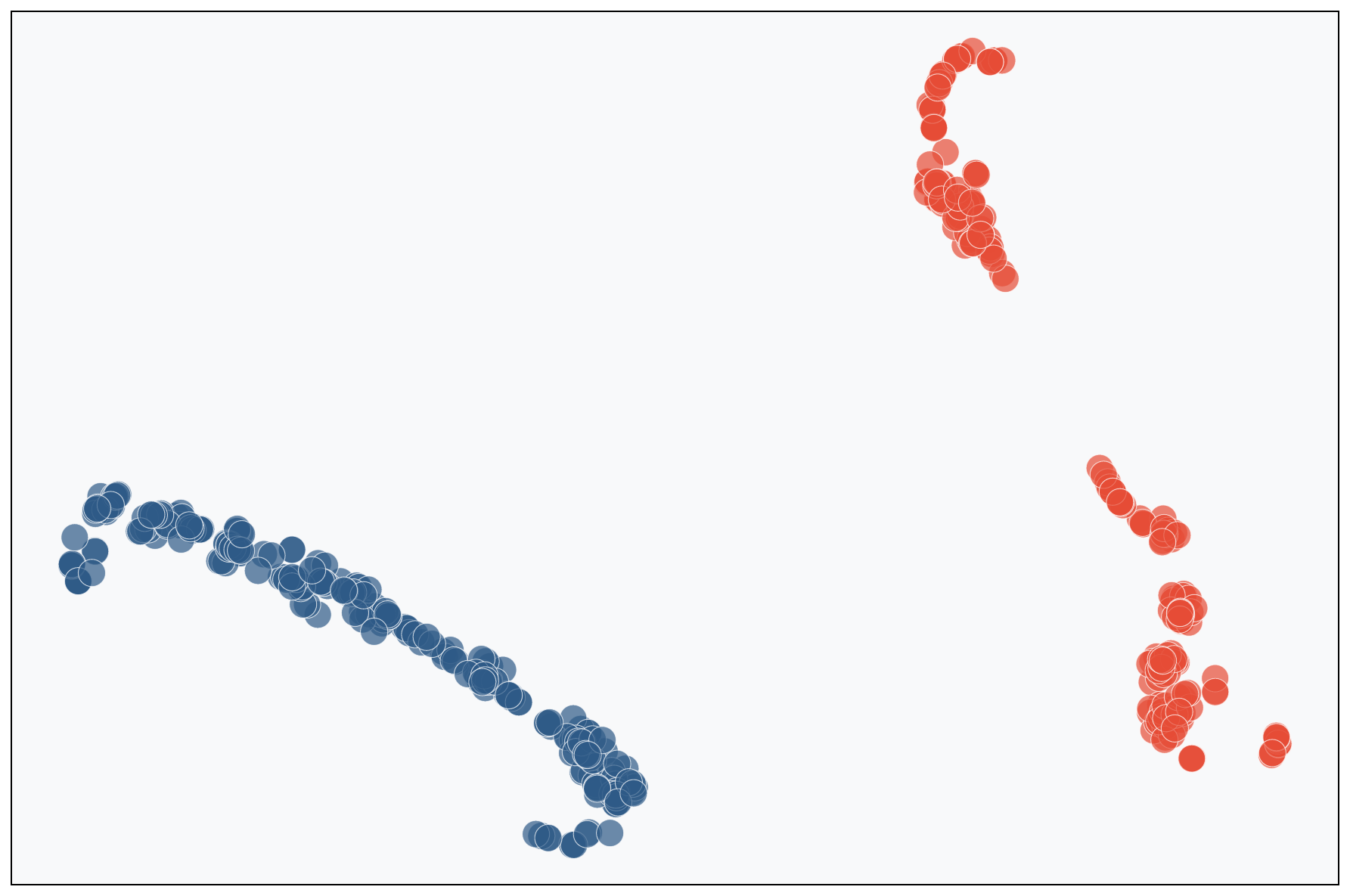}
        \caption*{k=17}
    \end{subfigure}\hfill
    \begin{subfigure}{0.24\linewidth}
        \includegraphics[width=\linewidth]{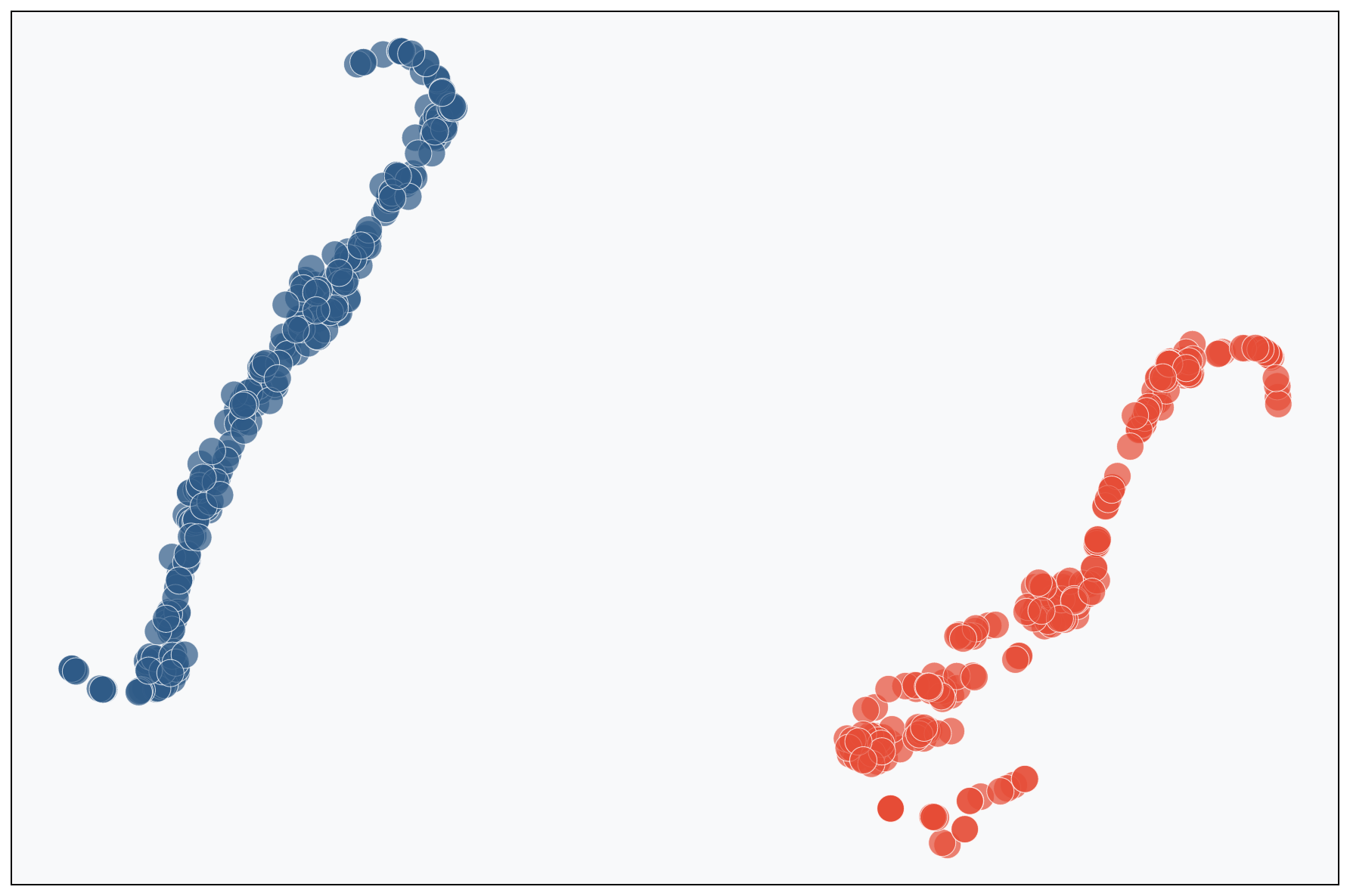}
        \caption*{k=18}
    \end{subfigure}\hfill
    \begin{subfigure}{0.24\linewidth}
        \includegraphics[width=\linewidth]{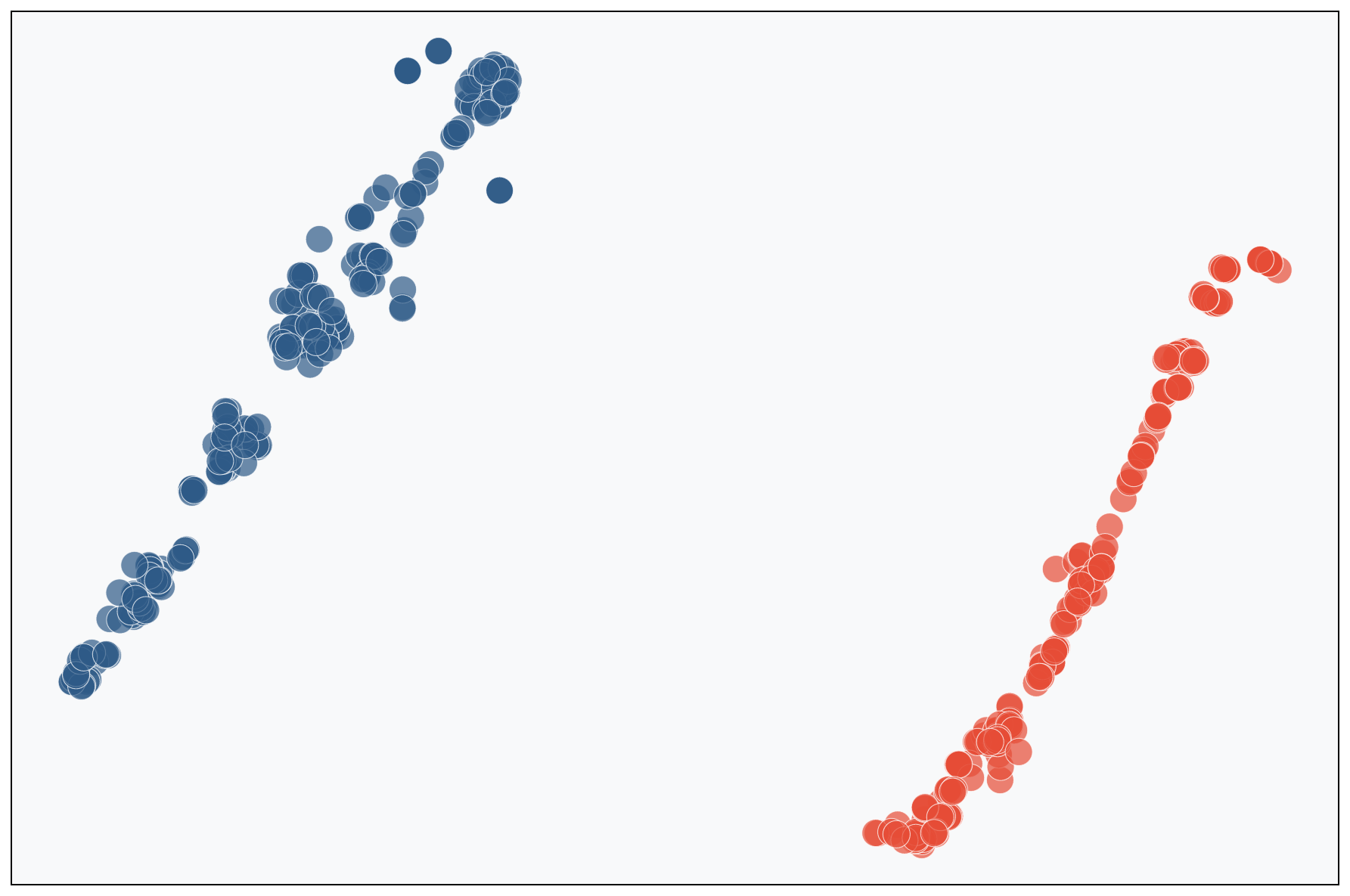}
        \caption*{k=19}
    \end{subfigure}\hfill
    \begin{subfigure}{0.24\linewidth}
        \includegraphics[width=\linewidth]{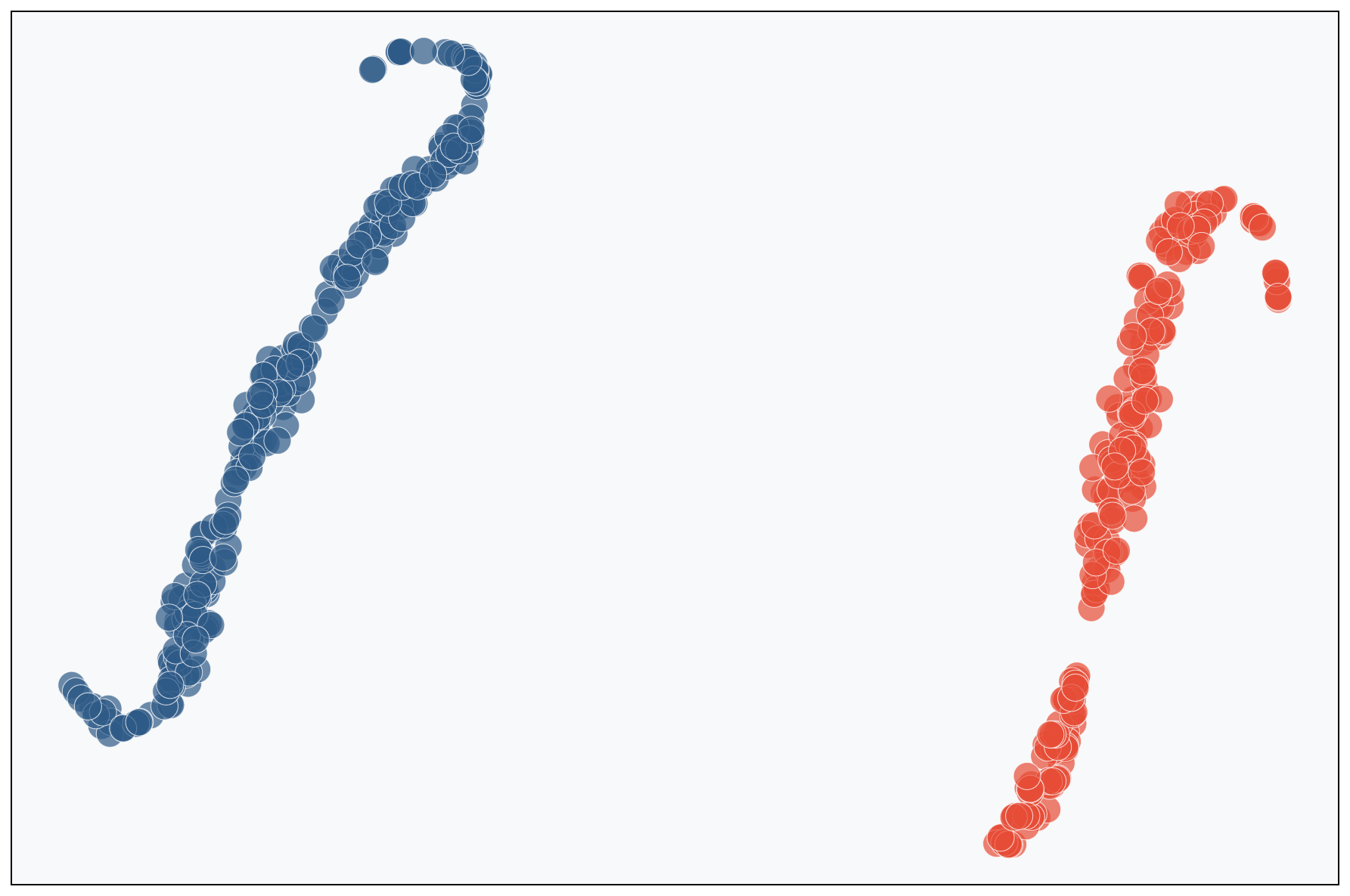}
        \caption*{k=20}
    \end{subfigure}
    
    \vspace{0.2em}
    
    \begin{subfigure}{0.24\linewidth}
        \includegraphics[width=\linewidth]{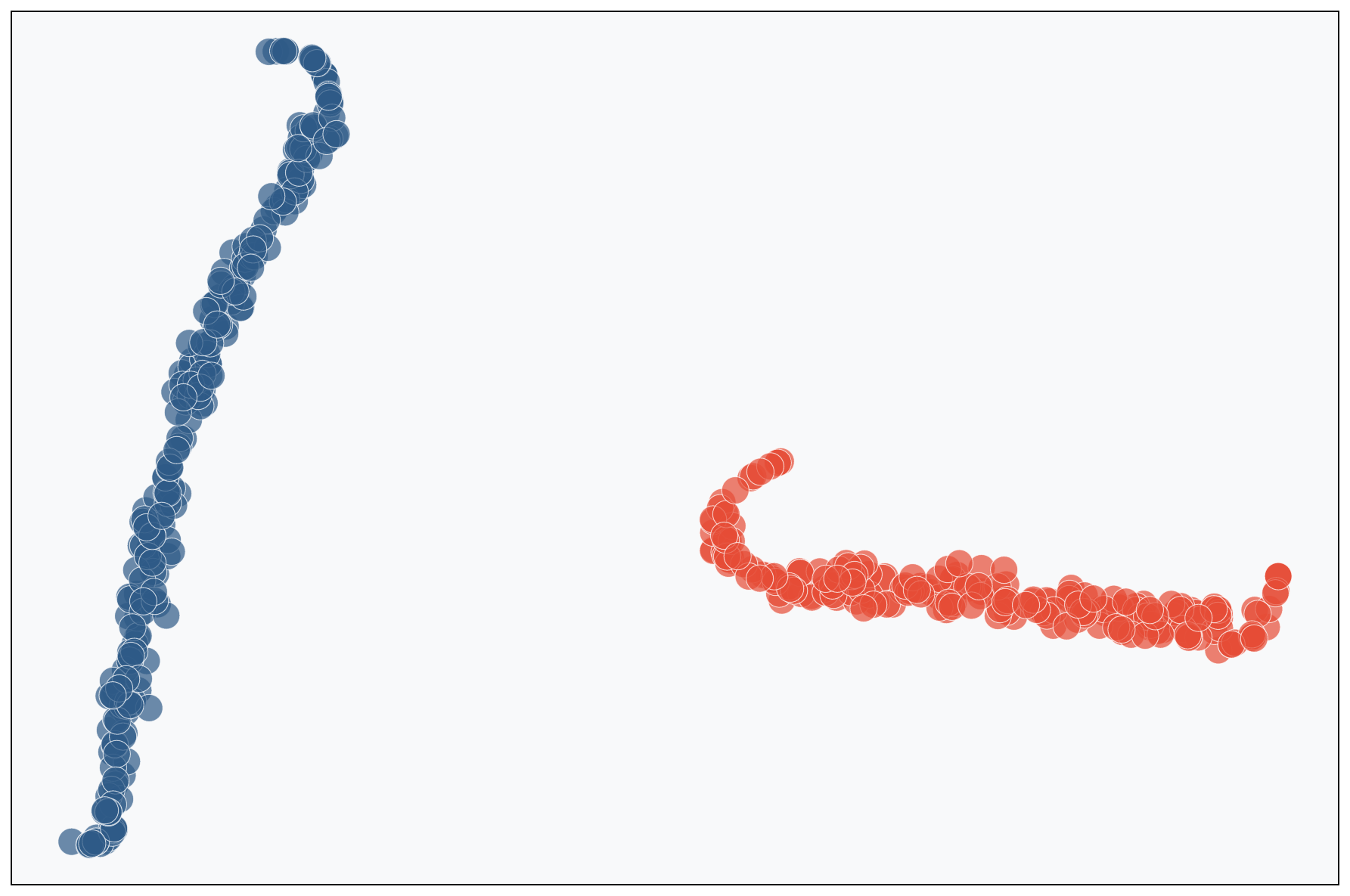}
        \caption*{k=21}
    \end{subfigure}\hfill
    \begin{subfigure}{0.24\linewidth}
        \includegraphics[width=\linewidth]{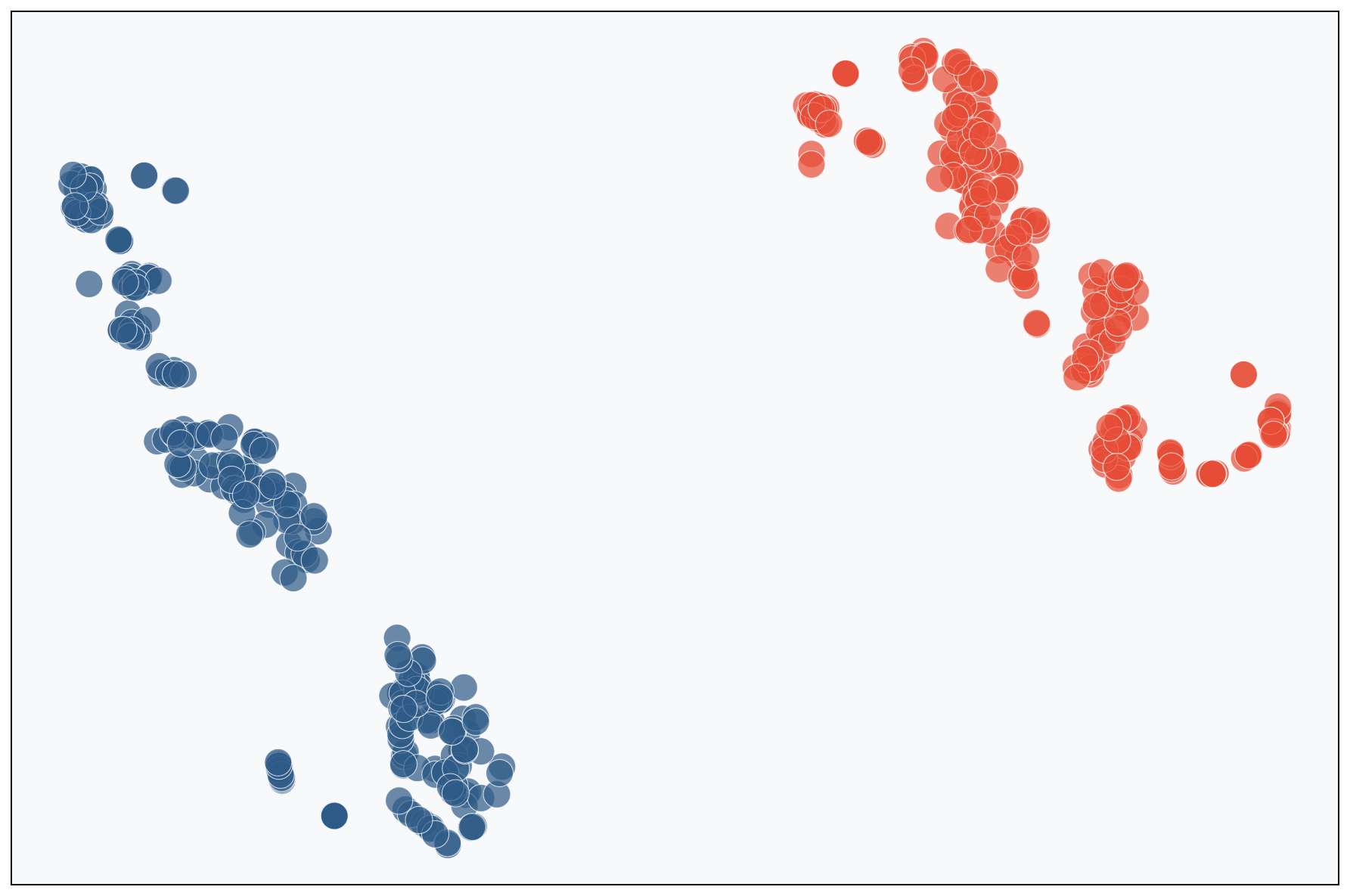}
        \caption*{k=22}
    \end{subfigure}\hfill
    \begin{subfigure}{0.24\linewidth}
        \includegraphics[width=\linewidth]{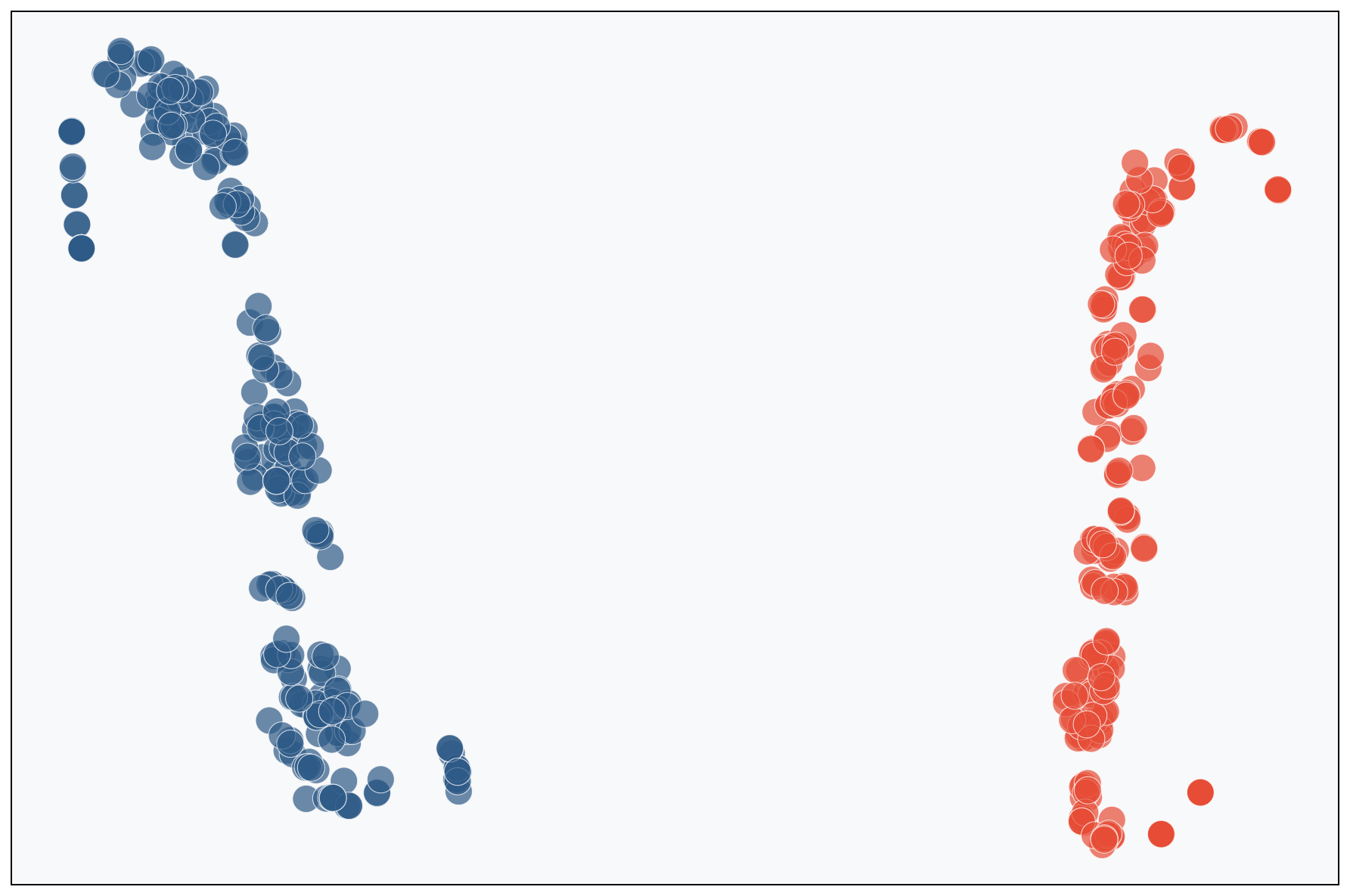}
        \caption*{k=23}
    \end{subfigure}\hfill
    \begin{subfigure}{0.24\linewidth}
        \includegraphics[width=\linewidth]{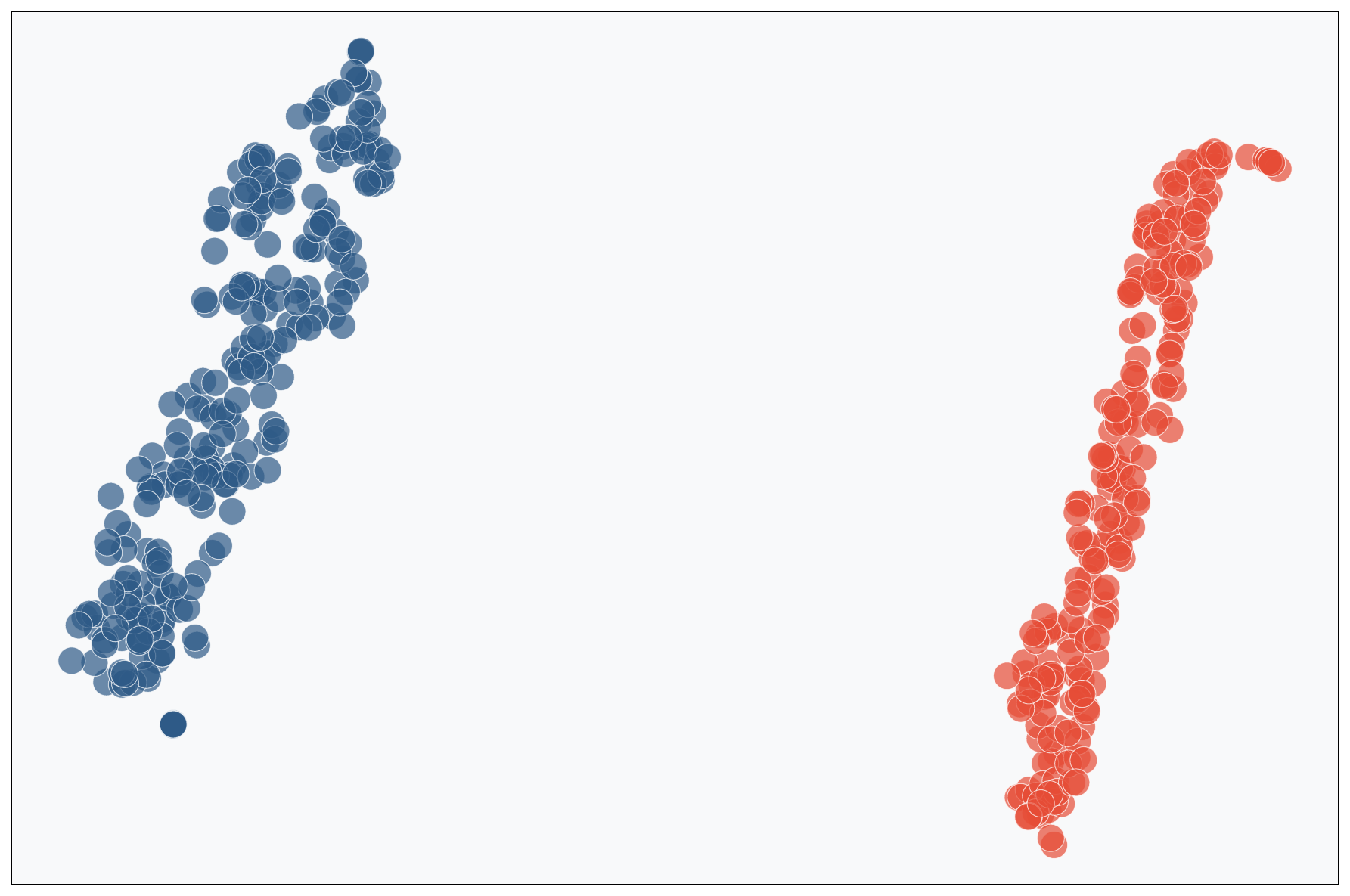}
        \caption*{k=24}
    \end{subfigure}

    \vspace{0.2em}
    
    \begin{subfigure}{0.24\linewidth}
        \includegraphics[width=\linewidth]{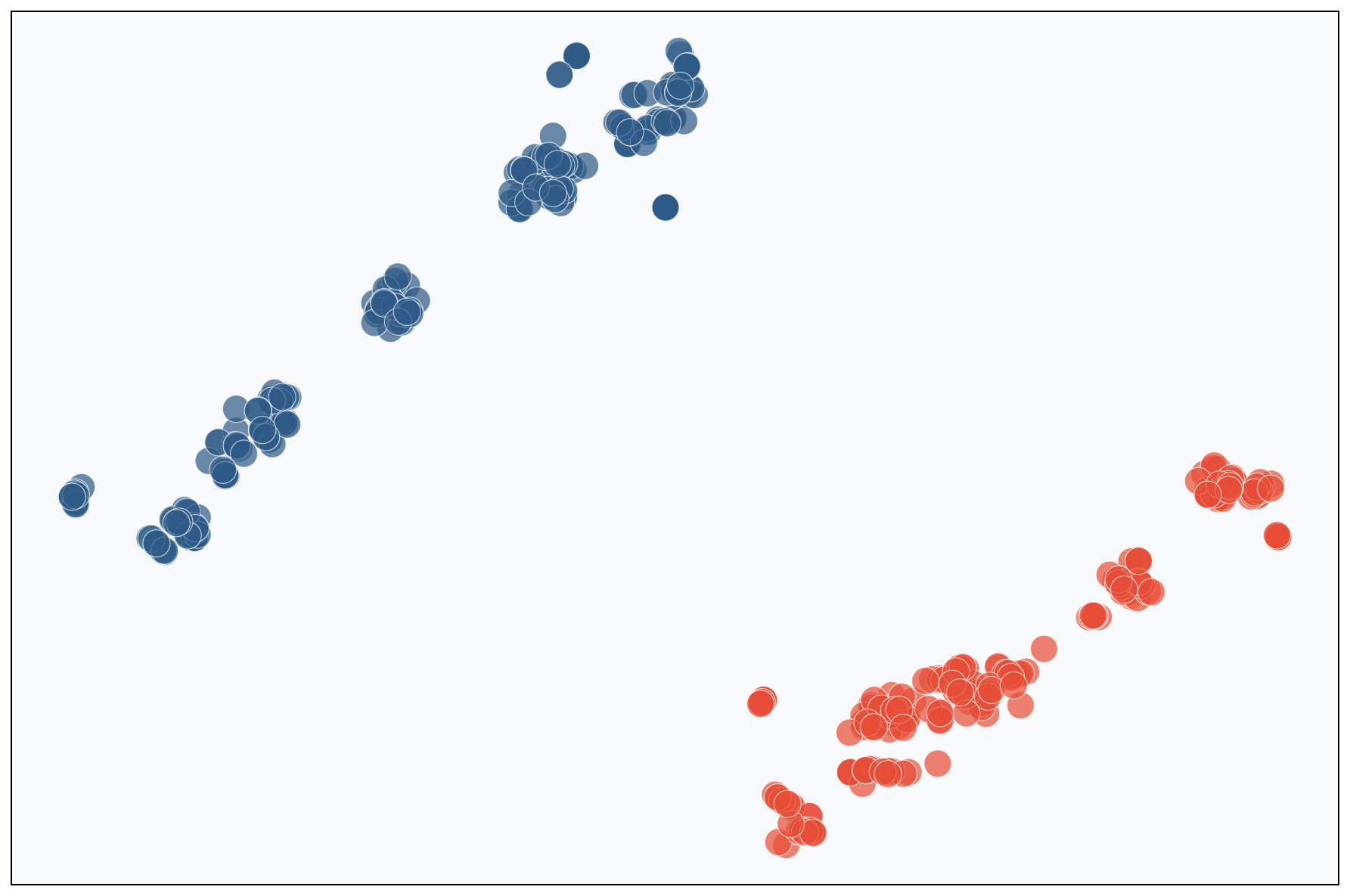}
        \caption*{k=25}
    \end{subfigure}\hfill
    \begin{subfigure}{0.24\linewidth}
        \includegraphics[width=\linewidth]{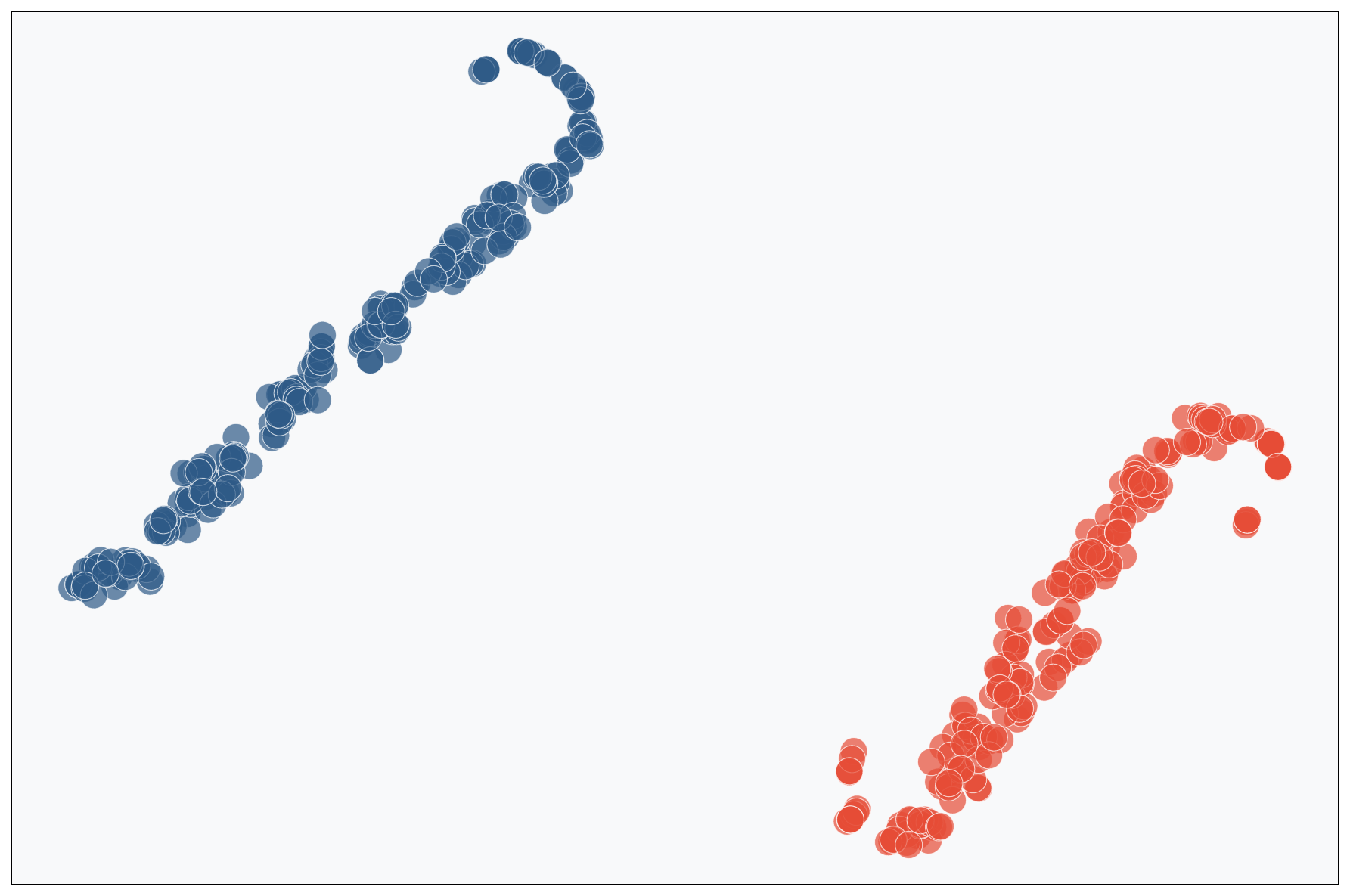}
        \caption*{k=26}
    \end{subfigure}\hfill
    \begin{subfigure}{0.24\linewidth}
        \includegraphics[width=\linewidth]{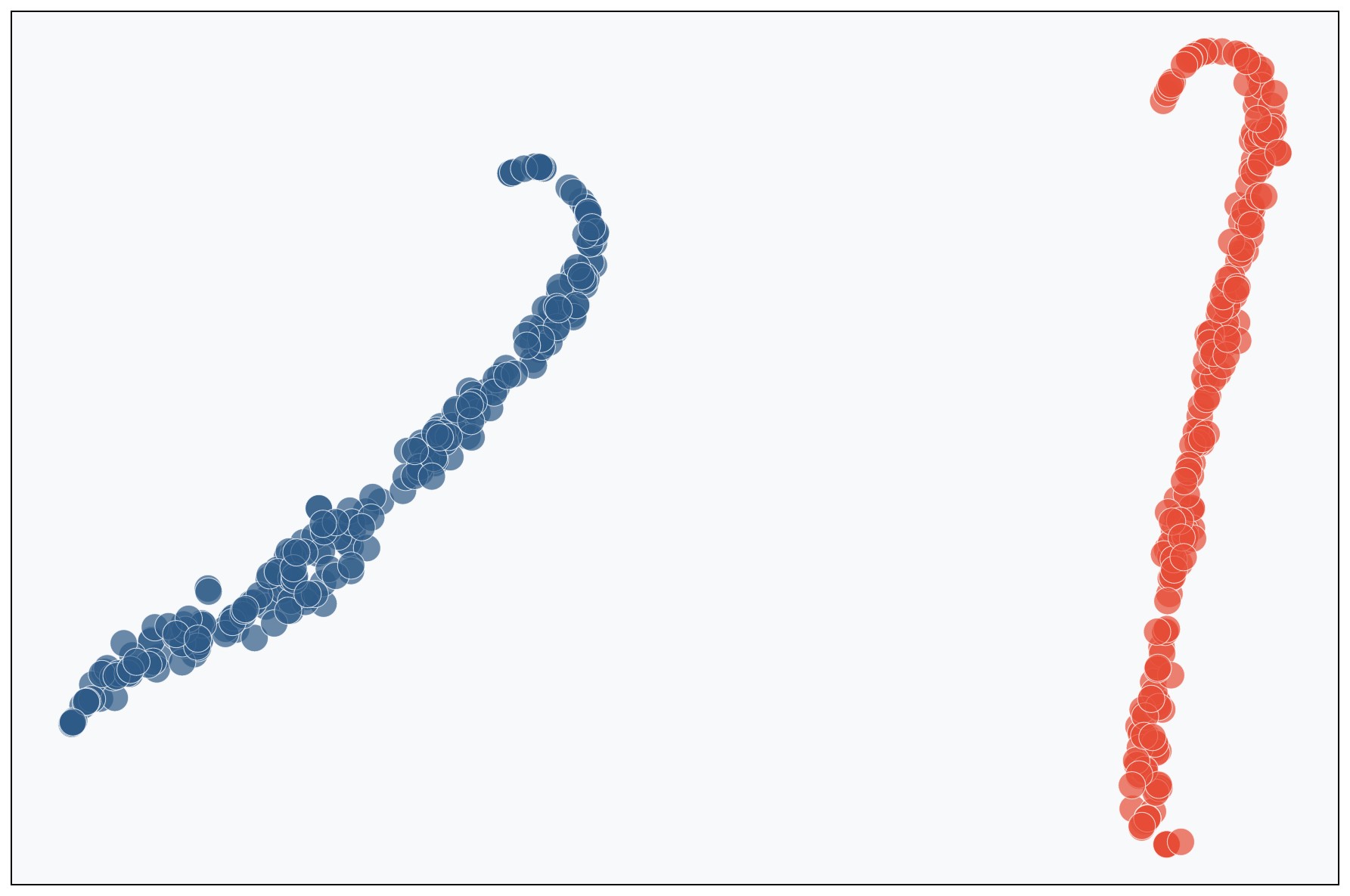}
        \caption*{k=27}
    \end{subfigure}\hfill
    \begin{subfigure}{0.24\linewidth}
        \includegraphics[width=\linewidth]{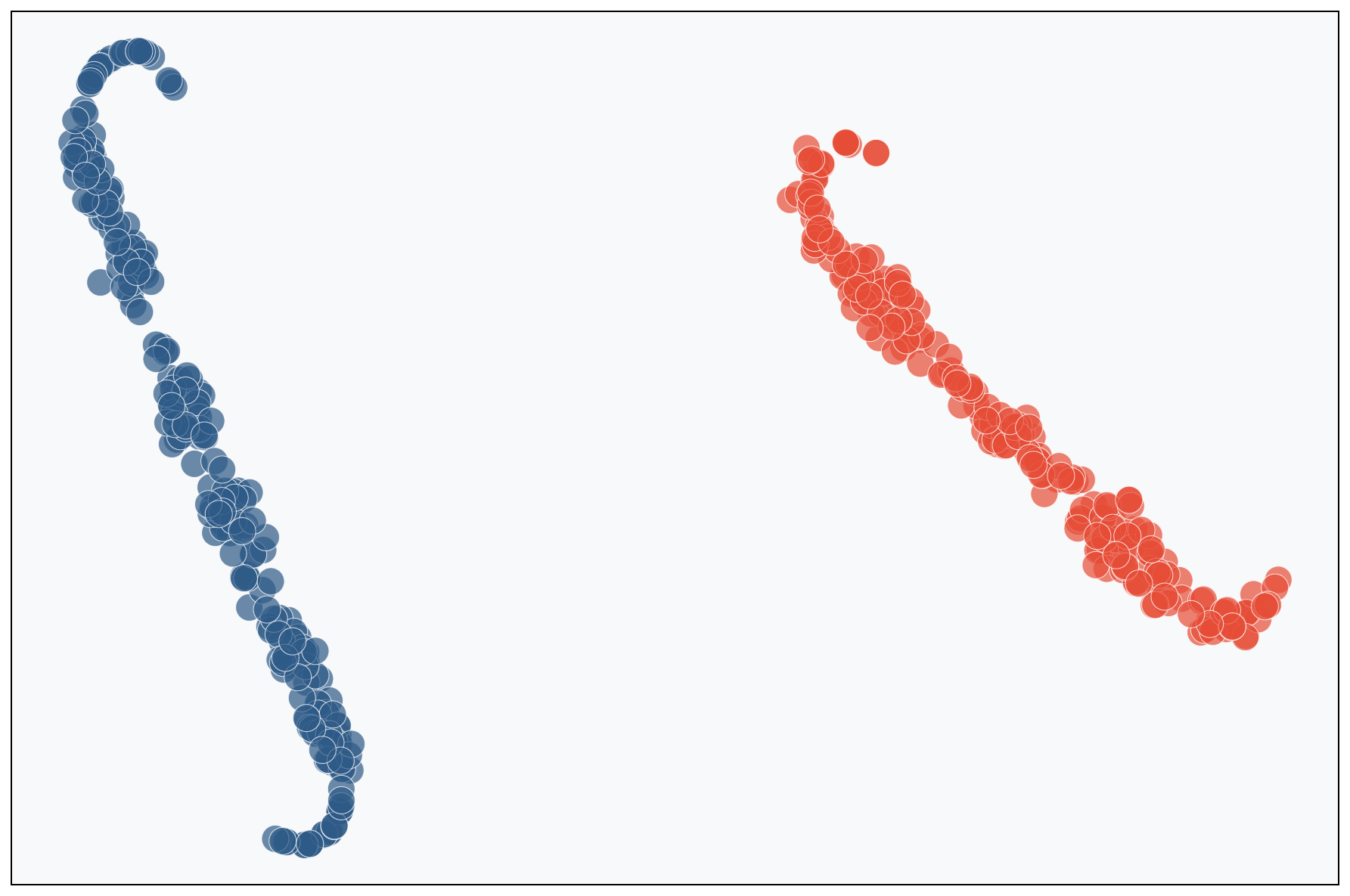}
        \caption*{k=28}
    \end{subfigure}

    \vspace{0.2em}
    
    \begin{subfigure}{0.24\linewidth}
        \includegraphics[width=\linewidth]{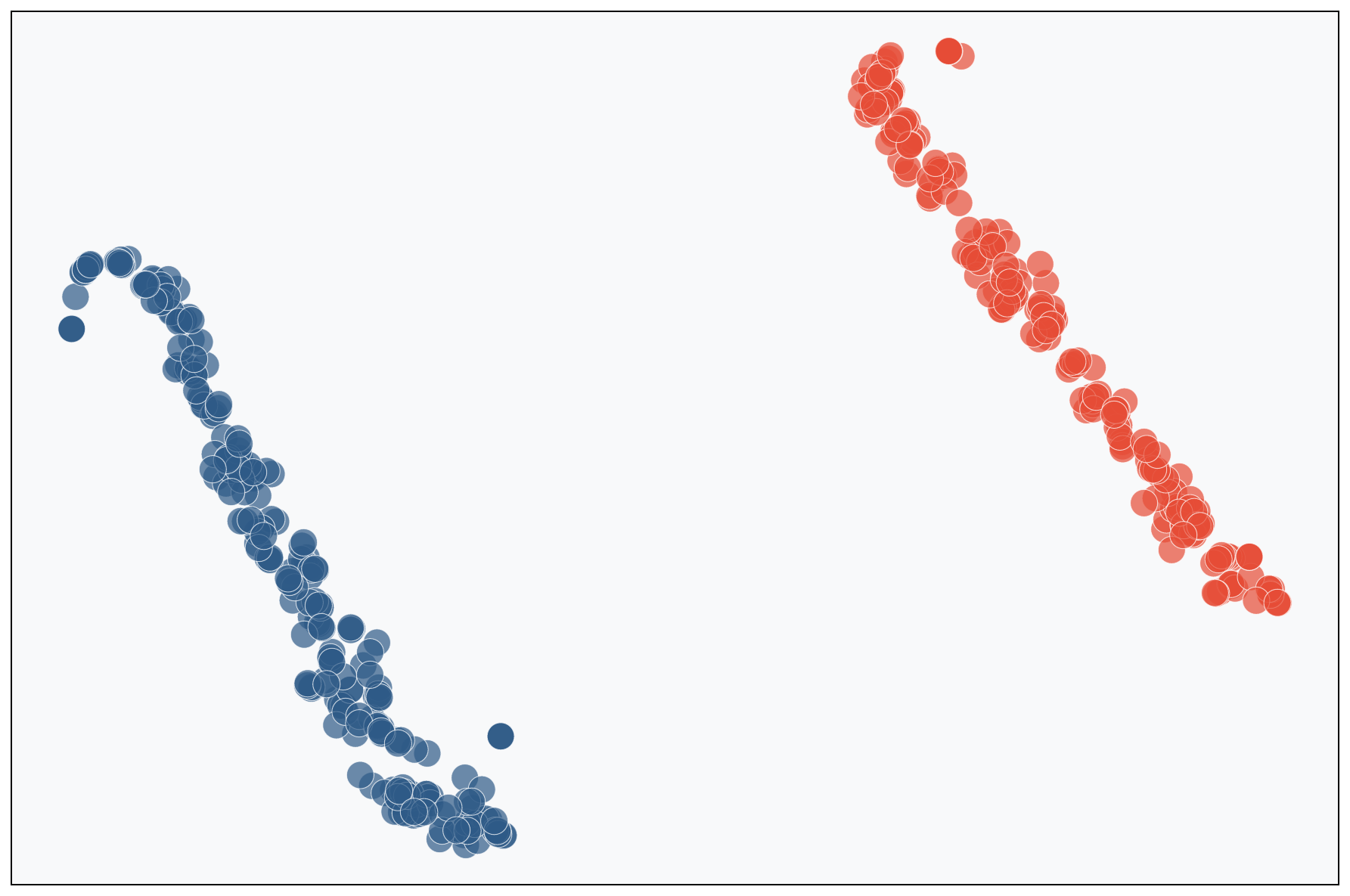}
        \caption*{k=29}
    \end{subfigure}\hfill
    \begin{subfigure}{0.24\linewidth}
        \includegraphics[width=\linewidth]{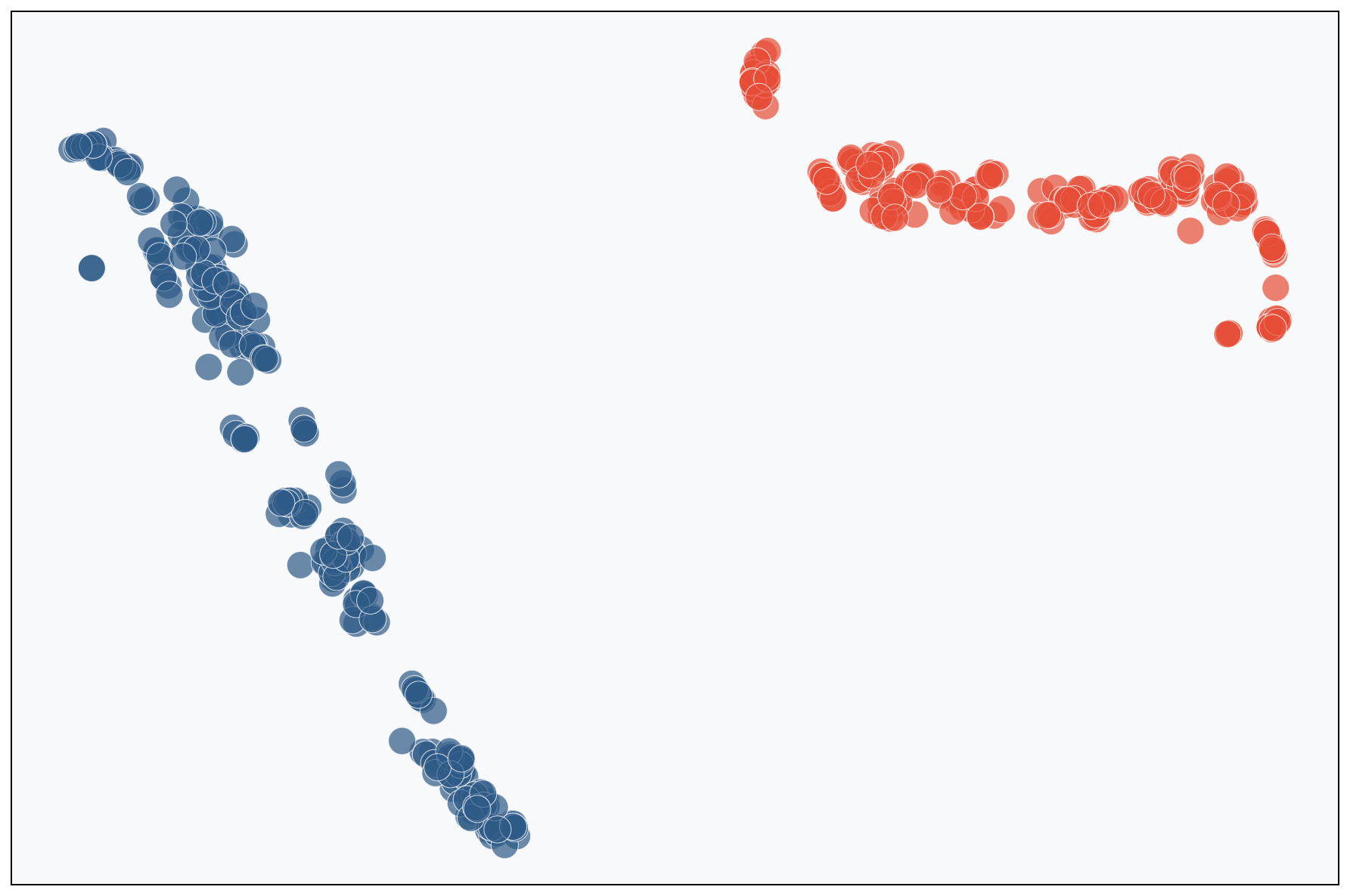}
        \caption*{k=30}
    \end{subfigure}\hfill
    \begin{subfigure}{0.24\linewidth}
        \includegraphics[width=\linewidth]{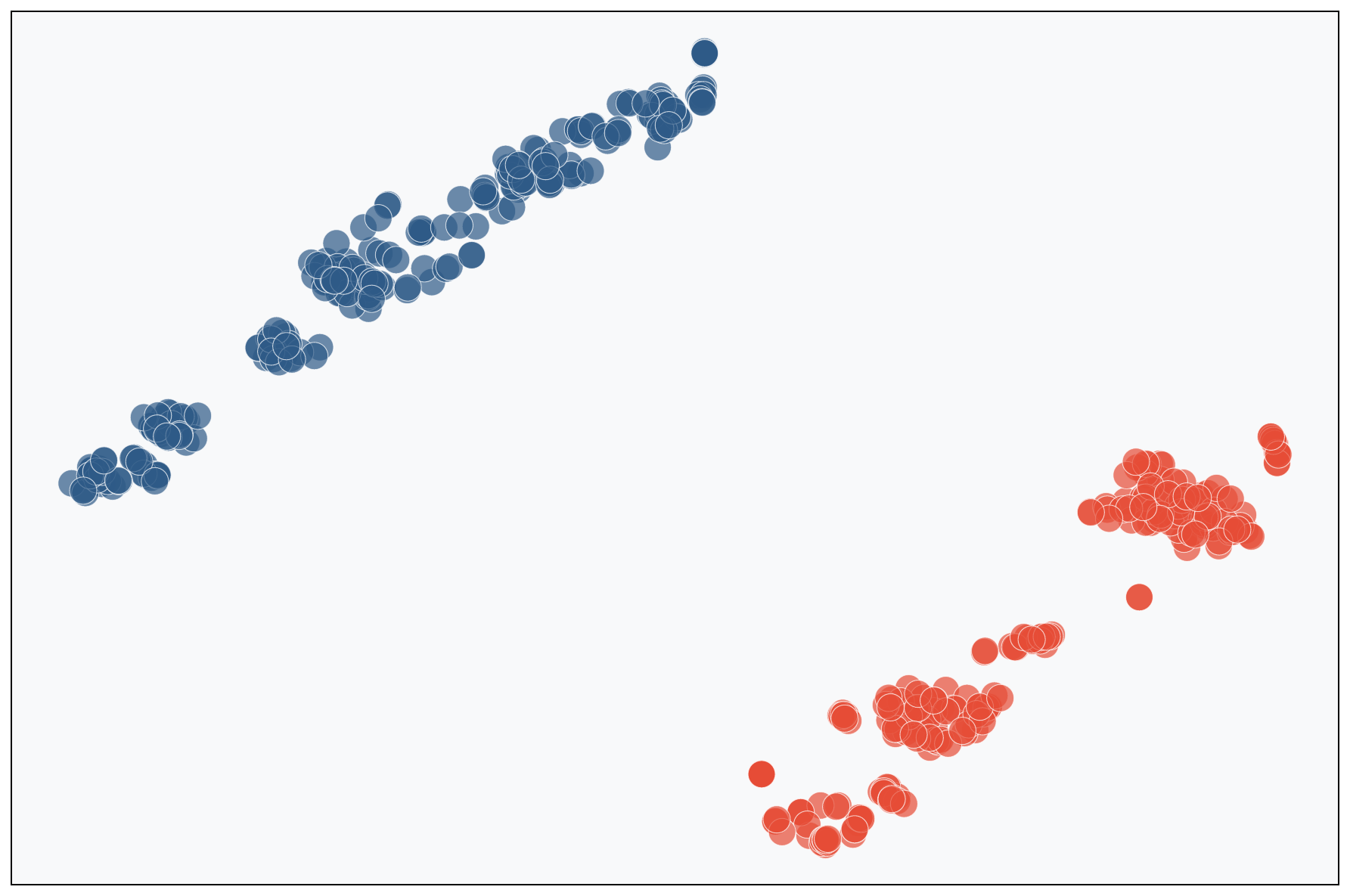}
        \caption*{k=31}
    \end{subfigure}\hfill
    \begin{subfigure}{0.24\linewidth}
        \includegraphics[width=\linewidth]{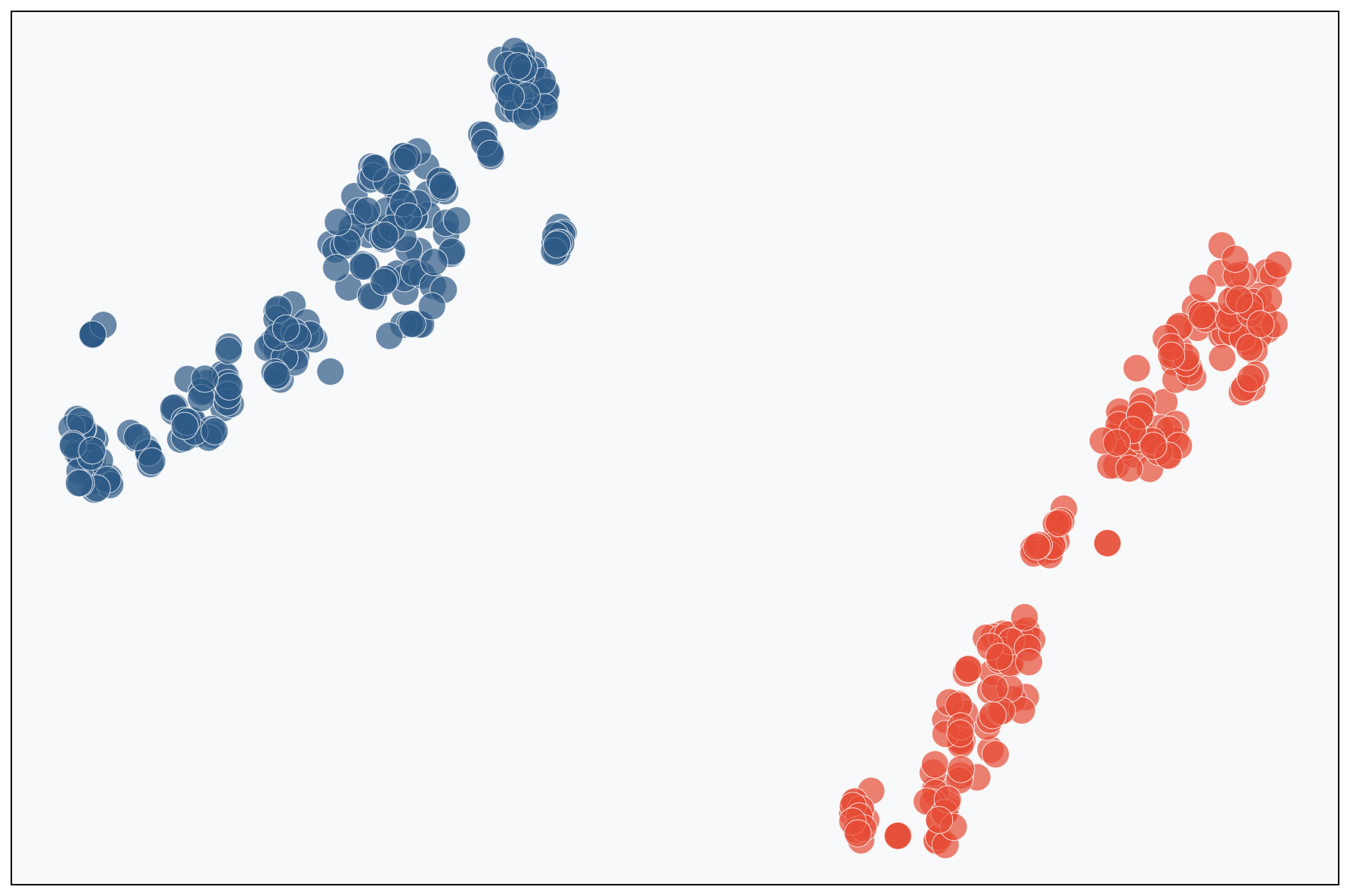}
        \caption*{k=32}
    \end{subfigure}
    \caption{t-SNE analysis on averaged $\beta$ on GYAFC (continued).}
    \label{fig:tsne_layers2_gyafc} 
\end{figure*}

\paragraph{GYAFC.} Figure \ref{fig:tsne_layers1_gyafc} and \ref{fig:tsne_layers2_gyafc} show the t-SNE plots of all 32 layers on GYAFC.

\begin{figure*}[htbp]
    \begin{minipage}{\textwidth}
        \centering
        \hspace{-5pt}
        \includegraphics[width=0.7\textwidth]{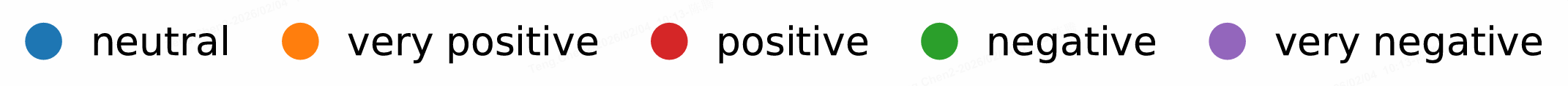}
    \end{minipage}
    \vspace{0.001em}
    \centering

    \begin{subfigure}{0.24\linewidth}
        \includegraphics[width=\linewidth]{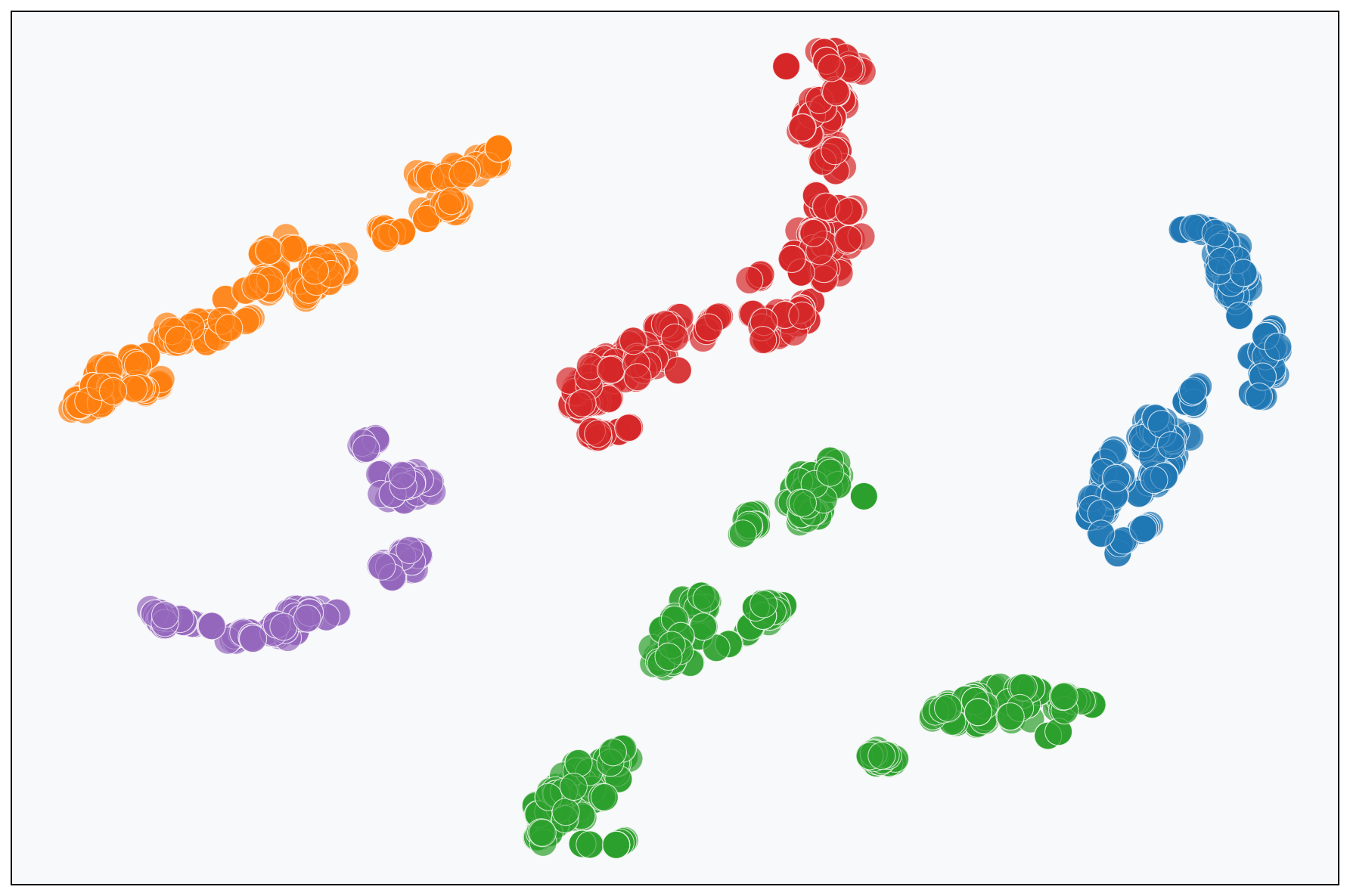}
        \caption*{k=1}
    \end{subfigure}\hfill
    \begin{subfigure}{0.24\linewidth}
        \includegraphics[width=\linewidth]{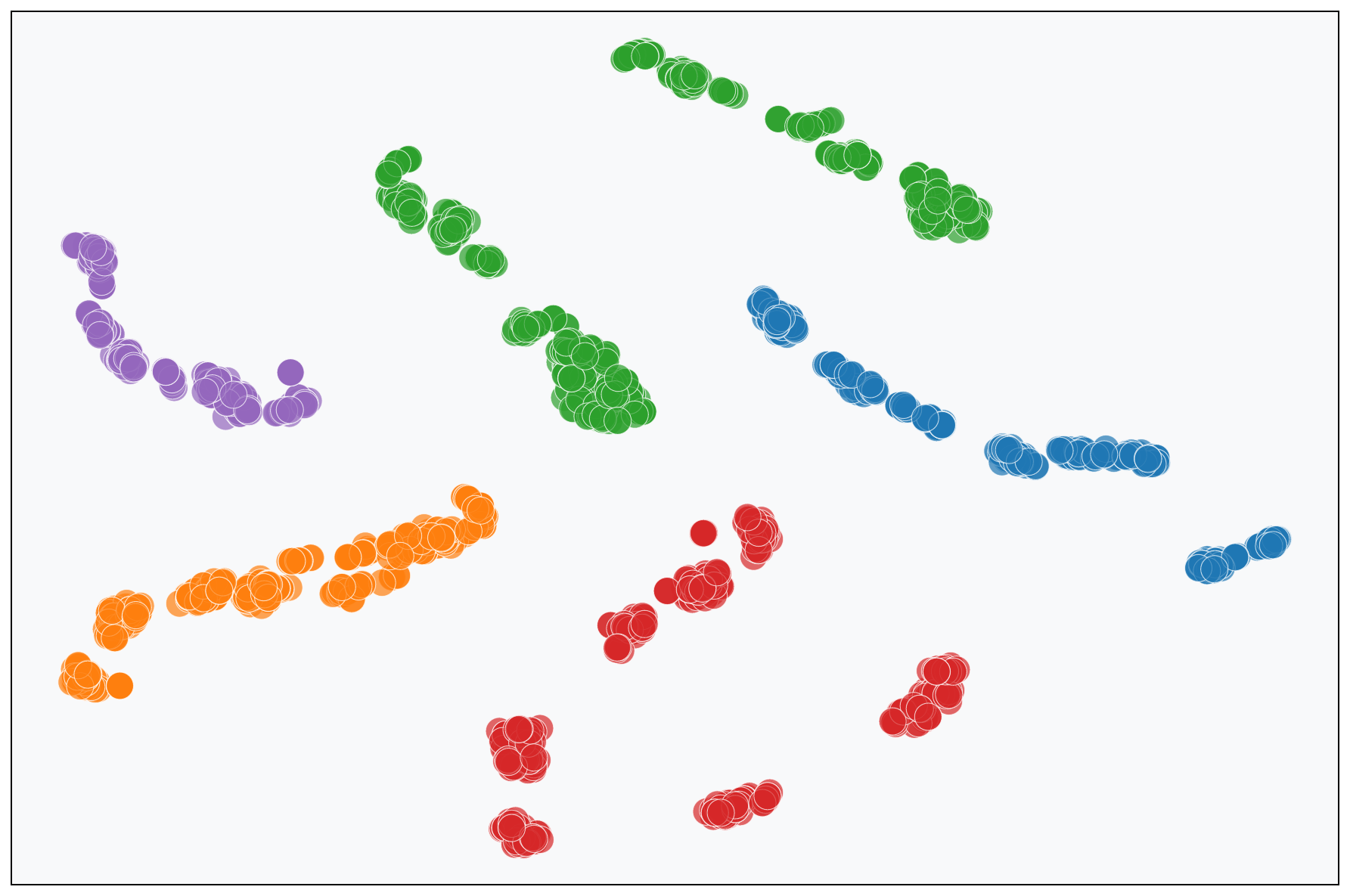}
        \caption*{k=2}
    \end{subfigure}\hfill
    \begin{subfigure}{0.24\linewidth}
        \includegraphics[width=\linewidth]{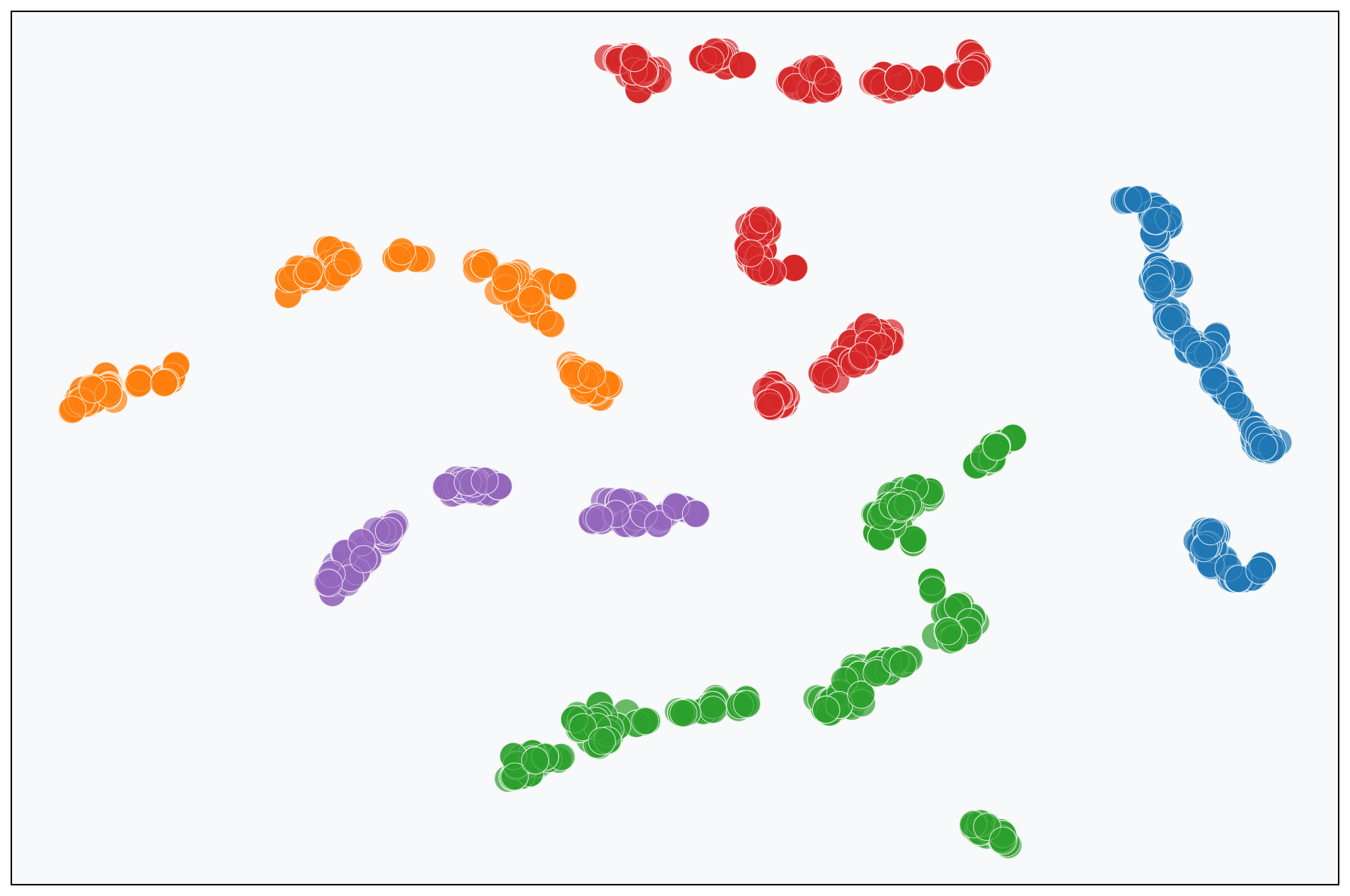}
        \caption*{k=3}
    \end{subfigure}\hfill
    \begin{subfigure}{0.24\linewidth}
        \includegraphics[width=\linewidth]{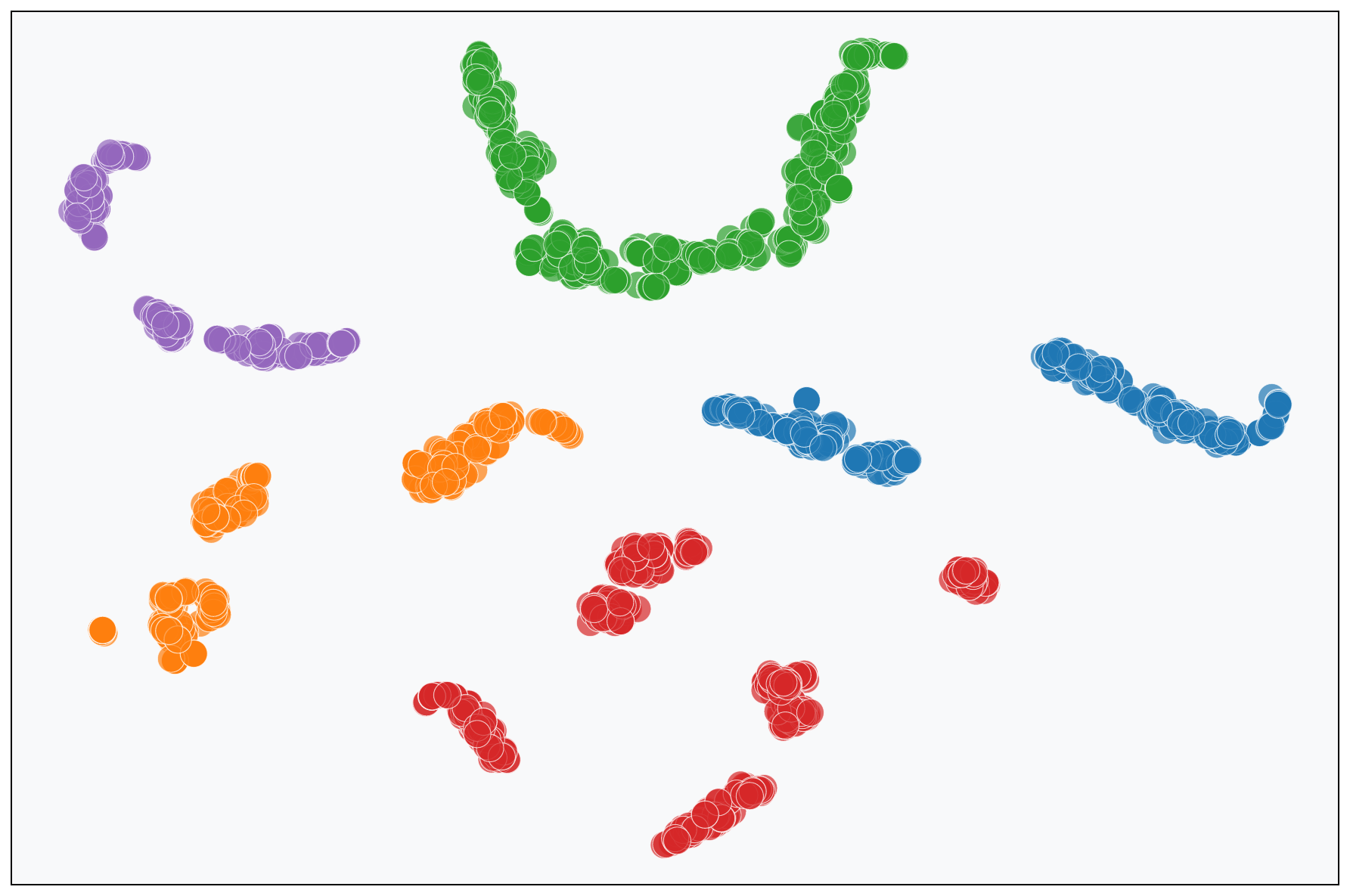}
        \caption*{k=4}
    \end{subfigure}
    
    \vspace{0.2em}
    
    \begin{subfigure}{0.24\linewidth}
        \includegraphics[width=\linewidth]{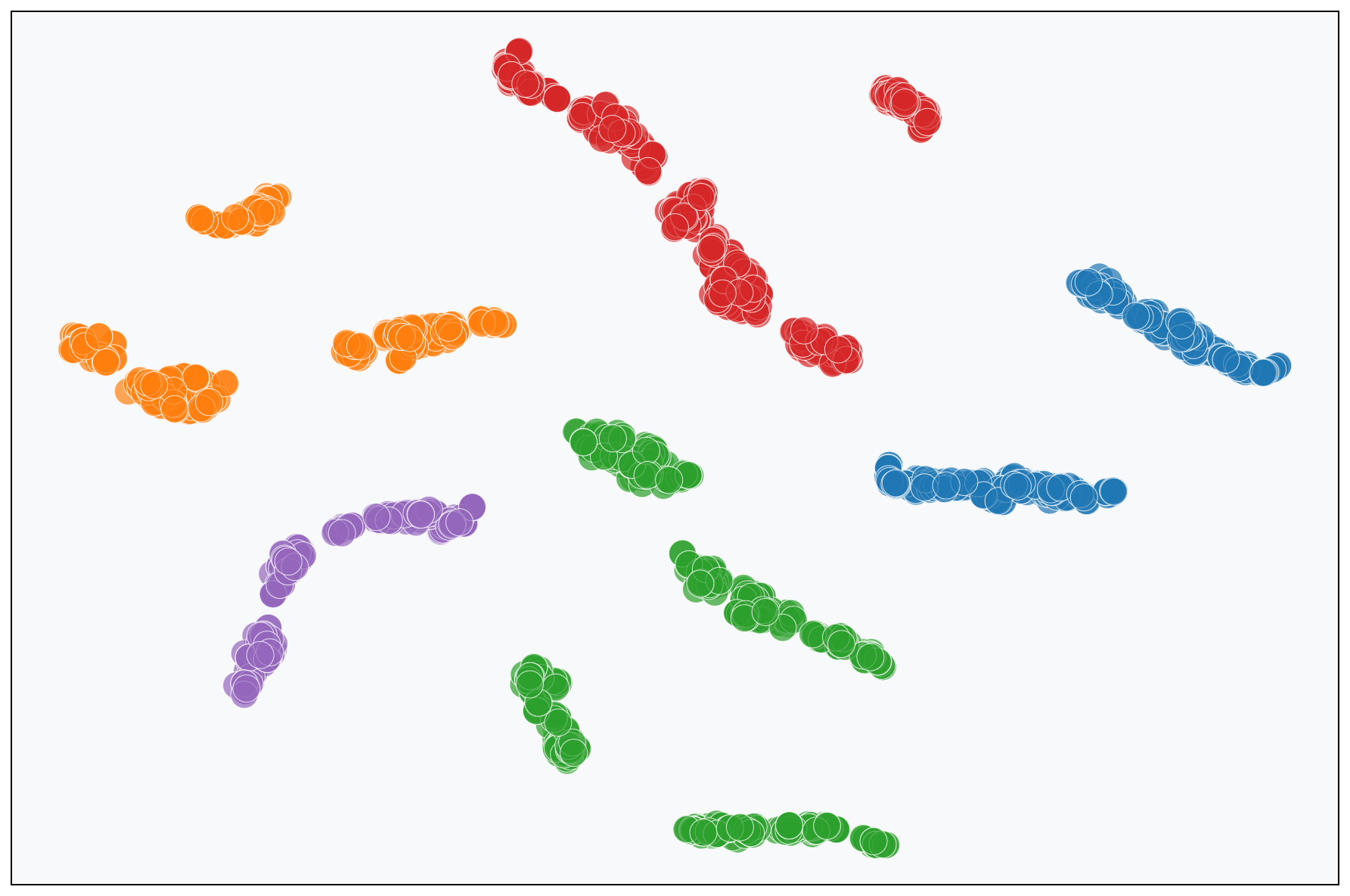}
        \caption*{k=5}
    \end{subfigure}\hfill
    \begin{subfigure}{0.24\linewidth}
        \includegraphics[width=\linewidth]{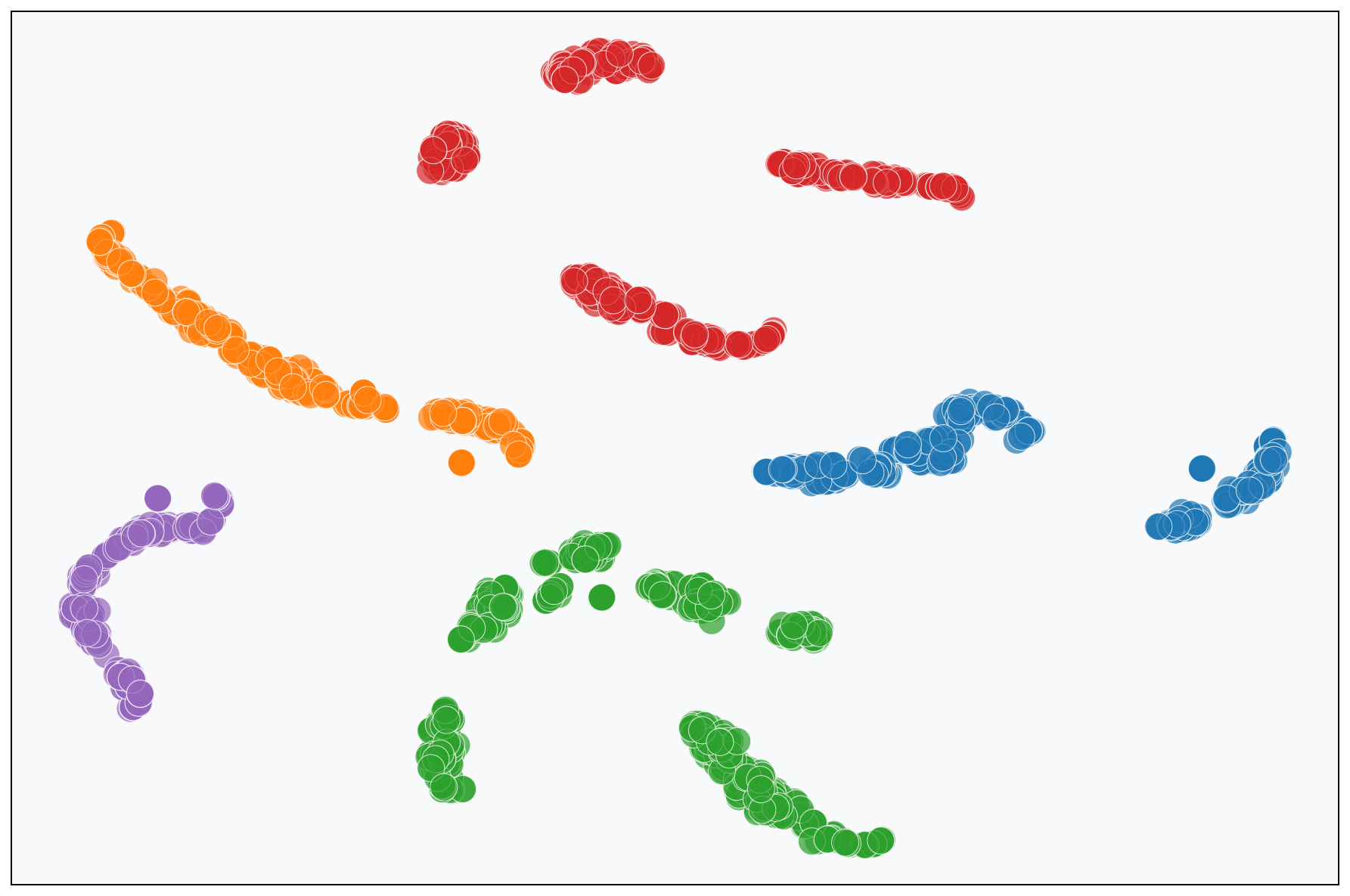}
        \caption*{k=6}
    \end{subfigure}\hfill
    \begin{subfigure}{0.24\linewidth}
        \includegraphics[width=\linewidth]{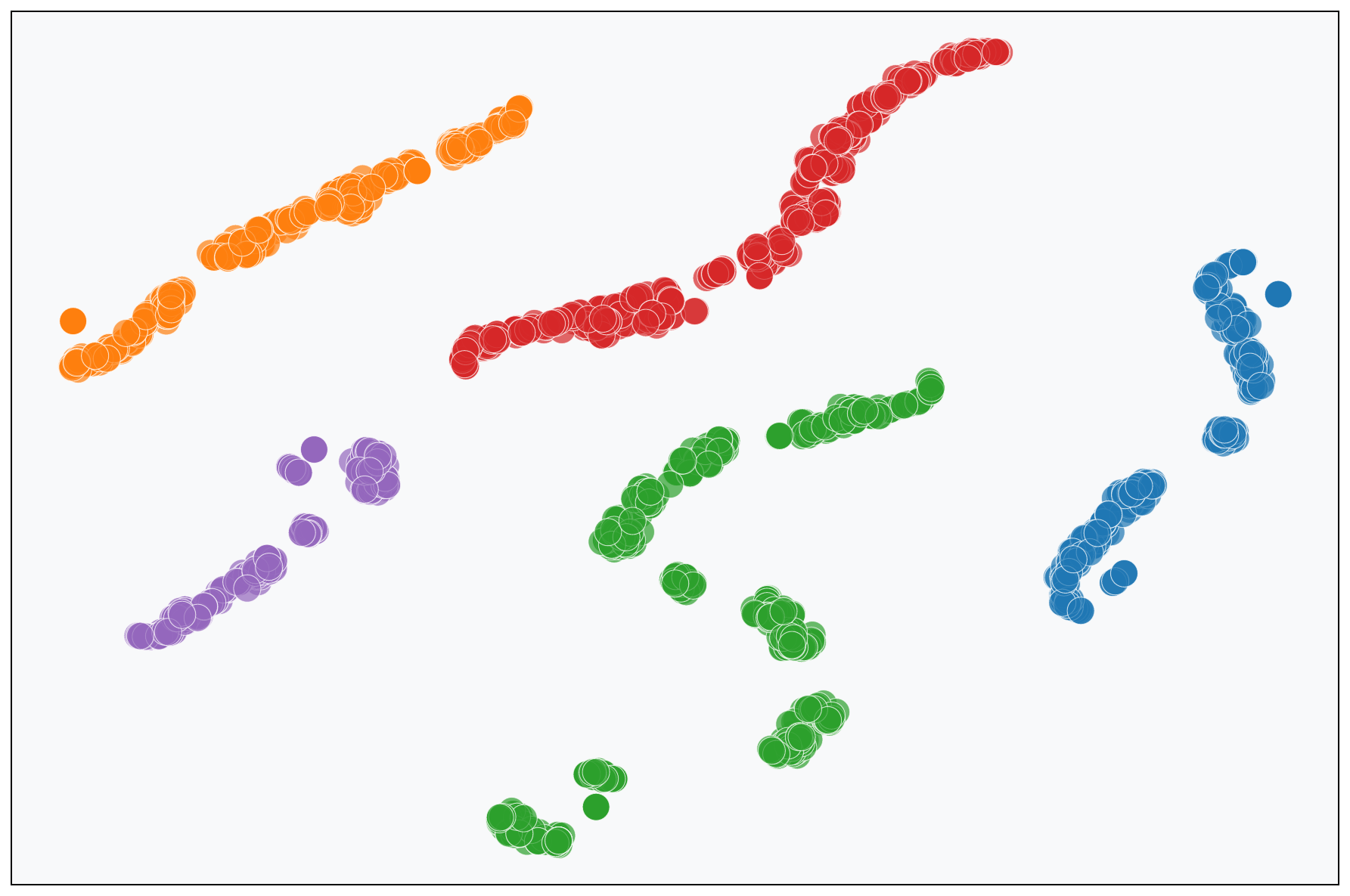}
        \caption*{k=7}
    \end{subfigure}\hfill
    \begin{subfigure}{0.24\linewidth}
        \includegraphics[width=\linewidth]{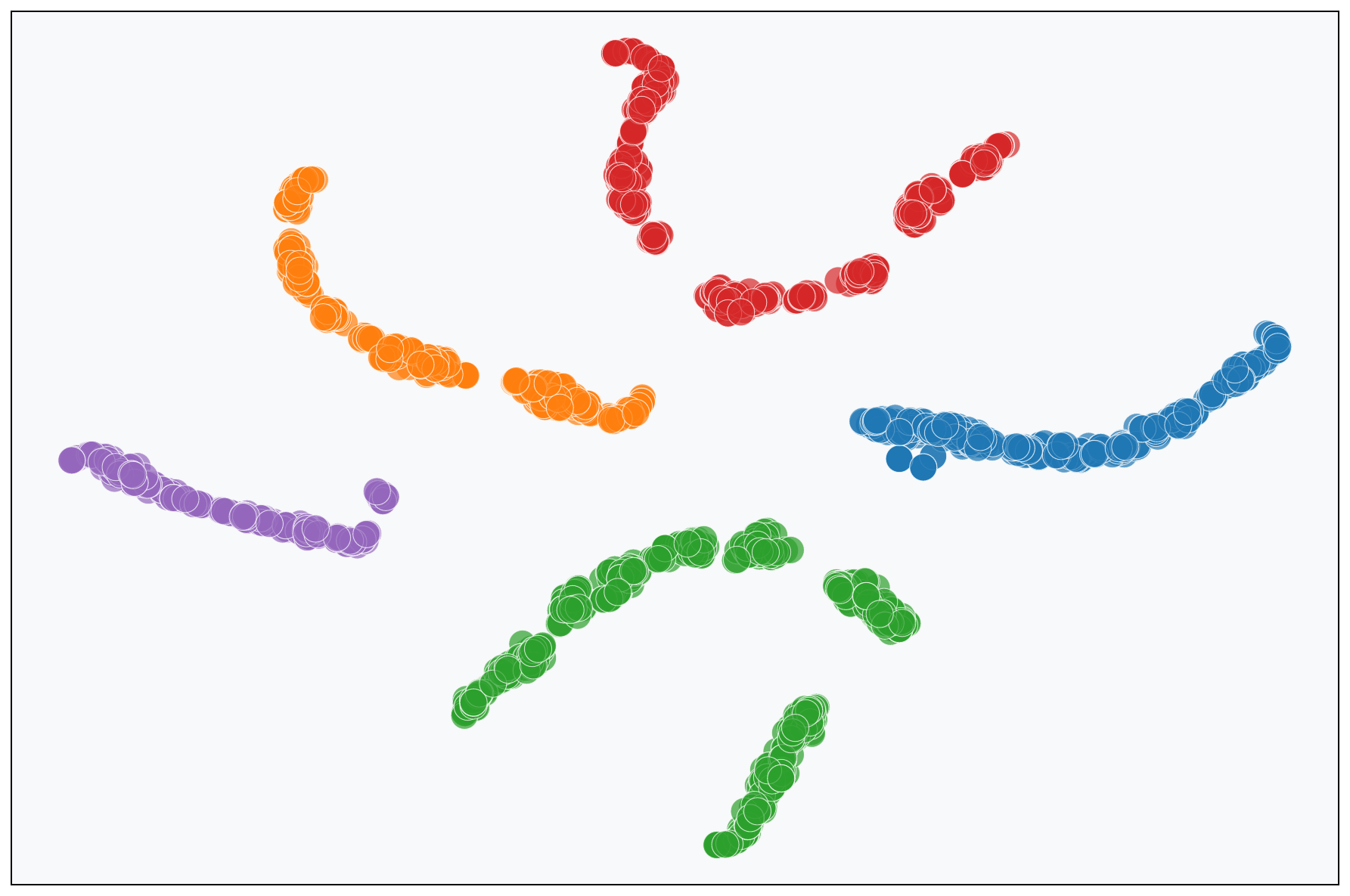}
        \caption*{k=8}
    \end{subfigure}
    
    \vspace{0.2em}
    
    \begin{subfigure}{0.24\linewidth}
        \includegraphics[width=\linewidth]{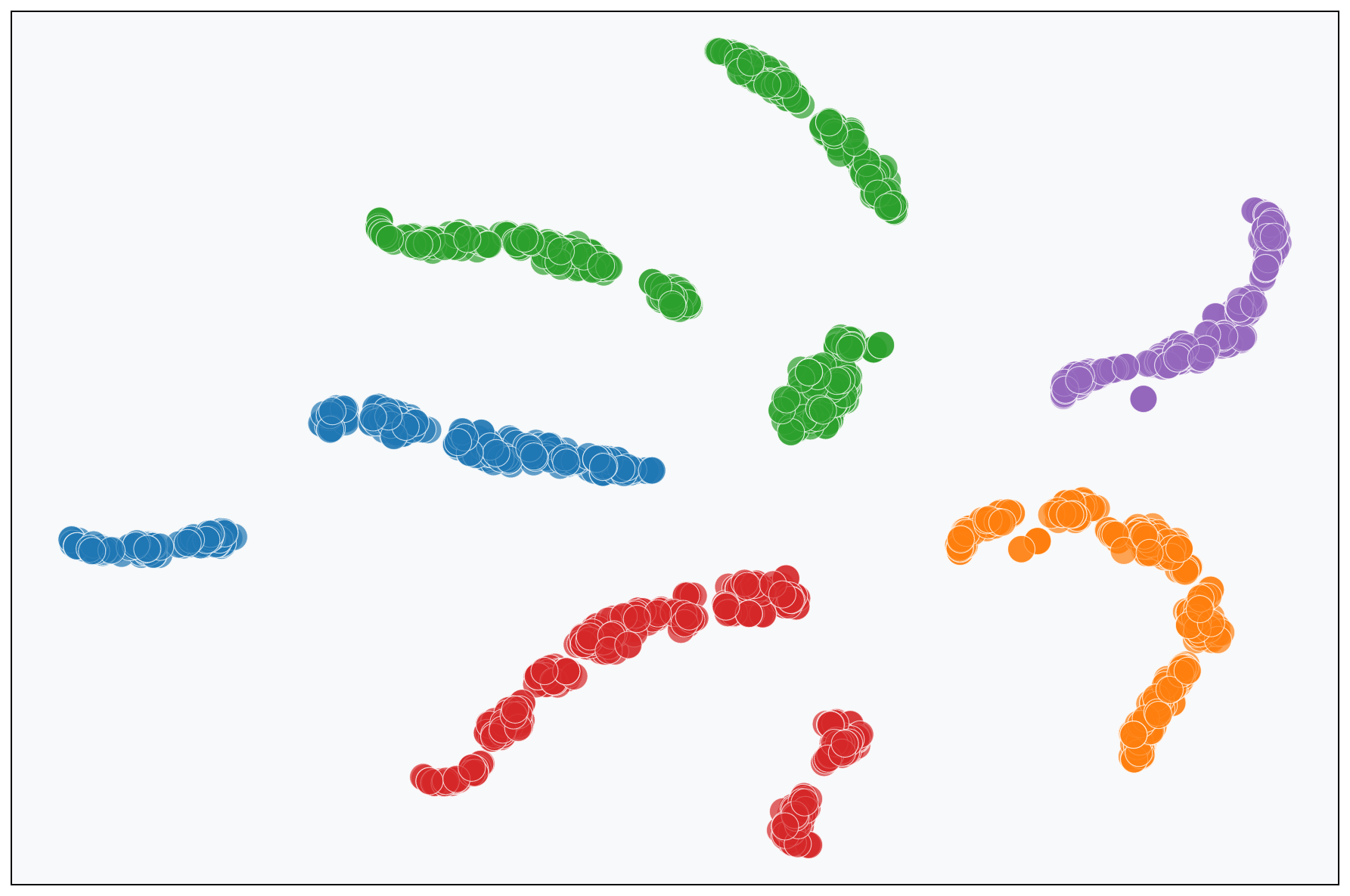}
        \caption*{k=9}
    \end{subfigure}\hfill
    \begin{subfigure}{0.24\linewidth}
        \includegraphics[width=\linewidth]{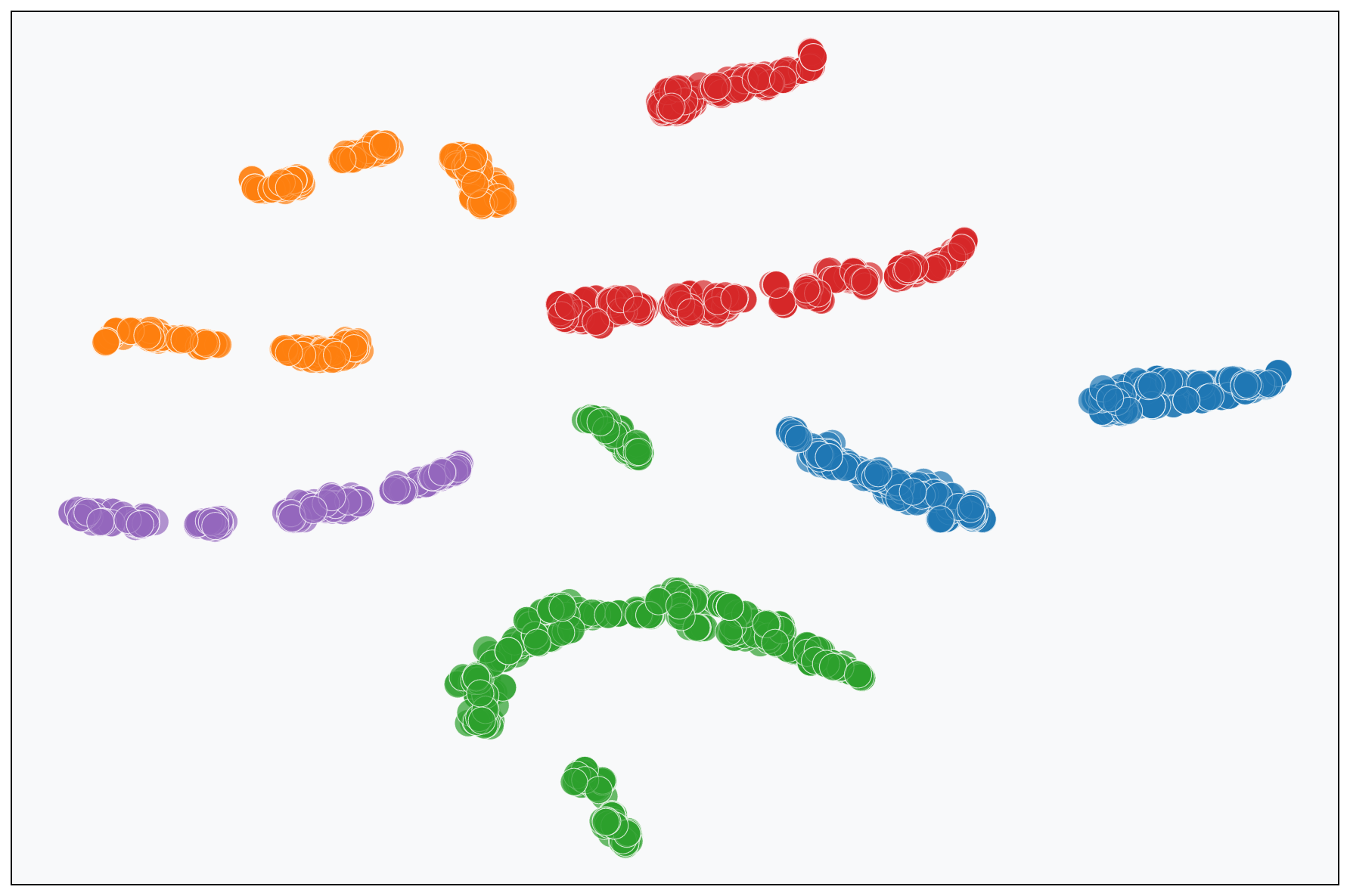}
        \caption*{k=10}
    \end{subfigure}\hfill
    \begin{subfigure}{0.24\linewidth}
        \includegraphics[width=\linewidth]{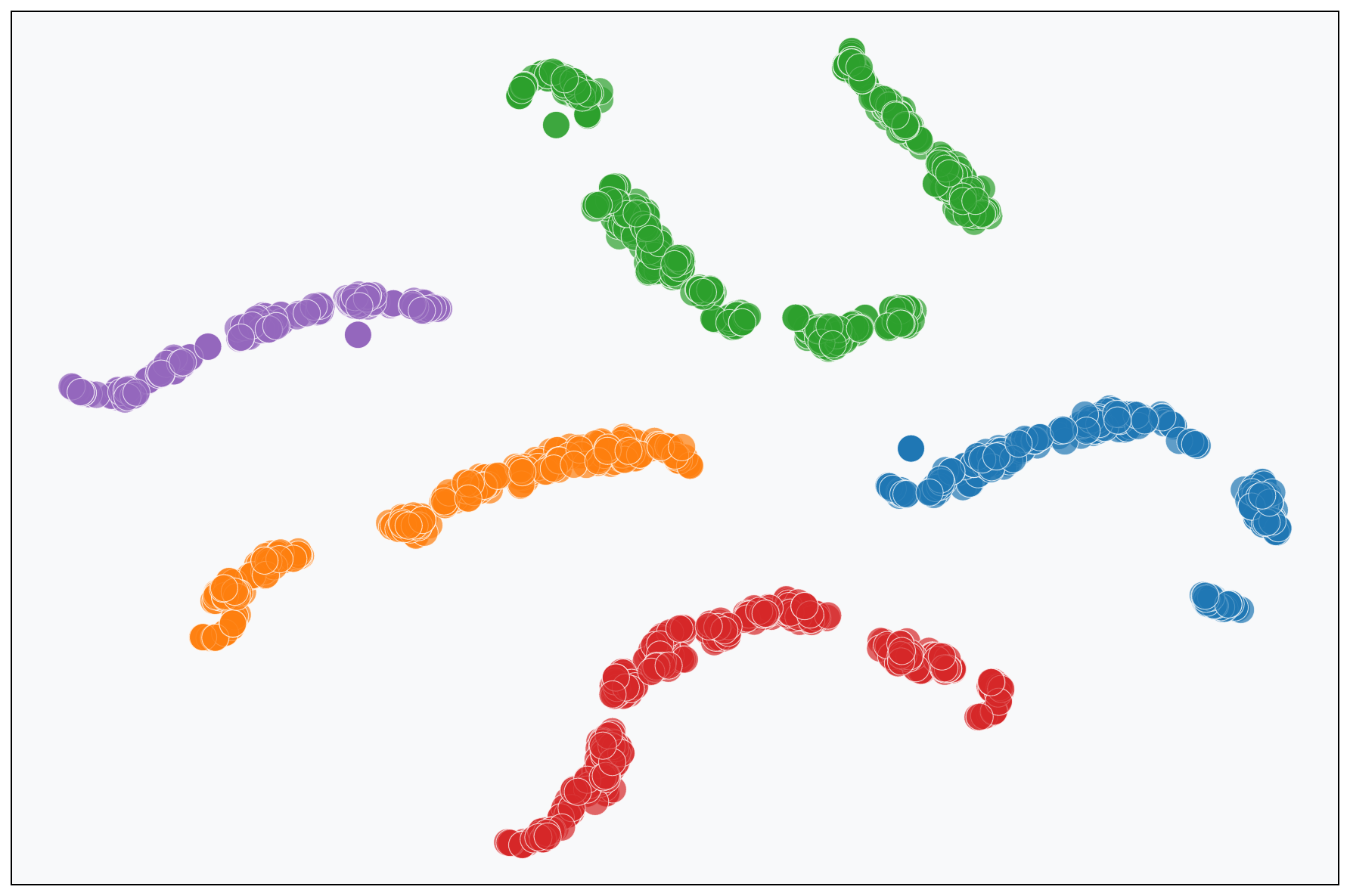}
        \caption*{k=11}
    \end{subfigure}\hfill
    \begin{subfigure}{0.24\linewidth}
        \includegraphics[width=\linewidth]{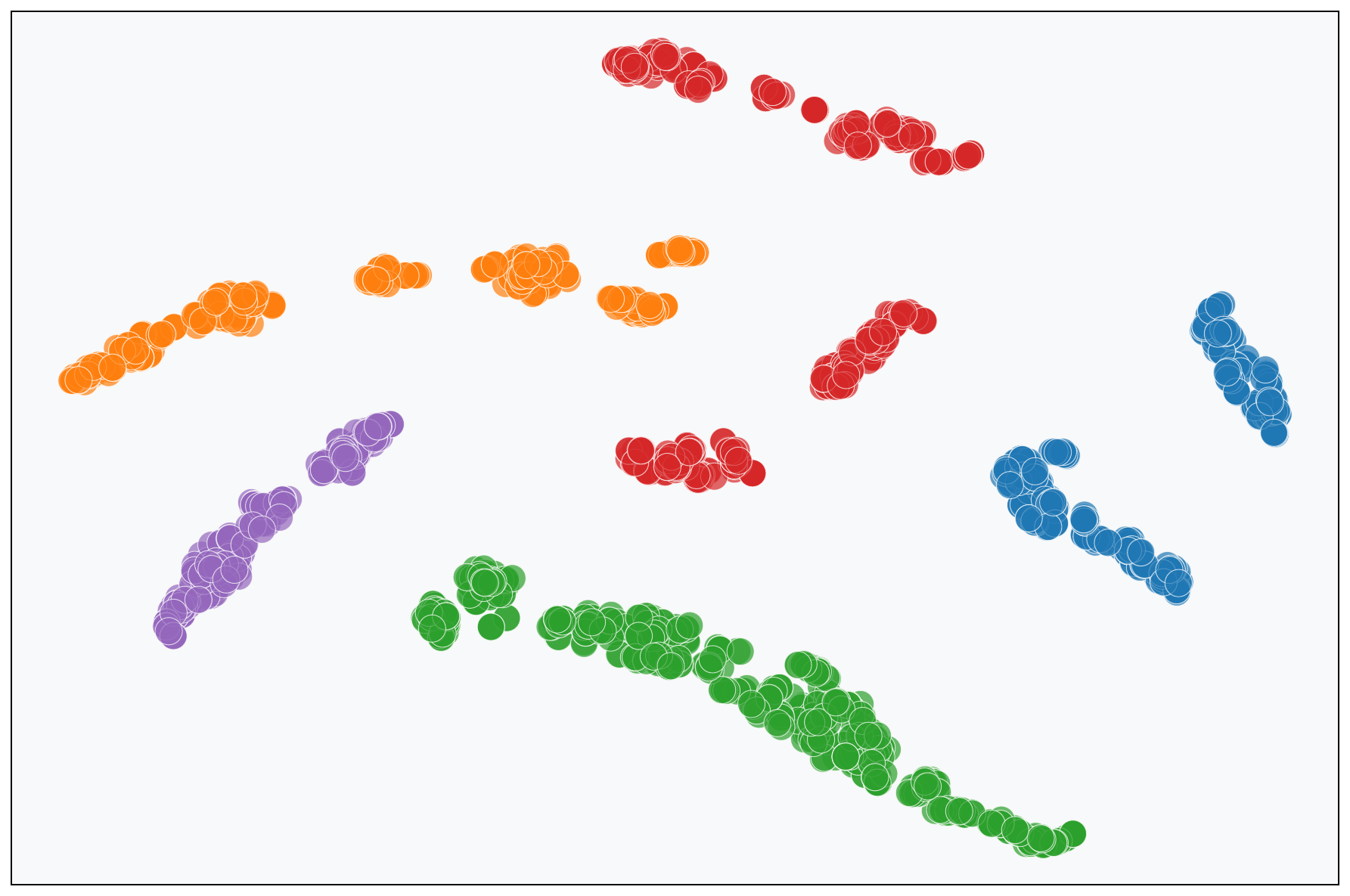}
        \caption*{k=12}
    \end{subfigure}
    
    \vspace{0.2em}
    
    \begin{subfigure}{0.24\linewidth}
        \includegraphics[width=\linewidth]{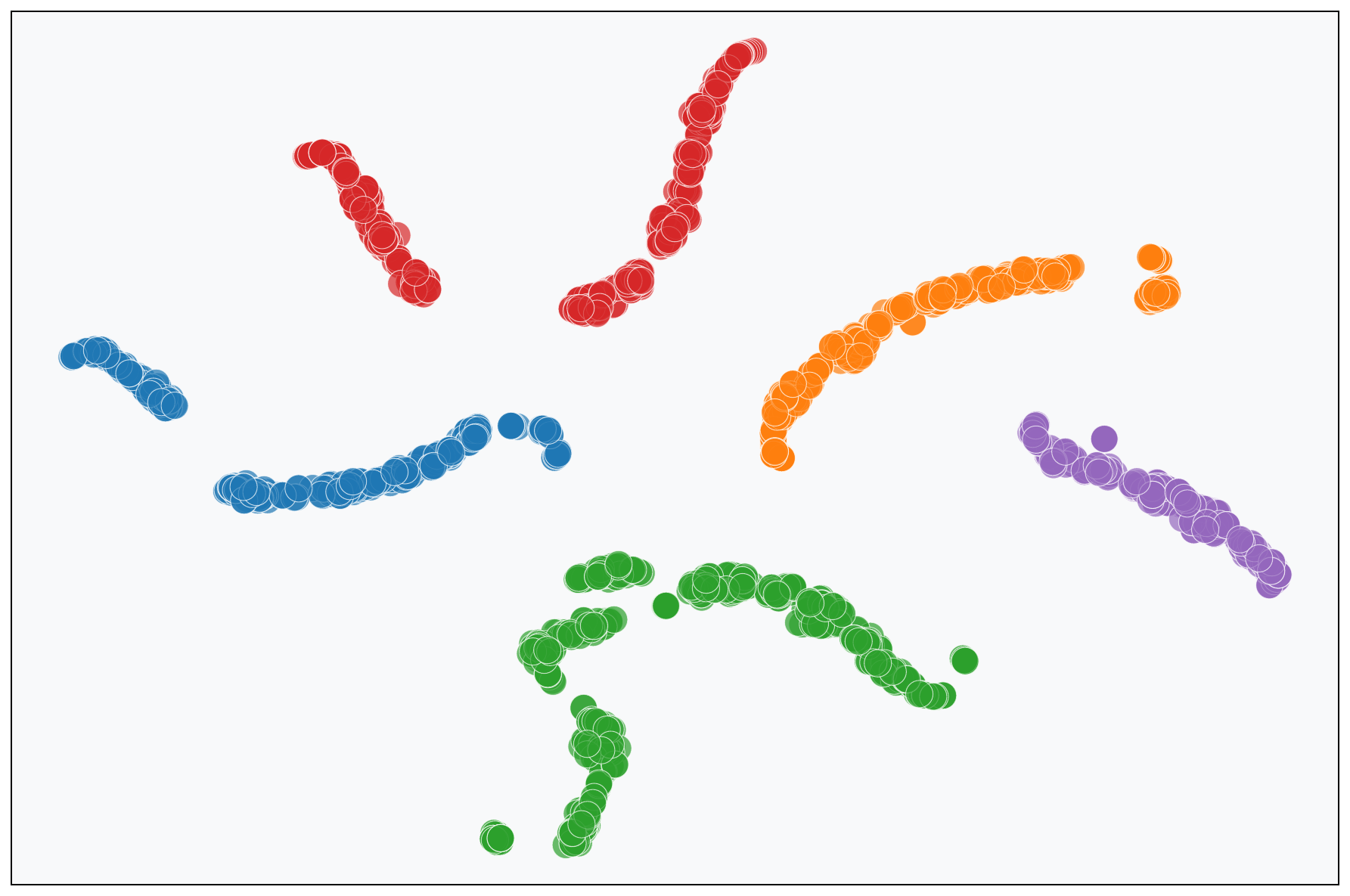}
        \caption*{k=13}
    \end{subfigure}\hfill
    \begin{subfigure}{0.24\linewidth}
        \includegraphics[width=\linewidth]{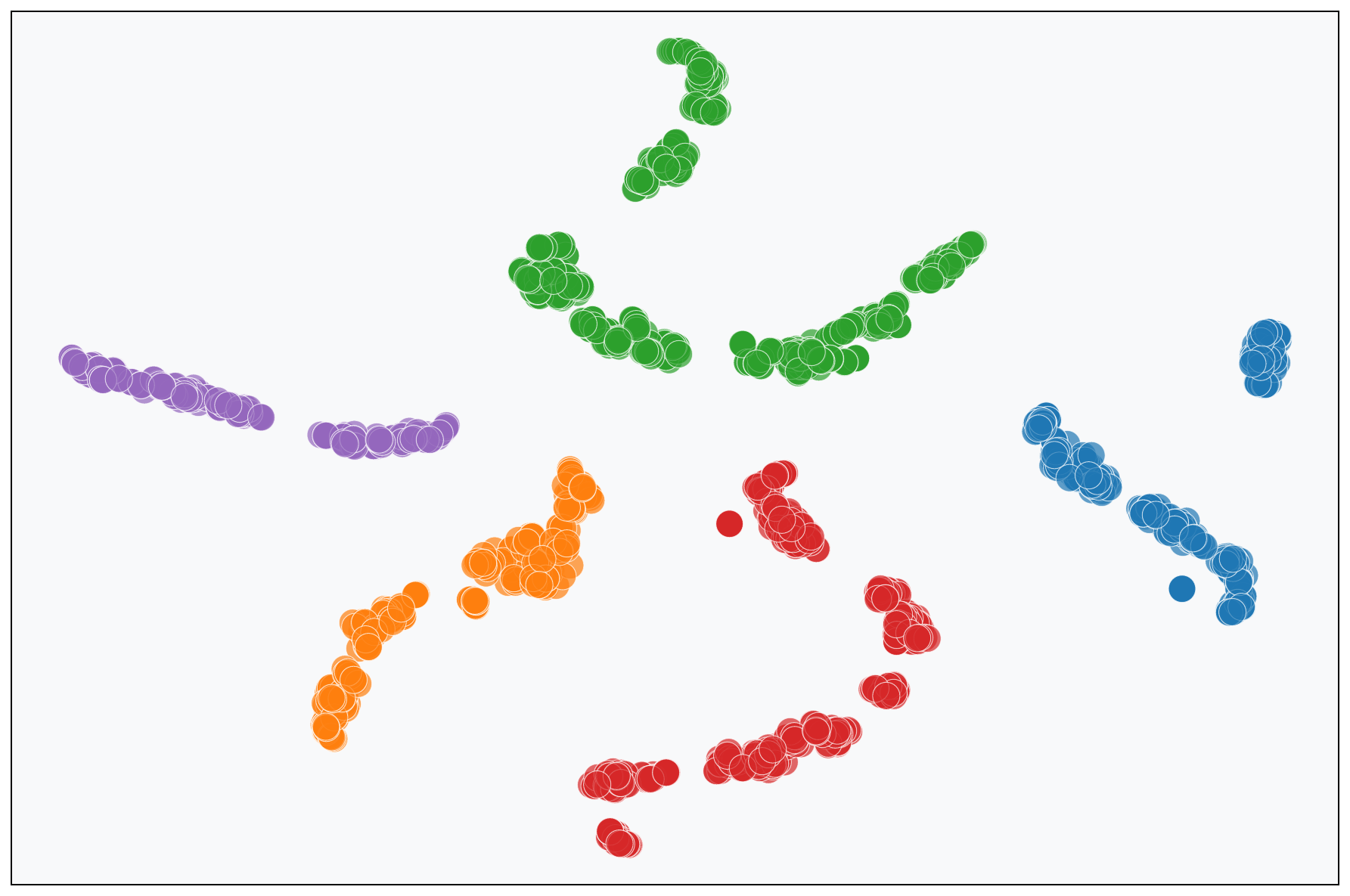}
        \caption*{k=14}
    \end{subfigure}\hfill
    \begin{subfigure}{0.24\linewidth}
        \includegraphics[width=\linewidth]{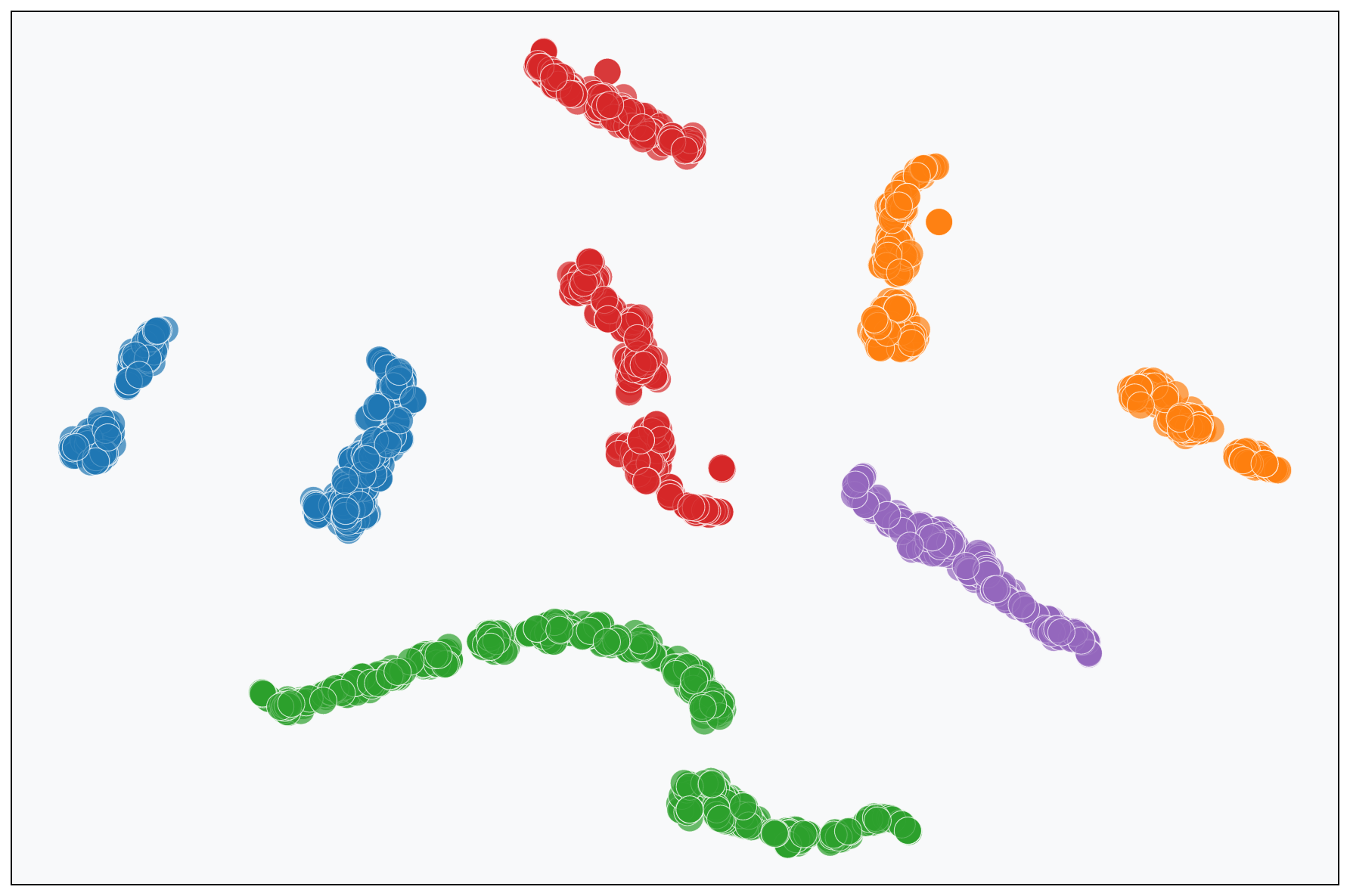}
        \caption*{k=15}
    \end{subfigure}\hfill
    \begin{subfigure}{0.24\linewidth}
        \includegraphics[width=\linewidth]{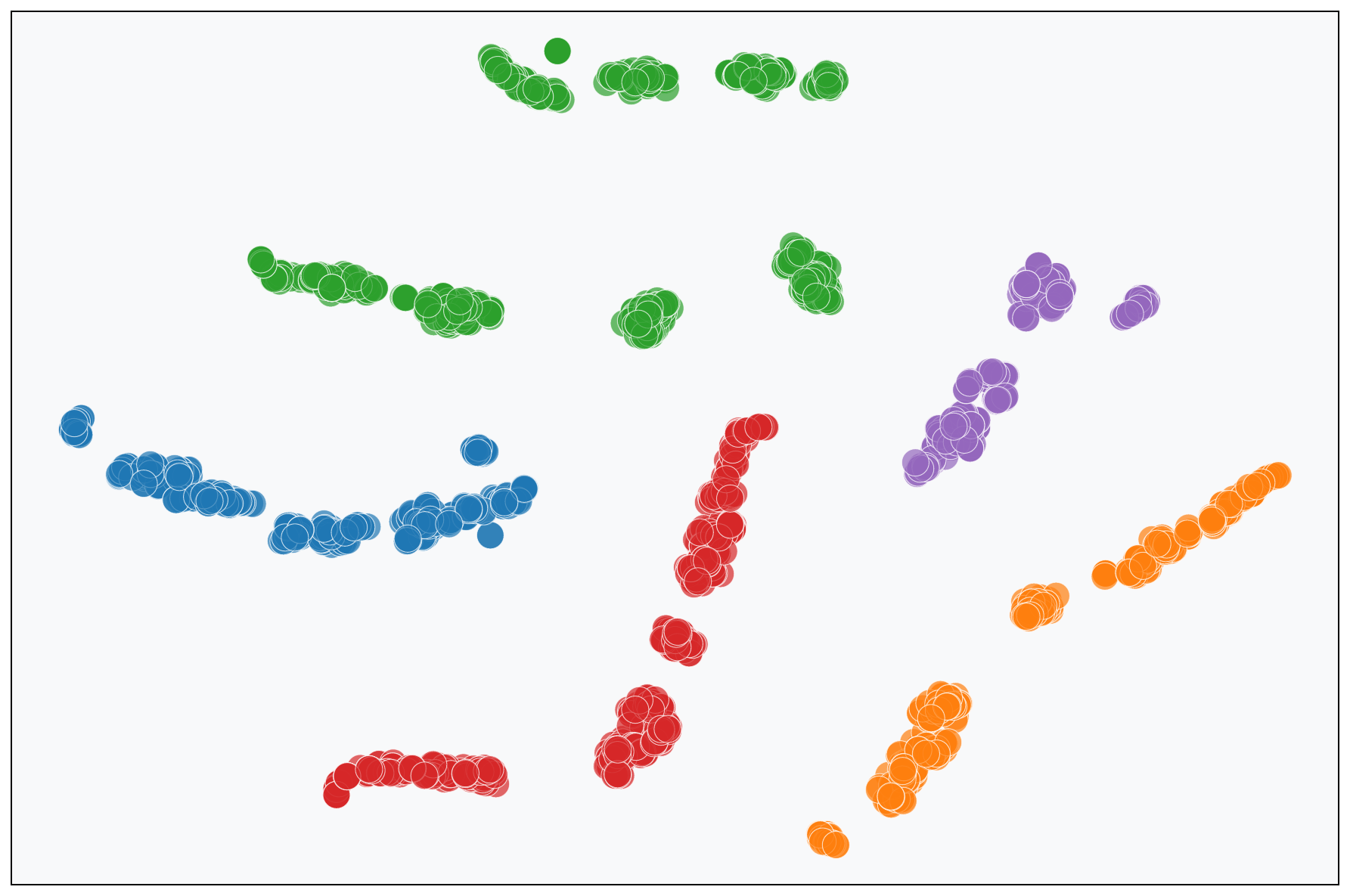}
        \caption*{k=16}
    \end{subfigure}
    
    \caption{t-SNE analysis of averaged $\beta$ on SST (first 16 layers).}
    \label{fig:tsne_layers1_sst} 
\end{figure*}

\begin{figure*}[htbp]
    \begin{minipage}{\textwidth}
        \centering
        \hspace{-5pt}
        \includegraphics[width=0.7\textwidth]{tSNE_SST/sst_legend.png}
    \end{minipage}
    \vspace{0.001em}

    \centering
    
    \begin{subfigure}{0.24\linewidth}
        \includegraphics[width=\linewidth]{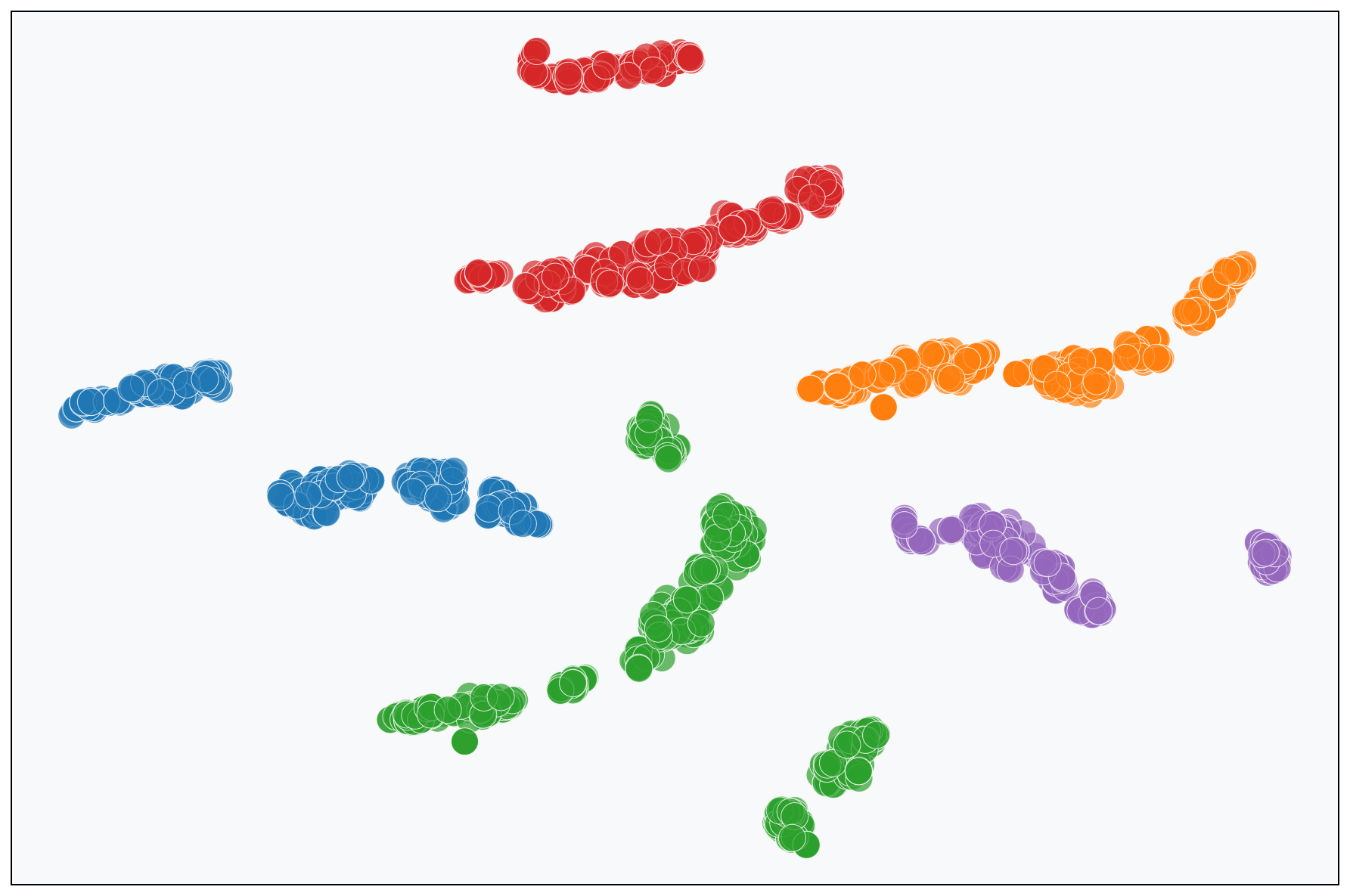}
        \caption*{k=17}
    \end{subfigure}\hfill
    \begin{subfigure}{0.24\linewidth}
        \includegraphics[width=\linewidth]{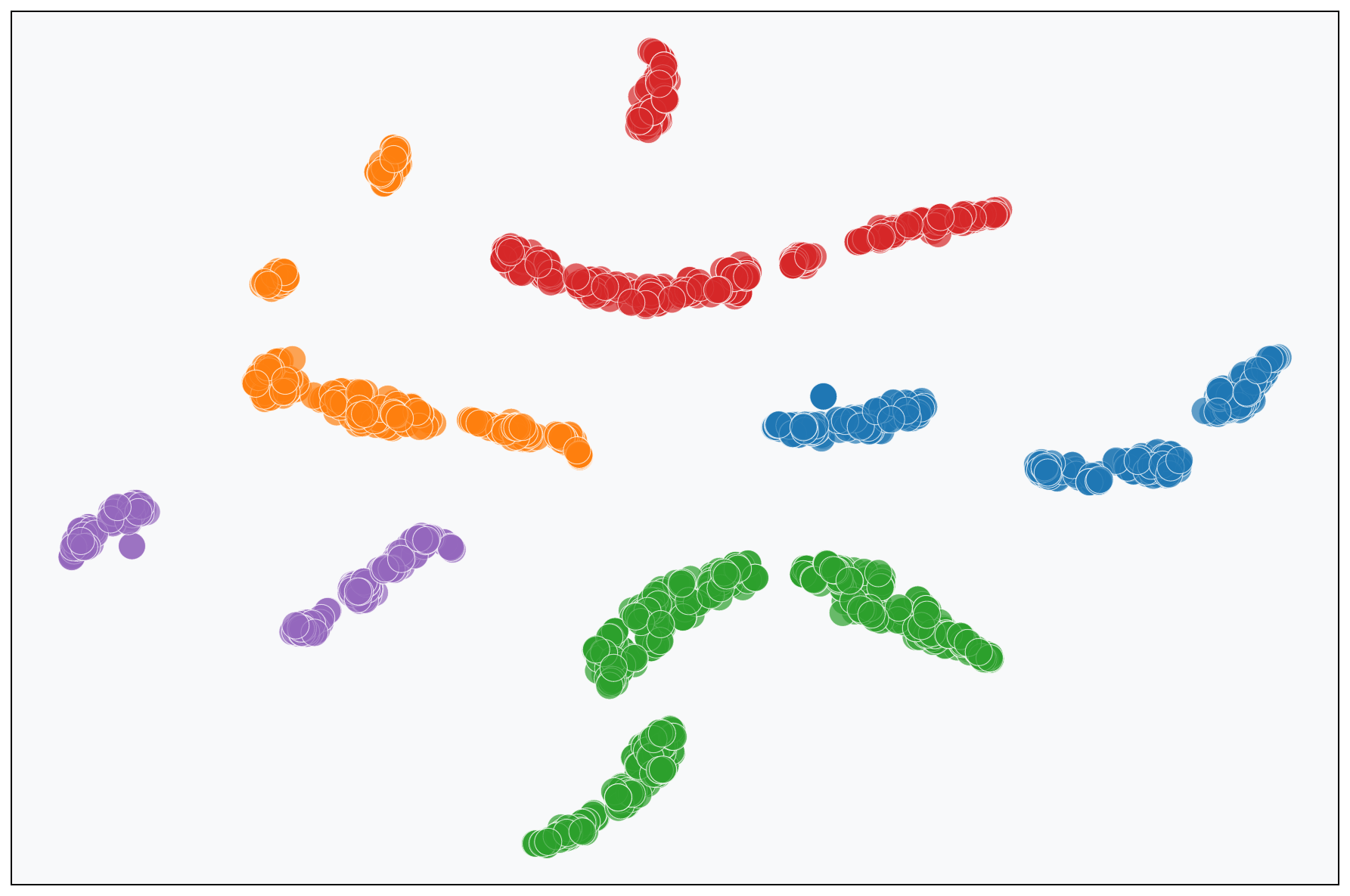}
        \caption*{k=18}
    \end{subfigure}\hfill
    \begin{subfigure}{0.24\linewidth}
        \includegraphics[width=\linewidth]{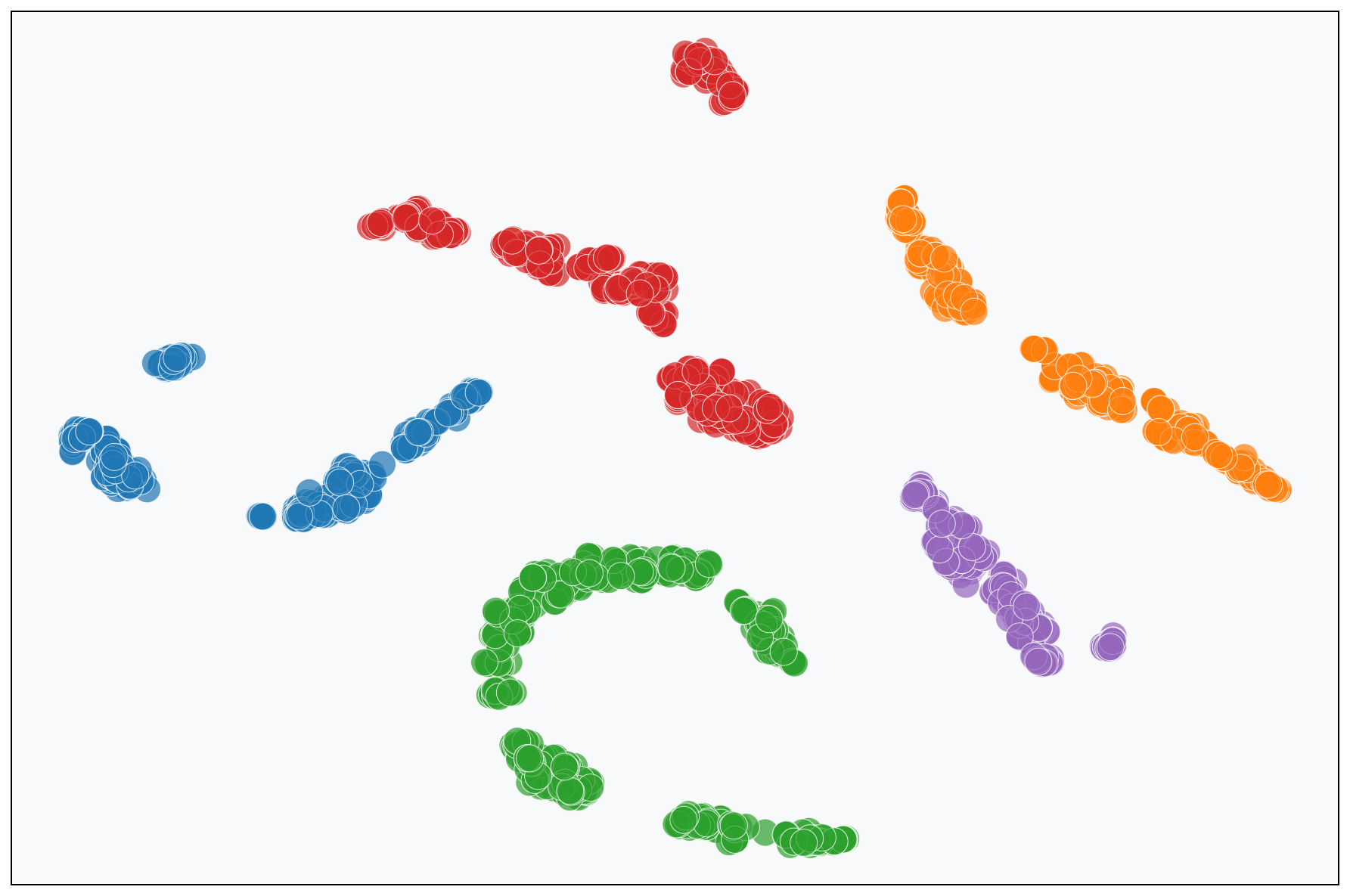}
        \caption*{k=19}
    \end{subfigure}\hfill
    \begin{subfigure}{0.24\linewidth}
        \includegraphics[width=\linewidth]{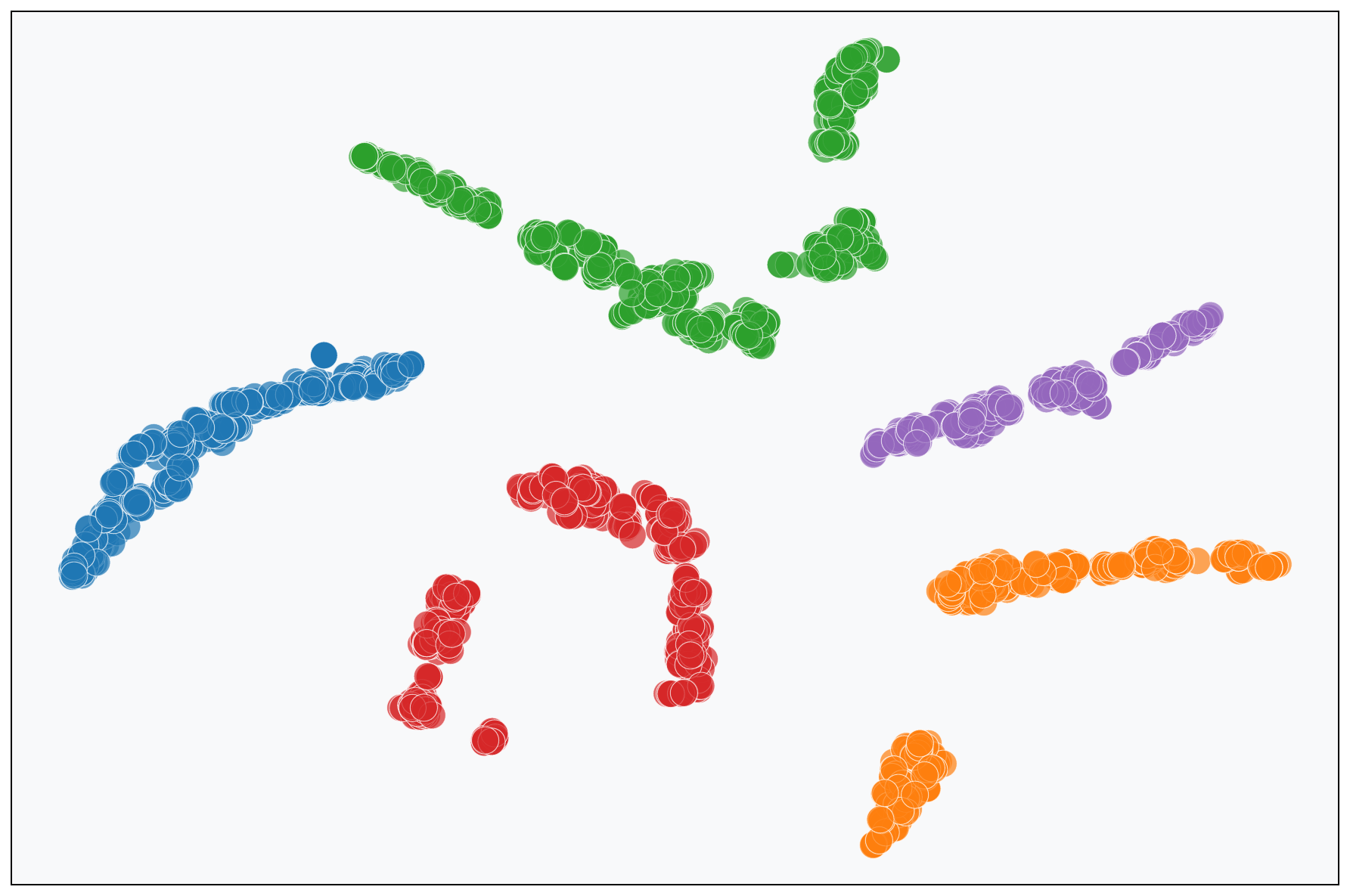}
        \caption*{k=20}
    \end{subfigure}
    
    \vspace{0.2em}
    
    \begin{subfigure}{0.24\linewidth}
        \includegraphics[width=\linewidth]{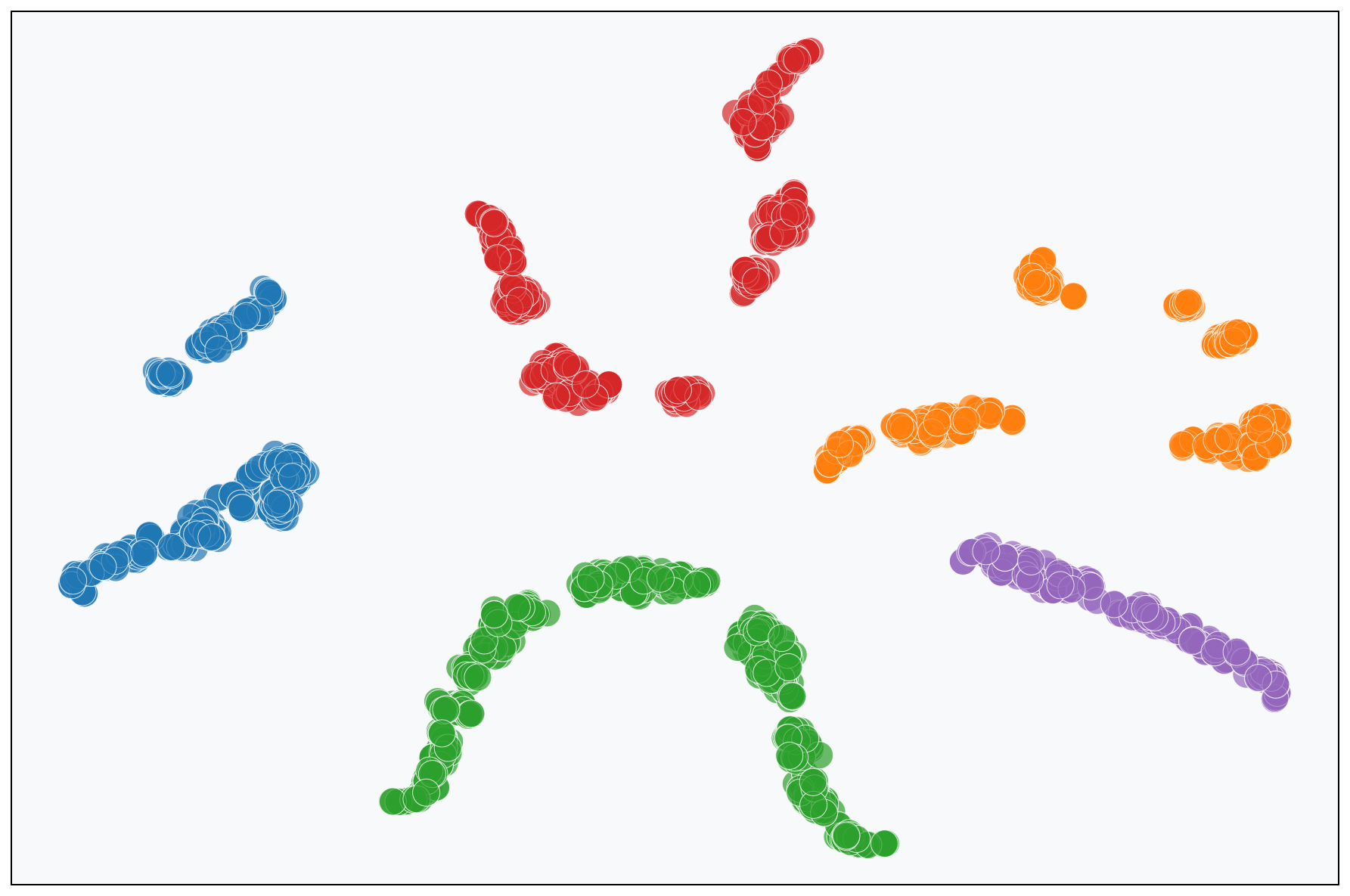}
        \caption*{k=21}
    \end{subfigure}\hfill
    \begin{subfigure}{0.24\linewidth}
        \includegraphics[width=\linewidth]{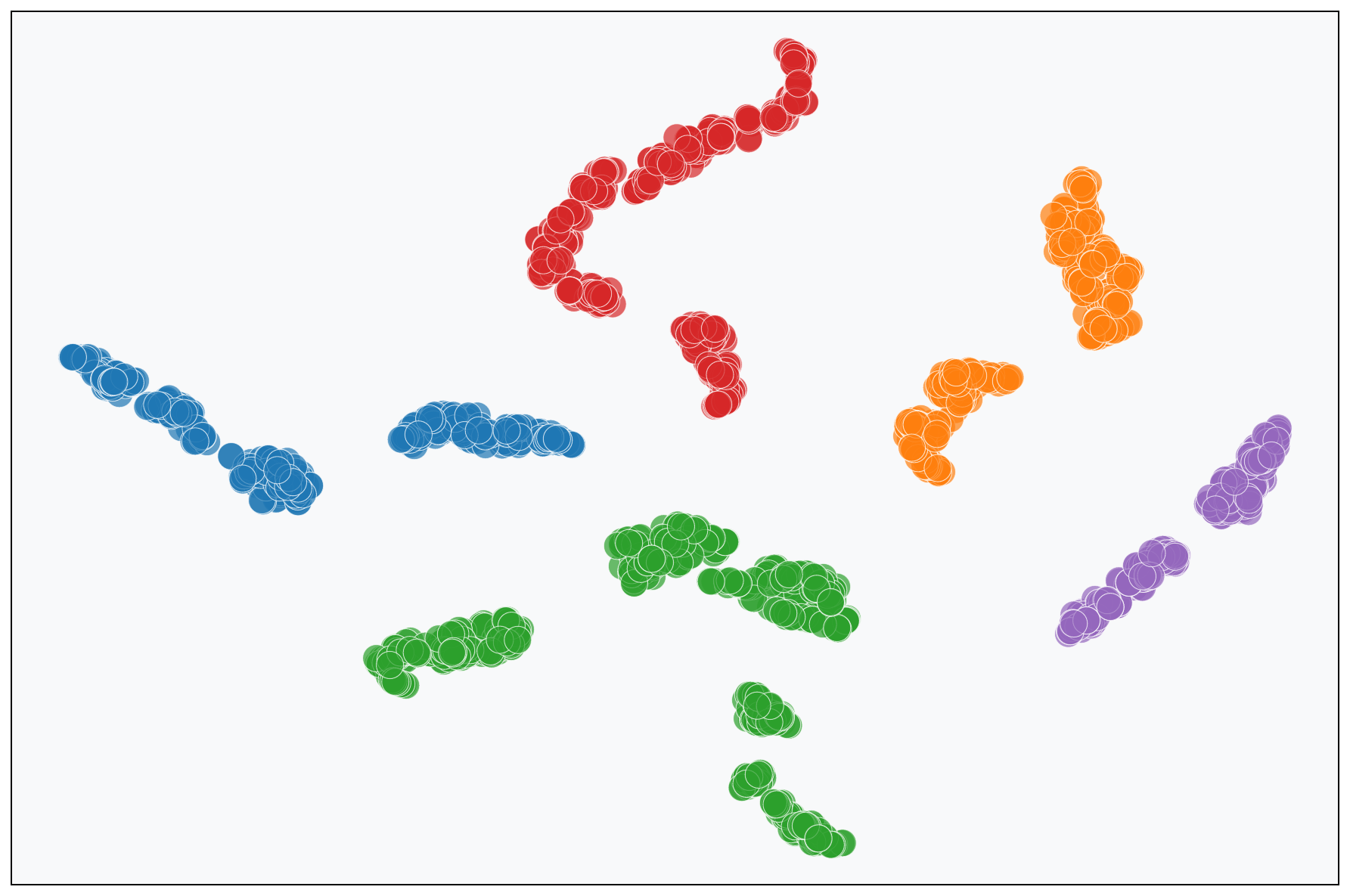}
        \caption*{k=22}
    \end{subfigure}\hfill
    \begin{subfigure}{0.24\linewidth}
        \includegraphics[width=\linewidth]{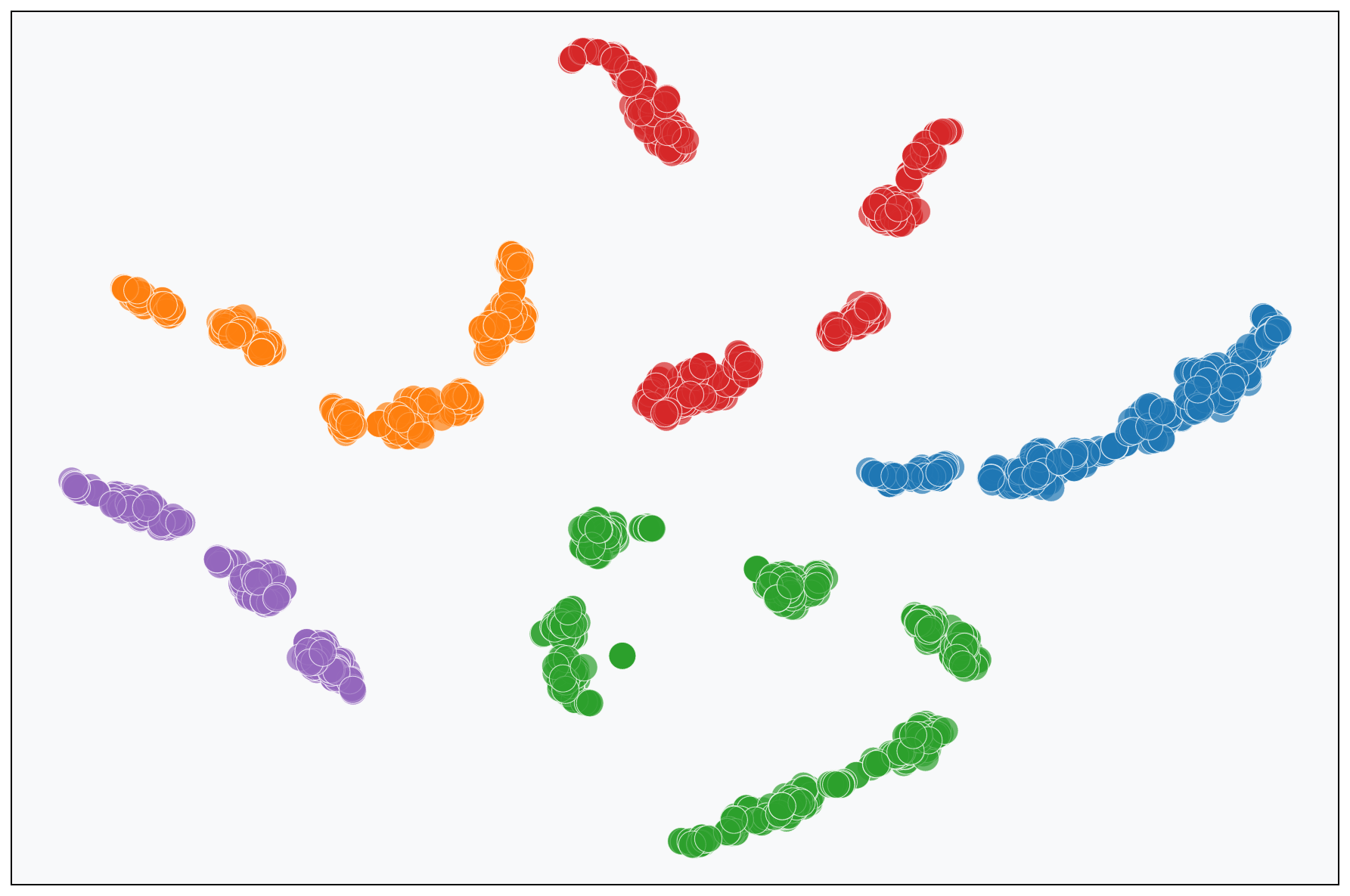}
        \caption*{k=23}
    \end{subfigure}\hfill
    \begin{subfigure}{0.24\linewidth}
        \includegraphics[width=\linewidth]{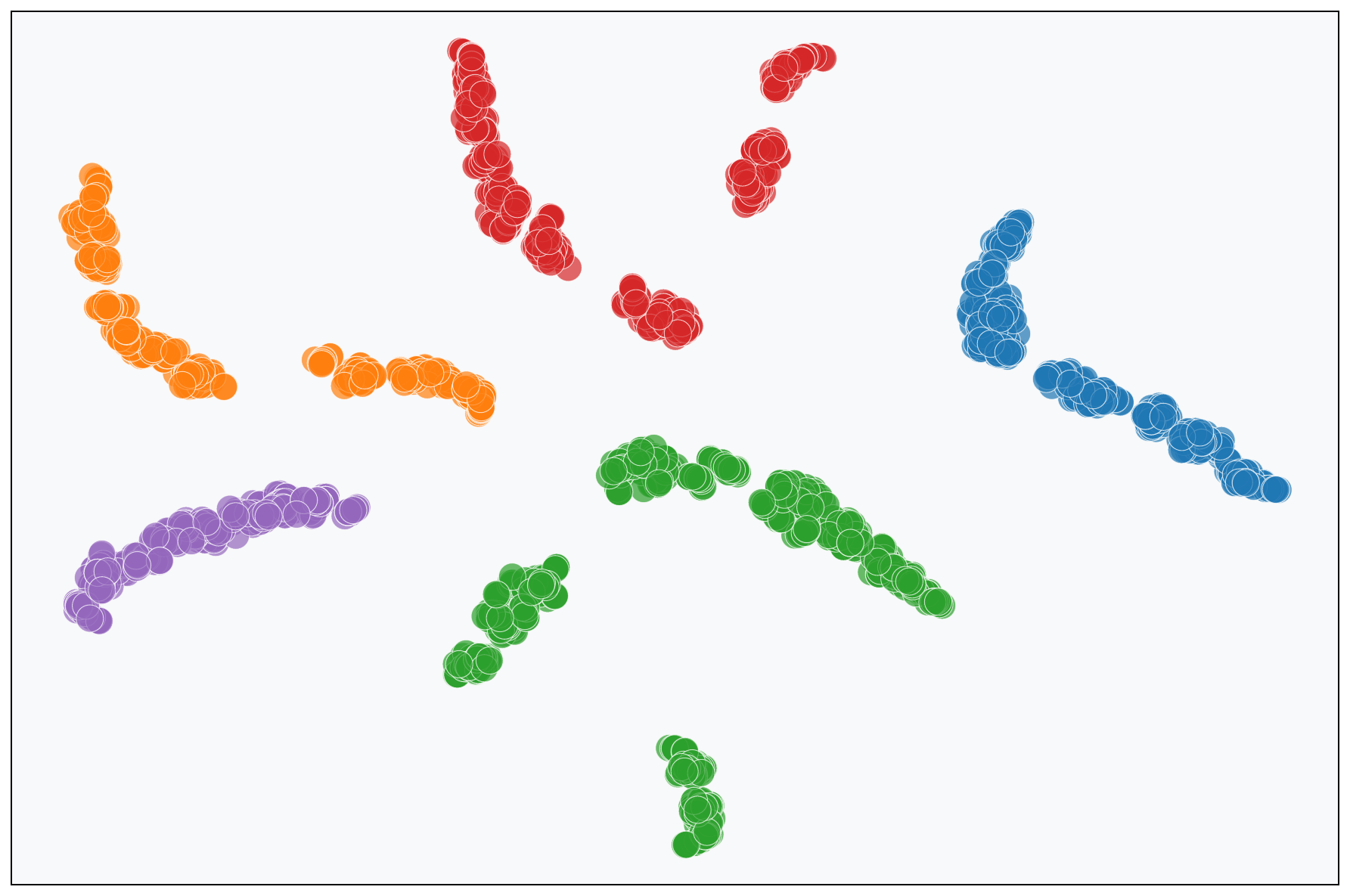}
        \caption*{k=24}
    \end{subfigure}

    \vspace{0.2em}
    
    \begin{subfigure}{0.24\linewidth}
        \includegraphics[width=\linewidth]{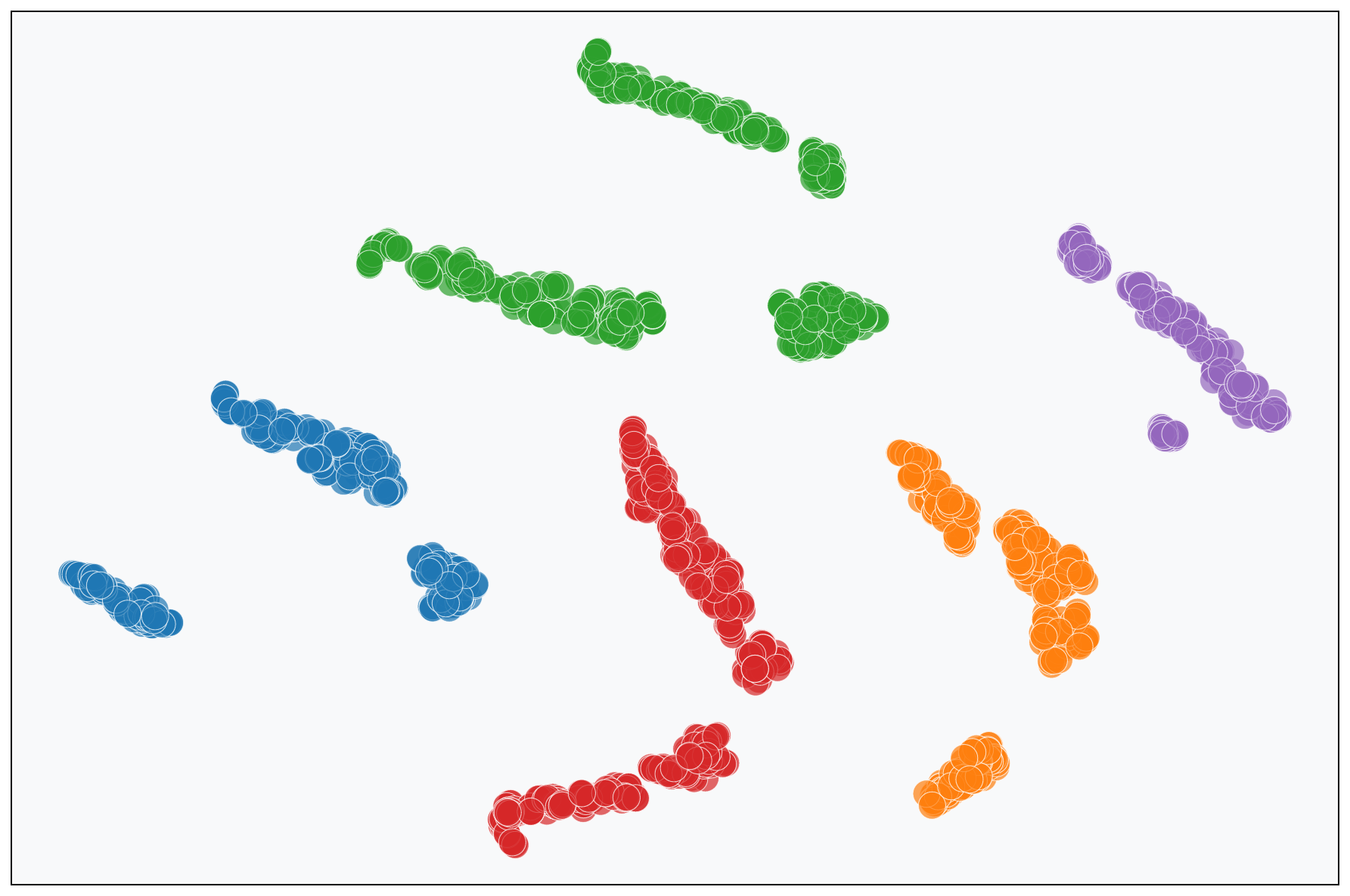}
        \caption*{k=25}
    \end{subfigure}\hfill
    \begin{subfigure}{0.24\linewidth}
        \includegraphics[width=\linewidth]{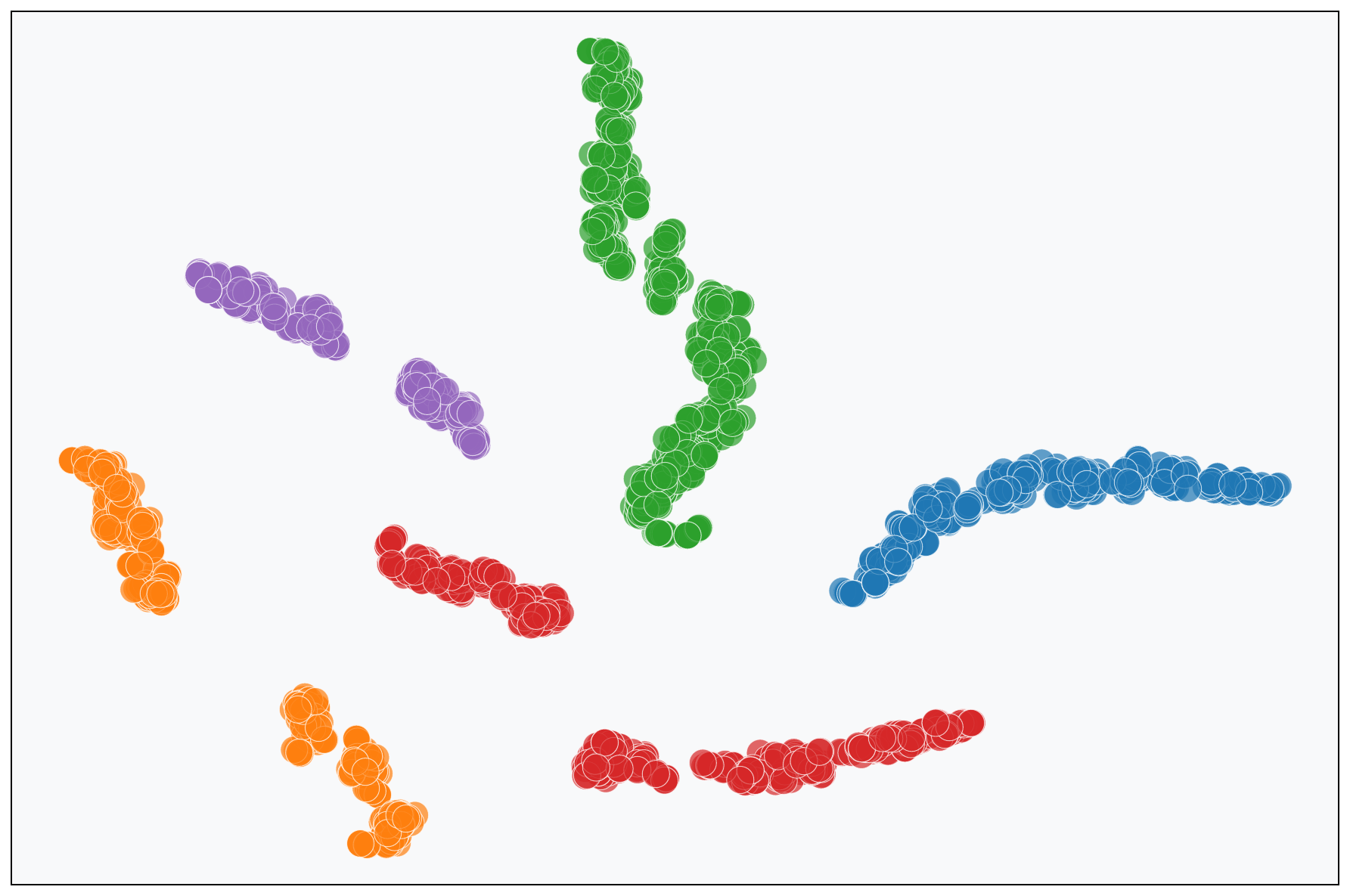}
        \caption*{k=26}
    \end{subfigure}\hfill
    \begin{subfigure}{0.24\linewidth}
        \includegraphics[width=\linewidth]{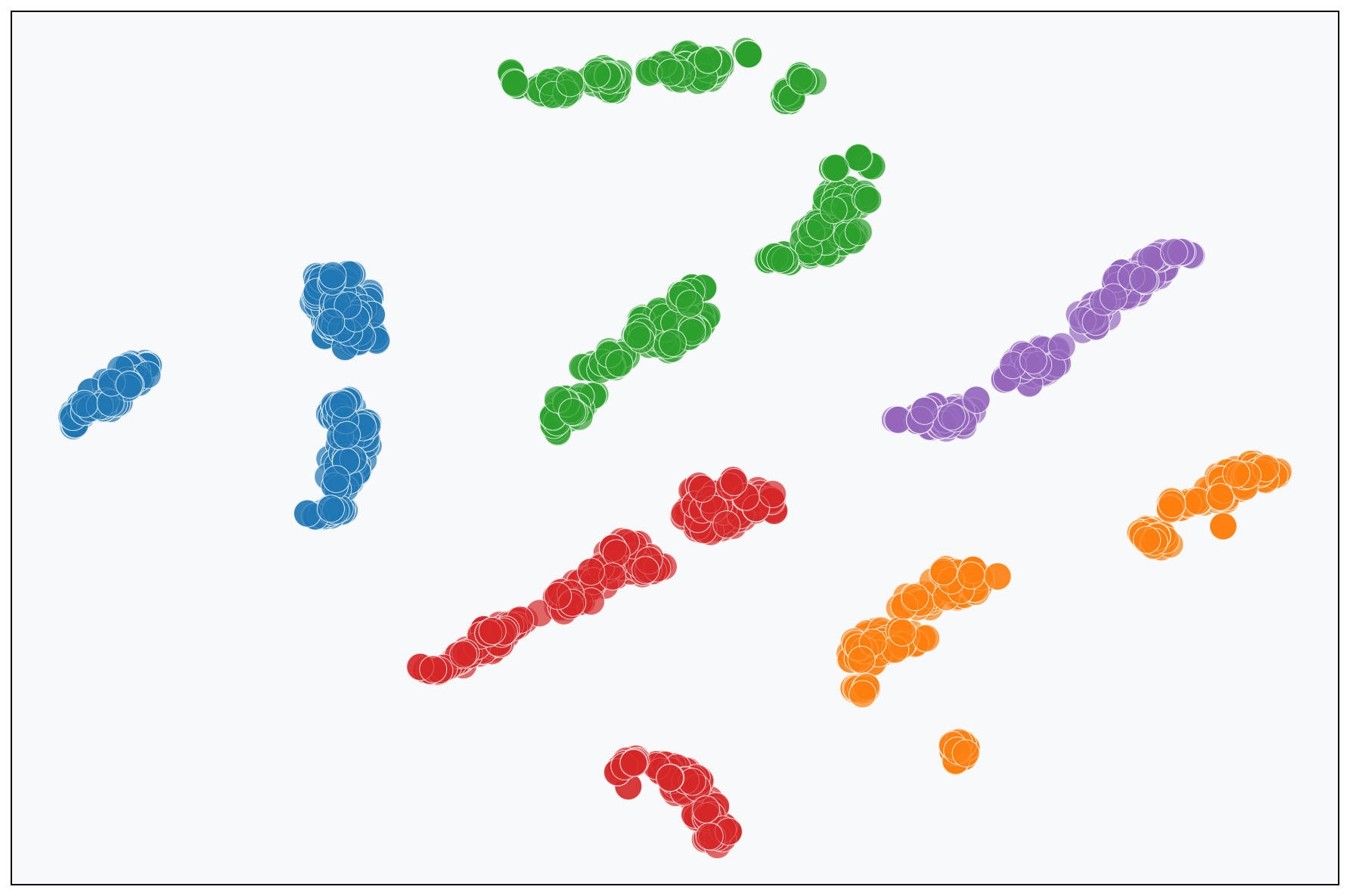}
        \caption*{k=27}
    \end{subfigure}\hfill
    \begin{subfigure}{0.24\linewidth}
        \includegraphics[width=\linewidth]{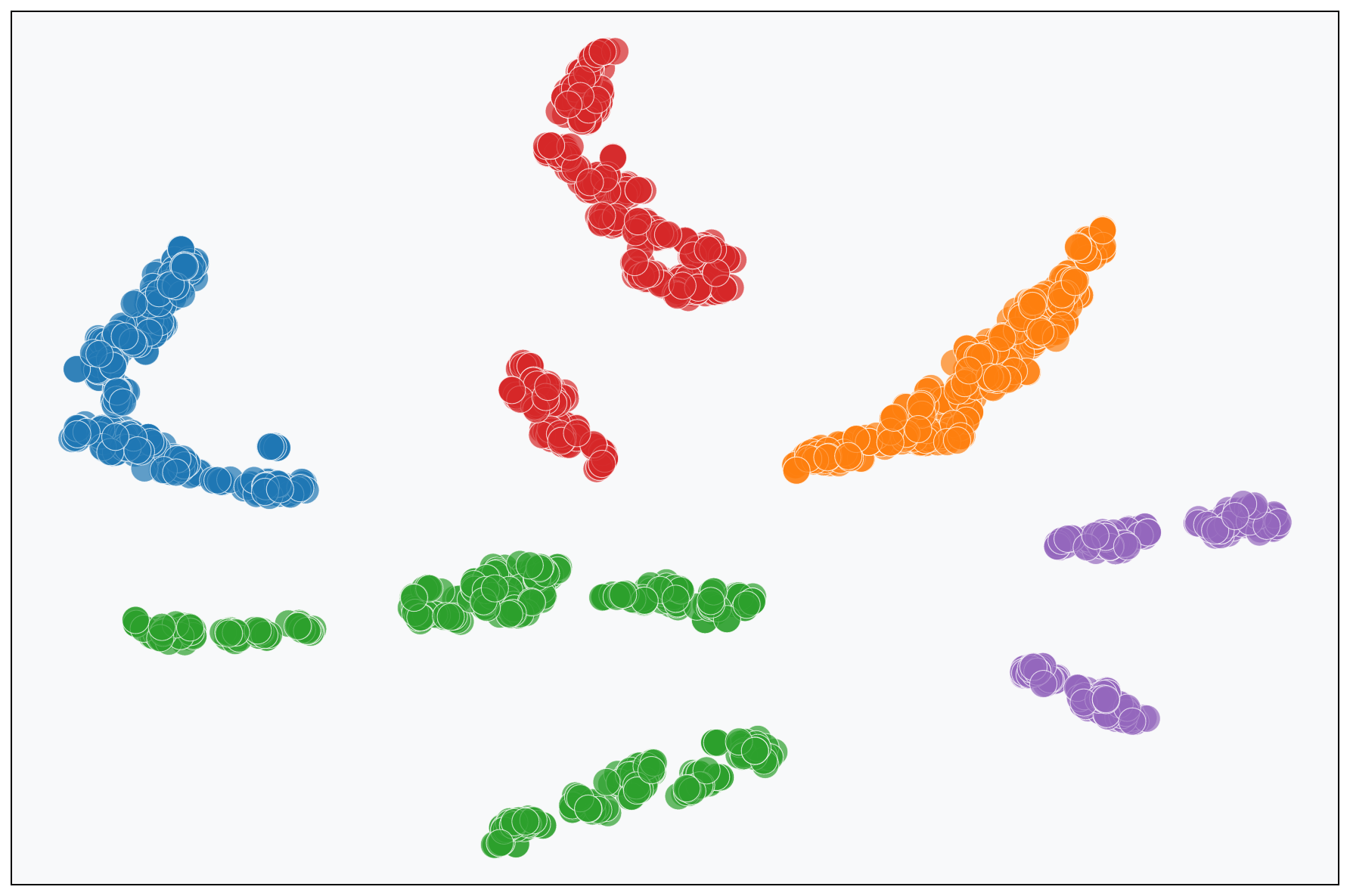}
        \caption*{k=28}
    \end{subfigure}

    \vspace{0.2em}
    
    \begin{subfigure}{0.24\linewidth}
        \includegraphics[width=\linewidth]{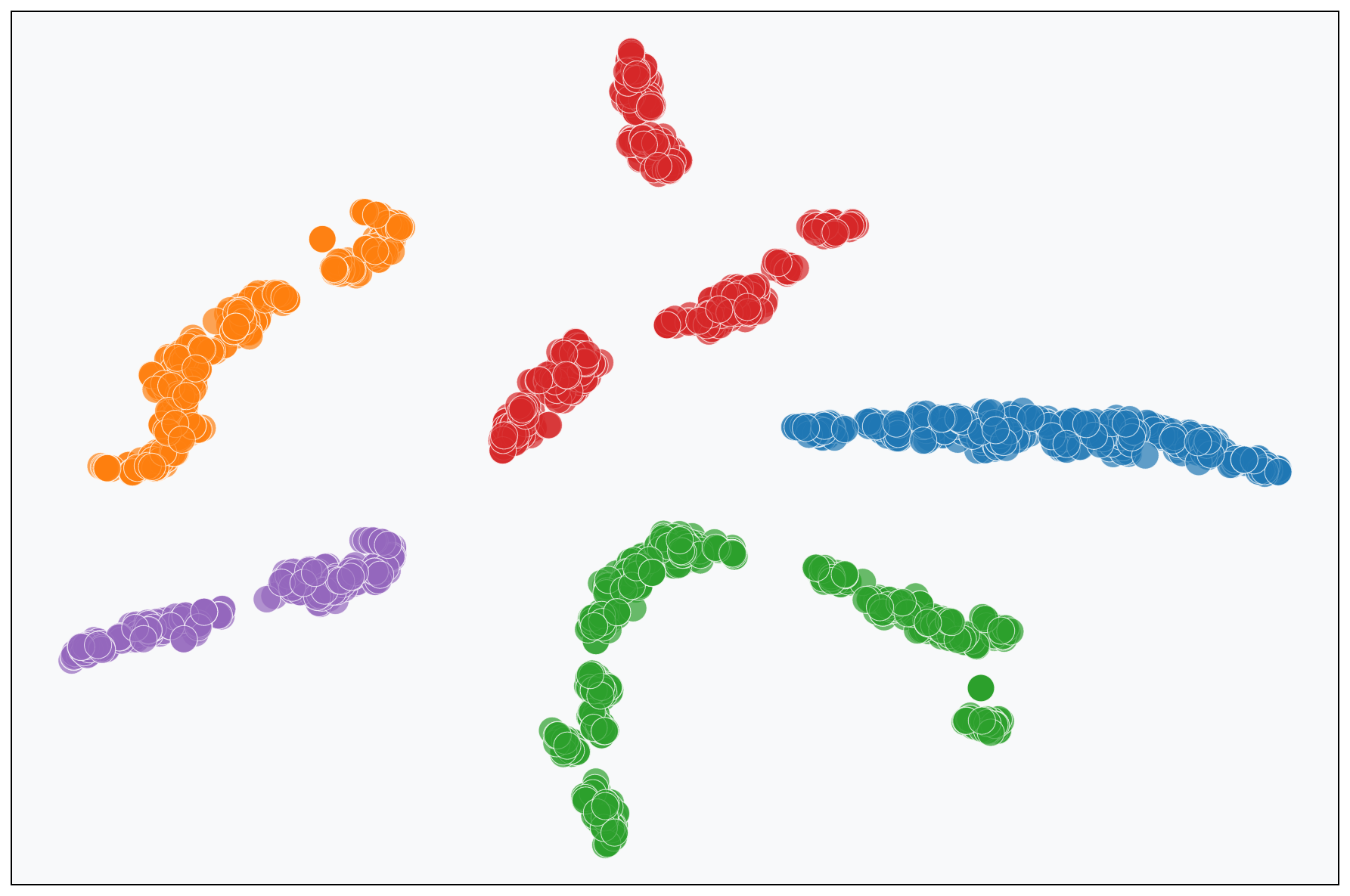}
        \caption*{k=29}
    \end{subfigure}\hfill
    \begin{subfigure}{0.24\linewidth}
        \includegraphics[width=\linewidth]{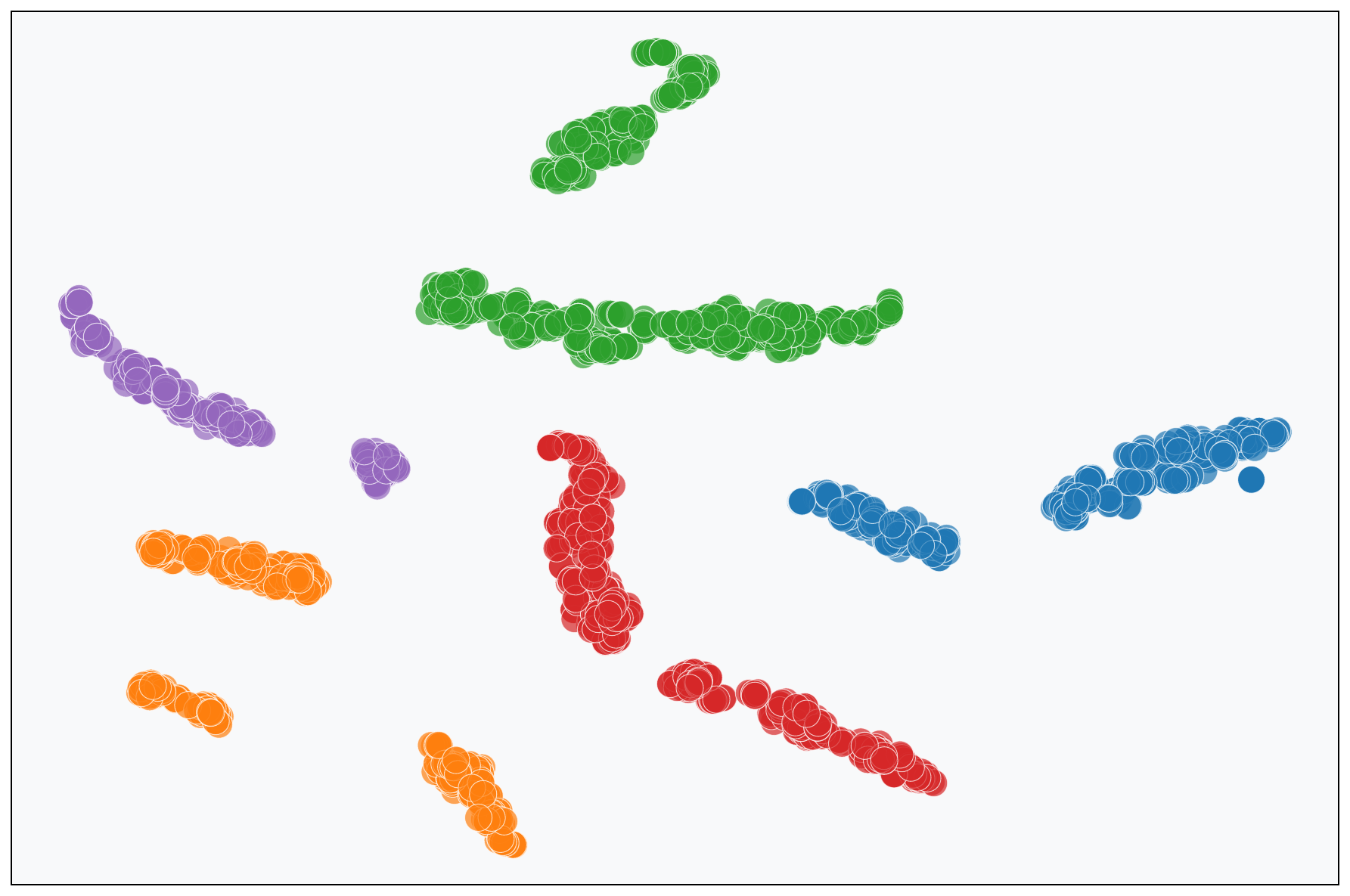}
        \caption*{k=30}
    \end{subfigure}\hfill
    \begin{subfigure}{0.24\linewidth}
        \includegraphics[width=\linewidth]{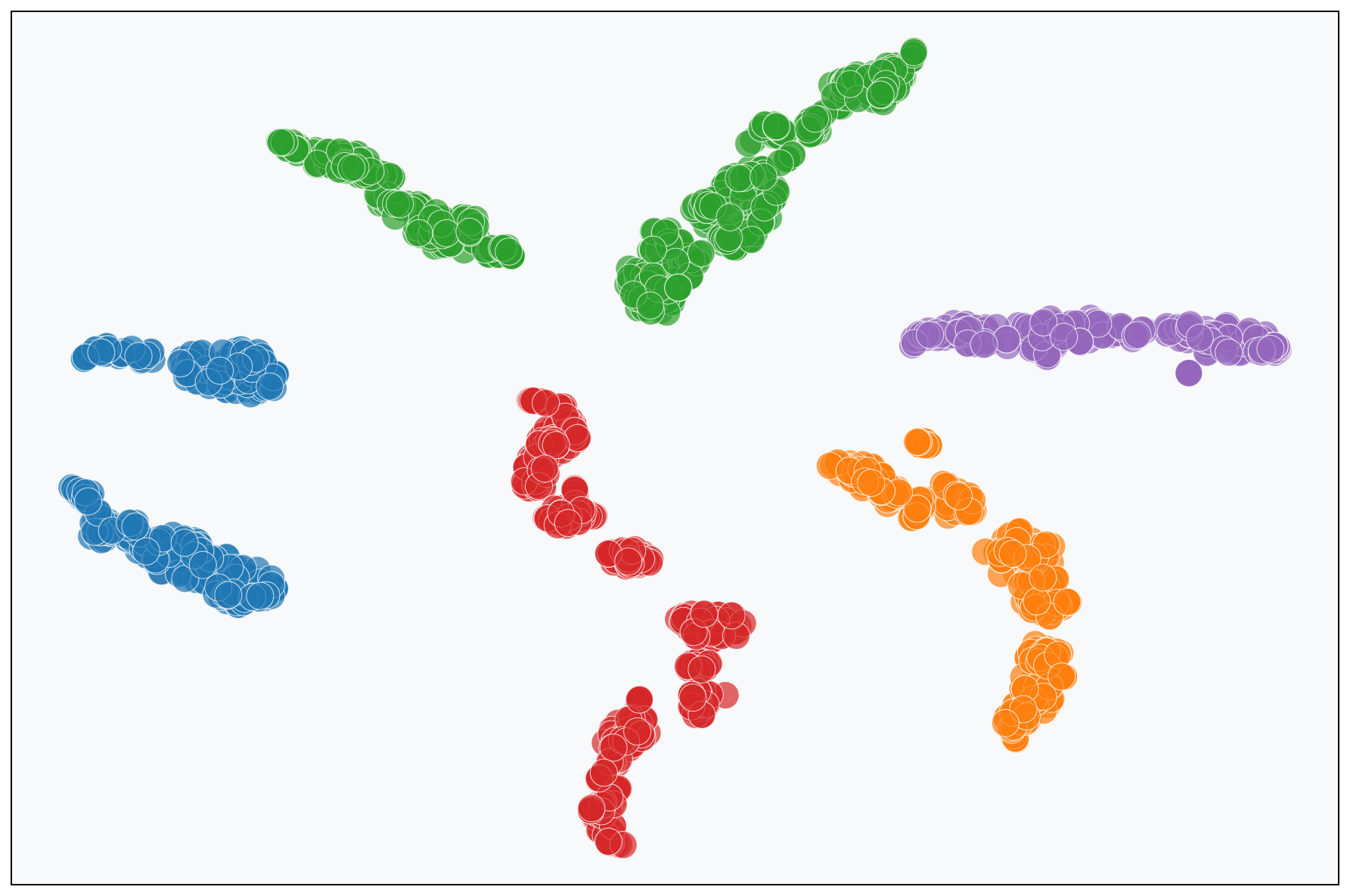}
        \caption*{k=31}
    \end{subfigure}\hfill
    \begin{subfigure}{0.24\linewidth}
        \includegraphics[width=\linewidth]{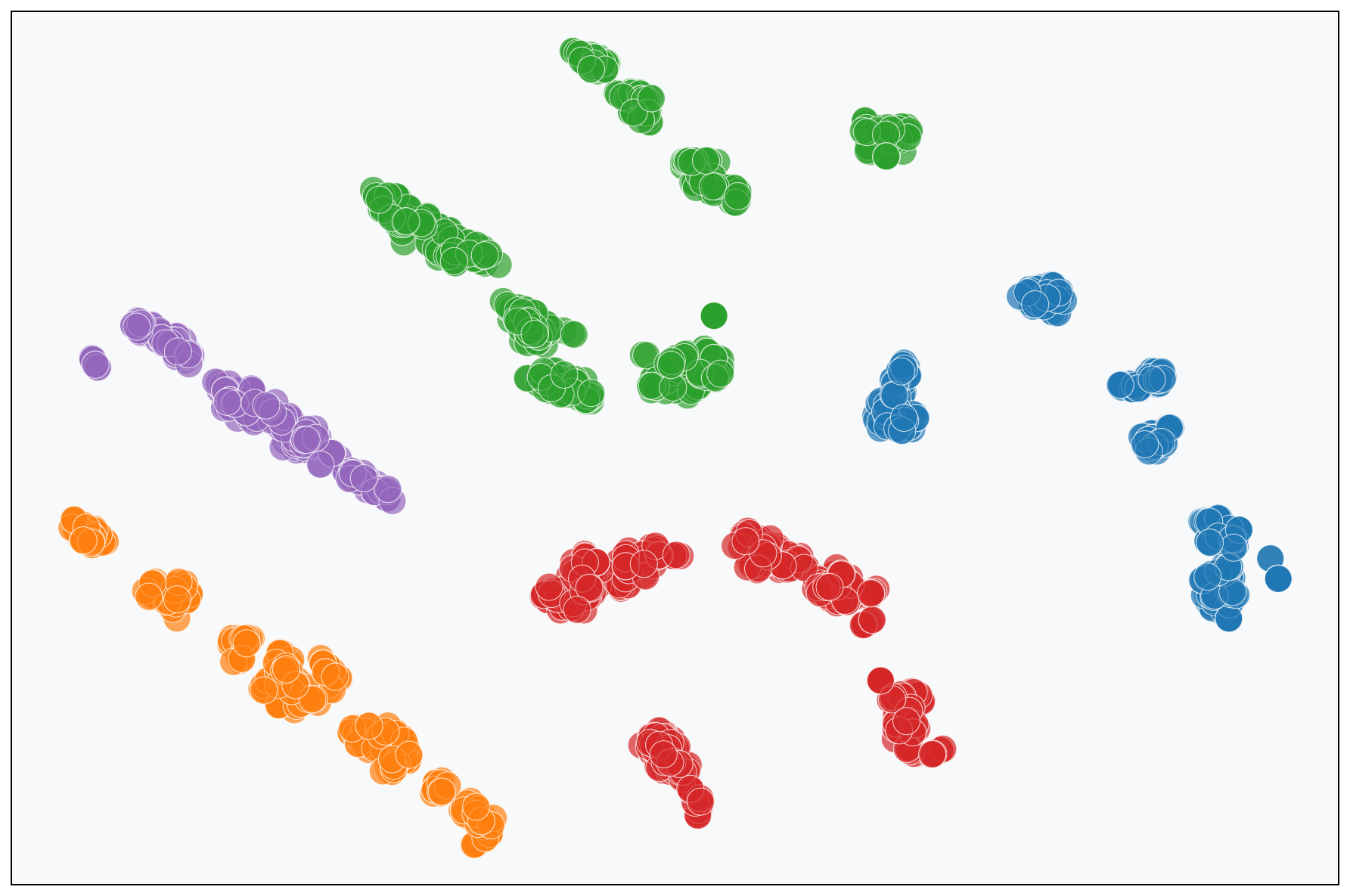}
        \caption*{k=32}
    \end{subfigure}
    \caption{t-SNE analysis on averaged $\beta$ on SST (continued).}
    \label{fig:tsne_layers2_sst} 
\end{figure*}

\paragraph{SST.} Figure \ref{fig:tsne_layers1_sst} and \ref{fig:tsne_layers2_sst} show the t-SNE plots of all 32 layers on SST.

\end{document}